\documentclass[a4,10pt,twoside,openright,italian,english]{book}

\usepackage[twoside=true]{geometry}

\usepackage[cam,center,a4,pdflatex,axes]{crop}

\usepackage{phdthesis}

\usepackage{fancyhdr}
\usepackage{color}
\usepackage{array}
\usepackage{mdwmath}
\usepackage{mdwtab}
\usepackage{amsmath,amssymb}
\usepackage{cite}
\usepackage{graphicx}
\usepackage{tabularx}
\usepackage{listings}
\usepackage{subfig}
\usepackage{booktabs}
\usepackage{latexsym}
\usepackage{color}
\usepackage{tcolorbox}
\usepackage{url}
\usepackage{bnf}
\usepackage{rotating}
\usepackage{multirow}
\usepackage{phdtitle}
\usepackage{paralist}
\usepackage{bibentry}
\usepackage[bookmarks=true,
bookmarksopen=true,hidelinks]{hyperref}
\usepackage[toc,acronym]{glossaries}
\usepackage{lscape}
\usepackage{algorithm}
\usepackage{algpseudocode}
\usepackage{longtable}
\usepackage[originalparameters]{ragged2e}
\usepackage[T1]{fontenc} 
\usepackage[utf8]{inputenc}
\usepackage[english,italian]{babel}
\usepackage{ulem}
\hypersetup{pdftitle={Title}, pdfauthor={Author}}

\nobibliography*

\department{ }
\author{Gianluca Carloni} 
\title{Human-aligned Deep Learning: Explainability, Causality, and Biological Inspiration}
\tutor{ }
\supervisor{ }
\phdyear{2025}
\phdmonth{April}
\phdcycle{ }

\newglossaryentry{matrix_channel}
{       name={$H^*$},
        description={Conjugate operation}
}
\newglossaryentry{trasp_x}
{       name={$[ x ]^{\rm T}$},
        description={transpose operator}
}
\newglossaryentry{vec_x}
{       name={\textbf{x}},
        description={vectors are in bold}
}
\newglossaryentry{floor_funct}
{       name={$\left\lfloor x \right\rfloor$},
        description={round to the lower integer of $x$}
}

\newacronym{label_CNN}{CNN}{Convolutional Neural Network}
\newacronym{label_Adam}{Adam}{Adaptive Moment Estimation}
\newacronym{label_AP}{AP}{Average Precision}
\newacronym{label_AUC}{AUC}{Area Under the Curve}
\newacronym{label_CT}{CT}{Computed Tomography}
\newacronym{label_DCNN}{DCNN}{Deep Convolutional Neural Network}
\newacronym{label_DL}{DL}{Deep Learning}
\newacronym{label_DNN}{DNN}{Deep Neural Network}
\newacronym{label_GPU}{GPU}{Graphical Processing Unit}
\newacronym{label_ILSVRC}{ILSVRC}{ImageNet Large Scale Visual Recognition Challenge}
\newacronym{label_LSTM}{LSTM}{Long Short-term Memory}
\newacronym{label_mAP}{mAP}{mean Average Precision}
\newacronym{label_ML}{ML}{Machine Learning}
\newacronym{label_MLP}{MLP}{Multilayer Perceptron}
\newacronym{label_MSE}{MSE}{Mean Squared Error}
\newacronym{label_PCA}{PCA}{Principal Component Analysis}
\newacronym{label_ReLU}{ReLU}{Rectified Linear Unit}
\newacronym{label_ResNet}{ResNet}{Residual Network}
\newacronym{label_RNN}{RNN}{Recurrent Neural Network}
\newacronym{label_ROC}{ROC}{Receiver Operation Characteristics}
\newacronym{label_AUROC}{AUROC}{Area Under the Receiver Operation Characteristics Curve}
\newacronym{label_SGD}{SGD}{Stochastic Gradient Descent}
\newacronym{label_SVM}{SVM}{Support Vector Machine}
\newacronym{label_MRI}{MRI}{Magnetic Resonance Imaging}
\newacronym{label_AM}{AM}{Activation Maximization}
\newacronym{label_ProtoPNet}{ProtoPNet}{Prototypical Part Network}
\newacronym{label_XAI}{XAI}{Explainable Artificial Intelligence}
\newacronym{label_AI}{AI}{Artificial Intelligence}
\newacronym{label_FFNN}{FFNN}{Feed Forward Neural Network}
\newacronym{label_NLP}{NLP}{Natural Language Processing}
\newacronym{label_CE}{CE}{Cross-Entropy}
\newacronym{label_LR}{LR}{Learning Rate}
\newacronym{label_PET}{PET}{Positron Emission Tomography}
\newacronym{label_TL}{TL}{Transfer Learning}
\newacronym{label_FT}{FT}{Fine Tuning}
\newacronym{label_TP}{TP}{True Positives}
\newacronym{label_FN}{FN}{False Negatives}
\newacronym{label_TN}{TN}{True Negatives}
\newacronym{label_FP}{FP}{False Positives}
\newacronym{label_CFE}{CFE}{Counterfactual Explanations}
\newacronym{label_CAB}{CAB}{Contextual Attention Block}
\newacronym{label_CoCoReco}{CoCoReco}{Connectivity-inspired Context-aware Recognition}
\newacronym{label_DAG}{DAG}{Directed Acyclic Graph}
\newacronym{label_TPR}{TPR}{True Positive Rate}
\newacronym{label_TNR}{TNR}{True Negative Rate}
\newacronym{label_FPR}{FPR}{False Positive Rate}
\newacronym{label_FNR}{FNR}{False Negative Rate}
\newacronym{label_CAM}{CAM}{Class Activation Mapping}
\newacronym{label_CPT}{CPT}{Conditional Probability Table}
\newacronym{label_iid}{i.i.d.}{Independent and Identically Distributed}
\newacronym{label_ID}{ID}{In-Distribution}
\newacronym{label_OOD}{OOD}{Out-Of-Distribution}
\newacronym{label_LLM}{LLM}{Large Language Model}
\newacronym{label_DG}{DG}{Domain Generalization}
\newacronym{label_BN}{BN}{Bayesian Network}
\newacronym{label_CD}{CD}{Causal Discovery}
\newacronym{label_GES}{GES}{Greedy Equivalence Search}
\newacronym{label_FCI}{FCI}{Fast Causal Inference}
\newacronym{label_JPD}{JPD}{Joint Probability Distribution}
\newacronym{label_SCM}{SCM}{Structural Causal Model}
\newacronym{label_ICM}{ICM}{Independence of Causal Mechanisms}
\newacronym{label_IRM}{IRM}{Invariant Risk Minimization}
\newacronym{label_CRL}{CRL}{Causal Representation Learning}
\newacronym{label_EHR}{EHR}{Electronic Health Record}
\newacronym{label_ITC}{ITC}{Infero-Temporal Cortex}
\newacronym{label_LGN}{LGN}{Lateral Geniculostriate Nucleus}
\newacronym{label_P_cell}{P cell}{Parvocellular}
\newacronym{label_M_cell}{M cell}{Magnocellular}
\newacronym{label_PFC}{PFC}{Pre-Frontal Cortex}
\newacronym{label_SC}{SC}{Superior Colliculus}
\newacronym{label_TV}{TV}{Total Variation}
\newacronym{label_WD}{WD}{Weight Decay}
\newacronym{label_CXR}{CXR}{Chest X-Ray}
\newacronym{label_KL}{KL}{Kullback-Leibler}
\newacronym{label_FFT}{FFT}{Fast Fourier Transform}
\newacronym{label_FC}{FC}{Fully Connected}
\newacronym{label_CV}{CV}{Cross Validation}
\newacronym{label_ES}{ES}{Early Stopping}
\newacronym{label_CF}{CF}{Counterfactual}
\newacronym{label_CLI}{CLI}{Command Line Interface}
\newacronym{label_GUI}{GUI}{Graphical User Interface}
\newacronym{label_SHAP}{SHAP}{SHapley Additive exPlanation}
\newacronym{label_PDP}{PDP}{Partial Dependency Plot}
\newacronym{label_PN}{PN}{Probability of Necessity}
\newacronym{label_PS}{PS}{Probability of Sufficiency}
\newacronym{label_GNN}{GNN}{Graph Neural Network}
\newacronym{label_T2w}{T2w}{T2 Weighted}
\newacronym{label_GS}{GS}{Gleason Score}
\newacronym{label_ISUP}{ISUP}{International Society of Urological Pathology}
\newacronym{label_FSL}{FSL}{Few-Shot Learning}
\newacronym{label_OSL}{OSL}{One-Shot Learning}
\newacronym{label_BAM}{BAM}{Bottleneck Attention Module}
\newacronym{label_RS}{RS}{Relational Scorer}
\newacronym{label_CL}{CL}{Contrastive Learning}
\newacronym{label_HCV}{HCV}{Human-inspired Computer Vision}
\newacronym{label_DM}{DM}{Diffusion Model}
\makeglossaries

\begin{document}
\selectlanguage{english}

\maketitle
\pagestyle{empty}
\cleardoublepage
\newpage

\thispagestyle{empty}
\vspace{\stretch{2}}\null


\pagestyle{empty}

\thispagestyle{empty}
    \null\vspace{\stretch {1}}
        \begin{flushright}
                "I don't quite know whether it is especially computer science or its subdiscipline Artificial Intelligence that has such an enormous affection for euphemism. We speak so spectacularly and so readily of computer systems that understand, see, decide, make judgments, and so on, without ourselves recognizing our own superficiality and immeasurable naivete with respect to these concepts. And, in the process of so speaking, we anesthetise our ability to evaluate the quality of our work and, what is more important, to identify and become conscious of its end use."\\  
                
                [...]\\
                
                "One can't escape this state without asking, again and again: "What do I actually do? What is the final application and use of the products of my work?" and ultimately, "am I content or ashamed to have contributed to this use?"\\
                
                [...] \\
                
                \textit{Joseph Weizenbaum\footnote{Born in Berlin to Jewish parents, Joseph Weizenbaum escaped Nazi Germany in 1936, immigrating to the United States. Later, he became a renowned computer scientist and professor at MIT, famous for creating ELIZA, the first chatbot, and being one of the pioneers of Artificial Intelligence.} from a talk given to the Gesellschaft für Informatik, at Karlsruhe, West Germany, on July 17, 1986.
                }
        \end{flushright}
\vspace{\stretch{2}}\null


\pagestyle{empty}
\setcounter{page}{1}
\pagenumbering{Roman}


\selectlanguage{english}


\pagestyle{fancy}

\selectlanguage{english}
\chapter*{Summary}
\lettrine{I}{n} recent years, artificial intelligence (AI) has become ubiquitous in everyday life, reaching beyond the technical community and into the popular consciousness. Indeed, it has entered mainstream media and applications, including ChatGPT-like conversational agents, Netflix recommendation systems, Instagram feed generators, and face recognition to unlock your device. Thus, similarly to electricity or computers, AI must be considered a general-purpose technology with varying applications.

In the realm of healthcare, particularly medical imaging, integrating AI, particularly deep learning (DL), holds immense potential. Daily, vast quantities of medical images are generated globally, necessitating new and efficient methods for analysis. Applying AI to these images could revolutionize diagnostics and patient care, for instance, by identifying high-risk patients, detecting diseases like cancer early, designing treatment plans, or developing personalized medicine biomarkers.

However, this area presents significant challenges, and the actual use of AI in medicine remains relatively limited. The heavy regulations, physical acquisition challenges, and privacy concerns behind medical data collection make it difficult to create large-scale datasets. Unlike humans, AI cannot work well on a low-data regime, and medical AI struggles even more to learn a robust data representation for downstream tasks. It falls short in distinguishing correlation from causation and ultimately learns shortcut paths to predict the outcome. This brings to unreliable behavior in post-deployment scenarios where the data distribution shifts, such as chest X-rays from a different hospital or machine. On top of that, the lack of interpretability and explainability of black box models complicates clinicians' ability to trust AI-generated insights.

This thesis seeks to align DL with humans' reasoning capabilities and needs to achieve a more efficient, explainable, and robust medical image classification. Specifically, we study and propose ways of tackling the limitations mentioned above from three perspectives: explainability, causality, and biological vision.
The thesis begins with an introduction to background notions regarding DL, medical image analysis, the field of eXplainable AI (XAI), causal DL, and the human visual system. Then, the first of the three perspectives begins - we study the effectiveness of neural networks' representation visualization on medical images and verify the applicability of an explainable-by-design solution for breast mass classification. This is followed by a comprehensive literature review at the intersection of XAI and causality, where we propose a general scaffold to cluster past and future research. That opens the doors to the second perspective, causality. We study and propose novel causality-driven modules to exploit feature co-occurrence in medical images and enable more effective and explainable predictions. The thesis then progresses to a deeper investigation of generalization capabilities, where we propose a new general framework that leverages causal concepts, contrastive learning, feature disentanglement, and injection of prior knowledge. Finally, we bridge to the third perspective, biological vision. We study how humans achieve object recognition and propose a connectivity-inspired neural network and an attention block that can model visual context.

Overall, our key findings indicate that: (i) simple activation maximization is not sufficient for getting visual insights into medical imaging DL models; (ii) prototypical-part learning is effective and its explanations are aligned with a radiologist's viewpoint; (iii) the concepts of explanation and causation, and the corresponding research fields of XAI and causal ML, are strongly intertwined; (iv) it is possible to exploit weak causal signals within medical images without a priori information and our module improves performance and explanations; (v) our causality-based framework effectively leverages information from multiple medical domains and attains robust generalization to out-of-distribution data; and, lastly, (vi) incorporating circuit motifs found in biological brains proved effective for a more human-aligned image recognition.

The ultimate goal of this thesis is to offer the scientific community insights into ways to render DL more aligned with human reasoning and needs and to propose promising research directions that can help bridge the gap between academic developments and practical applications. The implications of having more efficient, robust, explainable, and generalizable DL models are clinical usability and trust improvement, diagnostic error reduction, and safer adoption.
\selectlanguage{english}


\selectlanguage{english}


\selectlanguage{english}
\chapter*{List of publications}

\section*{International Journals}
\begin{enumerate}
    \item \textbf{Carloni, G.}, Berti, A., \& Colantonio, S. (2024). The role of causality in explainable artificial intelligence. \emph{Wiley Interdisciplinary Reviews: Data Mining and Knowledge Discovery}. Accepted DOI:10.1002/WIDM.70015, currently under production.

    \item \textbf{Carloni, G.}, Colantonio, S. (2024). Exploiting causality signals in medical images: A pilot study with empirical results. \emph{Expert Systems with Applications}, 249, 123433, Elsevier.
    
    \item \textbf{Carloni, G.}, Garibaldi, C., Marvaso, G., Volpe, S., Zaffaroni, M., Pepa, M., ... \& Jereczek-Fossa, B. A. (2023). Brain metastases from NSCLC treated with stereotactic radiotherapy: prediction mismatch between two different radiomic platforms. \emph{Radiotherapy and Oncology}, 178, 109424.
\end{enumerate}

\section*{International Conferences/Workshops with Peer Review}
\begin{enumerate}    
    \item \textbf{Carloni, G.}, Tsaftaris, S. A., \& Colantonio, S. (2024, October). CROCODILE: Causality aids RObustness via COntrastive DIsentangled LEarning. In \emph{International Workshop on Uncertainty for Safe Utilization of Machine Learning in Medical Imaging} (pp. 105-116), MICCAI. Cham: Springer Nature Switzerland.

    \item \textbf{Carloni, G.}, Colantonio, S. (2024, September). Connectivity-Inspired Network for Context-Aware Recognition. In \emph{International Workshop on Human-inspired Computer Vision}, ECCV. Cham: Springer Nature Switzerland. Accepted, not published yet.
    
    \item Xue, Y.*, Du, Y.*, \textbf{Carloni, G.}*, Pachetti, E.*, Jordan, C.*, \& Tsaftaris, S. A. (2023, October). Cine cardiac MRI reconstruction using a convolutional recurrent network with refinement. In \emph{International Workshop on Statistical Atlases and Computational Models of the Heart} (pp. 421-432), MICCAI. Cham: Springer Nature Switzerland.
    
    \item \textbf{Carloni, G.}*, Pachetti, E.*, \& Colantonio, S. (2023). Causality-Driven One-Shot Learning for Prostate Cancer Grading from MRI. In \emph{Proceedings of the IEEE/CVF International Conference on Computer Vision} (pp. 2616-2624).
    
    \item Berti, A.*, \textbf{Carloni, G.}*, Colantonio, S., Pascali, M. A., Manghi, P., Pagano, P., ...\& Barucci, A. (2022, September). Data models for an imaging bio-bank for colorectal, prostate and gastric cancer: the NAVIGATOR project. In \emph{2022 IEEE-EMBS International Conference on Biomedical and Health Informatics (BHI)} (pp. 01-04). IEEE.

    \item \textbf{Carloni, G.}*, Berti, A.*, Iacconi, C., Pascali, M. A., \& Colantonio, S. (2022, August). On the applicability of prototypical part learning in medical images: breast masses classification using ProtoPNet. In \emph{International Conference on Pattern Recognition} (pp. 539-557). Cham: Springer Nature Switzerland.
\end{enumerate}

\section*{Others}
\begin{enumerate}
    \item Berti, A., Buongiorno, R., \textbf{Carloni, G.}, Caudai, C., Conti, F. del Corso, G., ... \& Colantonio, S. (2024, May). From Covid-19 detection to cancer grading: how medical-AI is boosting clinical diagnostics and may improve treatment. In \emph{4th CINI National Lab AIIS Conference on Artificial Intelligence (Ital-IA 2024)}.

    \item Berti, A., Buongiorno, R., \textbf{Carloni, G.}, Caudai, C., del Corso, G., Germanese, D., ... \& Colantonio, S. (2023, May). Exploring the potential and challenges of AI in clinical diagnostics and remote assistance of individuals. In \emph{3rd CINI National Lab AIIS Conference on Artificial Intelligence (Ital-IA 2023)}.

    \item \textbf{Carloni, G.}, Marvaso, G., Garibaldi, C., Zaffaroni, M., Volpe, S., Pepa, M., ... \& Jereczek-Fossa, B. A. (2022, May). PO-1783 Leverage radiomic and clinical data in predicting SRS treatment outcomes in patients with brain mets. \emph{Radiotherapy and Oncology}, 170, Supplement 1.

    \item Righi, M., Leone, G. R., Carboni, A., Caudai, C., Colantonio, S., Kuruoglu, E. E., ... \& Moroni, D. (2022). SI-Lab Annual Research Report 2021, \emph{Technical report}. Consiglio Nazionale delle Ricerche.

    \item Zaffaroni, M., \textbf{Carloni, G.}, Volpe, S., Garibaldi, C., Marvaso, G., Gandini, S., ... \& Jereczek-Fossa, B. A. (2021). PO-1794 Features robustness in the radiomic workflow: the impact of software choice on feature variability. \emph{Radiotherapy and Oncology}, 161, S1519-S1520, Elsevier.
\end{enumerate}
*: shared first authorship. 
\selectlanguage{english}

\tableofcontents
\cleardoublepage
\newpage

\setcounter{page}{1}
\pagenumbering{arabic}

\cleardoublepage
\chapter{Introduction}
\label{chap:introduction}
\section{Rationale}
Imagery is prevalent in our lives as it effectively communicates information rapidly and transcends linguistic or cultural obstacles.
Every day, we collectively create, share, and get a remarkable volume of digital photos or videos. 
On the other hand, digital images are not only \textit{natural} images (e.g., photographs), but also \textit{medical} images. Over the past decade, the amount of data generated by radiological imaging modalities and digital pathology has skyrocketed, creating a data explosion that is transforming healthcare. The leading modality is X-rays, with over 3.6 billion examinations performed globally each year, followed by ultrasounds (1.5 billion), \acrshort{label_CT} scans (450 million), and \acrshort{label_MRI}s (about 150 million)\footnote{\url{https://www.grandviewresearch.com/industry-analysis/medical-imaging-systems-market}}. 
Given this situation, there is a growing interest in developing automated tools for managing medical visual data, thus aiding radiologists and pathologists in identifying abnormalities, measuring biomarkers, and forecasting outcomes.
Early methods utilized Computer Vision to develop image descriptors based on low-level, manually defined features of images, including edge distribution and color patterns.
However, this is a subjective and labor-intensive process that demands specialized knowledge and is not scalable to meet the current trends in image creation.
On top of that, these descriptors often fail to encompass the high-level concepts that human annotators assign to images (e.g., a radiologist annotating an X-ray scan).
In this regard, classical Machine Learning (\acrshort{label_ML}) methodologies represent an exciting way of overcoming hand engineering and significantly enhancing the performance of models by assisting with selecting, modulating, and merging features.
Still, identifying and extracting finely-tuned, problem-specific features remain crucial in complex perception tasks like vision, especially in high-stakes applications such as medical image analysis.

To address these limitations, techniques shifted towards inferring the visual semantics of images based solely on their visual content, utilizing information that machines can automatically extract from raw pixels and represent in numerical formats.
This is the era of Deep Learning (\acrshort{label_DL}), a family of \acrshort{label_ML} that allowed researchers to automate the perception and interpretation of visual data by extracting highly abstract information from raw pixels, significantly reducing the need for human labor.
Accordingly, \acrshort{label_DL} seeks to autonomously learn a hierarchy of feature extractors from data, transforming the input into a high-level feature space specifically designed to address a particular task.
Convolutional Neural Networks (\acrshort{label_CNN}s), in particular, transformed feature engineering and visual comprehension, surpassing manually crafted models across various vision tasks, including object detection, segmentation, and image classification.
Accordingly, the last ten years have seen a substantial adoption of Artificial Intelligence (\acrshort{label_AI}), which has set the new state of the art in numerous applications, such as medical image analysis for disease classification.

However, implementing \acrshort{label_DL}-based solutions presents significant engineering challenges. In this thesis, we study and propose ways of rendering such solutions more aligned with human users' reasoning, capabilities, and demands. Indeed, we tackle critical limitations encountered in using deep models and propose their adoption in novel approaches for efficient, explainable, and robust medical image classification.

\section{Objectives and Contributions}
\begin{figure}[h]
\centering
 \includegraphics[width=0.99\textwidth]{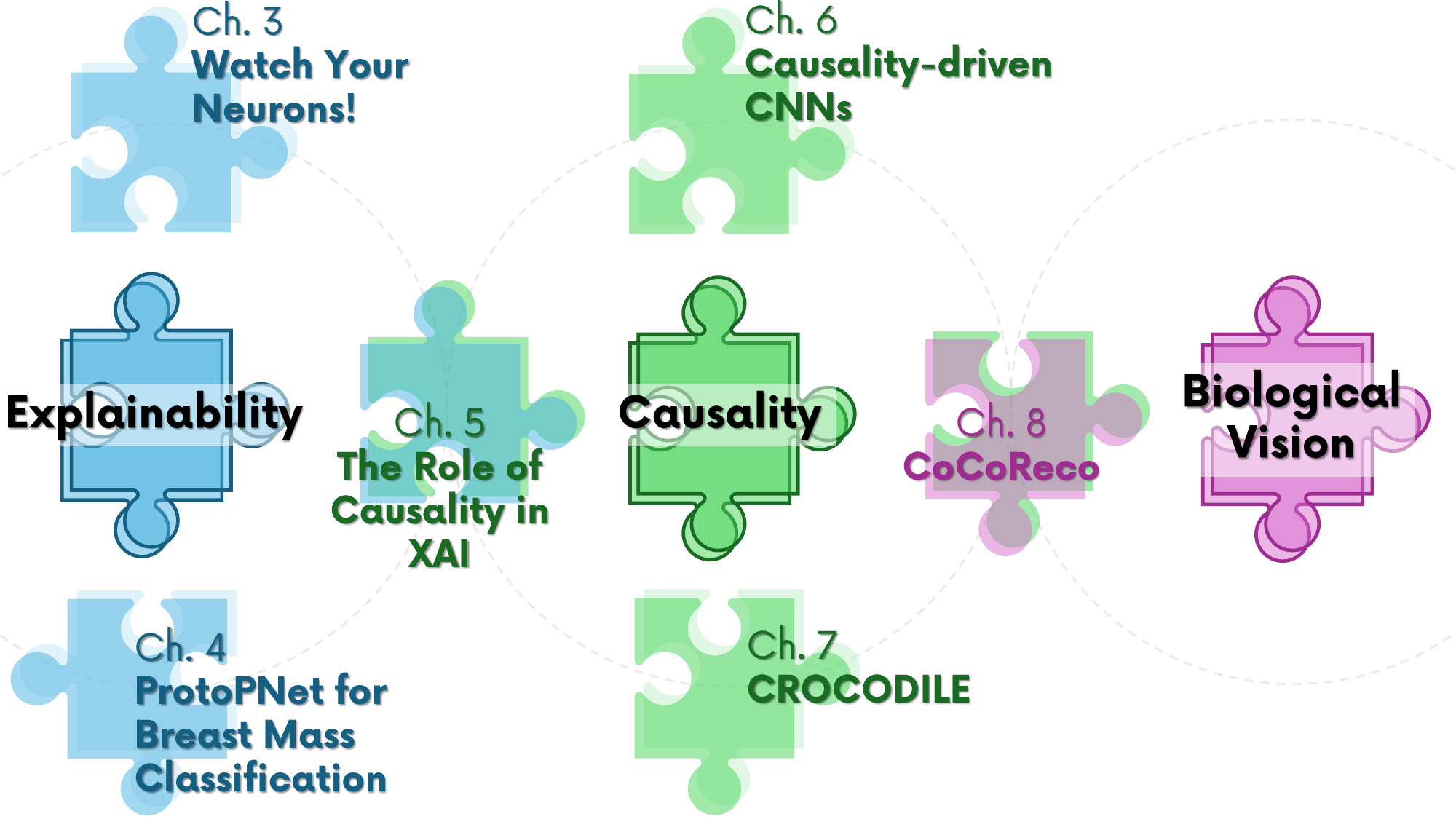}
\caption{\textbf{Dissertation Road-map}.}
\label{fig:intro_roadmap}
\end{figure}

This thesis investigated human-aligned \acrshort{label_DL} from three perspectives: explainability, causality, and biological inspiration. Accordingly, excluding the introduction (\ref{chap:introduction}), background (\ref{chap:background}), and conclusion (\ref{chap:conclusions}) chapters, this dissertation is divided into three logical parts. Figure \ref{fig:intro_roadmap} shows a road map including the three spheres of interest and the chapters pertaining to them. 
\newpage

In the following, we highlight this research's original contributions to those fields, including new findings, methodologies, and theoretical advancements.
\begin{figure}
\centering
 \includegraphics[width=0.99\textwidth]{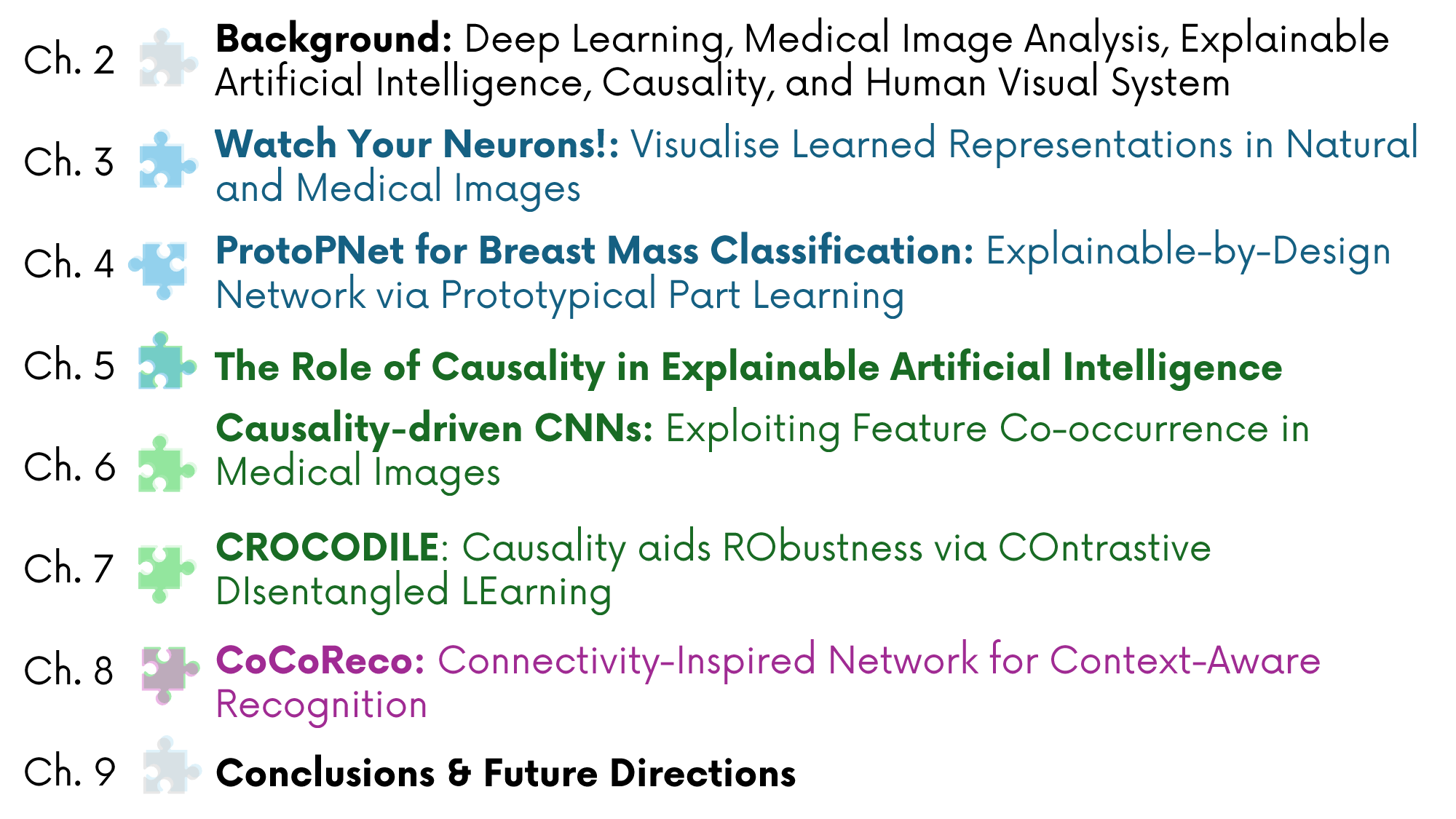}
\caption{\textbf{Chapters' Organization}. Click these numbers to skip to Chapter: \ref{chap:background}, \ref{chap:XAI_wyn}, \ref{chap:XAI_protopnet}, \ref{chap:review_XAI_causality}, \ref{chap:mulcat_ICCV_ESWA}, \ref{chap:crocodile}, \ref{chap:cocoreco_ECCV}, \ref{chap:conclusions}.}
\label{fig:intro_roadlist}
\end{figure}
\subsection{Part One: Explainability}
In the first part of the thesis (Chapters \ref{chap:XAI_wyn} and \ref{chap:XAI_protopnet}), we focus on the \textbf{explainability} of \acrshort{label_DL} models, investigating the need for interpretations arising in human end-users when presented with black-box decisions. Then, Chapter \ref{chap:review_XAI_causality} acts as a bridge section towards the Part Two of this thesis.

\subsubsection{Contributions in Chapter 3}
We start our journey by exploring an interesting feature-visualization approach to gain insights into the representations learned in the final neurons on a trained deep CNN: \textit{activation maximization} (\acrshort{label_AM}). Our first contribution is to apply \acrshort{label_AM} on natural images to study the effect of varying optimization settings. For instance, we make the novel observation of \textit{compositionality} in the visualization produced when the input image is exaggerated in size w.r.t. its original value, or we show how the optimization stochasticity could be exploited to reveal different facets of the representation manifold. Our following contribution is pushed by the \acrshort{label_ML} deficiency regarding spurious correlations. We deploy \acrshort{label_AM} to determine possible sources of bias and shortcut learning in pre-trained image recognition models. With motivating findings, we investigate whether this compelling and professedly simple explanation method could work on medical images. That is where we make our most important contributions. Indeed, we give empirical evidence of the difficulty in employing such methods effectively in medical imaging scenarios, showing how accuracy can rise at the expense of interpretability. We then propose novel mitigation strategies inspired by medical image regularities and discuss possible causes of their failure. All in all, this starting chapter served as a \textit{Watch out}, made us realize the urgency of more complex and capable explanation methods to be satisfactory for humans, and set the footing for what follows.

\subsubsection{Contributions in Chapter 4}
In Chapter \ref{chap:XAI_protopnet}, we investigate a more well-established explainable AI (\acrshort{label_XAI}) method for computer vision, called \acrshort{label_ProtoPNet}, which is an explainable-by-design (i.e., ante-hoc) model based on prototypical part learning. This type of reasoning for medical images was still in its infancy at the time of this study, and our first contribution was to study its applicability for clinical diagnosis rigorously. Specifically, we frame it as an automatic breast mass classification system from mammogram images to resemble radiologists' patch recognition and comparison behavior when presented with a new case. In addition to classification metrics, we were interested in assessing the ability of the model to provide end-users with plausible explanations.
From that stems the second main contribution of this chapter. We involve an experienced radiologist and ask her for clinical feedback on three aspects - (i) the quality and relevance of learned prototypes to characterize classes, (ii) the clinical significance of prototype activations, and (iii) the degree of satisfaction in the way the model combines (i) and (ii) to deliver specific explanations.

\subsubsection{Contributions in Chapter 5}
Another crucial aspect necessary for a human-aligned \acrshort{label_AI} is \textbf{causality}, that is, the study of cause-effect relationships. After investigating explainability techniques in the previous two chapters, we realize causality can closely relate to \acrshort{label_XAI}. In this regard, we bridge these two fields in Chapter \ref{chap:review_XAI_causality}. We conduct and present a systematic literature review jointly covering the two fields and investigating the role(s) of causality in the world of \acrshort{label_XAI}. Although the concepts of causation and explanation have been jointly investigated since ancient times, they have evolved in diverse ways in computer science. At the time of this study, there was no clear vision of whether there was a dependent relationship between those two fields in the realm of \acrshort{label_AI}. Our main contribution in this chapter is to answer that question in a structured way, unveiling dependencies, synergies, and limitations from theoretical and methodological viewpoints for the first time.
We first perform a high-level analysis of the final records cohort regarding keyword co-occurrence. The produced bibliographic networks provide insights into how the literature relates different research concepts.
Then, we extract information from the publications to answer our research question and identify three main perspectives: \textit{Critics to \acrshort{label_XAI} Under the Causality Lens}, \textit{\acrshort{label_XAI} for causality}, and \textit{Causality for \acrshort{label_XAI}}. Those clusters (and their sub-clusters) provide a valuable scaffold to systematically organize past (and future) literature. An updated analysis repeating the search $16$ months later and the structured collection of relevant software solutions used to automate causal tasks constitute two other of our contributions.

\subsection{Part Two: Causality}
The previous chapter marks the transition to the second logical part of the thesis (Chapters \ref{chap:mulcat_ICCV_ESWA} and \ref{chap:crocodile}), where we make multiple contributions proposing technical novelties inspired by causal concepts to boost \acrshort{label_DL} and rendering it more aligned with human reasoning and capabilities.
\subsubsection{Contributions in Chapter 6}
As we shall see in Section \ref{sec:back_causality}, the concepts of causal inference and reasoning have received increasing attention across the \acrshort{label_AI} community in recent years. Along this line of research, the processed data are often tabular, structured, simulated, and accompanied by information about the process that generated them. However, we typically do not have such a priori information in real-world image analysis applications like medical imaging.
Our main contribution in this chapter is to propose a novel method to automatically discover and exploit weak causal signals in medical images without requiring prior knowledge and use them to enhance \acrshort{label_CNN} classifiers. To enable “causality-driven” \acrshort{label_CNN}s, we operationalize the computation of feature co-occurrence into the concept of \textit{causality map}, design a new \textit{causality-factors extraction} module, and present a novel, attention-inspired scheme based on feature map enhancement. As part of our extensive empirical evaluations, we develop different architecture variants, integrate them with existing attention-based solutions, and test our methods in fully-supervised and few-shot learning. Our last contribution is to explore whether designing this causality-based reasoning could improve the explainability of these models. Results show our method improves classification and produces more robust predictions by focusing on the relevant parts of the image, thus enhancing reliability, trustworthiness, and user confidence.

\subsubsection{Contributions in Chapter 7}
In Chapter \ref{chap:crocodile}, we take a causal perspective on another fundamental problem of \acrshort{label_DL} image classifiers: domain shift bias. According to that, such systems experience significant performance degradation when applied to real-world, "in-the-wild" data, whose statistical distribution departs from the training one. In this chapter, we propose a novel framework for robust medical image classification under distribution shift. Specifically, our original contributions are as follows. We introduce the design of a dual-branched architecture that can leverage information from different datasets (i.e., domains) to learn how to disentangle causal and spurious features, operationalize a latent causal intervention, and remove the effect of confounding in medical imaging. We propose a new \textit{relational scorer} module that learns to compare pairs of latent representations from the two branches to attain cross-domain robustness. For instance, we align the causal features determining the disease and the spurious features that should not discriminate source domains. Also, we design multiple contrastive learning terms to enforce consistency/separation of representations of instances with same/different class labels.
Moreover, motivated by the hierarchical structure of radiological findings, we propose a new way to inject background (medical) knowledge into the model, effectively repurposing the \textit{causality map} tool proposed in Chapter \ref{chap:mulcat_ICCV_ESWA}. That helps the model disambiguate labels better, disregarding the confounding effect. This is a significant contribution since it enables capturing semantic priors without relying on data. The proposed framework fosters a medical image classifier's robustness, enhancing generalization to new data. By removing confounders, our bias-mitigation solution learns what to focus on/suppress by borrowing from multiple sub-disciplines. It is general and can be applied to tackle domain shift bias in many applications, fostering a safer and more human-ready \acrshort{label_AI}.

\subsection{Part Three: Biological Inspiration}
The third part of this dissertation focuses on the final area of interest we investigated to attain a more human-aligned DL: \textit{biological vision}. Indeed, by taking inspiration from how humans attain image recognition and how this visual information flows across brain regions, we investigate ways to advance human-inspired computer vision.

\subsubsection{Contributions in Chapter 8}
Specifically, in Chapter \ref{chap:cocoreco_ECCV}, we bridge the field of causality with the perspective of biological vision, and we focus on the effect of incorporating circuit motifs found in biological brains to address recognition.
The first main contribution is proposing a novel biologically motivated neural network for image classification. Our \textbf{Co}nnectivity-inspired and \textbf{Co}ntext-aware \textbf{Reco}gnition network (\acrshort{label_CoCoReco}) is a multi-branched convolutional model conceptually inspired by the mechanisms of human vision. Motivated by the connectivity of human (sub)cortical streams, we implement bottom-up and top-down modulations that mimic the extensive connections between visual and cognitive areas. We propose biologically plausible skip connections and draw inspiration from recent connectomic studies in deciding where to place them and how much to weigh their contribution.
The next main contribution is operationalizing the importance of biological visual context in machine vision. To achieve that, we advance our causal viewpoint on feature co-occurrence (Chapter \ref{chap:mulcat_ICCV_ESWA}) and propose a Contextual Attention Block (\acrshort{label_CAB}). \acrshort{label_CAB} can be added to any traditional feed-forward architecture to improve recognition by modeling feature co-occurrence (i.e., context) in the real world. We place our module at different bottlenecks to infuse a hierarchical context awareness. The findings of this preliminary study are promising, as they show performance and interpretability gains over baselines. 
\chapter{Background}
\label{chap:background}
This chapter provides the reader with the background notions on the research fields and applications this thesis investigates. We thus review and discuss the basic concepts of \acrshort{label_DL}, medical image analysis and classification, \acrshort{label_XAI}, causality, and biological vision. 

In Section \ref{sec:back_DL}, we introduce \acrshort{label_DL}, specifically focusing on multilayer perceptrons (\acrshort{label_MLP}s), \acrshort{label_CNN}s, and transformers. Also, popular loss functions of interest for the experiments in this thesis are reviewed, together with the basics of gradient-based optimization.
Section \ref{sec:back_MIA} overviews medical image analysis and describes traditional methods (e.g., feature engineering and radiomics) as well as \acrshort{label_DL}-based transfer learning with pre-trained models. Moreover, the image classification problem with \acrshort{label_DL} is rigorously defined, and valuable evaluation metrics are reported.
In Section \ref{sec:back_XAI}, we describe \acrshort{label_XAI} with post-hoc and ante-hoc explanation methods, presenting the background notions of relevant approaches, such as class activation mapping, prototypical learning, counterfactual explanations, and feature visualization.
Section \ref{sec:back_causality} informs the reader about the study of cause-effect relationships, the importance of integrating causal modeling, and how to move from association to causation from a technical and implementation viewpoint. The section concludes with hot directions of causal \acrshort{label_ML} and its benefits in medical imaging.
Finally, we take a cognitive neuroscientific step in Section \ref{sec:back_biological} and summarize key concepts of visual information processing in humans. We outline the ventral and dorsal streams, top-down modulations, and the importance of visual context in object recognition.

\section{Deep Learning}
\label{sec:back_DL}
\acrshort{label_DL} is a specialized area within \acrshort{label_AI}, particularly related to \acrshort{label_ML} and representation learning. It involves learning a hierarchy of data representations (or features) from data to address specific tasks \cite{bengio2007scaling,goodfellow2016deep}.
Drawing inspiration from nature, \acrshort{label_DL} models are typically realized as Deep Neural Networks (\acrshort{label_DNN}s). These computational models consist of numerous layers of processing units that replicate the structure and interconnections of neurons found in the mammalian brain\cite{rosenblatt1958perceptron}.
Notably, the representations acquired through \acrshort{label_DL} techniques mirror the signal structures found in our brain's neocortex, which are essential for executing intelligent behaviors. This observation implies significant parallelism between these two fields \cite{cadieu2014deep,kubilius2016deep}.

In their most general form, \acrshort{label_DNN}s are structured as a graph of non-linear parametric transformations, referred to as layers, which function as feature extractors. Beginning with raw data, each layer identifies significant patterns in its input and delivers a more abstract representation of the data to the subsequent layer.

Given an input $\mathbf{x}$, we can formally express the output $\mathbf{y}$ of a \acrshort{label_DNN} as
\begin{equation}
\label{eq:DL_DNN}
    \mathbf{y}=f(\mathbf{x},\Theta),
\end{equation}
\noindent where $f(\cdot)$ represents any composition of parametric transformations (layers) and $\Theta$ denotes the collection of all parameters, referred to as the \textit{weights} of the \acrshort{label_DNN}.

Given an input-target couples training set $\mathbf{X}=\{(\mathbf{x}_i, \mathbf{y}_i^*),i=1,\dots,N\}$, the goodness of a certain configuration of parameters is quantified by a \textit{loss function} $\mathcal{L}(X;\Theta)$ that measures how much the model predictions $\mathbf{y}$ differ from the targets $\mathbf{y}^*$ provided by $\mathbf{X}$. The specific formulation of $\mathcal{L}$ depends on the task and is further discussed in Section \ref{sec:back_DL_LossFunctions}. In the end, the learning problem boils down to the optimization problem
\begin{equation}
\label{eq:DL_optimizationProblem}
    \Theta^*=\underset{\Theta}{\arg\min} L(\mathbf{X}, \Theta),
\end{equation}
\noindent in which we search the best parameter configuration $\Theta^*$ that minimizes the loss function $\mathcal{L}(\mathbf{X};\Theta)$.

The \textit{architecture} (or computation graph) of a \acrshort{label_DNN} is defined by the specific layers used and their interconnections. We can roughly categorize \acrshort{label_DNN}s into Feed-Forward Neural Networks (\acrshort{label_FFNN}s) and Recurrent Neural Networks (\acrshort{label_RNN}s).
Contrary to \acrshort{label_FFNN}s, where information flows from inputs to outputs in a non-recursive fashion, \acrshort{label_RNN}s are stateful neural networks presenting feedback loops in their computation graphs. Therefore, the output depends not only on the input but also on the current state of the network. This type of networks is naturally able to process sequences of inputs and is a natural enabler of textual and language representation tasks. Since the aim and scope of this thesis concern image data, particularly medical images, we will not review popular \acrshort{label_RNN} architectures.
In the following sections, we will briefly review some practical and successful formulations of \acrshort{label_FFNN}s and provide the reader with the basics of gradient-based optimization of Equation \ref{eq:DL_optimizationProblem}.

\subsection{Multilayer Perceptrons and Convolutional Neural Networks}
\label{sec:back_DL_MLP_CNN}
\acrshort{label_FFNN}s are networks whose computation graph can be expressed as a directed acyclic graph (\acrshort{label_DAG}), i.e., there are no feedback loops, and information flows from inputs to outputs in a non-recursive cascade of computations. Thus, when computing the whole chain from inputs to outputs (i.e., the network's \textit{forward} pass), each transformation defined by layers is calculated only once.
The two most popular \acrshort{label_FFNN}s are the \acrshort{label_MLP} and the \acrshort{label_CNN}s, which we describe below.

\subsubsection{Multilayer Perceptron}
The \acrshort{label_MLP} is the most basic type of artificial neural network in the \acrshort{label_DL} field. It consists of a series of inner-product (or fully connected) layers, which are the fundamental components of \acrshort{label_DNN}s. Each inner-product layer performs a linear transformation of the input, followed by a typically non-linear element-wise activation function. Formally, given an input comprised of $n$ features $\mathbf{x}\in\mathbb{R}^n$, the output $\mathbf{y} \in \mathbb{R}^m$ of the layers is obtained as 
\begin{equation}
    \mathbf{y}=\varphi(\mathbf{W}\mathbf{x}+\mathbf{b}),
\end{equation}
\noindent where $\mathbf{W}\in \mathbb{R}^{n\times m}$ and $\mathbf{b} \in \mathbb{R}^m$ are learnable parameters of a linear projection to a $m$-dimensional space. Commonly used activation functions $\varphi:\mathbb{R} \rightarrow \mathbb{R}$ are the rectified linear unit (\acrshort{label_ReLU}), the sigmoid $\sigma(\cdot)$, and $tanh(\cdot)$. The dimension of the \textit{input features} is $n$ and the dimension of the \textit{output features} is $m$.

Traditionally, this layer consisted of a collection of $m$ perceptrons. The perceptron, as introduced by Rosenblatt in 1958 \cite{rosenblatt1958perceptron}, is a fundamental component of artificial neural networks designed to emulate the structure of biological neurons.
As a neuron cell, it is composed of $n$ inputs $\mathbf{x}_i (i = 1\dots n)$, usually connected to the output of other neurons, and a single output $\mathbf{y}$ (axon). Each input connection has an associated weight $\mathbf{w}_i (i = 1\dots n)$, which determines the extent to which 
the signal passing through that connection is amplified or suppressed. In a perceptron, these weights represent the strength of the connections between neuron cells. The neuron 'fires' when the sum of its weighted inputs exceeds a certain threshold. This behavior is modeled in the perceptron, where the output is defined by applying the activation function $\varphi$ to the inner product of the inputs and weights.
In \acrshort{label_MLP}s, one layer's outputs are densely connected to each input neuron of the next layer, which is why it's called a “fully-connected layer”. The output of an \acrshort{label_MLP} with $L$ layers is defined as
\begin{equation}
    y=\varphi(\mathbf{W}_L(\dots \varphi(\mathbf{W}_2(\varphi(\mathbf{W}_1\mathbf{x}+\mathbf{b}_1))+\mathbf{b}_2))+\mathbf{b}_L),
\end{equation}
\noindent where $(\mathbf{W}_i, \mathbf{b}_i)$ are the parameters of the $i$-th fully-connected layer in the network.

\subsubsection{Convolutional Neural Network}
\label{sec:back_DL_MLP_CNN_cnn}
A \acrshort{label_CNN} is an \acrshort{label_FFNN} composed of at least one \textit{convolutional} layer. This layer computes the cross-correlation\footnote{Cross-correlation, also known as the sliding inner product, is commonly used in signal processing to identify matches between two signals. Essentially, a small signal, referred to as a \textit{filter}, which contains the prototype we aim to match, is moved across a larger input signal. At each position, the inner product of the overlapping sections of the two signals evaluates the quality of the match.}
between the input and a set of learnable filters. Given the significant similarity between convolution and cross-correlation operations, this layer is frequently described as "convolutional" in deep learning literature. Therefore, we will use this terminology throughout the thesis. Due to its widespread use in image-related fields relevant to this work, we present the formulation of the two-dimensional (2D) version of cross-correlation.

Given a 2D input matrix $\mathbf{x} \in \mathbb{R}^{H\times W}$ and a 2D filter $\mathbf{w}\in\mathbb{R}^{K_1\times K_2}$, the cross-correlation $\mathbf{y}\in\mathbb{R}^{H'\times W'}$ between $\mathbf{x}$ and $\mathbf{w}$ is given by:
\begin{equation}
\label{eq:DL_crosscorrelation}
    \mathbf{y}_{u,v} = \sum_{m=1}^{K_1} \sum_{n=1}^{K_2}\mathbf{w}_{i,j}\cdot \mathbf{x}_{i+u-1,j+v-1}, 
\end{equation}
\noindent where $u=1,\dots,H'$ and $v=1,\dots,W'$. Conceptually, the filter $\mathbf{w}$ is overlaid on the input $\mathbf{x}$, and at each possible position $(u, v)$, the scalar product between the overlapping input and the filter is calculated. The output dimensionality varies based on the padding $P$ added to each side of the input and the stride $S$ of the filter application, according to the following relationships:
\begin{equation}
\label{eq:DL_H_and_W_formula}
H' = \left\lfloor \frac{H - K_1 + 2P}{S} \right\rfloor + 1,
W' = \left\lfloor \frac{W - K_2 + 2P}{S} \right\rfloor + 1
\end{equation}
Inputs and outputs of convolutional layers are referred to as \textit{feature maps} because high values in the two-dimensional map typically indicate the presence of a feature that a specific filter has learned to recognize. 

Equation \ref{eq:DL_crosscorrelation} defines the cross-correlation operation for inputs and outputs having a single feature map. Images instead are represented as 3D tensors having $C$ channels (e.g., $C$ = $3$ for RGB data like natural images, $C$ = $1$ for grayscale data like many medical images) and two spatial dimensions $H$ and $W$; therefore, the definition of 2D cross-correlation is extended to 3D tensors, allowing the filters to span the depth of the input tensor. This means that filters are still applied over the two spatial dimensions, $H$ and $W$, but each output value is influenced by all the input feature maps at a specific spatial position. A convolutional layer typically consists of a set of $K$ filters. Each filter is applied to the input, producing $K$ output feature maps stacked along the depth dimension. The obtained output is a new 3D tensor $\mathbf{y} \in \mathbb{R}^{H'\times W'\times K}$ that is commonly followed by an element-wise non-linear activation $\varphi(\cdot)$.

The primary distinction between fully connected and convolutional layers lies in their use of weights. Fully connected layers assign a unique weight to each pair of input and output features. In contrast, convolutional layers share the weights of their filters across spatial dimensions, enabling them to inherently learn translation-invariant features.

Pooling layers are frequently employed in \acrshort{label_CNN}s to decrease the volume of data processed by subsequent layers. As the name implies, this layer groups data and aggregates it using a non-parametric function, such as maximum or average. These groups are defined by fixed-size windows that slide across one or more data dimensions, similar to the cross-correlation operator. In the two-dimensional case, the input and output sizes follow Equation \ref{eq:DL_H_and_W_formula}. Convolutional layers with small strides often generate redundant local information in their output, so a max-pooling layer is typically used to reduce the resolution of intermediate feature maps.

Convolutional layers are stacked to build \textit{deep} \acrshort{label_CNN}s, which are one of the core tools of \acrshort{label_DL} for image perception and analysis \cite{krizhevsky2012imagenet,sermanet2013overfeat,simonyan2014very,he2015delving,he2016deep,xie2017aggregated}. Once trained, filters in a deep \acrshort{label_CNN} typically form a hierarchy of detectors. Layers near the input identify simple features in the data, while subsequent layers build on these to detect more complex features. A notable example of \acrshort{label_CNN}s' representational power is object recognition. The visual structure of objects in images is hierarchical: objects can be broken down into parts, parts into patches, patches into textures, textures into edges or blobs, and finally into pixels. \acrshort{label_CNN}s trained for object recognition often develop a hierarchy of detectors that mirrors this visual decomposition.
Since 2010, deep \acrshort{label_CNN}s have gained popularity as solutions for competitions on benchmark datasets that assess algorithms for large-scale object detection and image classification. The most notable example is the ImageNet Large Scale Visual Recognition Challenge (\acrshort{label_ILSVRC})\footnote{https://www.image-net.org/challenges/LSVRC/index.php}, which has been instrumental in advancing computer vision and \acrshort{label_DL} research. The ImageNet data is available for free to researchers for non-commercial use\footnote{https://www.image-net.org/download.php}. 

\subsection{Transformers and Attention}
\label{sec:back_DL_TransformerAttention}
Transformers are a type of neural network architecture introduced by Vaswani et al. in 2017\cite{vaswani2017attention}. They have revolutionized natural language processing (\acrshort{label_NLP}) and have been increasingly applied to various other domains, including computer vision and medical imaging. Unlike traditional neural networks, Transformers rely entirely on the attention mechanism to draw global dependencies between input and output.
The attention mechanism allows the model to focus on different parts of the input sequence when generating each part of the output sequence. This is particularly useful for tasks where the relationship between input and output is complex and non-linear.

In \textbf{self-attention}, each element of the input sequence attends to all other elements, allowing the model to capture dependencies regardless of their distance in the sequence. This is achieved through the computation of attention scores, which are used to weigh the importance of different elements.
On the other hand, the \textbf{scaled dot-product attention} is a specific type of attention mechanism used in Transformers. It involves three main components: \textit{queries}, \textit{keys}, and \textit{values}. The attention scores are computed as the dot product of the \textit{query} and \textit{key} vectors, scaled by the square root of the dimension of the \textit{key} vectors, and then passed through a \textit{softmax} function (ref. Eq. \ref{eq:softmax}) to obtain the attention weights. These weights are then used to compute a weighted sum of the \textit{value} vectors.

The Transformer architecture consists of an encoder and a decoder, each composed of multiple self-attention layers and \acrshort{label_FFNN}s. The encoder processes the input sequence and generates a set of attention-based representations, which are then used by the decoder to produce the output sequence.
Each \textbf{encoder} layer consists of a multi-head self-attention mechanism followed by a position-wise \acrshort{label_FFNN}. Layer normalization and residual connections are applied to stabilize and enhance the training process.
The \textbf{decoder} layers are similar to the encoder layers but include an additional multi-head attention mechanism that attends to the encoder's output.

Lastly, \textbf{Vision Transformers} (ViTs) are an adaptation of the Transformer architecture to the domain of computer vision. Introduced by Dosovitskiy et al. (2020)\cite{dosovitskiy2020image}, ViTs have demonstrated impressive performance on various image recognition tasks. Unlike \acrshort{label_CNN}s, ViTss divide an image into fixed-size patches. Each patch is then flattened and linearly embedded into a vector, similar to the tokenization process in \acrshort{label_NLP}. Since Transformers do not inherently understand the spatial relationships between patches, positional encodings are added to the patch embeddings to retain spatial information. The core of ViTs is the self-attention mechanism, which allows the model to weigh the importance of different patches when making predictions. This mechanism enables ViTs to capture long-range dependencies and contextual information. However, ViTs typically require large amounts of training data to achieve optimal performance. This can be a limitation in domains where labeled data is scarce, such as medical imaging. Also, training ViTs can be computationally intensive, requiring significant hardware resources and longer training times compared to \acrshort{label_CNN}s.

\subsection{Loss Functions}
\label{sec:back_DL_LossFunctions}
In \acrshort{label_ML} methods, loss functions are vital constituents. Their function is to quantitatively evaluate a model's effectiveness for the specific task at hand.
Given a training set $\mathbf{X}=\{(\mathbf{x}_i,\mathbf{y}_i^*),i=1,\dots,N\}$ comprised by $N$ couples of inputs and desired outputs, the quality of a particular configuration of parameters $\Theta$ is quantitatively defined by a \textit{loss function} $\mathcal{L}(\mathbf{X};\Theta)$ that measures how much predictions and targets differ. A loss function is usually defined as one or more terms summed together. The primary term $\mathcal{L}_d(\mathbf{X};\Theta)$ is defined as the average of the individual loss values computed on each sample of the dataset $\mathcal{L}(\mathbf{y}_i;\mathbf{y}_i^*)$, and we denote it the \textit{data term}. An optional secondary term, $\mathcal{L}_r(\Theta)$, often referred to as the \textit{regularization term}, is used to regularize the network and depends solely on the model parameters $\Theta$. Regularization involves adding constraints to the model to prevent undesirable properties during training, such as \textit{overfitting}—where the model performs well on known data but poorly on unseen data. The regularization term is typically multiplied by a hyperparameter $\alpha$ and then added to the loss function:
\begin{equation}
\mathcal{L}(\mathbf{X};\Theta)=\frac{1}{N}\sum_{i=1}^{N}\mathcal{L}(f(\mathbf{x}_i;\Theta),\mathbf{y}_i^*) + \alpha\mathcal{L}_r(\Theta)
\end{equation}
In the following, we briefly present some of the most used formulations of $\mathcal{L}(y_i;y_i^*)$ and $\mathcal{L}_r(\Theta)$ of interest for this thesis.

\subsubsection{Mean Squared Error Loss}
In regression problems, our goal is to make predictions that are close to one or more real-valued targets. For instance, in an age estimation task, we want our network to predict the precise age as a real number (e.g., 59.5). The mean squared error (\acrshort{label_MSE}) between predictions $\mathbf{z}$ and targets $\mathbf{z}^*$ is a commonly used loss function to evaluate the accuracy of regressions:
\begin{equation}
    \mathcal{L}_{MSE}(\mathbf{z},\mathbf{z}^*)=(\mathbf{z}-\mathbf{z}^*)^2
\end{equation}
\subsubsection{Cross-entropy Loss}
Cross-entropy (\acrshort{label_CE}) loss is widely used in \acrshort{label_ML} to quantify the difference between two categorical distributions. It is particularly common in single-label classification tasks, where a single label must be assigned to a data point from a set of $N$ possible labels ($N \geq 2$). Let $\mathbf{z}$ and $\mathbf{z}^*$ be the probability masses of two $N$-way categorical distributions, i.e., $\mathbf{z}_i,\mathbf{z}_i^* \in [0,1], \sum_{i=1}^{N} \mathbf{z}_i^*\log \mathbf{z}_i$; the CE loss between the predicted distribution $\mathbf{z}$ and the target one $\mathbf{z}^*$ is defined as
\begin{equation}
    \mathcal{L}_{CE}(\mathbf{z},\mathbf{z}^*)=-\sum_{i=1}^{N}\mathbf{z}_i^*\log \mathbf{z}_i.
\end{equation}
Classification models are often designed to output an $N$-dimensional vector $\mathbf{z}$ that is mapped to a categorical distribution via the \textit{softmax} function
\begin{equation}
\label{eq:softmax}
    \textit{softmax}(\mathbf{z})_i=\frac{e^{\mathbf{z}_i}}{\sum_{j=1}^{N} e^{\mathbf{z}_j}}.
\end{equation}

\subsubsection{Contrastive (or Siamese) Loss}
Contrastive Loss, often used in Siamese Networks, is a loss function designed to learn representations by comparing pairs of samples. It aims to minimize the distance between similar pairs and maximize the distance between dissimilar pairs. Given a binary label $\mathbf{y}^*$ indicating whether the pair is matching ($\mathbf{y}^*=1$) or mismatching ($\mathbf{y}^*=0$), a metric distance $d(\cdot)$ between the feature vectors of the two samples (e.g., squared Euclidean), and a margin parameter $m$ that defines the minimum distance for mismatching pairs, we define the contrastive loss as
\begin{equation}
    \mathcal{L}(\mathbf{x}_1,\mathbf{x}_2,\mathbf{y}^*) = (\mathbf{y}^*\cdot d(f(\mathbf{x}_1),f(\mathbf{x}_2)) + (1 - \mathbf{y}^*)\cdot\max(0, m - d(f(\mathbf{x}_1),f(\mathbf{x}_2)))
\end{equation}
This helps in learning a feature space where similar samples are close together, and dissimilar samples are far apart, which is particularly useful in many representation learning tasks. The research line of \textit{Contrastive Learning} capitalizes on such loss terms.

\subsubsection{Kullback-Leibler Divergence Loss}
In deep learning, Kullback-Leibler (\acrshort{label_KL}) divergence measures how one probability distribution diverges from a second, reference distribution. Often used in training probabilistic models, it quantifies the difference between the predicted probability distribution  $q(x)$ and the true distribution $p(x)$. The \acrshort{label_KL} divergence is defined as:
\begin{equation}
\mathcal{L}_{\text{KL}}(p \parallel q) = \sum_{x} p(x) \log \frac{p(x)}{q(x)}
\end{equation}
Minimizing the \acrshort{label_KL} divergence in a loss function encourages the predicted distribution to match the true distribution closely. This is crucial for tasks such as variational autoencoders and in cases where probabilistic reasoning is essential for robust predictions.

\subsubsection{Weight Decay: $L_1$ and $L_2$ Norms}
In \acrshort{label_DL} literature, the most common regularization terms penalize parameters with large norms. This is typically achieved by defining the regularization term $\mathcal{L}_r(\Theta)$, which is added to the loss to be minimized, as the $p$-norm of the parameters. Practical implementations include:
\begin{itemize}
    \item $p = 1$ ($L_1$ weight decay) - Also named LASSO operator, it adds a penalty equal to the absolute value of the magnitude of coefficients to the loss function, thus encouraging sparsity in the weight configuration (i.e., many parameters having a null optimal value). This leads to a sparser model that is easier to interpret (automatic feature selection) but can be affected by bias.
    \begin{equation}
        \mathcal{L}_r(\Theta)=||\Theta||_1 = \sum_i|\theta_i|
    \end{equation}
    \item $p = 2$ ($L_2$ weight decay) - Also named Ridge regression, it adds a penalty equal to the square of the magnitude, thus increasing stability and preventing overfitting. This produces a more uniform utilization of the weights, penalizing under-utilized and over-utilized ones. Using all available parameters, the model must use all the features but is less affected by bias.
    \begin{equation}
        \mathcal{L}_r(\Theta)=\frac{1}{2}||\Theta||_2^2 = \frac{1}{2}\sum_i\theta_i^2
    \end{equation}
\end{itemize}

\subsection{Gradient-based Optimisation}
\label{sec:back_DL_GradientBasedOptim}
As said already, solutions to Equation \ref{eq:DL_optimizationProblem} are the parameter configurations $\Theta$ minimizing the loss function $\mathcal{L}$ defined over a given training set $\mathbf{X}$. Due to the non-linear non-convex nature of \acrshort{label_DL} models, however, closed-form solutions to that equation exist only in sporadic cases. This explains the reluctance in the past years to work with non-convex models due to their lack of theoretical guarantees on convergence. Despite their non-convex nature, these models possess greater capacity and representational power, leading to superior overall performance. The most common method for finding suboptimal (yet valuable) solutions to Equation \ref{eq:DL_optimizationProblem} is an iterative gradient-based optimization technique known as \textbf{Gradient Descent}.

In gradient descent, we seek a new solution following the direction of the gradient of the loss function $\nabla \mathcal{L}(\mathbf{X};\Theta)$ with respect to the parameters $\Theta$. This direction represents the steepest ascent of the loss surface in the parameter space, that is, the one that maximized the loss change locally. A new parameter configuration $\Theta^*$ is chosen by \textit{descending} the loss surface along the direction of maximum steepness with a step size of $\lambda$. The resulting update rule is
\begin{equation}
\label{eq:DL_GD_update_rule}
    \Theta^*=\Theta - \lambda \nabla \mathcal{L}(\mathbf{X};\Theta)
\end{equation}
This rule is iterated until a local (or global) minimum of the loss function is reached. In the \acrshort{label_ML} field, the $\lambda$ parameter is referred to as the \textit{learning rate} (\acrshort{label_LR}) of the gradient descent optimization.
Larger \acrshort{label_LR}s result in bigger steps along the loss surface, typically enhancing the convergence rate and helping to avoid local minima, though they come with a higher risk of oscillating around minimum points. Conversely, smaller \acrshort{label_LR}s lead to a slower convergence rate but can more effectively navigate the local topology of the loss function, achieving a better local solution.

As formalized in Equation \ref{eq:DL_DNN}, \acrshort{label_DNN}s consist of non-linear functions $f_i$, each with its own set of parameters $\theta_i$. To efficiently compute gradients $\nabla \mathcal{L}(\mathbf{X};\theta_i)$ for each $\theta_i \in \Theta$, the \textbf{Back-propagation} algorithm is used. This algorithm begins by calculating the gradient of the loss function from the final layer, which produces the ultimate predictions, and then propagates the error backward through the previous layers using the chain rule for derivatives of composite functions. The loss function $\mathcal{L}$ and intermediate layers $f_i$ are designed to be differentiable, ensuring that the entire model, being a composition of differentiable functions, is also differentiable, allowing back-propagation to be applied.

So far, the presented gradient-based optimization has involved computing the gradient while considering the whole training dataset. This can restrict \acrshort{label_DL} applications that require large training sets to train models, thereby increasing the computational cost needed to compute the \textit{exact} value of $\mathcal{L}(\mathbf{X};\theta_i)$. To address this issue, \textbf{Stochastic Gradient Descent} (\acrshort{label_SGD}), particularly mini-batch \acrshort{label_SGD}, has been introduced to compute an \textit{estimate} of the loss function and its gradients at a reduced computational cost.
In mini-batch \acrshort{label_SGD}, only a small random sample of the entire training set, known as a \textit{batch} or \textit{mini-batch}, is used to estimate the loss function. This estimate is then utilized in back-propagation to approximate its gradient and update the parameters as described in Equation \ref{eq:DL_GD_update_rule}. The mini-batch size is a hyperparameter that balances the computational cost of computing the loss and the accuracy of the loss estimate. Determining the optimal value for this hyperparameter can be challenging, as it heavily depends on the specific application and available hardware resources.

Other successful proposals were introduced to improve the parameter update rule. Examples include \textbf{\acrshort{label_SGD} with Momentum}, which adds to the current direction given by the loss gradient a fraction of the direction computed in the previous iteration (i.e., inertia), and \textbf{Adaptive Moment Estimation} (\acrshort{label_Adam}). The latter computes an adaptive \acrshort{label_LR} for each parameter to be optimized based on the second moment of the gradients and estimates the first moment as in \acrshort{label_SGD} with Momentum. The resulting update rule for the \acrshort{label_Adam} optimizer for each $\theta_i \in \Theta$ is:
\begin{equation}
    \theta_i^{(k+1)} = \theta_i^{(k)} - \lambda \frac{\hat{m}_i^{(k)}}{\sqrt{\hat{v}_i^{(k)}} + \epsilon}
\end{equation}
where the first moment $\hat{m}_i^{(k)}$ is governed by a hyperparameter $\beta_1$ which typically assume value $0.9$, and the second moment $\hat{v}_i^{(k)}$ is governed by $\beta_2$, typically assuming value of $0.999$.

Another form of regularization is \textbf{Dropout}. Instead of regularizing the loss terms, as seen above, dropout consists of deactivating (i.e., drop) some neurons of an internal layer (by setting their activations to zero) during the training phase and therefore allowing the back-propagation of the error only in the active neurons. The selection of neurons to deactivate is random: each neuron subjected to dropout has a probability $p$ of remaining active. This technique typically increases the number of training iterations required for a model to converge, as only the weights of the active neurons are updated in each optimization step. However, by training a random subset of neurons at each iteration, dropout reduces neuron co-adaptation, encouraging more diverse and independent weight configurations. This enhances the capacity of larger models and helps prevent overfitting. All neurons are utilized during testing, effectively combining the predictions of multiple independently trained models. The activations of layers trained with dropout are scaled by $p$ to approximate the combined prediction of this virtual ensemble of sub-networks.

\section{Medical Image Analysis and Classification}
\label{sec:back_MIA}
\subsection{Overview of Medical Image Analysis}
\label{sec:back_MIA_Overview}
Over the last half a century, there has been increasing automation in medical data production, management, and analysis. Biomedical and health informatics \cite{miller2000opportunities,mantas2010recommendations,panayides2020ai}, data models for bio-banks \cite{roden2008development,coppola2019biobanking,berti2022data}, clinical diagnostics and remote assistance of individuals \cite{sublett1995design,berti2023exploring,jeddi2020remote}, and biomedical image indexing and retrieval \cite{brodley1999content,murala2013local,karthik2021deep} are only a few examples.

Medical image analysis is a rapidly evolving field that leverages advanced computational techniques to extract meaningful information from medical images. These images, ranging from X-rays/ultrasounds to \acrshort{label_CT}/\acrshort{label_MRI} scans or histology slides, provide invaluable insights into a patient's health and aid in diagnosis, treatment planning, and disease monitoring \cite{greenspan2016guest,litjens2017survey,shen2017deep,esteva2019guide,kang2023benchmarking}.
The primary goal of medical image analysis is to develop algorithms that can automatically identify, quantify, and characterize abnormalities or patterns within medical images. This can involve object detection, segmentation, registration, and classification tasks. By automating these processes, medical image analysis can improve diagnostic accuracy, reduce the workload of healthcare professionals, and enable more personalized and effective patient care.

While medical image analysis has advanced considerably, it has been shaped by a progression of techniques that have evolved in parallel with computational and algorithmic advancements. Traditional methods, such as manual segmentation and handcrafted feature extraction, laid the groundwork for automating medical image interpretation \cite{gillies2016radiomics}. The advent of \textit{radiomics}, which focuses on extracting many features from medical images to inform clinical decision-making, marked a significant leap in precision medicine by leveraging quantitative imaging data \cite{lambin2017radiomics}. As computational power increased and data availability grew, \acrshort{label_DL} methods have revolutionized the field, allowing end-to-end learning systems that can automatically detect and classify patterns within images with unprecedented accuracy \cite{litjens2017survey}. The following sections will explore this progression in more detail, starting with traditional approaches and leading into the transformative impact of DL on medical image classification and evaluation.

\subsection{Traditional Methods, Feature Engineering, and Radiomics}
\label{sec:back_Traditional_Engineering_Radiomics}
\subsubsection{Edge detection, segmentation, and morphological operations}
Edge detection, segmentation, and morphological operations are fundamental techniques in medical image analysis that enhance extracting relevant features from medical images. These methods are crucial in identifying anatomical structures, lesions, and other significant areas within images, facilitating accurate diagnosis and treatment planning.

\textbf{Edge detection} is a technique used to identify the boundaries of objects within an image \cite{ziou1998edge}. In medical imaging, edges often correspond to significant transitions in intensity, which can indicate the presence of anatomical structures or pathological changes. Standard edge detection algorithms include (i) the \textit{Sobel Operator}, which uses convolution with Sobel kernels to compute the gradient magnitude, highlighting regions of high spatial frequency \cite{pratt2007digital}; (ii) the \textit{Canny Edge Detector}, which involves several steps to detect edges at multiple scales \cite{canny1986computational}, and (iii) the \textit{Laplacian of Gaussian} technique, that combines Gaussian smoothing with Laplacian filtering to detect edges. It is sensitive to noise but can effectively identify fine details in medical images \cite{marr1980theory}.

\textbf{Segmentation} is partitioning an image into meaningful regions, typically corresponding to different structures or tissues. Effective segmentation is critical for subsequent analysis and classification tasks \cite{pham2000current}. Common segmentation techniques include (i) Global or adaptive thresholding, to differentiate between foreground (e.g., tumors) and background (e.g., healthy tissue) \cite{otsu1975threshold}; (ii) Region-based segmentation (e.g., region growing) to group pixels based on, for instance, intensity similarity or connectivity \cite{adams1994seeded}; (iii) Active contours (i.e., snakes), which uses energy-minimizing curves to evolve towards object boundaries and adapt to complex shapes \cite{kass1988snakes}.

\textbf{Morphological operations} are non-linear image processing techniques that process images based on their shapes. These operations are instrumental in medical imaging for refining segmentation results and enhancing image features \cite{serra1983image,soille1999morphological}. Common morphological operations include Dilation (i.e., adding pixels to the boundaries of objects) and Erosion (removing pixels from the boundaries) to close small gaps in segmented regions or to separate connected objects \cite{haralick1987image}, together with their combinations, Opening (i.e., erosion followed by dilation) and Closing (i.e., dilation followed by erosion) \cite{vincent1991watersheds}.

\subsubsection{Handcrafted features, texture analysis, shape analysis}
Feature engineering is a critical step in the medical image analysis pipeline, where raw image data is transformed into meaningful representations that can enhance the performance of classification algorithms. This process involves extracting relevant features that capture the essential characteristics of the data, enabling more accurate predictions \cite{guyon2006introduction}.

\textbf{Handcrafted features} refer to manually designed attributes derived from medical images based on expert knowledge and domain expertise. These features are tailored to capture specific characteristics relevant to the medical context \cite{kumar2012radiomics}. Common types of handcrafted features include (i) \textit{Intensity Features}, which represent pixel intensity values and their distributions, such as mean, median, variance, and histogram-based features \cite{haralick1973textural}; (ii) \textit{Gradient Features}, which capture the changes in intensity across the image, providing information about the edges and transitions between different regions \cite{tang1998texture}, and (iii) the \textit{Statistical Features} such as skewness, kurtosis, and entropy that can be computed from the intensity histogram of an image. 

\textbf{Texture analysis} focuses on quantifying the spatial arrangement of pixel intensities in an image, providing valuable information about the underlying tissue characteristics. Texture features can be extracted using various methods, including (i) \textit{Gray-Level Co-Occurrence Matrix}, which analyzes the spatial relationship between pixel pairs at various distances and orientations. Those include contrast, correlation, energy, and homogeneity, which can reveal information about tissue texture and structure \cite{haralick1973textural}; (ii) \textit{Local Binary Patterns}, that encodes local patterns by comparing each pixel with its neighboring pixels, thus capturing micro-patterns \cite{ojala1996comparative}; and (iii) \textit{Wavelet Transform}, which decomposes an image into different frequency components, allowing for multi-resolution analysis, providing insights into both local and global texture characteristics \cite{unser1995texture}.

\textbf{Shape analysis} involves the extraction and characterization of the geometric properties of objects within images. This analysis is crucial for identifying and classifying anatomical structures and lesions in medical images. Key shape features include, for instance, \textit{contour features} describing the boundary of an object (perimeter, area, and compactness) \cite{zhang2004review}, and \textit{Fourier descriptors} representing the shape of an object in the frequency domain to capture its global shape characteristics \cite{persoon1986shape}.

\subsubsection{Radiomics: seeing more within medical images}
Traditional methods like the ones reviewed above often rely on predefined rules and are limited in capturing the full complexity of medical images. They may miss subtle variations and patterns crucial for accurate diagnosis and prognosis. Therefore, kick-started by the need for more comprehensive, quantitative, and data-driven approaches in medical image analysis, the field of \textbf{Radiomics} emerged in the 2010s, also facilitated by the growing availability of big data and ML techniques \cite{lambin2012radiomics,aerts2014decoding,lambin2017radiomics,depeursinge2020standardised}. The term "radiomics" combines "radiology" with the suffix "-omics," which is commonly used in genomics and proteomics to denote comprehensive data analysis. In a nutshell, radiomics transforms medical images into high-dimensional data, extracts a wide range of quantitative features, and uses them to fit a predictive model (e.g., traditional statistical model, \acrshort{label_ML} model) to provide insights into tissue characteristics that are not readily visible to the human eye. This enhances the potential for personalized medicine and improved diagnostic accuracy.
Successful examples exist, such as automatic prognosis in head and neck cancer \cite{andrearczyk2021multi,meng2023radiomics,xu2023radiomics}, predicting radiotherapy outcomes from \acrshort{label_MRI} data \cite{zaffaroni2021po,carloni2021study,carloni2022po}, improving the diagnosis of myocardial infarction based on late gadolinium enhancement radiomics \cite{santinha2024rami}, and using radiomics models in nuclear medicine \cite{zwanenburg2019radiomics}, to name a few.

Despite their demonstrated gains over traditional statistical tools, such radiomic-based \acrshort{label_ML} models suffer from poor interpretability. The models' inherent complexity and the abstract nature of the underlying radiomic features hinder the ability to elucidate their internal workings. On top of that, people have been raising concerns about such methods' generalization and trustworthiness. For instance, studies have shown considerable variability in the models' performance depending on the partitioning of retrospective cohorts \cite{cannella2023performances}, and the brittleness of their predictive power when features are extracted with different software platforms \cite{carloni2023brain}.

\subsection{Deep Learning for Medical Imaging}
\label{sec:back_MIA_CL4MedicalImaging}
Over the last few years, \acrshort{label_AI} has prominently entered the research field in many application domains, including healthcare and biomedical imaging.
Compared to traditional methods and radiomics, \acrshort{label_DL} techniques, particularly \acrshort{label_CNN}s, have shown remarkable success in automatically learning features from raw image data without requiring extensive handcrafted feature engineering. 

In medical imaging, \acrshort{label_DL} features are utilized across various applications in addition to \textit{image classification} for diagnosing diseases (which is the focus of this thesis; ref. Sec. \ref{sec:back_MIA_ImageClassification}). Some key areas include medical image segmentation, enhancement and reconstruction, and predictive analysis.

Medical image \textbf{segmentation} involves partitioning an image into meaningful segments, such as separating different tissues or organs. Applications include brain tumor segmentation in \acrshort{label_MRI} scans \cite{khan2024brain,sivagami2024bts,vavekanand2024deep,xu2024novel}, oral ulcer segmentation \cite{jiang2024high}, retinal vessel segmentation to diagnose diabetic retinopathy using fundus images \cite{radha2024retinal}, skin lesion segmentation \cite{babu2024optimized,sonia2024segmenting}, multiple sclerosis lesion segmentation \cite{wiltgen2024lst}, and breast tissue and cancer segmentation in \acrshort{label_MRI} data \cite{ahn2023artificial,kim2023application,abo2024advances,forghani2024breast}.

Image \textbf{enhancement} and \textbf{reconstruction} in medical imaging are crucial for improving the quality and usability of images. Successful examples include noise reduction to enhance the clarity and detail of positron emission tomography (\acrshort{label_PET}) images \cite{manoj2024utilizing,xiao2024medical}, super-resolution techniques to reconstruct high-resolution images from low-resolution scans \cite{yu2023super,shin2024super}, accelerated MRI for reconstructing high-quality images from undersampled data (thus reducing scan times) \cite{xue2023cine,lyu2024state}, artifact reduction, and image synthesis.

Finally, \textbf{predictive analysis} involves using image features to predict patient outcomes or the likelihood of disease recurrence. These applications demonstrate the versatility and power of DL in transforming medical imaging, making it a critical tool in modern healthcare \cite{lee2023enhancing,ryu2023ocelot}. For instance, \cite{bang2024artificial} proposed an \acrshort{label_AI}-powered immune phenotype based on pre-treatment histology images to predict the treatment efficacy outcomes effectively; \cite{ahn2024deep} developed a \acrshort{label_DL}-based chest \acrshort{label_CT} prediction model to predict treatment response in non-small cell lung cancer patients and combined imaging data with histopathologic biomarkers; finally, \cite{moon2024artificial} developed \acrshort{label_ML}-based models of histopathology features from a whole-slide image to predict the recurrence risk, which is helpful for adjuvant treatment decisions and patient selection for novel treatments.

\subsubsection{Transfer Learning and Pre-trained Models}
Transfer learning (\acrshort{label_TL}) is a powerful technique in \acrshort{label_DL} where a model developed for a particular task is reused as the starting point for a model on a second task. This approach is particularly beneficial in medical imaging due to the often limited availability of labeled medical data. Indeed, there are several benefits of \acrshort{label_TL} in the medical domain, including data scarcity, reduced training time, and improved performance.
Medical datasets are often small and expensive to annotate\cite{kim2022did}. \acrshort{label_TL} allows models pre-trained on large datasets (like ImageNet) to be fine-tuned on smaller medical datasets, leveraging the learned features from the source domain.
Pre-trained models have already learned low-level features such as edges and textures. Fine-tuning (\acrshort{label_FT}) these models on medical images significantly reduces the training time compared to training from scratch. \acrshort{label_TL} can improve the performance of models on medical imaging tasks by providing a good initialization point, which helps in faster convergence and often leads to better generalization \cite{pan2009survey,krizhevsky2012imagenet,guo2019spottune}.

In a nutshell, \acrshort{label_TL} involves pre-training on large-scale datasets of natural images (to help the model learn general features that are useful across various tasks) and then \acrshort{label_FT} on the target medical datasets, basically adjusting the weights of the model to better fit the specific characteristics of medical images. \acrshort{label_FT} can be done by freezing early layers (keeping the weights of the initial layers fixed and only training the later layers) or by total \acrshort{label_FT}, where one allows all layers to be updated during training (generally, this leads to better performance but requires more resources).

Accordingly, many novelties in computer vision have been used and adapted for medical imaging settings for each of the above-mentioned areas. For instance, common models and architectures that are then fine-tuned are SegNet and U-Net for medical image segmentation \cite{badrinarayanan2017segnet,ronneberger2015u}, autoencoders and generative adversarial networks (GANs) for image reconstruction \cite{goodfellow2020generative}, and ResNets and DenseNets for image classification \cite{simonyan2014very,he2016deep}.
In recent years, the so-called Foundation Models (e.g., GPT-3, BERT, and CLIP) have been central to the research and industrial community. They are large-scale pre-trained models that serve as a base for various downstream tasks, and the principles of \acrshort{label_TL} and \acrshort{label_FT} are integral to developing and applying these models.

\subsection{Image Classification with Deep Learning: Problem and Evaluation}
\label{sec:back_MIA_ImageClassification}
Many real-world problems in medical image understanding can be formalized as classification problems, such as disease recognition, anomaly detection, automatic tagging, etc. Therefore, we formally define the problem of image classification in this section.

\subsubsection{Problem Setting}
Image classification entails assigning one or more labels to an image $\mathbf{x}$ from a finite set of $L$ labels. When only one label is selected from the $L$ available options, it is termed \textit{single-label L}-way classification. If $L = 2$, this is known as \textit{binary} classification.
When multiple labels can be assigned to a single piece of data, it is referred to as \textit{multi-label} classification. A \textit{multi-label L}-way classification problem is typically broken down into $L$ independent binary classification sub-problems, each aimed at predicting the presence or absence of one of the $L$ available labels.

In image classification, classifiers rely on raw pixel data, typically organized in a tensor with the shape \textit{(H, W, C)}, where \textit{H} represents the height, \textit{W} the width, and \textit{C} the number of channels of the image.
Let $I$ be the image space of all valid raw pixel values; given an image to be classified $\mathbf{x} \in I$, a single-label L-way classifier is a function $f$ mapping $\mathbf{x}$ into the label space. Classification is commonly \textit{probabilistic}, meaning that \textit{soft} assignments to the available labels are considered rather than \textit{hard} ones. In this case, an image $\mathbf{x}$ is mapped into its posterior probability $p_i = p(\mathbf{y}=i|\mathbf{x})$ of belonging to category $i$, for each category $i \in \{1,\dots,L\}$.
Thus, the classifier is a function $f:I\rightarrow \mathbf{p}$ which maps an image to the parameters $\mathbf{p}=(p_1,\dots,p_L),\sum_{i=1}^{L}p_i=1$ of a categorical distribution defined over the label space. One of the key advantages of soft assignments over hard assignments is their differentiability, which enables and facilitates gradient-based learning for models of $f$.

\subsubsection{Evaluation Metrics}
In the following, we report some commonly used evaluation metrics to measure the quality of classification models relevant to this thesis's scope and application use cases.
Assume $\mathbf{Y} = (y_1,\dots,y_n)$ is a set of predictions and $\tilde{\mathbf{Y}} = (\tilde{y}_1,\dots,\tilde{y}_n)$ the set of the ground-truth labels for a single-label L-way classification problem, with $y_i,\tilde{y}_i\in \{1,\dots,L\}$. To evaluate a set of label assignments, the \textbf{Confusion Matrix} is typically used, since it compactly reports the co-occurrence of predicted and ground-truth labels. A confusion matrix comprises four quadrants/elements, and the element $c_{ij}$ indicates the number of times the model predicted the class $i$ for a sample belonging to class $j$. For the binary case ($L=2$), many useful metrics can be computed from the confusion matrix. In this regard, the following terminology holds: \textit{True Positives} (\acrshort{label_TP}) is the number of correct hits, \textit{True Negatives} (\acrshort{label_TN}) is the number of correct rejections, \textit{False Positives} (\acrshort{label_FP}) is the number of false alarms, and \textit{False Negatives} (\acrshort{label_FN}) is the number of misses. Together, \acrshort{label_TP} and \acrshort{label_FN} are the positive samples (P), while TN and FP are the negative samples (N).

The classification \textbf{Accuracy} is defined for the binary case as
\begin{equation}
    \text{Accuracy}=\frac{\text{TP+TN}}{\text{P+N}}=\frac{\text{TP+TN}}{\text{TP+FN+TN+FP}}
\end{equation} representing the fraction of correctly classified samples. With more than two classes, the accuracy is given by the sum of the elements on the diagonal of the confusion matrix divided by the total number of classified samples:
\begin{equation}
    \text{Accuracy}=\frac{\sum_{i=1}^{L}c_{ij}}{|Y|}
\end{equation}
The accuracy metric is straightforward and intuitive for evaluating classifiers, but it may be misleading when employed on unbalanced test sets\footnote{
Consider a binary classification example with 95 positive and 5 negative samples; A classifier always predicting the positive class (a trivial acceptor) achieves an accuracy of 0.95 while behaving very poorly on negative classes.}. In this scenario, other ratios derived from the classification matrix are used to characterize a classifier's performance better. For instance, the \textbf{True Positive Rate} (\acrshort{label_TPR}), also known as \textit{Recall} or \textit{Sensitivity}, is the ratio between correctly predicted positive samples and the number of actual positive samples; it thus indicates the fraction of positive samples correctly accepted as positive. Similarly, the \textbf{True Negative Rate} (TNR), or \textit{Specificity}, is the fraction of negative samples correctly accepted as negative.
Complementary to \acrshort{label_TNR}, the \textbf{False Positive Rate} (\acrshort{label_FPR}) indicates the fraction of negative samples incorrectly classified as positive (i.e., false alarms), while the \textbf{False Negative Rate} (\acrshort{label_FNR}) is complementary to TPR and indicate the fraction of positive samples incorrectly classified as negative (i.e., miss rate).

The \textbf{Receiver Operating Characteristic} (\acrshort{label_ROC}) curve is a graphical representation that illustrates the diagnostic ability of a binary classifier as its discrimination threshold is varied. It plots the \acrshort{label_TPR} against (y-axis) the \acrshort{label_FPR} (x-axis) at various threshold settings. Curves lying closer to the optimal (0; 1) point are considered better classifiers independently of the particular threshold chosen.
To quantitatively compare classifiers on the \acrshort{label_ROC} plane, the \textbf{Area Under the Curve} (\acrshort{label_AUC}) is often computed as a unique threshold-independent metric; the higher the \acrshort{label_ROC} curve, the closer the \acrshort{label_AUC} is to 1. An \acrshort{label_AUC} score of 1 indicates perfect classification, while a score of 0.5 suggests no discriminative power, equivalent to random guessing. The \acrshort{label_AUC} of the \acrshort{label_ROC} curve is sometimes acronymized as \acrshort{label_AUROC}.

\section{Explainable Artificial Intelligence}
\label{sec:back_XAI}
Today's world of information research is largely dominated by \acrshort{label_AI} and \acrshort{label_DL} technologies, that are being deployed transversely across many sectors, where they can be an added value to humans. However, concerns have been raised about the transparency of their decisions, especially in the image domain. In this regard, \acrshort{label_XAI} has recently gained popularity.

\subsection{The Need for Explanations}
\label{sec:back_XAI_need4expla}
Although \acrshort{label_DL} models usually outperform humans at many levels, performance is not all we need. In the era of complex \acrshort{label_ML}/\acrshort{label_DL} models, industrial and research communities demand more explainability and trustworthiness. Understanding and trusting \acrshort{label_AI} decisions has become imperative, and this is further enforced, for instance, by guidelines for trustworthy AI by the
European Commission\footnote{\url{https://digital-strategy.ec.europa.eu/en/library/ethics-guidelines-trustworthy-ai}}.
These needs emerge from the user's difficulty in understanding the internal mechanisms of an intelligent agent that led to a decision. Based on this degree of understanding, the user often decides whether to trust the output of a model.
In this scenario, \acrshort{label_XAI} plays a pivotal role. \acrshort{label_XAI} aims to provide humans with explanations to understand the reasoning behind an \acrshort{label_AI} system and its decision-making process. In other words, the goal of \acrshort{label_XAI} is to enable end-users to understand the underlying explanatory factors of why an \acrshort{label_AI} decision is taken. 
The term \acrshort{label_XAI} was first introduced in \cite{van2004explainable}, but its popularity has spread across the literature only after the DARPA's \acrshort{label_XAI} program \cite{gunning2019darpa}, reaching a certain degree of maturity to date \cite{guidotti2018survey,du2019techniques,carvalho2019machine,rudin2019stop,arrieta2020explainable,molnar2020interpretable_book,baniecki2024adversarial,hassija2024interpreting,mersha2024explainable}. 

In healthcare, \acrshort{label_XAI} has been utilized for medical image analysis, acute critical illness prediction, intraoperative decision support systems, drug discovery, and treatment recommendations \cite{van2022explainable,lauritsen2020explainable,gordon2019explainable,jimenez2020drug,carloni2022applicability,berti2024explainable,sadeghi2024review,salih2024review}. 
Regarding finance, popular applications of \acrshort{label_XAI} are credit risk management and prediction, loan underwriting automation, and investment advice \cite{bussmann2021explainable,moscato2021benchmark,sachan2020explainable,yang2021multiple,vcernevivciene2024explainable,weber2024applications}.
In education, \acrshort{label_XAI} has been applied in automatic essay scoring systems, educational data mining, and adaptive learning systems \cite{kumar2020explainable,alonso2019explainable,khosravi2022explainable,gunasekara2024explainability,krishnan2024impact}, while digital forensics for law enforcement context represents an example in the legal domain \cite{hall2022explainable,richmond2024explainable}.

Regardless of the application field, \acrshort{label_XAI} is driven by the idea of making the reasoning process of AI transparent and, therefore, AI models more intelligible to humans. Accordingly, when it comes to explaining the logic of an inferential system or a learning algorithm, four pillar motivations for deploying \acrshort{label_XAI} were established \cite{adadi2018peeking}:
\begin{itemize}
    \item explain to \textit{justify} (i.e., provide justifications for particular decisions to make sure they are not unfairly yielded by bias)
    \item explain to \textit{control} (i.e., understand the system behavior for debugging vulnerabilities and potential flaws)
    \item explain to \textit{improve} (i.e, understand the system behavior for enhancing its accuracy and efficiency)
    \item explain to \textit{discover} (i.e., learn from machines their knowledge on relationships and patterns).
\end{itemize}

\subsection{Post-hoc and Ante-hoc Explanations}
\label{sec:back_XAI_postHoc_anteHoc}
\begin{figure}
\centering
 \includegraphics[width=0.99\textwidth]{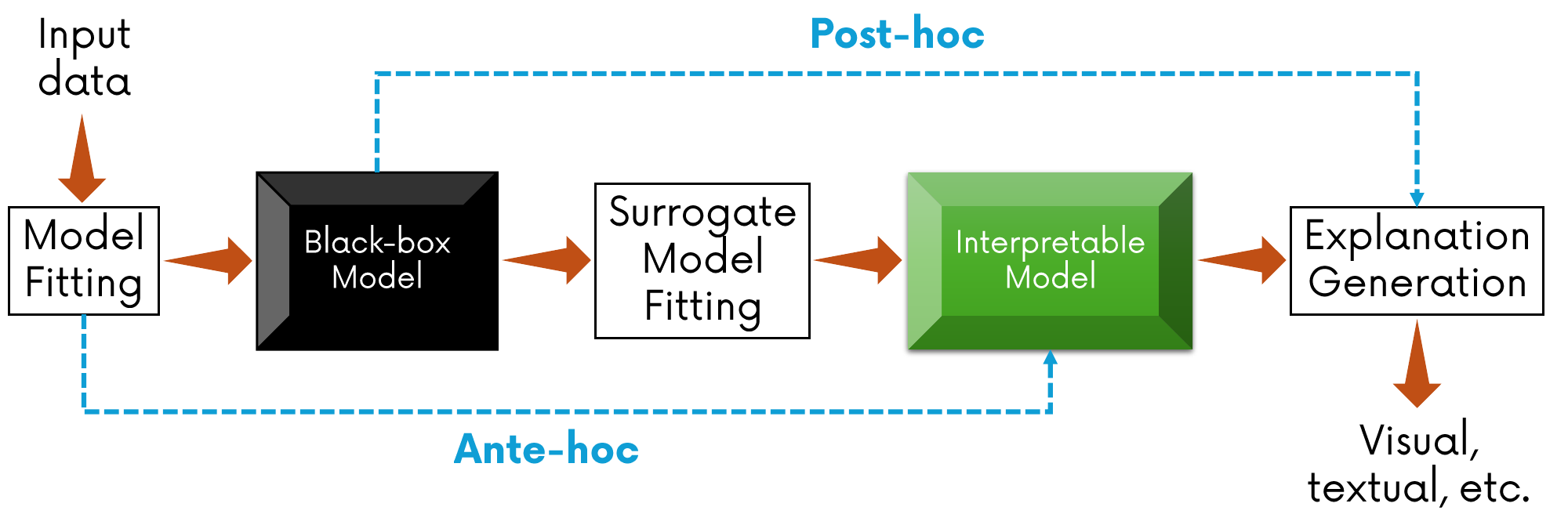}
\caption{\textbf{Post-hoc and Ante-hoc XAI}. In post-hoc methods, the outcome of a black-box model is passed through an additional, surrogate, interpretable model that generates an explanation. Ante-hoc methods avoid using a black-box model in the first place and instead build an inherently transparent model. So, together with an outcome, it already provides an explanation - the reasoning process is transparent.}
\label{fig:back_xai_postante}
\end{figure}

Explanation methods developed so far can be broadly divided into two classes: \emph{post-hoc} explanations and \emph{ante-hoc} explanations (see Figure \ref{fig:back_xai_postante}).
The first class comprises solutions based on separate models that are supposed to replicate most of the behavior of the black-box model. Indeed, post-hoc explanations are generated after the model has made a decision. They aim to provide insights into the model's behavior without altering its architecture. Their major advantage is that they can be applied to an already existing and well-performing model. However, in approximating the outcome, they may not reproduce the exact calculations of the original model. Common post-hoc techniques include global/local approximations \cite{ribeiro2016should,greenwell2017pdp}, saliency maps \cite{simonyan2013deep,bach2015pixel,sundararajan2017axiomatic}, and derivative-based methods for scoring feature importance\cite{lundberg2017unified}.

By contrast, the second class of explanation methods, also known as \emph{explaining by design}, comprises inherently interpretable models that provide their explanations in the same way the model computes its decisions. This does not mean that ante-hoc models are always simple, such as decision trees, linear models, and rule-based systems. Indeed, training, inference, and explanation of the outcome are intrinsically linked, potentially in a complex network setting. Such methods include Deep k-Nearest Neighbors~\cite{papernot2018deep}, Logic Explained Networks~\cite{ciravegna2021logic}, and prototype learning \cite{rudin2022interpretable,chen2019looks,li2018deep}.

\subsection{Popular Methods in XAI}
\label{sec:back_XAI_methods}
In this subsection, we touch upon some popular methods in \acrshort{label_XAI} that may be useful to the reader in understanding the contents of this thesis better. 

\subsubsection{Class Activation Mapping}
\label{sec:back_XAI_methods_CAM}
\begin{figure}
\centering
 \includegraphics[width=0.99\textwidth]{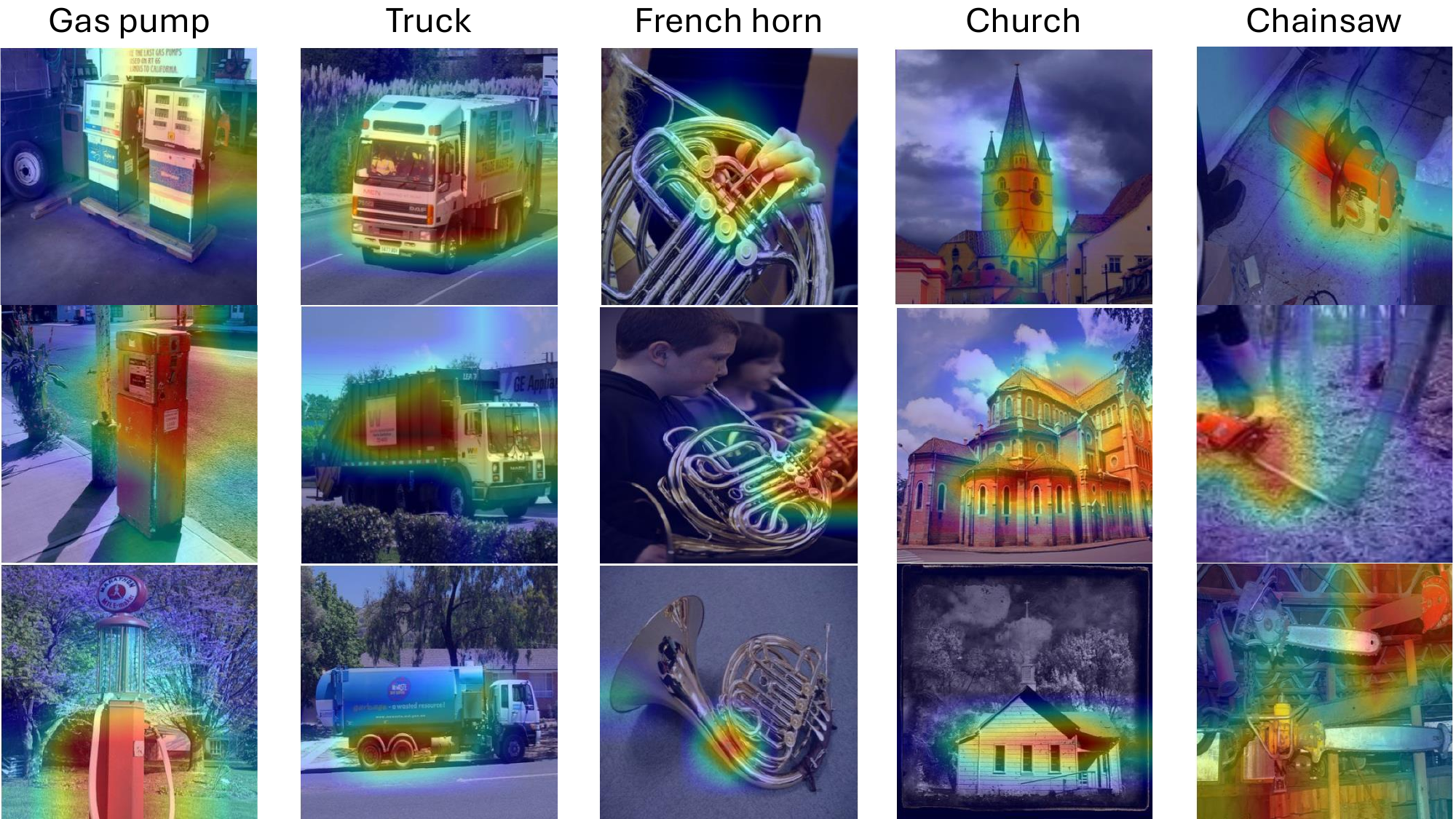}
\caption{\textbf{Class Activation Maps} for five popular categories of ImageNet. For each class, three examples are shown. The Grad-CAM maps are superimposed on the original image as heatmaps: higher activation values are represented in dark red; the lower the value, the bluish the pixel.}
\label{fig:back_gradCAM}
\end{figure}

Class Activation Mapping (\acrshort{label_CAM}) techniques, such as Grad-CAM \cite{selvaraju2017grad,selvaraju2020grad}, are post-hoc \acrshort{label_XAI} methods used for visual explanations in \acrshort{label_CNN}s. Grad-CAM generates heatmaps highlighting the regions of an input image most relevant to the model's prediction. This helps understand which parts of the image contribute to a particular class decision, making it easier to interpret and trust the model's output (see Figure \ref{fig:back_gradCAM}).

Grad-CAM decodes the importance of each feature map for a specific class by computing and analyzing gradients of the predicted class scores flowing into the final convolutional layer of the \acrshort{label_CNN}. By focusing on how these gradients impact class predictions, Grad-CAM is class-discriminative - it provides visualizations specific to particular predicted classes, enhancing interpretability.

Other works have revealed different disadvantages and limitations of such a method \cite{chattopadhay2018grad,viering2019manipulate}. Sometimes, Grad-CAM can struggle with precise object localization, where only certain portions of the object of interest are deemed important, missing the others (e.g., partial identification of the truck and French horn in the last row of Figure \ref{fig:back_gradCAM}), or with multiple occurrences of the same class, where it may not effectively handle images with multiple instances of the category of interest class (e.g., several chainsaws in the last row of Figure \ref{fig:back_gradCAM}).
Nevertheless, \acrshort{label_CAM} techniques, specifically those based on Grad-CAM, are valuable tools to interpret \acrshort{label_DNN}s' decisions and aid model debugging and improvement, transfer learning and fine-tuning, visual question answering, and image captioning. We often deployed Grad-CAM in our investigations regarding medical imaging as reported in Chapters \ref{chap:mulcat_ICCV_ESWA} and \ref{chap:cocoreco_ECCV}.

\subsubsection{Prototypical Learning}
\label{sec:back_XAI_methods_prototypes}
When it comes to image classification, one of the most familiar approaches humans exploit is to analyze the image and, by similarity, identify the previously seen instances of a particular class. The \textit{prototypical learning} is a research line focusing on building models that mimic this type of reasoning, leading to ante-hoc \acrshort{label_XAI} methods for image data.
Traditional prototypical learning methods emphasize the creation of a metric space where classification is performed by computing distances to these prototypes \cite{rudin2022interpretable,li2018deep}. This is particularly effective in few-shot learning scenarios, where the goal is to recognize new categories using very few labeled samples.

\begin{figure}
\centering
 \includegraphics[width=0.99\textwidth]{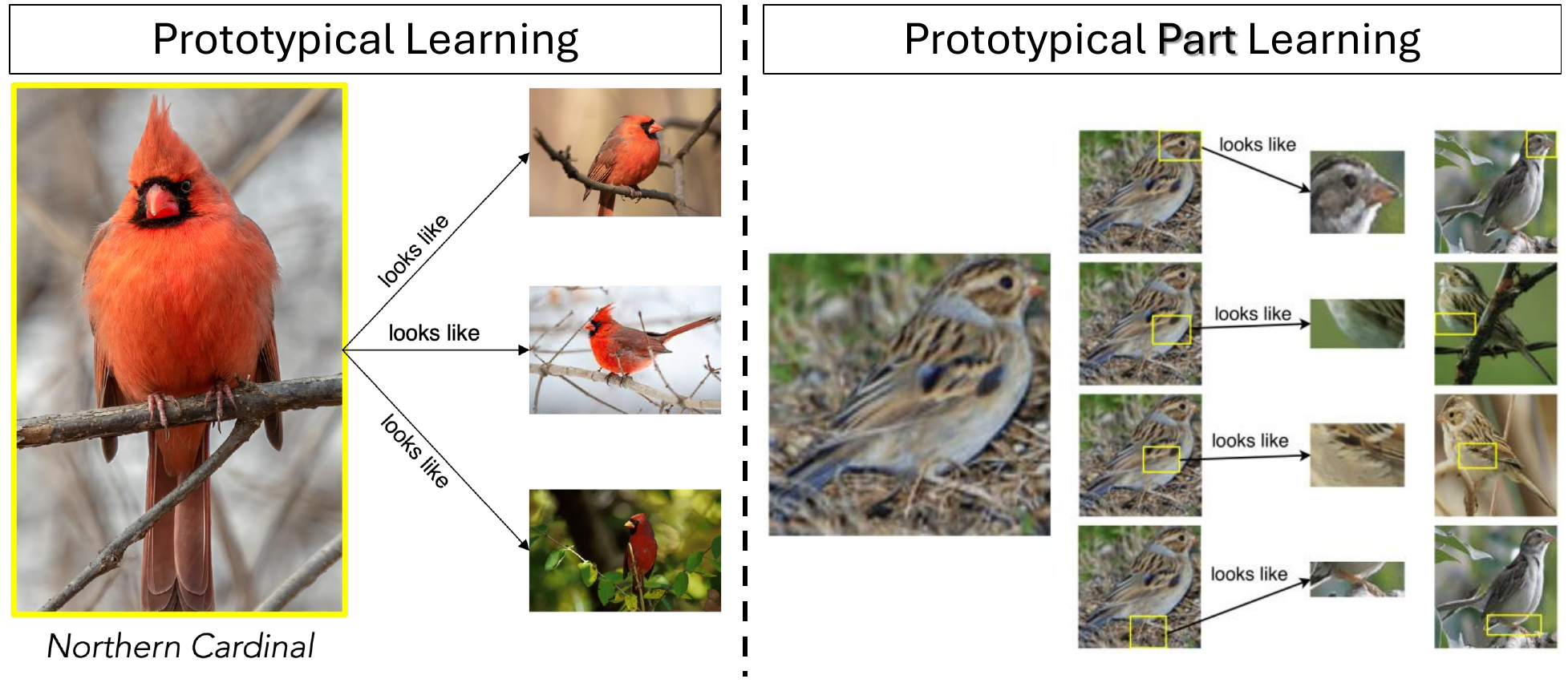}
\caption{\textbf{Difference between Prototypical Learning and Prototypical Part Learning}. While the former (left panel) compares one whole image to another whole image, the latter (right panel) seeks the relevant parts of the input image that led to a specific class prediction. In other words, parts of observations could be compared to parts of other observations. Adapted from \cite{chen2019looks}.}
\label{fig:back_prototypicallearning}
\end{figure}

The prototypical \textit{part} learning, introduced in \cite{chen2019looks}, extends this concept by focusing on learning prototypical parts of the input data. Indeed, the prototypical part network (\acrshort{label_ProtoPNet}) breaks down an image into prototypes and uses evidence gathered from the prototypes to qualify the image. Thus, the model's reasoning is qualitatively similar to that of ornithologists, physicians, and others on the image classification task (see Figure \ref{fig:back_prototypicallearning}).
\acrshort{label_ProtoPNet} dissects images into parts, learns a set of prototypes for each class, and uses them to make classifications. Indeed, at training time, the network uses only image-level labels without fine-annotated images. Then, at inference time, the network predicts the image class by comparing its patches with the learned prototypes.
Since the release of \acrshort{label_ProtoPNet}, an array of extensions of the original method have been
developed. ProtoPool~\cite{rymarczyk2022interpretable} and ProtoPShare \cite{rymarczyk2020protopshare} introduce a pooling layer that aggregates features from multiple prototypes to capture the variability in data better. ProtoTree~\cite{nauta2021neural} combines prototype learning with decision trees, while ProtoPNeXt~\cite{willard2024looks} modifies the training process.
Those models explain visually by indicating the most informative parts of the image w.r.t. the output class. By showing which prototypes influenced the prediction, this approach allows the user to qualitatively evaluate how reasonable and trustworthy the prediction is according to the user's domain knowledge.
We investigate the effectiveness of this prototypical part learning when applied to medical imaging in Chapter \ref{chap:XAI_protopnet}.

\subsubsection{Counterfactual Explanations}
\label{sec:back_XAI_methods_CFE}
Counterfactual explanations (\acrshort{label_CFE}s) were first introduced as an interpretation method by \cite{wachter2017counterfactual}.
The idea behind this post-hoc, local, model-agnostic method is to explain a prediction by examining which features would need to be changed to achieve a desired prediction. Therefore, it is an example-based method, much like the prototypical learning. However, unlike prototypes, counterfactuals do not have to be actual instances from the training data but can be a new combination of feature values. Also, they are mainly used for tabular data, with few recent stretches to high-dimensional data. 

\acrshort{label_CFE}s provide insights by identifying the minimal changes needed to an input to alter the model's prediction. For example, in a loan approval scenario, a \acrshort{label_CFE} might indicate that increasing the applicant's income by a certain amount would change the decision from rejection to approval. Besides, there are usually multiple different \acrshort{label_CFE}s; each counterfactual tells a different “story” of how a particular outcome was reached.

Over the years, \acrshort{label_CFE}s have demonstrated value for understanding decision boundaries and providing helpful individual insights to users \cite{mothilal2020explaining}. More recent research moves beyond local \acrshort{label_CFE}s embracing global, systemic explanations \cite{wielopolski2024unifying}, and studies \acrshort{label_CFE}s' probabilistic plausibility and guarantees \cite{stkepka2024counterfactual}.

\subsubsection{Feature Visualization}
\label{sec:back_XAI_methods_FeatureViz}
What if we could visualize the image objects that specific parts of a machine vision system are “looking for”? In other words, is it possible to mitigate ML opacity via a post-hoc method that visualizes what a DNN has learned?

Making the learned features explicit is actually a method, and it is called \textit{Feature Visualization} \cite{erhan2009visualizing}. It has been revealed to be an exciting technique to visually represent the data patterns and characteristics that the network has identified during training. Indeed, feature visualization, and specifically the optimization entailed by \textbf{activation maximization (\acrshort{label_AM})}, can be used to visualize what a neural network has learned by generating images that maximize the activation of specific neurons or layers.\cite{mordvintsev2015inceptionism,olah2017feature,olah2018building}.
Basically, feature visualization is an optimization problem. We exploit the differentiable nature of \acrshort{label_DNN}s w.r.t their input and want to find the optimal input image that would cause the firing of a specific neuron (i.e., its activation) and use derivatives to iteratively manipulate our input to attain the goal \cite{erhan2009visualizing,olah2017feature,nguyen2019understanding}.

This optimization procedure turns the way \acrshort{label_DNN}s are trained. Instead of iteratively altering the model's weights upon incoming \textit{fixed} images, we fix weights of the trained model and iteratively \textit{alter} the input image to maximally activate the neuron of interest.
\begin{figure}
\centering
 \includegraphics[width=0.98\textwidth]{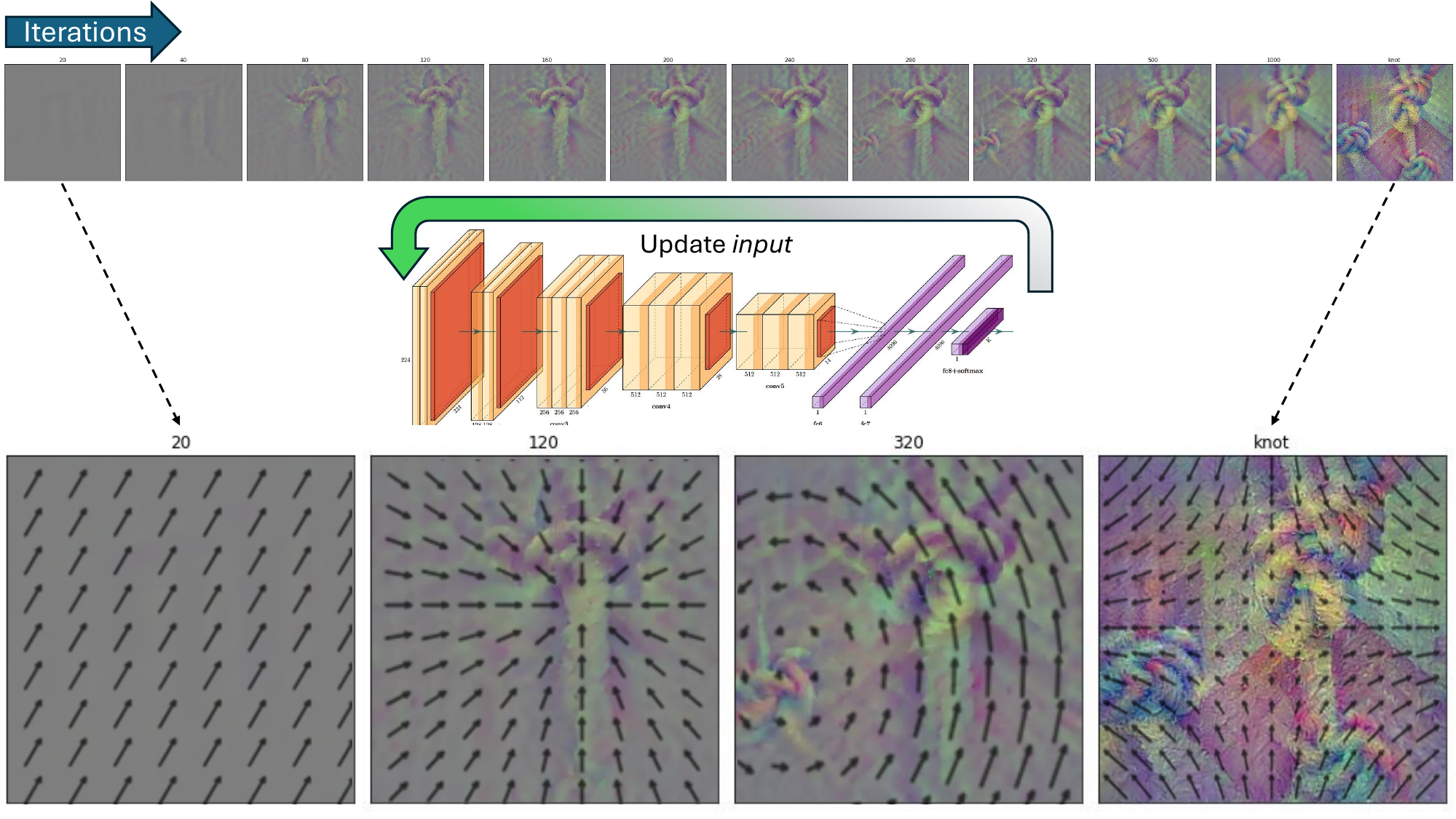}
\caption{\textbf{Feature Visualization}. The idea behind Activation Maximization (\acrshort{label_AM}) is to leverage the differentiable nature of neural networks w.r.t. their inputs. The goal is to compute the input image that maximally activates a specific neuron (e.g., the \textit{'knot'} class) in a trained model. Starting from pure noise, a gradient \textit{ascent} is performed via an iterative optimization to manipulate the raw pixels of the input. At the top of the figure, the \acrshort{label_AM}s at subsequent iterations are shown. At the bottom, vector fields are superimposed to four exemplar \acrshort{label_AM}s to represent the pixels' alteration.}
\label{fig:back_XAI_vectorfield_AM}
\end{figure}
As Figure \ref{fig:back_XAI_vectorfield_AM} shows, to visualize what a neuron in a pre-trained deep \acrshort{label_CNN} has learned, the following steps are involved.
After selecting a pre-trained model and freezing all its parameters, we need to select a neuron of interest --- be it a specific class activating neuron in the final fully connected layer (e.g., the \textit{'knot'} neuron) or a group of neurons at intermediate layers.
Second, we initialize a random noise sample which acts as a starting image of the desired shape (e.g. $224\times224\times3$).
We then pass this input image through the network's layers up to the hidden layer containing the neuron of interest.
Differently from the training/prediction stages, for visualization, we are interested in the "activation" rather than in a "prediction". Therefore, the individual response of a single neuron to a specific input image when it reaches the neuron’s layer is now interpreted as the loss function, thus driving the learning process.

By setting the optimizer (e.g., \acrshort{label_SGD}) on the raw input pixels rather than on the model's parameters, this new loss flows back through the network beyond the input layer and is used to alter the values of the input, which is then used again as the input image during the next iteration, and so on. After completion of the process --- either by reaching a predefined number of iterations or a gradient value ---  an image that highly activates one specific neuron is obtained. We deem this image a \textbf{prototype} for that specific class.

\section{Causality}
\label{sec:back_causality}
\subsection{\textit{What?} The Study of Cause-Effect Relationships}
\label{sec:back_causality_what}
Causality is a fundamental notion in science and engineering. It can be defined as the study of how things influence each other, i.e., how causes lead to effects, and how we can intervene, manipulate, or imagine different scenarios based on causal knowledge. Causality is important for understanding the world, making decisions, and designing policies.

Even though the wide literature on causality spans different interpretations, such as the Wiener-Granger causality \cite{granger1969investigating}, causal potential theory \cite{xu2018machine}, Lorentzian causality \cite{minguzzi2019lorentzian}, and quantum causality \cite{goswami2020experiments}, the one by computer scientist Judea Pearl is popularly associated with \acrshort{label_AI}. Pearl identifies some major obstacles still undermining the ability of \acrshort{label_AI} systems in reasoning in a way akin to humans, to be overcome by equipping machines with causal modeling tools \cite{pearl2019seven}. Among those obstacles, is the lack of robustness of AI systems in recognizing or responding to new situations without being specifically programmed (i.e., adaptability), as well as their inability to grasp cause-effect relationships. Instead, those abilities are innate features of human beings, who can communicate with, learn from, and instruct each other since all their brains reason in terms of cause-effect relationships \cite{pearl2018theoretical}.

Pearl was the first to develop graphical causal inference, which came from research on \acrshort{label_AI} and was not related to \acrshort{label_ML} for a long time.
Then, in recent years, some scholars have reasoned about how the fields of causality and \acrshort{label_ML}/\acrshort{label_DL} have been and should be connected \cite{scholkopf2021toward,berrevoets2024causal,feuerriegel2024causal,zhang2024causal}, claiming that the difficult unsolved problems of \acrshort{label_ML} and \acrshort{label_AI} have a lot to do with causality \cite{scholkopf2022causality,jones2024causal}.

According to Pearl's famous \textit{Ladder of Causation} \cite{pearl2018book}, humans organize their knowledge of the world according to three distinct levels (i.e., the ladder's \textit{rungs}) of cognitive ability (see Table \ref{tab:ladder_of_causation}): \textit{association}, \textit{intervention}, and \textit{counterfactual}.
\textbf{Association} is the ability to passively observe and learn from data, and to find correlations or patterns. Reasoning on this level could not distinguish the cause from the effect and it is where conventional \acrshort{label_AI} approaches to classification or regression stand today. 
\textbf{Intervention} is the ability to manipulate and change the data-generating process and to estimate causal effects. Thus, this second rung involves not just viewing what exists, but also changing it. However, reasoning on this rung cannot reveal what will happen in an imaginary world where some observed facts are bluntly negated. To this end, we need to climb to the third rung of the ladder, i.e., \textbf{Counterfactuals}. It is the ability to imagine and explain alternative scenarios, and to attribute causes to effects. This level involves imagination since to answer counterfactual queries one needs to go back in time and change history. For instance, we may wonder whether it was, indeed, \textit{turning the heating system on} that caused a \textit{warm apartment} or, rather, for instance, the outdoor weather.
\begin{table}
\caption{The Ladder of Causation by Pearl \cite{pearl2018book}.}
\label{tab:ladder_of_causation}
\begin{tabularx}{\textwidth}{XXXX}
\textbf{Level (rung)} & \textbf{Cognitive ability (activity)} & \textbf{Typical question} & \textbf{Example}\\
\hline
Association & Seeing, observing (i.e., recognizing recurrent patterns in an environment) & “What if I see \dots?” & "What is the probability that an apartment is warm if I see the heating system being on?"\\
\hline
Intervention & Doing (i.e., predicting the effect(s) of multiple intentional actions on the environment and choosing the best to produce a desired outcome) & "What if I do \dots?" & "What is the probability that the apartment will get warm if I turn on the heating system?".\\
\hline
Counterfactuals & Imagining, reasoning in retrospection, and understanding & "What if I had done \dots?" & "What would have happened to the indoor comfort of the apartment if I had kept the heating system off?".\\
\hline
\end{tabularx}
\end{table}
Pearl's \textit{ladder} shows that each level requires more causal knowledge and assumptions than the previous one, but enables more causal questions and answers than the previous one.

In general, building models that represent causal relationships among variables from observations may be challenging without relying on assumptions that are hard to verify in practice, such as the absence of unmeasured confounding between the variables \cite{robins1999impossibility, greenland2015limitations}. 
Nevertheless, Pearl's work was groundbreaking in that it transformed causality from a notion clouded in mystery into a concept with logical foundations and defined semantics. The formalization of causality in mathematical terms within an axiomatic framework allowed the development of automatic computational systems for causal modeling. In this regard, we give some notations and terminology in Section \ref{sec:back_causality_how}.

\subsection{\textit{Why?} Current ML issue with non-I.I.D. Data: Predictive vs. Causal Models}
\label{sec:back_causality_why}
Over the last few decades, \acrshort{label_ML} models (particularly \acrshort{label_DL}) have demonstrated their ability to be powerful \textit{predictive} devices. Traditionally, they have been focused on pattern recognition, operating under a correlation-based framework. By using ideas such as minimizing the error between predictions and reality (i.e., \textit{empirical risk minimization}), updating the weights of the network based on the error (i.e., \textit{back-propagation}), and choosing the proper structure for the network (i.e., \textit{inductive bias} \cite{ulyanov2018deep}), \acrshort{label_ML} has made significant progress in solving problems in image recognition, language understanding and graph analysis \cite{kaddour2022causal}. Nevertheless, \acrshort{label_ML} lacks the ability to understand or infer \textit{causal} relationships, which are essential for generalization, reliable decision-making, and robust predictions\cite{scholkopf2022causality}. 

What the community tends to forget sometimes is that although \acrshort{label_ML} is much more flexible than traditional statistics in the sense that we need much fewer assumptions to work, ML is not assumption-free. \acrshort{label_ML} will lead to effective predictions only if we meet certain assumptions, and the main assumption is that the data are \textbf{independent and identically distributed (\acrshort{label_iid})} - i.e., the underlying joint distribution that we use to predict some variable of interest is static, it does not change over time (it is a multivariate static distribution). This is why \acrshort{label_ML}/\acrshort{label_DL} models excel in \textit{predictive} tasks on in-distribution (\acrshort{label_ID}) data.

However, we do not care about predictions; we care about \textit{generalizable} predictions. In this sense, a purely \textbf{predictive model} would give the most accurate predictions given all the information available (over all the available data), but would fail to generalize well on data where the learned input-output correlations are no longer present, i.e., out-of-distribution (\acrshort{label_OOD}) data \cite{robey2021model,usynin2023sok,li2023whac}. For instance, performance greatly drops when the data changes, \acrshort{label_LLM}s have difficulty inferring causal relations \cite{joshi2022all,joshi2024llms}, and predictions are unfair and discriminate against some groups of people \cite{wadsworth2018achieving}.

In contrast, a \textbf{causal model} aims to be accurate when predicting the outcome in response to changes in the treatment or exposure\cite{peters2016causal,scholkopf2022causality,jiao2024causal}. The causal model should give better answers to counterfactual, or hypothetical conditions that may be \acrshort{label_OOD} or simply rare in the data. Therefore, if one cares about being able to predict the effect of interventions or hypotheticals, a causal model is needed. If one only wants to predict outcomes similar to the current data, an ordinary predictive model is sufficient.

The research field of \textit{domain generalization} (\acrshort{label_DG}) aims to train models that can perform well on datasets from unseen environments (\acrshort{label_OOD} data). This is typically done by identifying features that are invariant across different domains \cite{nam2021reducing,bi2023mi,nguyen2023causal,zhou2023learning,bai2024hypo,guo2024investigating}, embedding prompts into vision transformers \cite{yan2023epvt}, or merging individual-client and cross-client learning in a federated setting \cite{chen2023federated}. Improving \acrshort{label_DG} also often intertwines with feature disentanglement/decoupling, for instance in transformers \cite{kim2023dimix}, generative models \cite{gowal2020achieving}, autoencoders \cite{li2021generalized,an2023causally,zhao2023disentangling}, and \acrshort{label_CNN}s \cite{meng2020mutual,liu2022decoupled}.

However, when moving from experimental test beds to real-world settings, the structures and distributions cannot be deemed as static and can naturally change. This is much true in rapidly evolving fields such as medicine and marketing. In this sense, causal inference can be helpful. Conversely, \acrshort{label_ML} is a perfect, powerful match for the estimation (inference) part in complex systems. This is why it would be important to match the idea of thinking structurally about the data-generating process (from causality) and then use the \acrshort{label_ML} power for the estimation part. Integrating causal reasoning in \acrshort{label_ML}/\acrshort{label_DL} holds promises toward robustness to exogenous changes and accurate modeling of counterfactual scenarios that are core to scientific experimentation, understanding, and decision-making.

\subsection{\textit{How?} Causal Modelling: From Association to Causation}
\label{sec:back_causality_how}
\subsubsection{Directed Acyclic Graph (DAG)}
\label{sec:back_causality_how_DAG}
\begin{figure}
    \centering
    \includegraphics[width=0.5\textwidth]{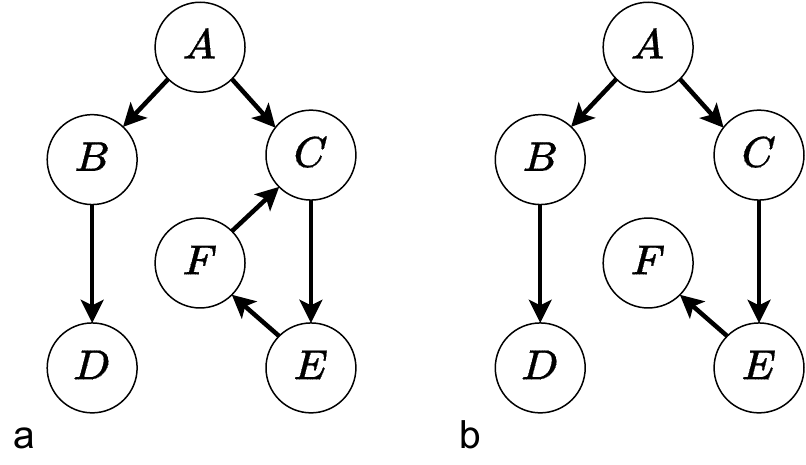}
    \caption{Examples of directed graphs: (a) directed cyclic graph, (b) directed acyclic graph (\acrshort{label_DAG}).}
    \label{fig:simple_dag}
\end{figure}

Much of the discussion of causality and qualitative modeling is occupied by \acrshort{label_DAG}s \cite{pearl2009causality}. \acrshort{label_DAG}s are graphical models that show the causal relationships between variables using nodes and arrows. The arrows indicate the direction of causation, and the graph has no cycles, meaning that no variable can cause itself. \acrshort{label_DAG}s are important tools because they can help us understand the causal structure and effects of a system, and how to intervene or manipulate it.
From graph theory, a \textit{graph} consists of a set $V$ of vertices (i.e., variables) and a set $E$ of edges (i.e., relationships) that connect some pairs of vertices. A graph is \textit{directed} when all the edges are directed (i.e., marked by a single arrowhead).
In a directed graph, an edge goes from a \textit{parent} node to a \textit{child} node.
A \textit{path} in a directed graph is a sequence of edges such that the ending node of each edge is the starting node of the next edge in the sequence (e.g., nodes $A$, $B$, $D$ in Fig.~\ref{fig:simple_dag}). A \textit{cycle} is a path in which the starting node of its first edge equals the ending node of its last edge (e.g., nodes $C$, $E$, $F$ in Fig.~\ref{fig:simple_dag}a), and this represents mutual causation or feedback processes.
When a directed graph does not include directed cycles, it is called a \acrshort{label_DAG}.

\subsubsection{Bayesian Network (BN)}
\label{sec:back_causality_how_BN}
Some other important tools for causality are \textit{Bayesian networks} (\acrshort{label_BN}s) \cite{pearl1985bayesian}. \acrshort{label_BN}s are graphical models that represent the probabilistic relationships between variables using nodes and edges. A BN can be used to factorize the joint distribution of the variables, and to perform inference and learning from data.
Typically, a \acrshort{label_BN} consists of two parts: a qualitative one based on a \acrshort{label_DAG}, representing a set of variables and their dependencies, and a quantitative one based on local probability distributions for specifying the probabilistic relationships.

Let $\textbf{X}=[X_1, X_2, \dots, X_m]$ be a data matrix  with $n$ samples and $m$ variables. In the \acrshort{label_DAG} $G=(V, E)$ of a \acrshort{label_BN}, each node $V_k \in V$ represents the random variable $X_k$ in $\textbf{X}$, $k \in \{1,2, \dots, m\}$, and each edge $e \in E$ describes the conditional dependency between pairs of variables. The absence of an edge implies the existence of conditional independence. 

A \acrshort{label_BN} can also encode causal information if the edges represent direct causal influences. This is called a \textit{causal \acrshort{label_BN}}, which can be used to answer questions about interventions (i.e., changing the value of a variable, regardless of its causes), counterfactuals (i.e., what would have happened if something was different), and causal effects (i.e., the difference in the outcome between two scenarios, one with the intervention and one without).

The structure of the \acrshort{label_DAG} can be constructed either manually, with expert knowledge of the underlying domain (knowledge representation), or automatically learned from a large dataset. In this regard, causal discovery (\acrshort{label_CD}) denotes a broad set of methods aiming at retrieving the topology of the causal structure governing the data-generating process, using the data generated by this process. Popular \acrshort{label_CD} algorithms include the PC algorithm \cite{spirtes1991algorithm}, the Fast Causal Inference (\acrshort{label_FCI}) algorithm \cite{spirtes2000causation}, and the Greedy Equivalence Search (\acrshort{label_GES}) \cite{chickering2002optimal}.

Regarding the quantitative part of which a \acrshort{label_BN} consists, the local probability distributions can be either \textit{marginal}, for nodes without parents (root nodes), or \textit{conditional}, for nodes with parents. In the latter case, the dependencies are quantified by Conditional Probability Tables (\acrshort{label_CPT}s) for each node given its parents in the graph. These quantities can be estimated from data in a process known as Parameter Estimation, two popular examples of which are the Maximum Likelihood approach and the Bayesian approach.

\begin{figure}
    \centering
    \includegraphics[width=0.75\textwidth]{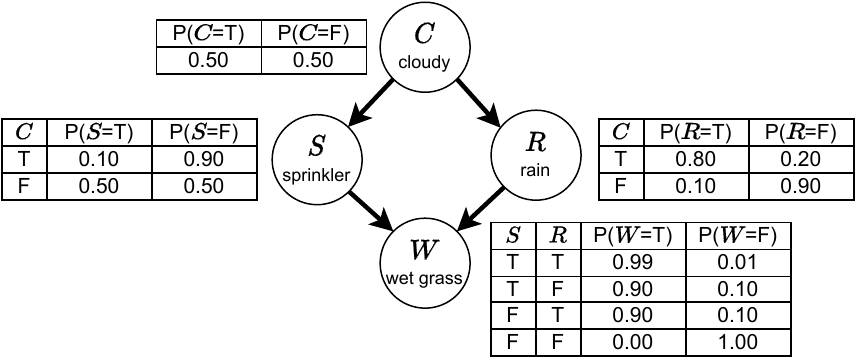}
    \caption{Example of a fully specified BN which models the probability of observing wet grass. In this (simplified) real-world scenario, grass can be wet either by turning on a sprinkler or by rainfall, and both can be influenced by the presence of clouds in the sky.}
    \label{fig:simple_BN}
\end{figure}

Once the \acrshort{label_DAG} and \acrshort{label_CPT}s are determined, a BN is fully specified and compactly represents the Joint Probability Distribution (\acrshort{label_JPD}). An example of a fully specified \acrshort{label_BN} is shown in Fig.~\ref{fig:simple_BN}. 
According to the \textit{Markov condition}, each node is conditionally independent of its non-descendants, given its parents. As a result, the \acrshort{label_JPD} can be expressed in a product form:
\begin{equation}
    \label{eq:conditional_prob}
p(X_1, X_2, \dots, X_m) = \prod_{k=1}^{m} p(X_k | \mathbb{X}_{pa(k)})
\end{equation}
Where $\mathbb{X}_{pa(k)}$ is the set of parent nodes of $X_k$ and $p(X_k | \mathbb{X}_{pa(k)})$ is the conditional probability of $X_k$ given $\mathbb{X}_{pa(k)}$.
Thus, such a \acrshort{label_BN} can be used for predictions and inference, that is, computing the posterior probabilities of any subset of variables given evidence about any other subset.

\subsubsection{Structural Causal Model (SCM)}
\label{sec:back_causality_how_SCM}
What really made it possible to formalize causality in mathematical terms within an axiomatic framework was the concept of \textit{Structural Causal Models} (\acrshort{label_SCM}s) \cite{pearl2009causality,scholkopf2021toward}. \acrshort{label_SCM}s consist of a set of variables, a set of structural equations that define how each variable is determined by its causes, and a set of exogenous variables that represent the external factors that are not modeled. \acrshort{label_SCM}s can represent both deterministic and probabilistic causality and can handle various types of interventions, counterfactuals, and causal effects. \acrshort{label_SCM}s can also be represented graphically, using causal diagrams or graphs, which show the causal relationships between the variables using nodes and edges.

Consider the set $\textbf{X}$ of variables associated with the vertices of a \acrshort{label_DAG}. When each of them appears on the left-hand side (i.e., the dependent variable) of an equation of the type:  
\begin{equation}
\label{eq:scm}
X_k = f_k(\mathbb{X}_{pa(k)}, U_k), \quad k = 1, \dots, m
\end{equation}
that represents an autonomous mechanism, then the model is called an \acrshort{label_SCM} \cite{pearl2009causality, scholkopf2021toward}.
In this equation, $f_k$ represents a deterministic function depending on the $X_k$'s parents in the graph (i.e., $\mathbb{X}_{pa(k)}$), and on $U_k$, which represents the exogenous variables (i.e., errors or noises due to omitted factors).
These noises are assumed to be jointly independent, and hence, ensure that each structural equation can represent a general conditional distribution $p(X_k | \mathbb{X}_{pa(k)})$,
Recursively applying Eq.~\ref{eq:scm}, when the distributions of $U = \{U_1,\dots,U_m\}$ are specified, allows the computation of the entailed observational joint distribution $p(X_1, X_2,\dots,X_m)$, which, in turn, can be canonically factorized into Eq.~\ref{eq:conditional_prob}.

The advantages of using the \acrshort{label_SCM} language include modeling unobserved variables (i.e., latent variables and counfounders), easily formalizing interventions, and computing counterfactuals. Indeed, the causal Markov kernels $p(X_k | \mathbb{X}_{pa(k)})$ typically capture independent physical processes in the sense of the “Independence of Causal Mechanisms” (\acrshort{label_ICM}) principle \cite{peters2017elements}:
\begin{quote}
    \textit{
    The causal generative process of a system’s variables is composed of autonomous modules that do not inform or influence each other. In the probabilistic case, this means that the conditional distribution of each variable given its causes (i.e., its mechanism) does not inform or influence the other conditional distributions.
    }
\end{quote}

\subsubsection{Computing Interventions and Counterfactuals}
\label{sec:back_causality_how_Interventions_Counterfactuals}
An important implication of \acrshort{label_ICM} is \textit{modularity}. From the manipulation or intervention viewpoint, since the causal mechanisms are autonomous modules, changing one of the $p(X_i | \mathbb{X}_{pa(i)})$ will not affect the other causal conditionals $p(X_j | \mathbb{X}_{pa(j)})$ on the right-hand side of Equation \ref{eq:conditional_prob}, but will leave the rest of the system invariant.

Interventions and counterfactuals are defined through a mathematical concept called $do(\cdot)$ operator, which simulates physical interventions by modifying a subset of structural equations (e.g., replacing them with a constant), while keeping the rest of the model unchanged.
For instance, an intervention that fixes the value of one of the $X_i$ to some constant $x_i$ is denoted $do(X_i = x_i)$. Due to the assignment character of structural equations, interventions are naturally expressed within the \acrshort{label_SCM} framework. To model an intervention, one simply replaces the corresponding structural equations, which gives rise to a new interventional \acrshort{label_SCM}.

To go a step further, counterfactuals differ from interventions for a piece of background information - observing the factual outcome (i.e., what actually happened) provides information about the background state of the system (as captured by the noise terms) which is used to reason about alternative, counterfactual outcomes.
Specifically, to compute the probability of counterfactuals, Pearl proposes a three-step procedure (i.e., a recipe). Given a known SCM $M$ over the set $\textbf{X}$ of variables, let $x_{factual} = [X_1=x_1,X_2=x_2,\dots,X_m=x_m]$ be the evidence. To compute the probability of a counterfactual instance $x_{counterfactual}$, one needs to:
\begin{enumerate}
    \item \textit{Abduction}: infer the values of exogenous variables in $U$ for $x_{factual}$, i.e., calculate $P(U|x_{factual})$. This is needed to update the noise distribution to its posterior given the observed evidence.
    \item \textit{Action}: intervene on $X=x_{factual}$ by replacing (some of) the equations by the equations $X=x_{counterfactual}$, where $x_{counterfactual} = [X_1=x_1',X_2=x_2',\dots,X_m=x_m']$, and thus obtain a new SCM $M'$. This is to manipulate the structural equations to capture the hypothetical intervention.
    \item \textit{Prediction}: use $M'$ to compute the probability of $P(x_{counterfactual}|x_{factual})$; This corresponds to using the modified \acrshort{label_SCM} to infer the quantity of interest.
\end{enumerate}

\subsubsection{Back-door Adjustment}
\label{sec:back_causality_how_BackdoorAdj}
A common problem in causal inference is spurious relationships due to confounding, which occurs when two variables appear to be causally related but, in fact, they are both influenced by a third variable called a \textbf{confounder}, that is not observed or controlled. For example, \textit{ice cream sales} and \textit{shark attacks} may seem causally related, but they are both confounded by the \textit{summer season}, which affects both of them. Unfortunately, this is where conventional AI approaches to classification or regression stand today.

In causal inference and DAG theory, the \textit{back-door adjustment} is a popular method to estimate causal effects in observational research, and involves conditioning on possible confounders between a treatment/exposure variable $X$ and an outcome variable $Y$.
The term "adjustment" refers to changing the \textit{adjustment set} $Z$, the set of variables included as covariates/features in a model. In this regard, the back-door adjustment method involves controlling for all variables that are causes of both the treatment $X$ and the outcome $Y$. This method identifies the set of nodes to condition on that collectively block all “back-door paths” between $X$ and $Y$. The intuition is that if these paths are blocked, any systematic correlation between $X$ and $Y$ reflects the effect of $X$ on $Y$.

To calculate interventional distributions under these settings, we can estimate the effect of $X$ on $Y$ by the formula:
\begin{equation}
    P(y|do(X=x)) = \sum_z P(y|x,z)P(z)
\end{equation}
which can be interpreted as dividing the data into categories by the values of $Z$ and $X$ (this is also called \textit{stratifying}) and calculating the weighted average of the \textit{strata} (i.e., the data categories). By conditioning on these two variables, we make the strata independent of each other. As $Z$ blocks the back-door paths, conditioning on $X$ is the same as $do(X=x)$.
 
\subsection{\textit{Where?} Hot Directions of Causal ML}
\label{sec:back_causality_where}
Where is the research on integrating causal reasoning in \acrshort{label_ML} and \acrshort{label_DL} systems moving? The following perspectives utilize the causal tools illustrated above to different extents, and we indicate where the contribution made in this thesis can be mapped into those hot directions.

\textbf{Causal supervised learning} integrates causal inference principles into traditional supervised learning tasks to improve generalization and robustness. Predictive generalization is enhanced by learning invariant features or mechanisms, both aiming at deconfounding models’ reliance w.r.t. spurious associations, tackling distribution shifts via causal understanding of data generation. Based on the concept of causal inference, \textit{Invariant Risk Minimization} (\acrshort{label_IRM}) is an approach designed to learn representations that are invariant across different environments, i.e., invariant causal predictors \cite{arjovsky2019invariant}, while \textit{causal regularization} uses causality as an inductive bias to regularize the learning process \cite{rojas2018invariant}. By embedding causal knowledge, such as relationships derived from causal graphs or domain expertise, this approach penalizes the model for relying on correlations and encourages it to focus on genuine causal relationships \cite{janzing2019causal}.
More recent works have appeared on causal domain generalization \cite{sheth2022domain,sheth2023causal}, multi-task generalization via regularization of spurious correlation \cite{hu2022improving,kumar2024causal} or using auxiliary labels\cite{makar2022causally}, learning domain invariant causal supervised representation \cite{wang2022causal,cho2024learning,wang2024learning}, robust learning via conditional prevalence adjustment \cite{nguyen2024robust}, deploying variational back-door adjustment in high dimensional data \cite{israel2023high,li2023disentangle,tang2023causality}, discovering causally invariant features for \acrshort{label_OOD} generalization \cite{atzmon2020causal,sui2022causal,fan2023generalizing,xu2023multi,zhang2023out,sui2024enhancing,wang2024discovering}, and causal attention \cite{wang2021causal,rao2021counterfactual,zhang2023towards}.

Specifically to medical image analysis, \cite{qu2024causality} propose a generalizable method for pancreatic cancer diagnosis based on causal intervention, and \cite{fay2024towards} use non-imaging information to learn a multi-modal feature representation that can make predictions based on learned causal dependencies while avoiding spurious correlations. Also, \cite{ong2024shortcut} study the shortcut learning in medical \acrshort{label_AI} from a causal viewpoint, and \cite{fay2023avoiding} propose a method to avoid it by mutual information minimization in brain MRI for Alzheimer's status, age group, and gender classification. \cite{montenegro2022disentangled} propose a privacy-preserving generative model inspired by causality that explicitly disentangles medical and identity features in a biometric eye iris dataset and a chest radiograph dataset. 

\textbf{Causal representation learning} (\acrshort{label_CRL}) aims at recovering latent causal variables from high-dimensional observations to solve causal downstream tasks, such as predicting the effect of new interventions or more robust classification. In other words, \acrshort{label_CRL} encodes causal structures into the latent spaces of \acrshort{label_ML} models to attain invariant, robust, and interpretable representations under interventions and counterfactuals \cite{scholkopf2021toward,ahuja2023interventional,ahuja2024multi,guo2023causal,zhang2024causal}. \acrshort{label_CRL} can manifest in many facets, such as: causal discovery for observational data \cite{nauta2019causal,jaber2020causal,petersen2022causal}, learning or assuming causal structures for the data generation process \cite{heinze2021conditional,yin2023causal} (even supervised by \acrshort{label_LLM}s \cite{ban2023causal}), causal generative modeling and diffusion-based methods \cite{mamaghan2024diffusion,komanduri2023identifiable,ribeiro2023high,komanduri2024causal}, causal embeddings \cite{zhang2023personalized,roy2024directed}, structured data augmentations\cite{eastwood2023self}, or causal feature selection \cite{zhang2020causal,wang2024discovering}. Moreover, \acrshort{label_CRL} benchmarks and datasets are emerging \cite{liu2023causal,lynch2023spawrious}, together with open-source Python libraries \cite{zheng2024causal}. Other interesting lines relate to learning causal disentangled representations from soft interventions \cite{bagi2024implicit,komanduri2023learning,zhang2024identifiability} or for missing data \cite{chen2024cdrm}.
Notably, recent works have been investigating how to identify causal representations from unknown (possibly latent) interventions \cite{buchholz2024learning,von2024nonparametric}, or unify \acrshort{label_CRL} with the invariance principle \cite{yao2024unifying} and with foundation models \cite{rajendran2024learning}. 
An interesting recent discussion is also provided by \cite{richens2024robust}, which shows how any agent capable of adapting to a sufficiently large set of distributional shifts must have learned a causal model of the data-generating process.

On the medical side, works have been started to describe the importance of considering causal \acrshort{label_ML} and \acrshort{label_CRL} in healthcare systems and medical image analysis \cite{sanchez2022causal,lu2022survey,blaas2023considerations}, and learning disentangled representations in the imaging domain \cite{pei2021disentangle,zhou2021chest,liu2022learning,fragemann2022applying}. In \cite{mao2022towards}, they propose learning causal structure and representation in COVID-19 diagnosis from chest X-rays. \cite{shen2022leads} propose an active \acrshort{label_CRL} in the ECG heartbeat classification field, \cite{luo2022deep} propose a causal embedding method to learn representations for prediction models from \acrshort{label_EHR}s, via removing dependencies between diagnoses and procedures by a weighting technique, while \cite{abdulaal2024causal} synergizes the metadata-based reasoning capabilities of \acrshort{label_LLM}s with the data-driven modeling of deep \acrshort{label_SCM}s for Alzheimer’s applications. 
In \cite{gu2023biomedjourney}, the authors present a novel method for counterfactual medical image generation by instruction image editing (instruction-learning) from multimodal patient journeys. To improve their model generalization performance, \cite{li2023causality} impose a causal contrastive mechanism to suppress non-causal features unrelated to the clinical diagnosis, while \cite{ouyang2022causality} and \cite{qiu2023causality,nie2023chest} act similarly for medical image segmentation and classification, respectively.
As for causal structure learning, \cite{yang2023lung} proposes building a network between the computational and semantic features of lung nodules through causal discovery algorithms and detecting lung nodules based on this network.

\textbf{Causal explanations} explain model predictions while accounting for the causal structure of either the model mechanics or the data-generating process \cite{beckers2022causal,carloni2023role,baron2023explainable}.
A first family of causal explanations is \textit{feature attribution} methods, which quantify the causal impact of input features. Some works have proposed models that learn to estimate to what degree specific inputs cause outputs in another \acrshort{label_ML} model \cite{schwab2019cxplain}, causality-inspired model interpreters \cite{wu2023causality}, methods defined over an \acrshort{label_SCM} for a dual-stage (factual and interventional) perturbation test \cite{chen2024feature}, etc. Other lines of research encompass generative causal explanations, where the latent factors involved in a black-box classification are learned and then included in a causal model \cite{o2020generative}, causal explanations for reinforcement learning models \cite{madumal2020explainable}, or drawing argumentative explanations from causal models \cite{rago2023explaining}. Moreover, \cite{wu2023causal} explored using approximate counterfactual training data to build more robust causal explanations, mimicking a black-box model's factual and counterfactual behaviors. \cite{breuer2024cage} present a global causal explanation framework based on Shapley values, which requires a predefined causal structure for features.
The second type of causal explanation is \textit{contrastive explanations} representing altered instances achieving the desired outcome \cite{lipton1990contrastive,robeer2018contrastive,molnar2020interpretable,stepin2021survey}. Such methods align and sometimes overlap with counterfactual explanations, which we described in Section \ref{sec:back_XAI}, and affect important areas such as interpretability \cite{konig2023if,kumar2023debiasing}, visual explanations with overdetermination\footnote{
Causal overdetermination is the situation where there are multiple segments in an image that are sufficient individually to cause the classification probability to be close to one.} \cite{white2023contrastive}, and ranking models \cite{castelnovo2024evaluative}.
Medical applications of causal explanations include counterfactual generation via a deep \acrshort{label_SCM} of the clinical and radiological phenotype of Alzheimer's \cite{wang2021interpretable,abdulaal2022deep}, psychiatric diagnoses \cite{maung2016diagnosis}, pregnancy prognostication via non-image data \cite{sufriyana2023human}, and localization of important brain regions for individual-level and group-level biomarkers for gender classification from heterogeneous brain measurements \cite{zhao2022explainable}.
%

Although they are outside the scope of this thesis, there are other areas where causal reasoning is entering the \acrshort{label_ML} field.
\textit{Causal generative modeling} supports sampling from interventional distributions to perform principled controllable generation, or from counterfactual distributions to sample editing tasks. Such methods learn the structural assignments, and some infer the causal structure from data \cite{goudet2018learning,reinhold2021structural,rasal2022deep,zhou2023opportunity,liu2022causalegm}.
\textit{Causal fairness} paves the way for criteria that assess a model’s fairness as well as mitigate harmful disparities w.r.t. causal relationships of the underlying data \cite{plecko2022causal,zhou2024counterfactual}.
\textit{Causal reinforcement learning} address RL considering the explicit causal structure of the decision-making environment into account, with benefits including deconfounding, intrinsic rewards, and data efficiency \cite{zeng2023survey, ruan2022causal,deng2023causal, gershman2017reinforcement}.

\subsection{\textit{Who?} Benefits and Implications of Causal ML in Medical Imaging}
\label{sec:back_causality_who}
Ultimately, who could be the beneficiary of all these approaches? Understanding the mechanisms and effects of interventions, manipulations, or counterfactual scenarios would facilitate every decision-making and policy design.
Causal \acrshort{label_ML}/\acrshort{label_DL} allows us to build accurate decision-support systems that estimate the effect of interventions under real-world constraints, such as messy observed data, imperfect causal knowledge, and computational constraints. This is why causal \acrshort{label_AI} has strong implications for medical image analysis \cite{vlontzos2022review,liu2023towards}.

However, do not expect causal \acrshort{label_ML} to be a "magic box" solution. Valuable projects demand domain expertise and an understanding of context (e.g., biological variations, medical imaging scanners, radiology protocols). Indeed, practitioners should combine causal \acrshort{label_AI} methods with expert domain knowledge to obtain the most effective results. Also, we should treat the results as hypotheses, not definite answers, when using causal discovery on data from high-stakes domains (such as healthcare)\cite{kaddour2022causal,ghosal2023contextual}.

Causal \acrshort{label_ML}/\acrshort{label_DL} can help us enhance the\acrshort {label_OOD} generalizability, robustness, and fairness of deep models across different domains and contexts \cite{liu2021learning,ouyang2022causality,tang2023causality,pavlovic2024improving}. For instance, healthcare facilities and hospitals would benefit from an \acrshort{label_AI} model that accounts for confounding features and selection/estimation bias \cite{bellamy2022structural,sanchez2022causal}.
In this regard, "\textit{causality matters in medical imaging}" \cite{castro2020causality}, as it can expose potential biases and form mitigation techniques to alleviate the scarcity of high-quality annotated data and mismatch between the development dataset and the target environment. Also, \cite{koccak2024bias} remark on the importance of proactively identifying and addressing \acrshort{label_AI} bias in medical imaging to prevent its adverse consequences from being realized later. 

\acrshort{label_ML} and \acrshort{label_DL} models that integrate causality can help us explain their behavior or decisions by identifying the relevant features, factors, or paths that led to the output or outcome of interest. 
\cite{harradon2018causal} develop an approach to explaining \acrshort{label_DNN}s by constructing Bayesian causal models on salient, human-understandable representations contained in the CNN.
\cite{chen2022c} propose a causal \acrshort{label_CAM} for weakly supervised semantic
segmentation under ambiguous boundaries and organ co-occurrence.
\cite{miao2023caussl} proposes a novel causal diagram to provide plausible explanations for the effectiveness of semi-supervised learning medical image segmentation. 
\cite{ribeiro2023high} propose using probabilistic causal models to generate high-fidelity image counterfactuals. 
All of that contributes to the \acrshort{label_XAI} goal and confirms how causality and explainability are indeed intertwined\cite{carloni2023role}.

Causal \acrshort{label_ML}/\acrshort{label_DL} can also help researchers discover new causal knowledge and insights from data by using deep models to learn flexible and expressive representations that capture the causal structure and semantics of the data \cite{lagemann2023deep}. Still, the challenge of encoding and structuring causal knowledge in the representation space remains. This direction could help scale causal effect estimation methods to high-dimensional and unstructured data such as text and images \cite{chu2023causal}. This would have important implications for scientific discovery and understanding of pursue-worthy experiments.

\section{Biological Visual System in Humans}
\label{sec:back_biological}
\subsection{Visual Information Processing}
\label{sec:back_biological_informationProcessing}
Historically, visual information processing in humans was described with the \textit{two-streams theory}: two anatomically distinct and functionally specialized cortical pathways exist, the \textbf{ventral} stream processes visual features like color, size, and dimension, and the \textbf{dorsal} stream primarily deals with the object's spatial features (location, orientation, and motion) \cite{goodale1992separate,whitwell2014two}.
\begin{figure}[tb]
  \centering
  \includegraphics[width=0.85\textwidth]{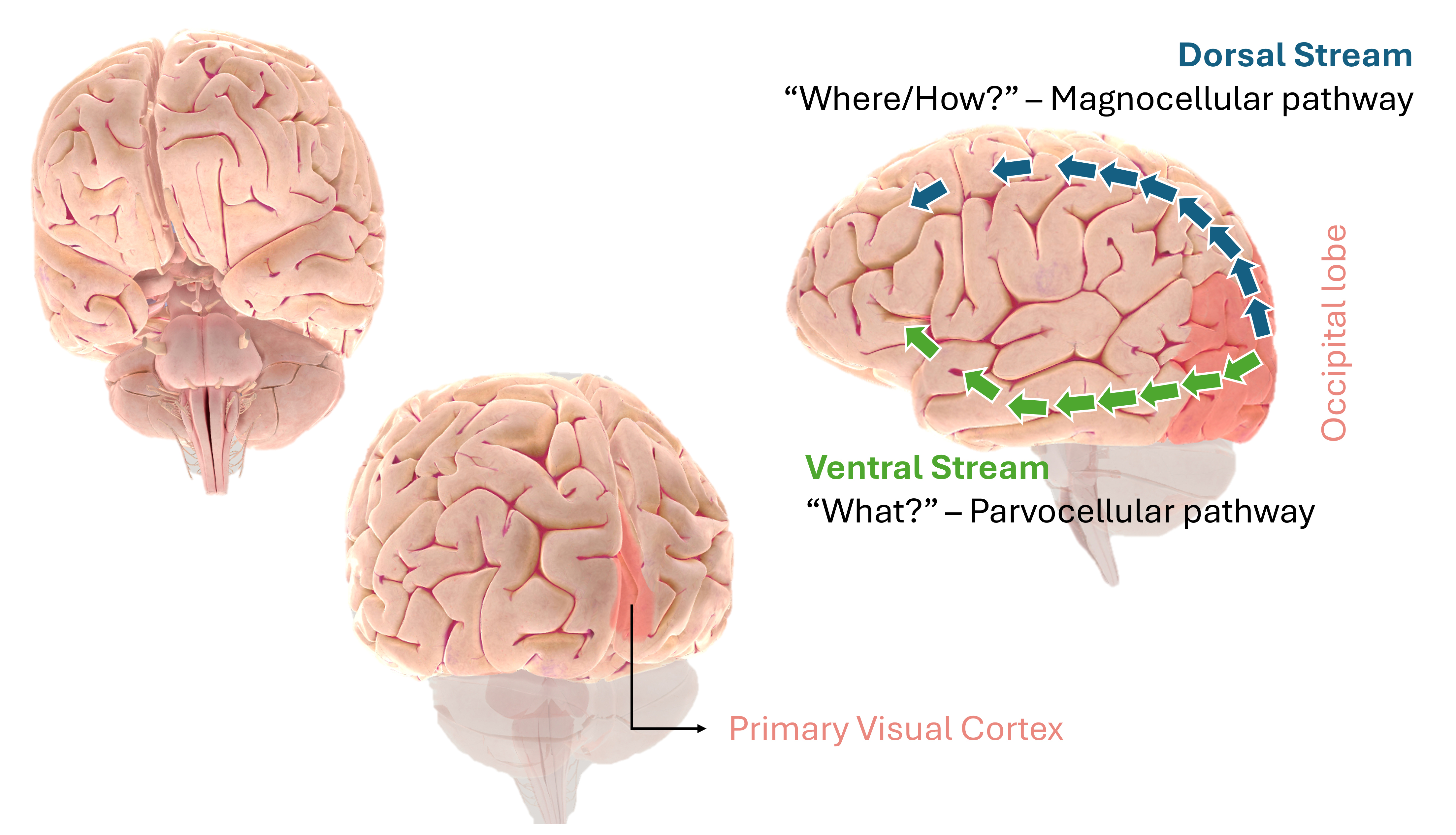}
  \caption{The ventral and dorsal visual pathways in human vision. Adapted from \url{https://www.brainfacts.org/} of the Society for Neuroscience (2017).}
  \label{fig:two_streams}
\end{figure}
The \textbf{ventral stream} runs from the primary visual cortex (V1) mainly through extrastriate visual areas II (V2) and IV (V4) to the inferotemporal cortex (\acrshort{label_ITC}), and then to the prefrontal cortex (\acrshort{label_PFC}), which is involved in linking perception to memory and action. This pathway is responsible for object recognition, categorization, and memorization, thus representing object shape and identity (i.e., the \textit{What?} of a visual scene).
Since it receives signals from the parvocellular cell (\acrshort{label_P_cell}) layers, which are sensitive to color and have a higher spatial resolution but lower temporal resolution, the ventral pathway processes the feature-rich information for fine local processing (i.e., textures and edges) in a bottom-up manner to form detailed representations of visual stimuli.
Indeed, information is processed sequentially with increasing complexities \cite{konen2008two}, and the invariance of those representations to position and scale increases. In parallel, there is also an increase in the size of the receptive fields as well as in the complexity of the optimal stimuli for the neurons \cite{gattass1981visual,gattass1988visuotopic}.

The \textbf{dorsal stream} is responsible for spatial perception, motion detection and attention, and thus represents spatial vision and visuomotor control, constituting the \textit{Where?} of the visual stimuli.
Anatomically and functionally, the dorsal stream runs from V1 to the parietal lobe. Since it receives retinal information from the magnocellular (\acrshort{label_M_cell}) layers, it involves fast processing of the information sensitive to luminance changes (high contrast gain) and low spatial frequencies.
Whereas visual representations in the ventral pathway are more invariant and reflect \textit{“what an object is”} (vision-for-\textit{perception}), those in the dorsal pathway are more adaptive and reflect \textit{“what we do with it”} (vision-for-\textit{action}).
At some level of neural processing, information about the identity and location of an object represented in the segregated pathways must be integrated. To this end, many schemes have been proposed in the literature \cite{perry2014feature,milleret2018beyond,choi2020proposal}, although no strict consensus on the ventral-dorsal modeling has been reached yet.

\subsection{Ventral and Dorsal Streams Communicate}
\label{sec:back_biological_ventralDorsal}
In the early 2000s, the general belief of a complete division of labor by two segregated pathways started to change. New evidence was calling for synergies between them: there exist “what” and “where” information in \textit{both} visual processing pathways.
Evidence supports the hypothesis that \textbf{dorsal and ventral visual areas communicate}, and there are shape-selectivity and non-action-based perceptual representations in the posterior regions of the dorsal pathway \cite{lehky2007comparison,konen2008two,takemura2016major}.

Moreover, the dorsal pathway itself is composed of several sub-pathways, where at least one has a functional, and probably necessary, role in object perception \cite{rousselet2004parallel}. 
Under this light, the two streams are not segregated but constitute an important symbiosis crucial in transmitting signals between regions. Shape encoding is thus performed \textit{also} in the dorsal pathway, and it is distinct from and not a mere duplication of that formed in the ventral pathway. This means that dorsal object representations are dissociable from those generated in the ventral pathway and play an independent and functional role in visual perception.
Therefore, visual perception should be studied not simply as a function of one (ventral) "what" pathway, but rather as the joint outcome of the processing and coordination of different "what" regions in both cortical visual pathways.

\subsection{A Fast Top-Down Modulation of Bottom-Up Representations}
\label{sec:back_biological_topDown}
In addition to the traditional bottom-up hierarchy of representation, new mechanisms of top-down processing were proposed \cite{bar2001cortical,bar2003cortical,bar2006top}. There exists a non-cortical, fast, "shortcut" stream in which early visual inputs are sent, partially analyzed, from the early visual cortex  (V1) to the \acrshort{label_PFC}. Possible interpretations of the crude visual input are generated in the PFC and then sent to the \acrshort{label_ITC}, subsequently activating relevant object representations, which are then incorporated into the slower, bottom-up process \cite{bar2006very,bar2007proactive}. Eventually, the coarse and global representations from the subcortical pathway guide and modulate the fine local representations from the ventral pathway in a \textbf{“top-down”} fashion to form more precise representations of the object category.
In this view, \acrshort{label_PFC} areas would receive coarse, low-resolution information via fast dorsal projections and generate predictions about object identity. This prediction would be feedback to the \acrshort{label_ITC}, facilitating recognition by limiting the number of possible object candidates \cite{paneri2017top}.
This “faster” subcortical pathway is parallel (or completed prior) to the ventral one.

In humans and other mammals, the two strongest pathways linking the eye to the brain are those projecting to the dorsal part of the lateral geniculostriate nucleus (\acrshort{label_LGN}) in the thalamus and to the superior colliculus (\acrshort{label_SC})\cite{goodale2013sight}.
From the former originate the above-described schema of ventral and dorsal streams. At the same time, the latter constitutes the other major retino–cortical visual pathway known as the tectopulvinar pathway, routing primarily through the \acrshort{label_SC} and thalamic pulvinar onto ventral area V4 and dorsal area V5/MT.

\subsection{Visual Context in Object Recognition}
\label{sec:back_biological_visualContext}
Context is of fundamental importance to both human and machine vision; e.g., an object in the air is more likely to be an airplane than a pig. The rich notion of context incorporates several aspects, including physics rules, statistical co-occurrences, and relative object sizes, among others \cite{bomatter2021pigs}.
Context is of critical importance for locating a target object in complex scenes as it helps narrow down the search area and makes the search process more efficient \cite{ding2022efficient}. Unsurprisingly, \acrshort{label_AI} and computer vision solutions embedding context information have emerged in the literature. Examples include zero-shot visual search \cite{ding2022efficient, bomatter2021pigs}, context-aware attention networks \cite{zhang2020putting}, and other computer vision models \cite{torralba2003contextbased,choi2012context,oliva2007role,mack2011object,kim2020learning}.
Unlike such solutions, as we shall see in Section \ref{sec:methods_cocoreco}, we propose a method that does not add computational overhead in terms of parameters, as it does not involve additional trainable parameters.
\chapter{\textit{Watch Your Neurons!}: Visualise Learned Representations for Natural and Medical Images}
\label{chap:XAI_wyn}
\begin{figure}[h!]
\includegraphics[width=\textwidth]{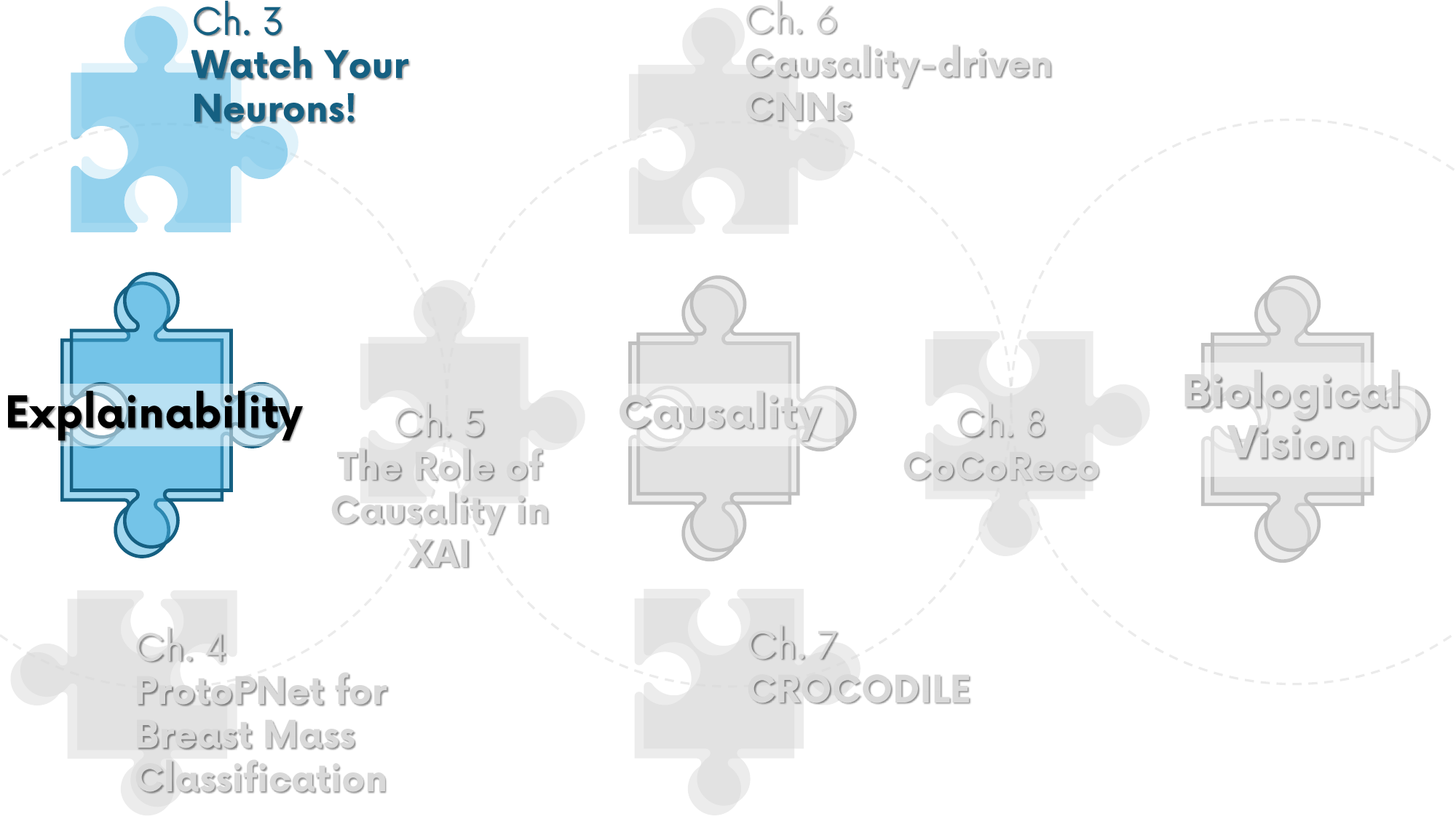}
\end{figure}

To attain a prediction via \acrshort{label_DNN}s, the input data goes through several steps, including multiplying by the weights the network has learned and applying some complex changes. Making just one prediction can require a considerable number of math calculations, and the exact number depends on how the neural network is built. Moreover, \acrshort{label_CNN}s learn abstract features and concepts in their hidden layers from raw image pixels, and we can't track every single step from the input data to the final prediction because so many factors are involved. This is why interpretability and explainability are necessary research fields in \acrshort{label_AI} (ref. Section \ref{sec:back_XAI}).

In Section \ref{sec:back_XAI_methods_FeatureViz}, we have introduced an interesting approach to visualize the learned representations of a trained \acrshort{label_DNN}. Indeed, \textit{feature visualization}, and specifically \acrshort{label_AM}, is a more computationally efficient method than post-hoc, model-agnostic solutions that look at the model “from the outside” (ref. Section \ref{sec:back_XAI_postHoc_anteHoc}).

The aim of this Chapter is threefold. First, we wanted to experiment with \acrshort{label_AM} on natural images to understand how feature visualization works at varying optimization settings. Then, motivated by the correlational brittleness of \acrshort{label_ML}, we wanted to study the feasibility of determining possible sources of bias in image datasets through \acrshort{label_AM} on a pre-trained model. Finally, we investigated whether this powerful and seemingly simple method of interpreting black-box models could work on medical images. 

The Chapter is organized as follows. In Section \ref{sec:XAI_wyn_AM}, we formalize the problem of \acrshort{label_AM} by outlining the gradient ascent optimization and discussing the importance of image prior regularizers.
Then, Section \ref{sec:XAI_wyn_ExpNatImg} presents our investigation of \acrshort{label_AM} methods applied to natural images (e.g., photographs). We experiment with exaggerating the input image size and number of iterations, highlighting unexplored behaviors. In the same section, we utilize \acrshort{label_AM} to expose potential dataset biases and shortcut features in image datasets, and we study the effect of stochastic variabilities in \acrshort{label_AM}.
Section \ref{sec:XAI_wyn_MedicalImages} presents our investigation on utilizing \acrshort{label_AM} for medical imaging applications, showing empirical evidence of how difficult it is to employ such methods in those scenarios. Next, we propose novel mitigation strategies and discuss potential reasons for their failure, before summarizing in Section \ref{sec:xai_wyn_summary}. 

\section{Activation Maximization: Problem Definition}
\label{sec:XAI_wyn_AM}
We have seen how feature visualization and \acrshort{label_AM} are basically optimization problems. We want to find the optimal input image that would cause the firing of a specific neuron. There are several strategies to tackle this goal. For instance, we might sift through our existing training images to find the ones that trigger the highest response in the network. While this method works, it has a drawback: the items in the training images might be related, making it unclear what the neural network focuses on. If the images that cause strong reactions in the network feature both a cat and a basket ball, it's uncertain whether the network responds to the cat, the basket ball, or perhaps to both.

\subsection{Gradient Ascent Optimization}
\label{sec:XAI_wyn_AM_gradientAscent}
Mathematically, \acrshort{label_AM} can be defined as:
\begin{equation}
    X^* = \arg\max_X a(X)
    \label{eq_AM}
\end{equation}
where $X^*$ is the optimal input, $X$ is the input variable, and $a(X)$ is the activation function of the unit of interest \cite{erhan2009visualizing,olah2017feature,nguyen2019understanding}. The corresponding update rule for the optimization procedure entails computing the gradient of the activation w.r.t the input:
\begin{equation}
    X^{(t+1)} = X^{(t)} + \gamma \frac{\partial a(X^{(t)})}{\partial X^{(t)}}
    \label{eq_optimization}
\end{equation}
where $t$ is the iteration, $\gamma$ is the step size, $a(\cdot)$ is the activation function.

\subsection{Too many solutions!}
\label{sec:XAI_wyn_AM_tooMany}
In its most general form, the optimization entailed by Equation \ref{eq_optimization} is a non-convex problem with too many solutions that cause high activations, especially in high dimensional image space. As a result, AM can often lead to unrealistic results in practice if it is not adequately bounded. The produced images are indecipherable, lack both visible structure and recognizable content, and ultimately fail to convey any discernible information to humans. Therefore, a mismatch exists between what the machine deems to be the optimal depiction of a particular neuron (the closest visual representation of its learned patterns) and what makes sense to a human observer. For more details on such perceptual bias, we refer the interested reader to Offert et al., 2021 \cite{offert2021perceptual}. 

\subsection{Image Priors for Regularization}
\label{sec:XAI_wyn_AM_imagePriors}
To overcome this problem and thus reduce the ubiquitous presence of very high frequency noise (i.e., fine textural patterns), we might inject boundary conditions into how the images are generated. It has been suggested that natural image priors be incorporated in the optimizing objective to act as regularizers that can be applied to the image before the optimization step. \cite{nguyen2019understanding}. Thanks to a regularization term $r(\cdot)$, Equation \ref{eq_AM} becomes:
\begin{equation}
    X^* = \arg\max_X (a(X)-r(X))
    \label{eq_AM_regularizer}
\end{equation}
The goal is to constrain the optimization to create an example that belongs to a distribution of a certain image kind. As a result, the update rule (Equation \ref{eq_optimization}) is extended by the regularizer:
\begin{equation}
    X^{(t+1)} = X^{(t)} + \gamma \frac{\partial a(X^{(t)})}{\partial X^{(t)}} - \lambda \frac{\partial r(X^{(t)})}{\partial X^{(t)}}
    \label{eq_optimization_regularizer}
\end{equation}
where the second right-hand-side (r-h-s) term causes activation to be maximized (as before), and the third r-h-s term causes regularization loss to be minimized.

For instance, reality reveals that natural images (i.e., photographs) are typically \textit{smooth} (i.e., do not contain abrupt changes in the pixel values) and barely have \textit{extreme} intensities in RGB feature space. This prior knowledge can be leveraged to conceive a regularizing loss, which is incorporated into the activation optimization to penalize pixels of extreme intensities and favor smoothness. One smoothness prior is the Total Variation (\acrshort{label_TV}), i.e., the sum of absolute differences for neighboring pixel values, and \acrshort{label_TV} filters have been largely used \cite{mahendran2015understanding}. Instead of computing the analytical gradient, the \textit{blur} trick is used in practice \cite{yosinski2015understanding}, where a blurred version $X_r$ of the input $X$ is used at every iteration step of the algorithm:
\begin{equation}
    X^{(t+1)} = X_r^{(t)} + \gamma \frac{\partial a(X^{(t)})}{\partial X^{(t)}}
    \label{eq_optimization_regularized}
\end{equation}

There is a plethora of other regularization techniques in addition to \textit{blur}. These include jitter (the image is randomly shifted a specified number of pixels) \cite{mordvintsev2015inceptionism}, bilateral filters \cite{tyka2016class}, stochastic clipping \cite{lipton2017precise}, Laplacian pyramids \cite{burt1987laplacian,mordvintsev2015deepdream}, etc.

\subsection{Neurons, Channels, and Layers}
\label{sec:XAI_wyn_AM_NCL}
The computation of feature visualizations by optimization does not refer solely to specific, selected neuronal units. With similar objectives, AM can be extended to groups of neurons such as channels (summing over rows and columns) and layers (summing over rows, columns, and channels, such as in DeepDream \cite{mordvintsev2015inceptionism}).
Therefore, AM empowers researchers with the ability to answer a varying spectrum of questions: what are the typical low-level or intermediate patterns the networks have learned? What is a typical example of this specific class (i.e., a \textit{prototype})? The latter is specifically the interest of this Chapter.

\section{Experiments with Natural Images}
\label{sec:XAI_wyn_ExpNatImg}
The first investigation of this study aimed at developing \acrshort{label_AM} methods to be applied to natural images and understanding how feature visualization works with varying optimization parameters.
Under these settings, we imagine the model has been previously trained and is now being deployed externally.
Specifically, we stress-tested different popular deep image classifiers (ResNet50, VGG16, Inception-V3) with ImageNet pre-trained weights for the automatic classification of $1000$ classes (\acrshort{label_ILSVRC}). 
For clarity of presentation and easiness of understanding, we selected some common and easily recognizable image categories out of the $1000$ available. Specifically, we chose varieties of animals, house objects, food, office-related objects, sports, and general/everyday objects, as follows:
\begin{verbatim}
    {'banana', 'bee', 'bison', 'boxer', 'burrito',
    'castle', 'cup', 'dalmatian carriage dog',
    'desk', 'hammer', 'knot', 'lemon', 'llama',
    'Maltese dog', 'mask', 'nail', 'nipple', 'ox',
    'plate', 'purse', 'rifle', 'screw', 'ski', 'sock',
    'sweatshirt', 'zebra'}.
\end{verbatim}

\begin{figure}
\centering
 \includegraphics[width=0.9275\textwidth]{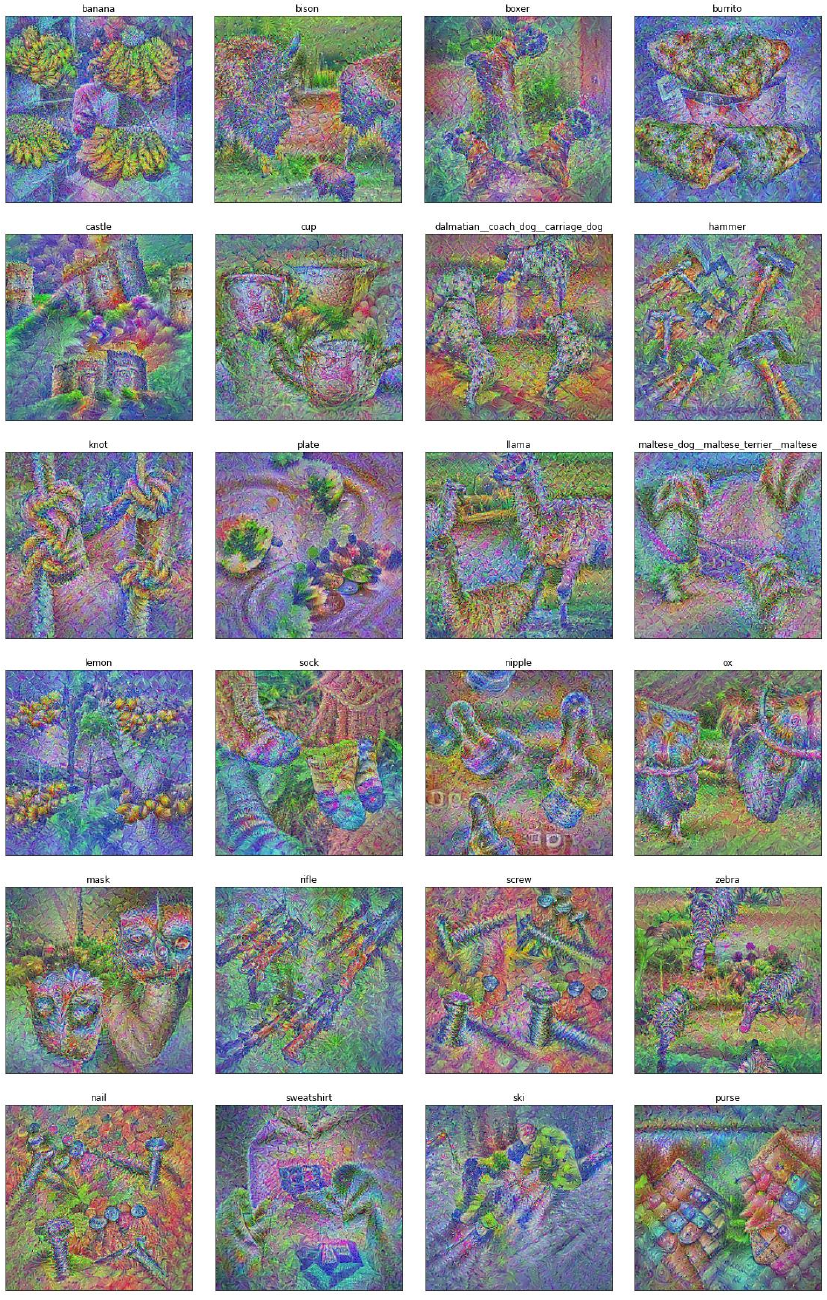}
\caption{Feature visualization through Activation Maximization on selected output neurons of the Inception-V3 model pre-trained on ImageNet (1000 classes). Each panel shows the obtained image that maximally activates the output neuron corresponding to that specific class (out of 1000).}
\label{fig_AMs}
\end{figure}

For our experiments, we chose an initial RGB image of normally distributed random noise of the same size of \acrshort{label_ILSVRC} training data ($224\times224\times3$).
We ran the learning process over $2000$ \acrshort{label_AM} iterations with different parameters of the optimizer:  \acrshort{label_LR} ranged from $0.1$ to $0.9$ and weight decay (\acrshort{label_WD}) from $1e-5$ to $1e-2$. We also applied the following regularizations: \acrshort{label_TV} filtering (either every $4$, $8$, or $16$ iterations), a Jitter of $0.06$ to $0.12$ \% pixels, and stochastic clipping between $0$ and $1$.
We found that the produced visualizations were broadly stable across \acrshort{label_WD} and regularization values, of moderate to high quality, and could be easily interpreted in accordance with the object category/class. For instance, Figure \ref{fig_AMs} shows the obtained \acrshort{label_AM} visualizations of specific output neurons of the last fully connected layer in the Inception-V3 model when using $2000$ \acrshort{label_AM} iterations, \acrshort{label_LR} of $0.4$, \acrshort{label_WD} of $1e-4$, \acrshort{label_TV} every $4$ iterations, a Jitter of $26$ pixels, and stochastic clipping.

\subsection{Exaggerating Input Image Size and Number of Iterations}
\label{sec:XAI_wyn_ExpNatImg_size}
Using an input size equal to the one on which the model was trained seems a natural choice. Similarly, utilizing $2000$ (or fewer) iterations was appropriate, as the loss function steadily reached a plateau by that number of iterations. 
However, we wanted to better understand the capabilities (and potential pitfalls) of feature visualization through \acrshort{label_AM}, and we asked: \textit{What if we exaggerate the size of the input image? What if we purposely increase the number of iterations?} 

\begin{figure}
\centering
\includegraphics[width=0.9\textwidth]{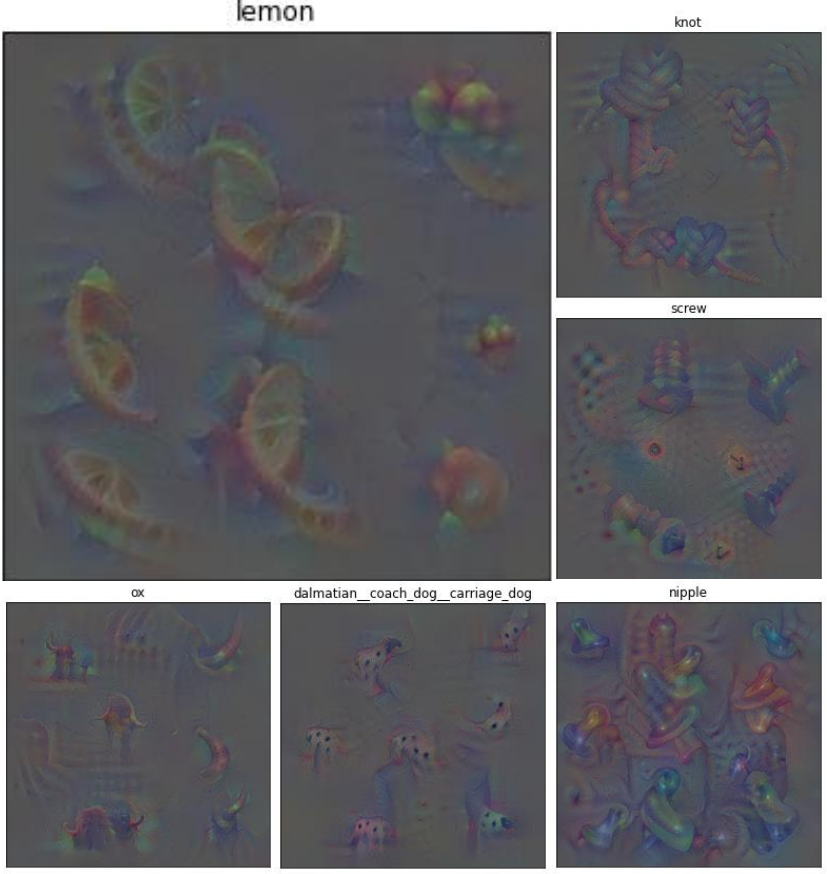}
\caption{\textbf{Exaggerating image size and number of iterations}. These visualizations are obtained from Inception-V3 with $768\times768$ input image and 10000 \acrshort{label_AM} iterations. The large image size of the input ($\times 3.5$ w.r.t original) results in a \textit{compositional} feature visualization, where several specific features (from different angles and scales) of the object category are present. The high number of \acrshort{label_AM} iterations yields a richer image in terms of detail and color shades. This image represents different categories: lemon (whose panel has been enlarged for visualization purposes), marine knot, metal screw, ox, dalmatian dog, and rubber nipple for babies.}
\label{fig_exaggeratingV}
\end{figure}

Therefore, according to our computational resources, we could increase the input image size by a factor of $\times3.5$ and the number of iterations by a factor of $\times5$. Figure \ref{fig_exaggeratingV} shows the results. 
The obtained output reveals a new, interesting scenario where colorful and detail-rich objects are represented \textit{compositionally} on a gray, average background. It seems that forcing a higher input size makes the \acrshort{label_AM} process generate an image where salient patterns from different angles and scales co-exist. This empowers us to get an understanding of the panel of possible features that activate the said neuron. For instance:
\begin{itemize}
    \item The \textit{lemon} neuron produces an AM with recognizable lemon objects, viewed both from far-away or in a close-up, and with both the lemon as a full-fruit and as a cut-out slice
    \item The \textit{screw} neuron activates with both small and large screws, seen from above (head only), oblique to the left and oblique to the right
    \item The \textit{ox} class produces an AM with close-ups on the horns, the hair, and the head of the animal.
\end{itemize}

\subsection{Exposing potential dataset biases and shortcut features}
\label{sec:XAI_wyn_ExpNatImg_bias}
Visualizations such as those provided in all the previous figures have an intrinsic potential for exposing biases within the dataset at hand and revealing shortcut features. These shortcut, easy-to-learn features might spuriously correlate with the target class label and, therefore, be leveraged by the classifier instead of the true, causal features that determine the object category. This problem is known as \textbf{shortcut learning}, and such spurious correlations that differ from true causation are the leading cause of the poor domain generalization abilities of traditional \acrshort{label_ML}/\acrshort{label_DL} systems when deployed in \acrshort{label_OOD} scenarios (ref. Section \ref{sec:back_causality_why}).

The field of shortcut learning is also intertwined with that of \textbf{feature disentanglement}. As we know, the \acrshort{label_CNN} channels learn new features as the training progresses, and such features become more and more abstract with the depth of the layer. At the final fully-connected layer, i.e., the classifier, we expect the network to have encoded certain concepts representing the classes to distinguish. Here lies the big hypothesis: units of a neural network (e.g., convolutional channels or classification neurons) learn disentangled concepts. Disentangled features mean that individual network units detect specific real-world concepts. For instance, convolutional channel 361 detects zebras, channel 127 cat snouts, channel 11 stripes at 30-degree angle, etc. Unfortunately, \acrshort{label_CNN}s are not perfectly disentangled, and several channels would contribute to recognizing a zebra.

\begin{figure}
\centering
\includegraphics[width=\textwidth]{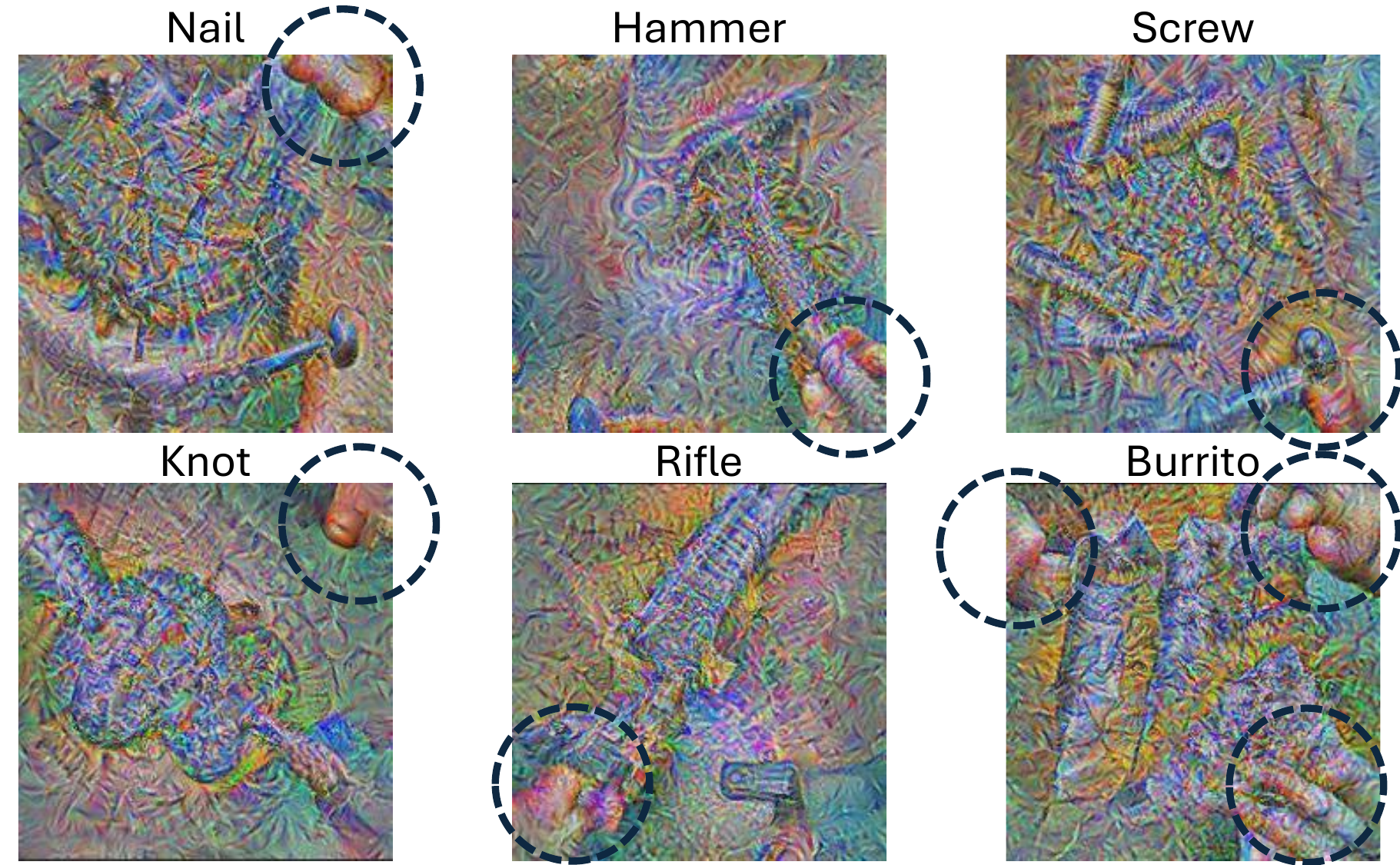}
\caption{\textbf{AM exposes dataset biases and shortcuts} (part 1). These visualizations, obtained from an Inception-V3 model with 1000 AM iterations, are exemplars of the spurious features that can be associated with the object category due to dataset bias. The AM optimization for these classes produced images that present human traits, such as fingers, hands, and arms (dashed circles), that typically co-exist in photos of human-handled objects.}
\label{fig_handsFingersArms}
\end{figure}

\begin{figure}
\centering
\includegraphics[width=\textwidth]{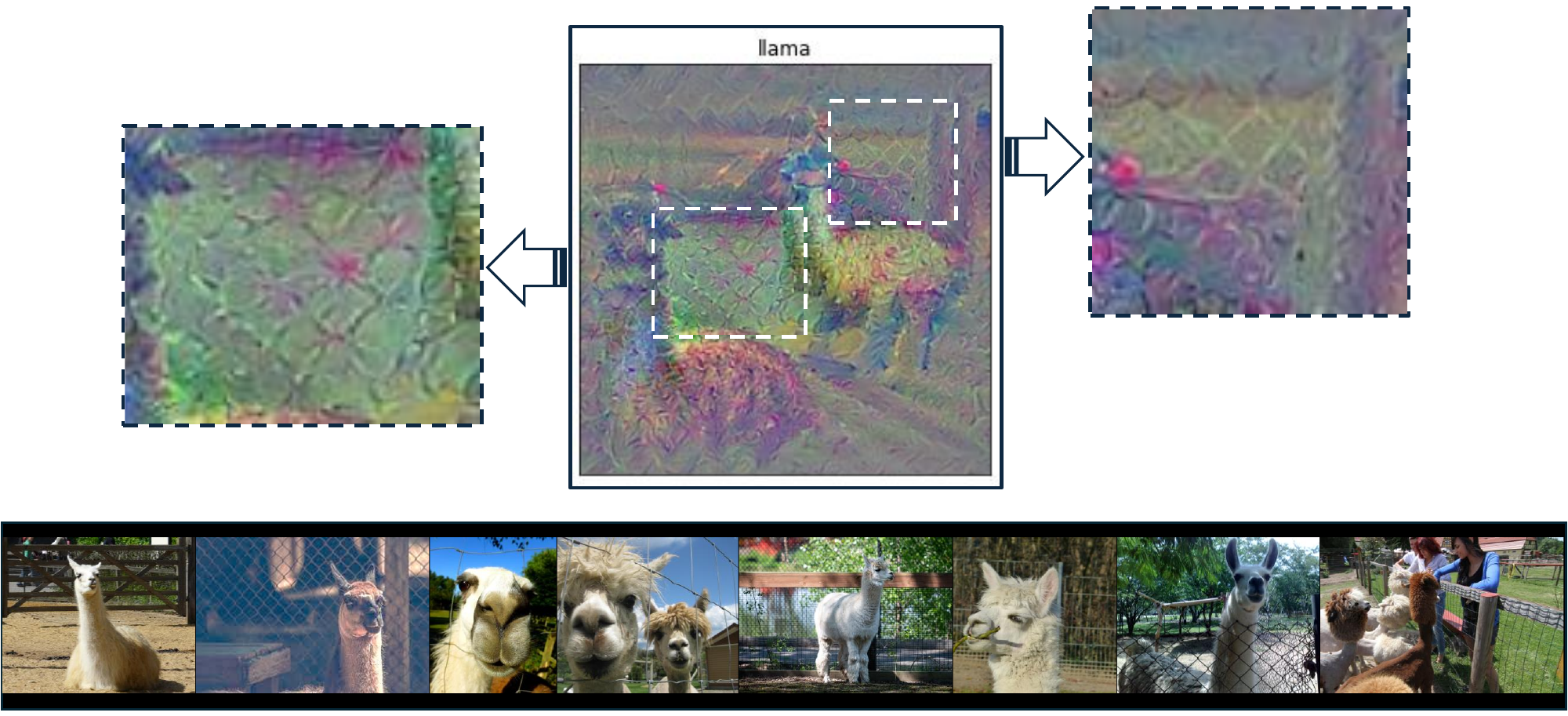}
\caption{\textbf{AM exposes dataset biases and shortcuts} (part 2). This visualization is another example of spurious features associated with the object category due to bias. The AM optimization for the \textit{'llama'} class produced feature visualizations showing structures of the diamond-shaped metal fences that typically co-exist in photos of llamas and alpacas (see some examples at the bottom).}
\label{fig_llama}
\end{figure}

Although \acrshort{label_AM} visualizations cannot prove that a unit has learned a certain, disentangled concept, they might shed some light on potential biases and shortcuts at the training dataset level. As illustrative examples, we computed the feature visualizations of peculiar object categories such as:
\begin{itemize}
    \item \textit{Nail}, \textit{hammer}, \textit{screw} and \textit{knot} --- may present features and patterns related to human hands, fingers and arms, that typically co-exist in photos of human-handled or human-grasped objects. This is also true for some food-related items, such as \textit{burrito}, or other human-actioned ones like \textit{rifle}. See Figure \ref{fig_handsFingersArms}. Other examples include the AM of class \textit{plate} containing representations of forks, knives, and glasses that typically co-exist on a dining table with the plates.
    \item \textit{Llama} and \textit{zebra} --- also show clear patterns and structures of diamond-shaped metal fences, vertical and horizontal slats of wooden fences, etc., that typically characterize photos of animals at the zoo. See Figure \ref{fig_llama}.
\end{itemize}
Those pieces of evidence show the difficulty of disentangling concepts that truly cause the label (e.g., shapes, sizes, appearance of carpentry parts or animal features) and concepts that are spuriously associated with the label (e.g., hands or fences). In such scenarios, common sense and general familiarity with the domain help us interpret within an \acrshort{label_AM} visualization what is class-related and what is a spurious concept. Below in this Chapter, we shall see that this hypothesis does not hold when dealing with more specialized domains, such as medical imaging.

\subsection{Stochastic Variability in Activation Maximization}
\label{sec:XAI_wyn_ExpNatImg_variability}
As with any data-driven, iterative process that entails computing gradient backpropagations, we observed an intrinsic variability in the output of the \acrshort{label_AM} learning procedure, even after setting a fixed seed for all the random processes and scripts. Instead of an issue, we think of this variability as a feature to exploit.

\begin{figure}
\centering
\includegraphics[width=\textwidth]{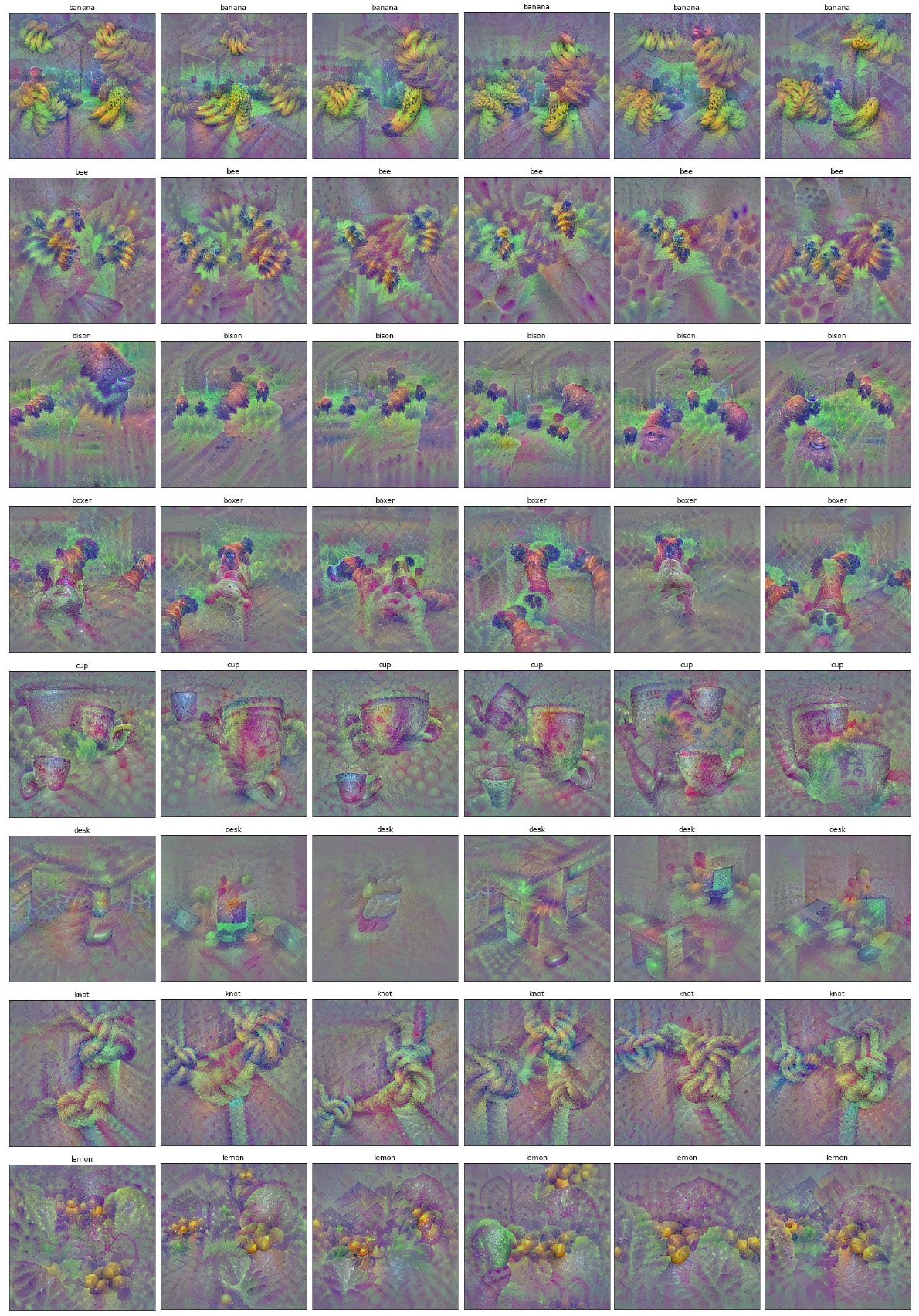}
\caption{Feature visualizations obtained by repeating the optimization process six times (columns) for each image class (rows) with the same learning settings.}
\label{fig_sixRuns1}
\end{figure}

Indeed, we argue that exposing a pre-trained model to different optimization processes (thus yielding similar yet different \acrshort{label_AM} images) is beneficial to visually explore the manifold of representations a neuron has learned when recognizing a specific class.
Therefore, we performed six parallel optimization processes starting from the same initial noise image and fixing all the \acrshort{label_AM} parameters. We present the \acrshort{label_AM} outputs for eight of the investigated classes in Figure \ref{fig_sixRuns1}. 

\section{Does Activation Maximization Work For Medical Imaging?}
\label{sec:XAI_wyn_MedicalImages}
Previous sections have revolved around pre-trained \acrshort{label_CNN}s for natural image classification, and the results have shown \acrshort{label_AM} is an easy-to-implement and insightful tool that can work well to understand the representations learned by an output neuron and potentially identifying bias sources.
Nevertheless, today, \acrshort{label_AI} largely permeates other domains, such as healthcare, with great applications to medical image understanding and analysis (ref. Section \ref{sec:back_MIA}). Aligned with the application domain of this thesis, we thus investigated if and to what extent applying AM techniques to \acrshort{label_CNN}-based medical image classifiers could bring benefits.

\subsection{Model Architecture and Training}
The first step was to train a network on medical data. Since optimizing the training scheme and proposing novel frameworks for improved classification is beyond the scope of this investigation, we resorted to regular architecture and training practices. We selected the popular VGG16 model and modified the first layer to accept a single channel input (i.e., grayscale medical image) and the last layer to have two output neurons. To properly assess the impact on \acrshort{label_AM}, we conceived two different training settings for the model: (i) \textbf{Transfer Learning} - the backbone is used as a "frozen" feature extractor with ImageNet pre-trained weights and only the last fully connected layer is trained; (ii) \textbf{Fine Tuning} - the weights are initialized to the pre-trained values but the whole network is retrained end-to-end. We trained the networks with early stopping with a standard \acrshort{label_Adam} optimizer with momentum $0.9$, a fixed learning rate of $5e-4$, and a batch size of $32$ tensors. As the evaluation metric, we considered the accuracy obtained on the test set.

\subsection{Medical Dataset}
As for the data, we selected the ChestX-ray14 dataset \cite{wang2017chestx}, a popular and freely available dataset of frontal chest X-ray (\acrshort{label_CXR}) images associated with labels describing the presence of up to 14 common thorax diseases (i.e., 13 radiological findings plus one label for \textit{no finding}). Since we were only interested in the \acrshort{label_AM} of classification neurons, we did not conceive a separate test set to assess generalization, and we used the official splits of training ($86524$) and validation ($25596$) images instead. We resized the input images to a consistent $448$x$448$ grid and adjusted their contrast in the $0$-$255$ range.

\subsection{Classification Tasks}
We investigated two types of classification tasks. The first aimed at predicting a medical condition and involved classifying input images as containing \textbf{finding} or \textbf{no-finding}, for which we aggregated the 13 possible disease labels. This could reflect the radiologist's need to rapidly screen thousands of \acrshort{label_CXR}s to spot those potentially needing further evaluation, thus increasing prioritization and efficiency.  
The second task, instead, aimed at uncovering potentially harmful biases and shortcuts from the data. By drawing on the (anonymized) metadata associated with patient scans, we have trained the networks on the trivial task of classifying \textbf{male} versus \textbf{female} subjects. 

\subsection{Accuracy Rises, Interpretability Falls}
\begin{figure}
\centering
\includegraphics[width=\textwidth]{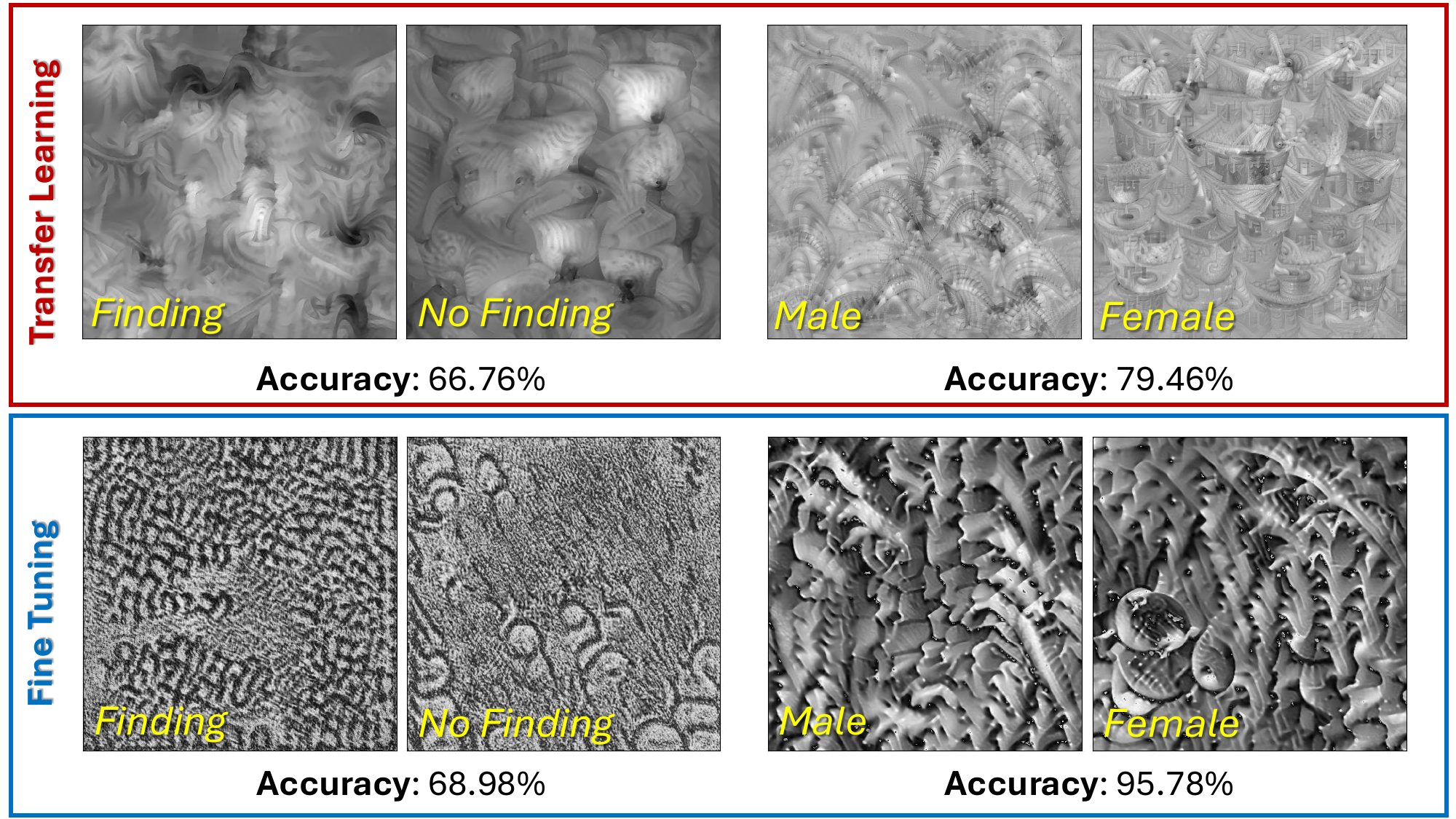}
\caption{Feature visualizations of output neurons on VGG16 models for radiology finding and patient gender classification from chest X-ray images. The top half of the figure refers to a \acrshort{label_TL} regime, where only the last fully connected layer is trained and the rest of the networks is a pretrained and frozen feature extractor. The bottom half of the figure refers to a \acrshort{label_FT} setting, where all the network is trained end-to-end. For each classification task, the accuracy of the corresponding classifier is reported together with the \acrshort{label_AM} images of its neurons. As the training passes from \acrshort{label_TL} to \acrshort{label_FT} regime, the classifiers perform moderately to significantly better in accuracy, but their \acrshort{label_AM} becomes more abstract and hard to interpret.}
\label{fig_wyn_medical}
\end{figure}

Our results in Figure \ref{fig_wyn_medical} show that the network performed moderately well on classifying \acrshort{label_CXR} images for the presence of thorax finding and much better for the male-vs-female prediction.
Specifically, when only the last fully connected layer was trained (i.e., the \acrshort{label_TL} setting), the best accuracy values we obtained were $66.76$\% (finding vs no finding) and $79.79$\% (male vs female). Those metrics increased to $68.98$\% and $95.81$\%, respectively, when the whole network was re-trained end to end (i.e., \acrshort{label_FT}).
Once obtained the best-performing models, we ran our feature visualization pipeline and obtained the \acrshort{label_AM}s for each class of interest in both tasks.

Due to, for instance, privacy concerns, we would not want an \acrshort{label_AI} to be able to trace the subject's gender from an anonymized medical image. However, evidence in the literature and our results point to this unfortunate direction. As argued by many and presented in Section \ref{sec:back_causality_why}, this is probably due to the models suffering from shortcut learning. We thus explored the \acrshort{label_AM} on the gender classification neurons to uncover which aspects of the image could potentially lead to gender recognition.
However, our findings (Figure \ref{fig_wyn_medical}) show that the learned representations are generally much more abstract and uninterpretable than those for natural images despite the networks' moderate to high classification performance.
Such visualizations impair our ability to spot potentially wrong, biased, and spurious aspects of the learned concepts.

Actually, Figure \ref{fig_wyn_medical} tells more.
The fact that \acrshort{label_AM}s on the final neurons gradually worsen and become more abstract when moving from "frozen" transfer learning (Figure \ref{fig_wyn_medical} first row) to end-to-end fine-tuning (Figure \ref{fig_wyn_medical} second row) supports our following two considerations.

First, unlike natural images, medical images are often grayscale (e.g., X-ray, \acrshort{label_CT}, diagnostic \acrshort{label_MRI}) and limited in number (e.g., due to low disease prevalence or privacy limitations). Naturally, this reduces the amount of information contained in the inputs, and thus learnable by the networks. 

\begin{figure}
\centering
\includegraphics[width=0.75\textwidth]{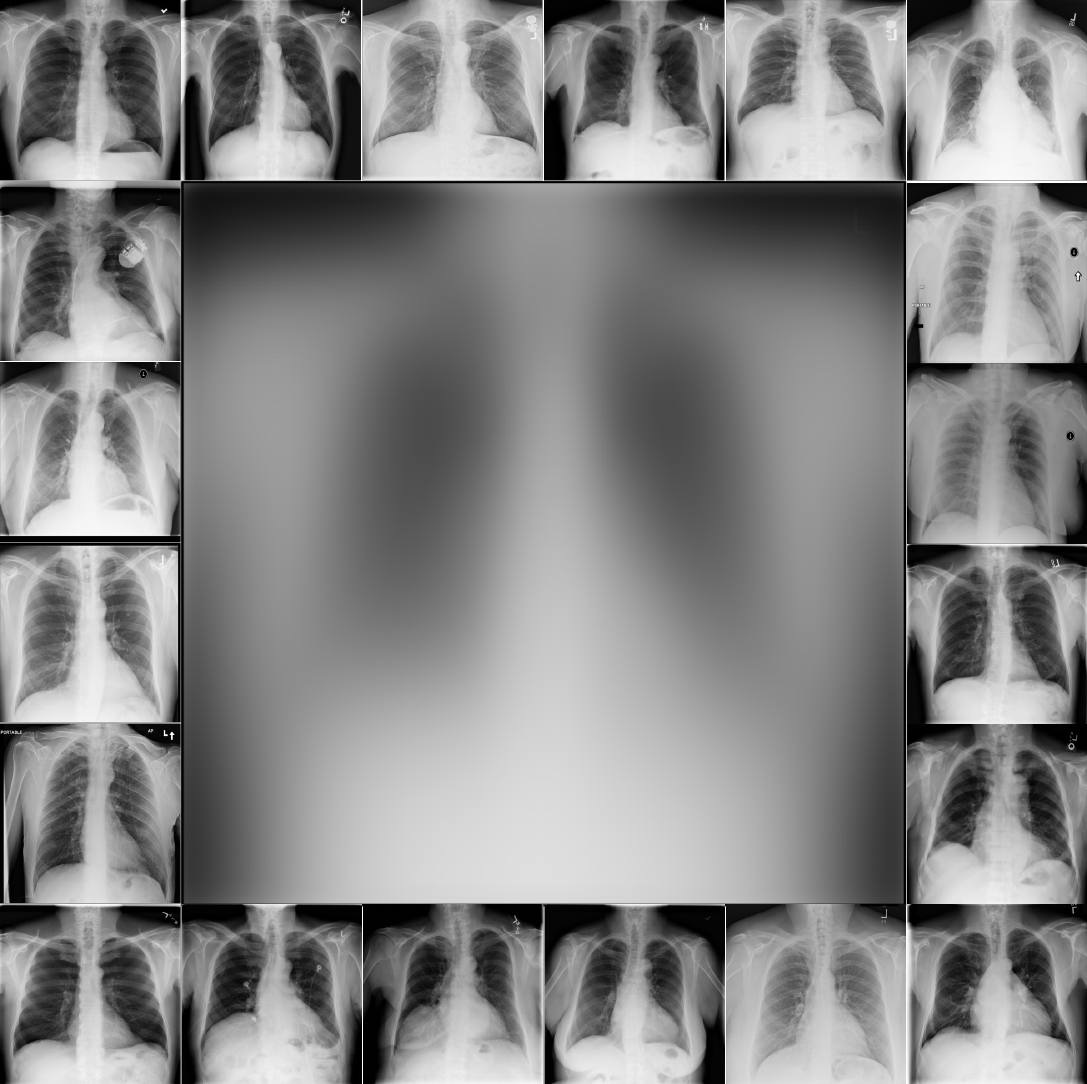}
\caption{The average chest X-ray (CXR) image and some CXRs from the training set of the NIH's ChestXRay14 dataset. The average image is obtained by computing the mean pixel intensities over more than $86000$ image samples, and this visualization helps understand how practically all \acrshort{label_CXR} images look, on average, alike. 
This structural regularity is one of the reasons we can still achieve reasonable performance with scarce data in medical \acrshort{label_AI}. On the other hand, this regularity implies that the network must learn more high-level and abstract features to tell different medical conditions apart.}
\label{fig_avgImage_CXR}
\end{figure}

Second, natural images' shape, layout, color, and appearance are highly variable. Photographs of people, animals, objects, buildings, and places can reveal many patterns, compositions, and magnifications that make the represented scene mostly inconsistent, even among image samples from the same class. Conversely, medical imaging is characterized by a high level of structure consistency due to anatomical regularities and radiological protocols that are followed when acquiring the scans. As a confirmation example, we computed the "average" appearance of a \acrshort{label_CXR} by averaging the over $86000$ image samples of the NIH's ChestXRay14 training dataset (Figure \ref{fig_avgImage_CXR}). As can be seen, despite the variability in patients' gender, demographics, and positioning in the scanner, practically all \acrshort{label_CXR}s are alike - a vertical structure at the center of the image (i.e., the spine and mediastinum), two darker zones symmetrical and opposite the vertical structure (i.e., the lungs), horizontal structures at the top of the image above the lungs (i.e., the clavicles and neck bones), a broad bright region at the bottom of the image (i.e., the abdomen), and a dark background of the figure (i.e., air).

\begin{figure}
\centering
\includegraphics[width=\textwidth]{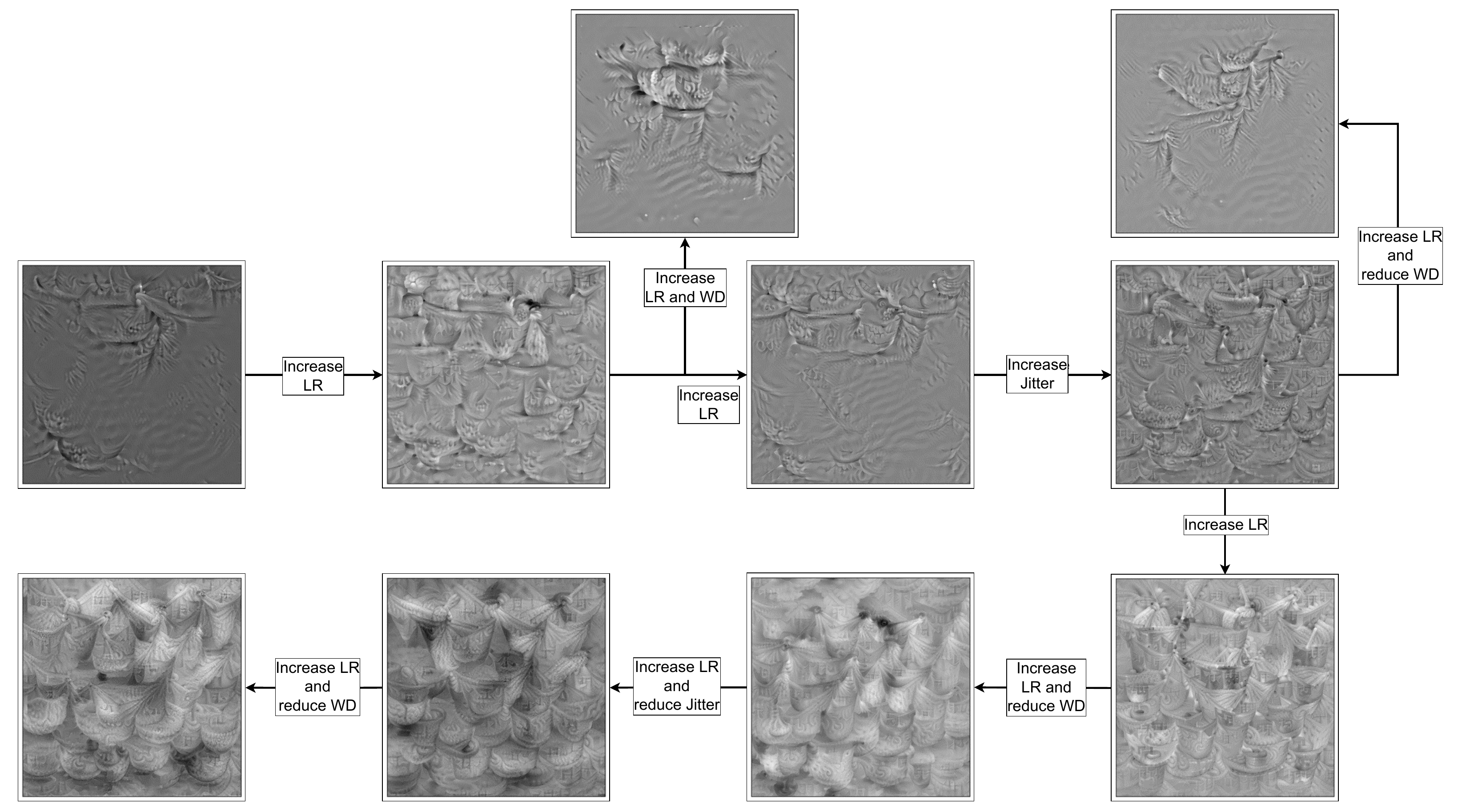}
\caption{Exploring the space of AM hyperparameters is challenging for medical images. The output greatly varies with the chosen values of \acrshort{label_LR}, \acrshort{label_WD}, and jittering and is hardly interpretable to humans. The example shown represents the AM images of the \textit{female} neuron of the VGG16 model trained in the \acrshort{label_TL} regime.}
\label{fig:wyn_medical_variability}
\end{figure}

On the one hand, this regularity in the general structure of images is probably one of the reasons we can still achieve reasonable performance with scarce data in medical \acrshort{label_AI}. However, on the other hand, this regularity is such that the features the network must learn to discern, for instance, \textit{pneumonia} from \textit{edema}, are much more specific, subtle, and consequently abstract than those distinguishing an airplane from a cat.
Moreover, the produced \acrshort{label_AM}s for these medical image cases fluctuate more depending on the \acrshort{label_AM} hyperparameters than the natural images case (see Figure \ref{fig:wyn_medical_variability}). 

\subsection{Exploiting Medical Image Regularities}
Inspired by the considerations above, we tried to find a way to leverage the regularities found in medical images and conceive new priors to guide the learning phase of \acrshort{label_AM}. At this stage, we acted at the loss function level.

Specifically, the total loss function used in the following experiments is a weighted sum of several loss components, designed to enforce various priors and constraints on the generated images. The total loss is given by:

\begin{equation}
    \mathcal{L} = \mathcal{L}_C +
    \lambda_{1}\cdot \mathcal{L}_H +
    \lambda_{2}\cdot \mathcal{L}_N +
    \lambda_{3}\cdot \mathcal{L}_S +
    \lambda_{4}\cdot \mathcal{L}_F
\end{equation}

Each component of the loss function serves a specific purpose:
\begin{itemize}
    \item Classification Loss ($\mathcal{L}_C$) - This term encourages the generated image to strongly activate a specific neuron (or class) in the trained CNN. It is defined as the negative activation of the target neuron.
    \item Symmetry Loss ($\mathcal{L}_S$) - This term enforces symmetry in the generated image by penalizing differences between the left and right halves. The right half is flipped horizontally, and the \acrshort{label_MSE} between the left half and the flipped right half is calculated. This encourages the image to be symmetrical.  This is particularly useful for medical images like \acrshort{label_CXR}s, where symmetry is expected (Figure \ref{fig_avgImage_CXR}).
    \begin{equation}
    \mathcal{L}_S = \left( I_{\text{left}} - \text{flip}(I_{\text{right}}) \right)^2
    \end{equation}
    \item Histogram Loss ($\mathcal{L}_H$): This term encourages the pixel intensity distribution of the generated image to match a predefined target distribution.  In this case, we use the mixture of Gaussians (GMM) defined over the average \acrshort{label_CXR} image as the target. The histogram of the image is computed and normalized. The target distribution is also computed and normalized. The symmetric \acrshort{label_KL} divergence between the image histogram and the target distribution is then calculated. This helps in achieving the desired visual appearance.
    \begin{equation}
        \mathcal{L}_{H}(I) = 0.5 \left( \text{KL} \left( \log(\text{hist(I)}), \text{GMM}  \right) + \text{KL} \left( \log(\text{GMM}), \text{hist(I)} \right) \right)
    \end{equation}  
    \item Noise Characteristics Loss ($\mathcal{L}_N$): This loss simulates quantum mottle, a type of noise that follows a Poisson distribution and is typical of radiography images. It does this by encouraging the local mean and variance of the image to be equal. The local mean and variance are computed using average pooling, and the \acrshort{label_MSE} between them is calculated. We selected a kernel size of $5$, stride of $1$, and padding of $2$ for the pooling operation. 
    \begin{equation}
    \mathcal{L}_{N}(I) = \left(\text{AvgPool2D}(I) - \left( \text{AvgPool2D}(I^2) - {\text{AvgPool2D}(I)}^2 \right) \right)^2
    \end{equation}
    
    \item Frequency Spectrum Loss ($\mathcal{L}_F$): This term enforces the generated image to have a similar frequency spectrum to a given average spectrum. We utilize the Fast Fourier Transform (\acrshort{label_FFT}) and employ the spectrum of the average \acrshort{label_CXR} image as the target. This helps in maintaining the overall features and structure of the image.
    \begin{equation}
    \mathcal{L}_{F}(I, \bar{I}) = \frac{1}{N} \sum_{i=1}^{N} \left( \left| \text{FFT}(I) \right| - \left| \text{FFT}(\bar{I}) \right| \right)^2
    \end{equation}
\end{itemize}

We tried to tune the hyperparameters of such new regularizing loss terms. Finding an optimal equilibrium between those regularizations proved to be hard, and we generally did not obtain more interpretable outputs than before. Taken together, those pitfalls undermine the applicability of simple AM techniques on deep \acrshort{label_CNN}s that have been fine-tuned for a medical imaging task, and call for more advanced methods.

\section{Summary}
\label{sec:xai_wyn_summary}
In this Chapter, we have first defined the problem of \acrshort{label_AM}, formalized the gradient ascent optimization, and discussed the need for image priors to narrow down the solutions to the optimization problem.
Then, we investigated \acrshort{label_AM} applied to natural images and found it works well with popular networks pre-trained on large datasets. Also, we revealed the previously unobserved effect of exaggerating input size and number of iterations. The output becomes a compositional representation of colorful and detail-rich objects, representing patterns from different angles and scales.

Next, we studied how \acrshort{label_AM} methods can determine possible sources of bias in image datasets and demonstrated the inability to disentangle concepts that indeed cause the label (e.g., appearance of carpentry parts or animal features) and concepts that are spuriously associated with the label (e.g., hands or fences).

Finally, the last part of the Chapter investigates whether \acrshort{label_AM} could work on medical images. Our findings reveal how difficult it is to make sense of medical image classifiers via simple \acrshort{label_AM}. We proposed novel ways to inject prior knowledge about medical images and conceived loss priors, even though we could not achieve a sufficient level of interpretation to the best of our methodological and optimization efforts.

Those significant results made us realize the need for more complex and powerful explanation methods. In the next Chapter, we show how we adopted an explainable-by-design method called \acrshort{label_ProtoPNet}, validated it on mammogram images, and involved a clinical expert to rate the radiology alignment of the produced explanations.
\chapter{ProtoPNet for Breast Mass Classification: Explainable-by-Design Network via Prototypical Part Learning}
\label{chap:XAI_protopnet}
\begin{figure}[h!]
\includegraphics[width=\textwidth]{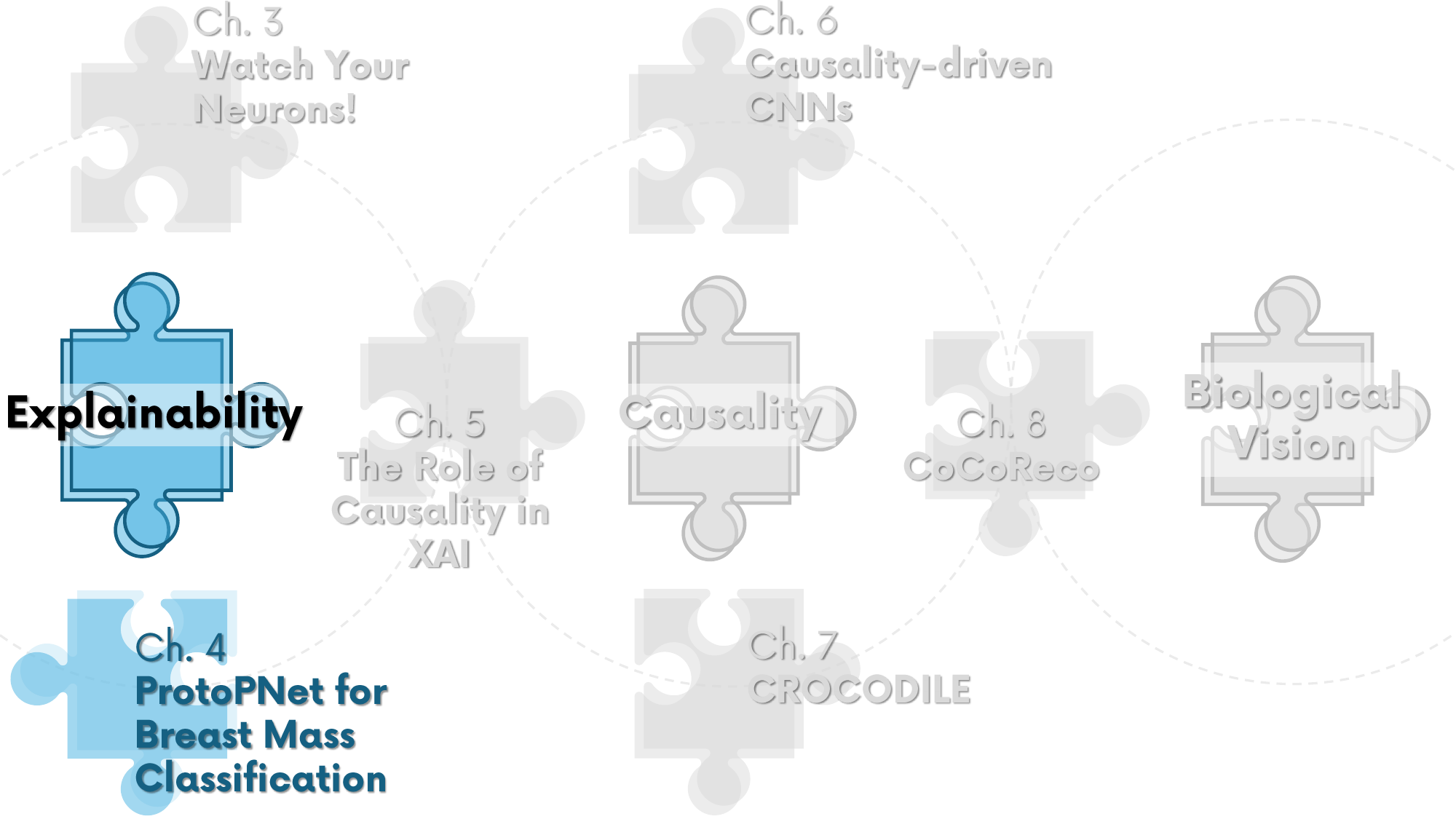}
\end{figure}

Presented with the difficulties of using simple \acrshort{label_AM} techniques on deep \acrshort{label_CNN}s fine-tuned for a medical imaging task (Chapter \ref{chap:XAI_wyn}), we take a step forward in this Chapter and investigate \textit{ante}-hoc \acrshort{label_XAI}. 
We develop an explainable-by-design model based on prototypical part learning for breast mass classification from mammogram images.
As we shall see, this method not only improves the model's classification performance but produces robust and compelling explanations for a radiologist. 

Section \ref{sec:intro} introduces the context and motivation of the study, highlighting the challenges posed by medical images and outlining the research questions.
In Section \ref{sec:related_works}, relevant literature is discussed, reviewing works that automatically classify breast masses from mammograms and the few ones that had explored applying ProtoPNet to medical images at the time of the study.
Then, in Section \ref{sec:materials_and_methods}, we describe the medical dataset we utilized, the functioning of the ProtoPNet, and the modifications we implemented. Also, the experimental settings, data splitting, hyperparameter tuning, and evaluation metrics are reported.
After presenting our results in Section \ref{sec:results_protopnet}, we dive into a crucial aspect of our investigation. Indeed, Section \ref{sec:clinical_viewpoint} illustrates how we involved an experienced radiologist to provide a clinical viewpoint on the quality of the learned prototypes, the patch activations, and the global explanations.
Finally, Section \ref{sec:discussion_protopnet} discusses our main findings and presents ideas for future work before pulling the threads in Section \ref{sec:conclusion}.
The content of this Chapter is based on the following
\begin{itemize}
    \item publication: Carloni, G., Berti, A., Iacconi, C., Pascali, M. A., \& Colantonio, S. (2022, August). "On the applicability of prototypical part learning in medical images: Breast masses classification using protopnet". In \textit{International Conference on Pattern Recognition (pp. 539-557). Cham: Springer Nature Switzerland.} \cite{carloni2022applicability}
    
    \item and code: \url{https://github.com/andreaberti11235/ProtoPNet}.
\end{itemize} 

\section{Context, Motivation, and Research Question}
\label{sec:intro}
In Section \ref{sec:back_XAI_methods}, we have seen that prototypical learning is a popular \acrshort{label_DL} research line ~\cite{rudin2022interpretable,li2018deep}. The key feature of this class of learning algorithms is to compare one \textit{whole} image to another \textit{whole} image.
Instead, one could wish to understand what are the relevant parts of the input image that led to a specific class prediction. In other words, \textit{parts} of observations could be compared to \textit{parts} of other observations.
In the attempt to build a \acrshort{label_DL} model that resembles this kind of logic, Chen et al.~\cite{chen2019looks} proposed the \acrshort{label_ProtoPNet}, as we have seen in Section \ref{sec:back_XAI_methods_prototypes}.

\acrshort{label_ProtoPNet} posed brilliant promises in classification domains regarding natural images (e.g., birds and cars classification~\cite{chen2019looks}, video deep-fake detection~\cite{trinh2021interpretable}).
On the other hand, the applicability of this type of reasoning to medical images was still in its infancy at the time of this study.
When presented with a new case, radiologists compare the images with previously experienced ones. They recall visual features that are specific to a particular disease, recognize them in the image at hand, and provide a diagnosis. For this reason, medical imaging seems to be suitable for prototypical part-based explanations.
Nevertheless, some critical issues can arise when bringing technologies from other domains – like computer vision - into the medical world. Unlike natural images, usually characterized by three channels (e.g., RGB, CYM), conventional medical images feature single-channel gray scales. For this reason, pixels contain a lower amount of information.
Furthermore, X-ray images represent a body's projection and, therefore, are flat and bi-dimensional. As a result, objects in the field of view could not be as separable and distinguishable as in real-world natural images. Such issues might be detrimental to the application of these methodologies.
In addition, the scarcity of labeled examples available for supervised training undermines the generalization capability achievable by complex models. This lack of labeled data is mainly due to the low prevalence of certain diseases, the time required for labeling, and privacy issues. Moreover, additional problems include the anatomical variability across patients and the image quality variability across different imaging scanners.

This work aimed to investigate the applicability of \acrshort{label_ProtoPNet} in mammogram images for the automatic and explainable malignancy classification of breast masses. The assessment of applicability was based on two aspects: the ability of the model to face the task (i.e., classification metrics) and the ability of the model to provide end-users with plausible explanations.
The novelty of this work stems from both the application of \acrshort{label_ProtoPNet} to the classification of breast masses without fine-annotated images and the clinical viewpoint provided for \acrshort{label_ProtoPNet}'s explanations.

\section{Related Works}
\label{sec:related_works}
Several works in the literature have applied \acrshort{label_DL} algorithms, and \acrshort{label_CNN}s in particular, to automatically classify benign/malignant breast masses from X-ray mammogram images. By contrast, at the time of this study, only a few works had explored the applicability of ProtoPNet to the medical domain and, more specifically, to breast mass classification. 

Concerning the use of \acrshort{label_CNN}s for this task on the Curated Breast Imaging Subset of DDSM (CBIS-DDSM)~\cite{lee2017curated} dataset some works follow. 
Tsochatzidis et al.~\cite{tsochatzidis2019deep} explored various popular \acrshort{label_CNN} architectures, by using both randomly initialized weights and pre-trained weights from ImageNet. With ResNet50 and pre-trained weights they obtained an accuracy of $0.749$.
Alkhaleefah et al.~\cite{alkhaleefah2020influence} investigated the influence of data augmentation techniques on classification performance. When using ResNet50, they achieved $0.676$ and $0.802$ before and after augmentation, respectively.
Arora et al.~\cite{arora2020deep} proposed a two-stage classification system. First, they exploited an ensemble of five CNN models to extract features from breast mass images and then concatenated the five feature vectors into a single one. In the second stage, they trained a two-layered feed-forward network to classify mammogram images. With this approach, they achieved an accuracy of $0.880$. They also reported the performance obtained with each individual sub-architecture of the ensemble, achieving an accuracy of $0.780$ with ResNet18. 
Ragab et al.~\cite{ragab2021framework} also experimented with multiple \acrshort{label_CNN} models to classify mass images. Among the experiments, they obtained an accuracy of $0.722$, $0.711$ and $0.715$ when applying ResNet18, ResNet50 and ResNet101, respectively.
Finally, Ansar et al.~\cite{ansar2020breast} introduced a novel architecture based on MobileNet and transfer learning to classify mass images. They benchmarked their model with other popular networks, among which ResNet50 led to an accuracy of $0.637$.

Regarding the application of \acrshort{label_ProtoPNet} to the medical domain, only a few attempts have investigated it by early 2022.
Mohammadjafari et al. \cite{mohammadjafari2021using} applied ProtoPNet to Alzheimer's Disease detection on brain magnetic resonance images from two publicly available datasets. As a result, they found an accuracy of $0.91$ with ProtoPNet, which is comparable to or marginally worse than that obtained with state-of-the-art black-box models.
Singh et al.~\cite{singh2021these,singh2021interpretable} proposed two works utilizing ProtoPNet on \acrshort{label_CXR} images of Covid-19 patients, pneumonia patients, and healthy people for Covid-19 identification. In~\cite{singh2021these}, they slightly modified the weight initialization in the model to emphasize the effect of differences between image parts and prototypes in the classification process, achieving an accuracy of $0.89$. In~\cite{singh2021interpretable}, they modified the metrics used in the model's classification process to select prototypes of varying dimensions and obtained the best accuracy of $0.87$.

To the best of our knowledge, the only application of prototypical part learning to the classification of benign/malignant masses in mammogram images at the time of the study was provided by Barnett et al.~\cite{barnett2021case}. They introduced a new model, IAIA-BL, derived from ProtoPNet, utilizing a private dataset with further annotations by experts in training data. They included both pixel-wise masks to consider clinically significant regions in images and mass margin characteristics (spiculated, circumscribed, microlobulated, obscured, and indistinct).
On the one hand, annotation masks of clinically significant regions were exploited at training time in conjunction with a modified loss function to penalize prototype activations on medically irrelevant areas.
On the other hand, they employed annotations of mass margins as an additional label for each image and divided the inference process into two phases: first, the model determines the mass margin feature and then predicts malignancy based on that information. For this purpose, they added a fully-connected (\acrshort{label_FC}) layer to convert the mass margin score to the malignancy score. With that architecture, they managed to achieve an \acrshort{label_AUROC} of $0.84$.

A search on bibliometric databases almost three years later reveals that prototypical part learning and \acrshort{label_ProtoPNet} have continued to evolve in medical imaging applications.
Examples include MAProtoNet, a multi-scale ProtoPNet for 3D \acrshort{label_MRI} brain tumor classification \cite{li2024maprotonet}, ProtoAL \cite{santos2024protoal}, prototype learning for explainable brain age prediction \cite{hesse2024prototype}, and broad-spectrum reviews on applications and challenges \cite{de2024part}. Additionally, our paper was cited in other subsequent research such as \cite{nauta2023co,an2024validation,choukali2024pseudo,sinhamahapatra2024enhancing,xiong2024towards}, confirming the significance and impact of our investigation.

\section{Methodology}
\label{sec:materials_and_methods}
In this work, we trained a \acrshort{label_ProtoPNet} model to classify benign/malignant breast masses from mammogram images on a publicly available dataset. We compared its performance to the baseline model on which ProtoPNet is based. We conducted a random search independently on both models with five-fold cross-validation (\acrshort{label_CV}) to optimize the respective hyperparameters.

\subsection{Breast Mammography Dataset}
\begin{figure}[t]
\centering
\includegraphics[width=0.99\textwidth]{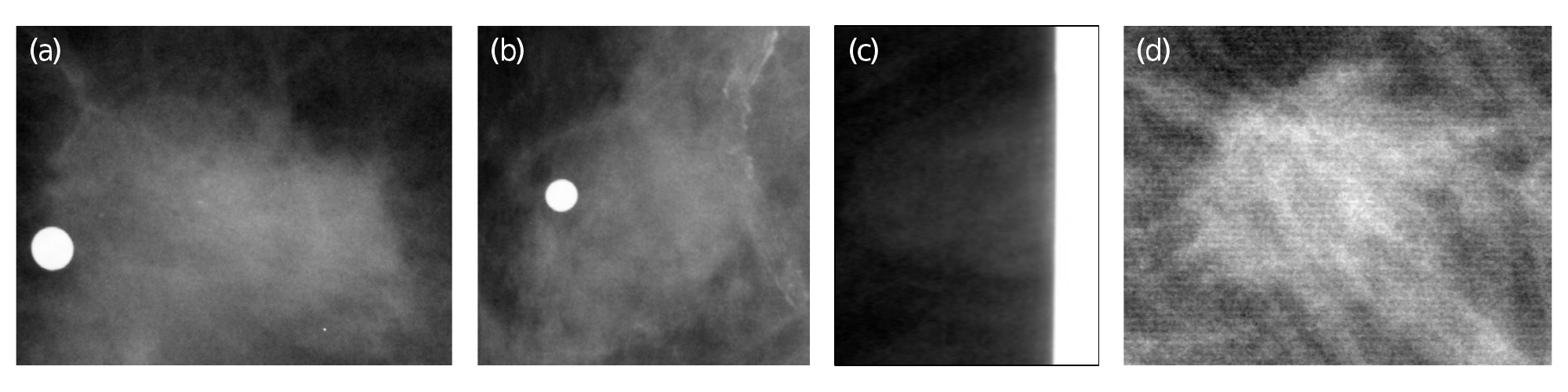}
\caption{Examples of images from the original CBIS-DDSM dataset that were removed due to artifacts. (a)-(b): annotation spot next to or within the mass; (c): white-band artifact; (d) horizontal-pattern artifact.}
\label{fig:bad_cases}
\end{figure}

In our study, we used images from CBIS-DDSM~\cite{lee2017curated}. The dataset is composed of scanned film mammography studies from $1566$ breast cases (i.e., patients). 
For each patient, two views (i.e., MLO and CC) of the full mammogram images are provided. In addition, the collection comes with the region of interest (ROI)-cropped images for each lesion. Each breast image has its annotations given by experts, including the ground truth for the type of cancer (benign, malignant, or no-callback) and the type of lesion (calcification or mass). Only the ROI-cropped images of benign and malignant masses for each patient were used in this study.
As a first step, we performed a cleaning process of the dataset by removing images with artifacts and annotation spots next to or within the mass region (Fig.~\ref{fig:bad_cases}). We then converted DICOM images of the cleaned dataset into PNG files.
The training and test split of the cohort was already provided in the data collection. To obtain a balanced dataset, we randomly selected the exceeding elements from the most numerous class and excluded them from the cohort.

\subsection{Prototypical Part Network}
\subsubsection{Architecture and functioning}
ProtoPNet comprises three main blocks: a \acrshort{label_CNN}, a prototype layer, and an \acrshort{label_FC} layer.
The \acrshort{label_CNN} block consists of a feature extractor, which can be chosen from many of the popular models competing on ImageNet challenges (VGGs, ResNets, DenseNets), and a series of add-on convolutional layers. This block extracts features from an input RGB image of size $224\times224$. Given this input size, the convolutional output has size $7\times7\times D$, where $D$ is the number of output filters of the \acrshort{label_CNN} block. \acrshort{label_ReLU} is used to activate all convolutional layers, except the last one that utilizes the sigmoid activation.
The prototype layer that follows comprises two $1\times1$ convolutional layers with \acrshort{label_ReLU} activation. It learns \emph{m} prototypes, whose shape is $1\times1\times D$. Each prototype embodies a prototypical activation pattern in one area of the convolutional output, which itself refers to a prototypical image in the original pixel space. Thus, we can say that each prototype is a latent representation of some prototypical element of an image.

At inference time, the prototype layer computes a similarity score as the inverted squared $L^2$ distance between each prototype and all patches of the convolutional output.
For each prototype, this produces an activation map of similarity score whose values quantify the presence of that prototypical part in the image. This map is up-sampled to the size of the input image and presented as an overlayed heat map highlighting the part of the input image that mostly resembles the learned prototype. The activation map for each prototype is then reduced using global max pooling to a single similarity score.
A predetermined number of prototypes represents each class in the final model. In the end, the classification is performed by multiplying the similarity score of each prototype by the weights of the \acrshort{label_FC} layer.

\subsubsection{Prototype learning process}
\begin{figure}[t]
\centering
\includegraphics[width=0.99\textwidth]{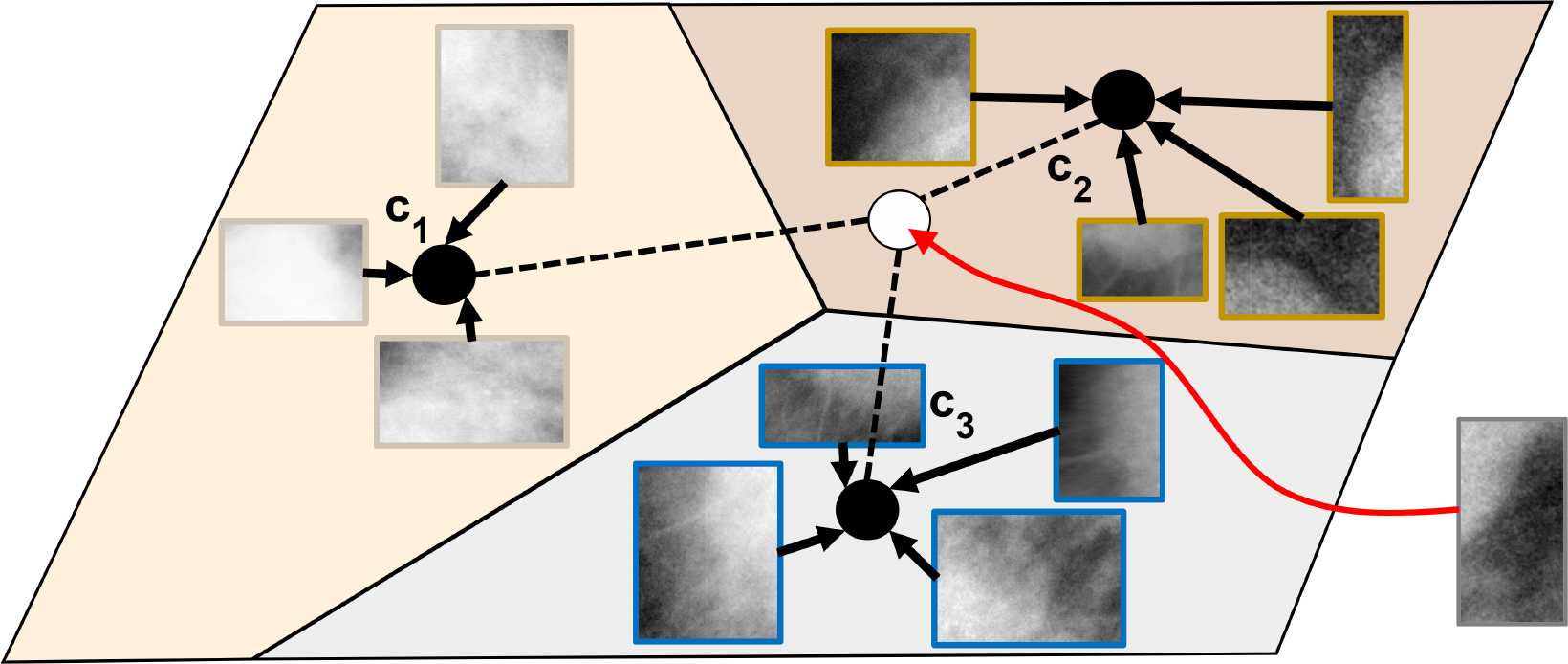}
\caption{\textbf{Learning the Prototypes}. Where do those prototypes come from? The training process of ProtoPNet consists of learning a meaningful latent representation where characteristic patches of training images become centroids in a clustering process, that is, prototypes.}
\label{fig:prototypes_schematic}
\end{figure}

The training process of \acrshort{label_ProtoPNet} consists of learning a meaningful latent representation where characteristic patches of training images (like a characteristic pattern or margin) become centroids in a clustering process, that is, prototypes (ref. Figure \ref{fig:prototypes_schematic}).

The training loss function consists of three terms: (i) the cross entropy penalizes wrong classifications; (ii) a \textit{clustering} term that encourages each training image to have some latent patch close to at least one prototype of its class; and (iii) a \textit{separation} term that encourages every latent patch of a training image to stay away from the prototypes of different classes.

In a nutshell, the learning process begins with the stochastic gradient descent of all the layers before the \acrshort{label_FC} layer (joint epochs). Then, prototypes are projected onto the closest latent representation of training images' patches. Finally, the optimization of the \acrshort{label_FC} layer is carried out. It is possible to cycle through these three stages more than once.

\subsubsection{Differences in our implementation}
Differences exist between the original paper introducing ProtoPNet \cite{chen2019looks} and our work.
Firstly and more importantly, we conceived a hold-out test set to assess the final models' performance, after the models were trained using \acrshort{label_CV}. In the original paper, instead, both the selection of the best model and the evaluation of its performance were carried out on the same set, i.e., validation and test sets were the same. 

\begin{figure}[t]
\centering
\includegraphics[width=0.99\textwidth]{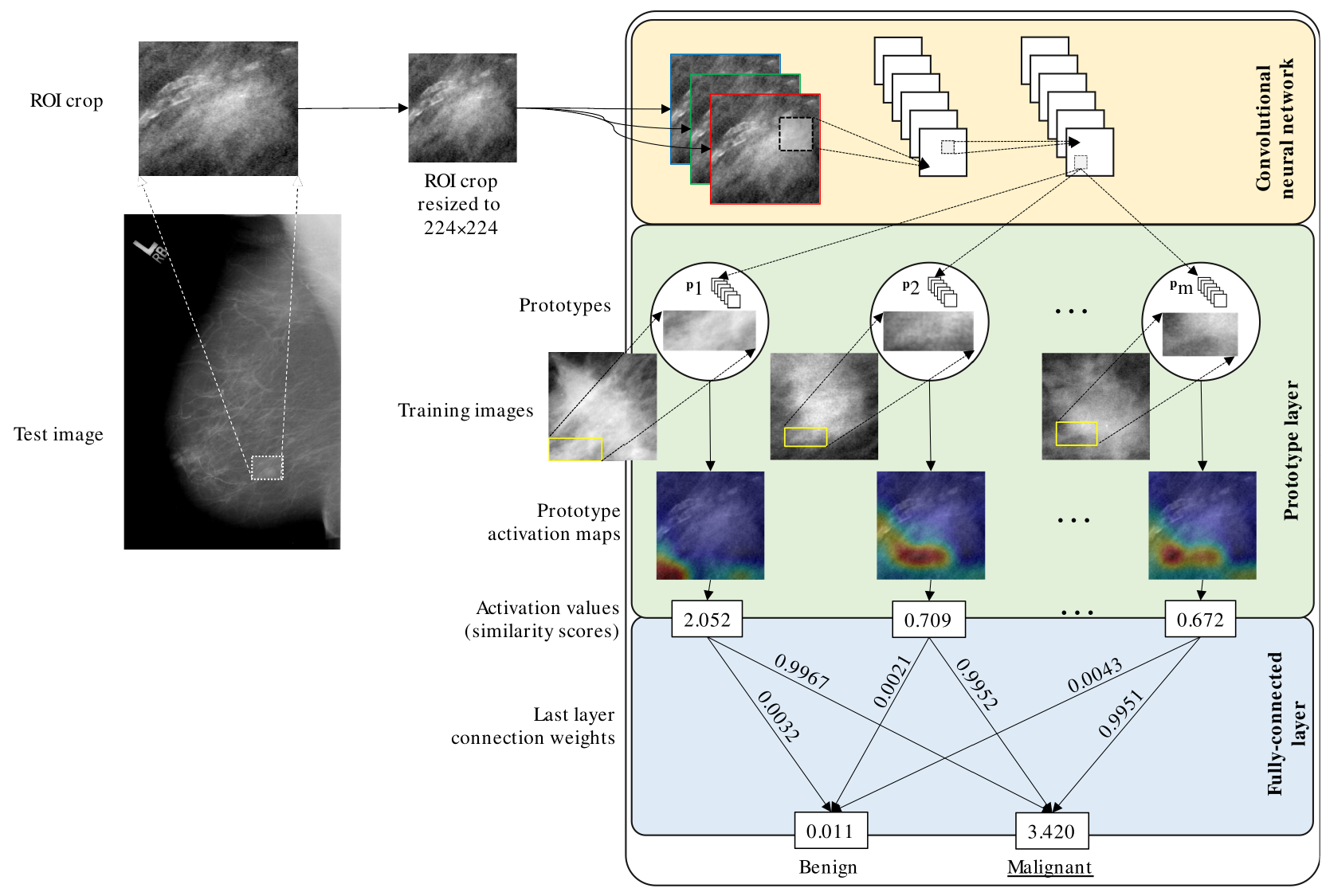}
\caption{Inference process through ProtoPNet: classification of a breast mass by means of the activation of pre-learned prototypes within the test image.}
\label{fig:ppnet}
\end{figure}

In addition, since \acrshort{label_ProtoPNet} works with three-channel images, we modified the one-channel gray-scale input images by copying the information codified in the single channel to the other two.
Then, we set the number of classes for the classification task to two instead of $200$.
Finally, to reduce overfitting when training a large model using a limited dataset, we introduced a 2D dropout layer and a 2D batch-normalization layer after each add-on convolutional layer of the model.
An overview of our implementation of \acrshort{label_ProtoPNet} architecture and its inference process is depicted in Fig.~\ref{fig:ppnet}, taking the classification of a correctly classified malignant mass as an example.

\subsection{Experimental Settings}
As for the \acrshort{label_CNN} block of ProtoPNet, the residual network ResNet18 with weights pre-trained on the ImageNet dataset was used in this experiment. 
Images were resized to a dimension of $224\times224$ pixels and their values were normalized with $mean$ and standard deviation ($std$) equal to $0.5$ for the three channels. 
As a result, image values range between $-1$ and $+1$ and this helps to improve the training process.

\begin{figure}[t]
\centering
\includegraphics[width=0.99\textwidth]{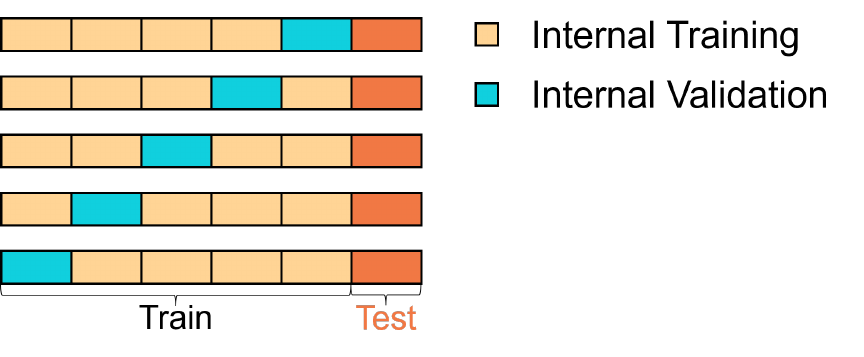}
\caption{\textbf{Five-fold Cross Validation}. The train set was split in 5 parts, one was used as the internal validation and the remaining 4 as the internal training. All the splits were patient based, in other words, masses of the same patients were in the same “internal” set (training/validation).}
\label{fig:5fold_CV_protopnet}
\end{figure}

We then performed a random search to optimize the model's hyperparameters. For each configuration, we built a five-fold \acrshort{label_CV} framework for training lesions, creating the internal-training and internal-validation subsets with an $80$-$20\%$ proportion (ref. Figure \ref{fig:5fold_CV_protopnet}). We performed the splitting in both class-balanced and patient-stratified fashion; this way, we maintained the balance between the classes and we associated lesions of the same patients to the same subset (internal-training or internal-validation) for each \acrshort{label_CV} fold. We employed the StratifiedGroupKFold function from the scikit-learn library~\cite{scikit-learn} for this purpose.

\begin{table}[t]
\caption{Values of the ProtoPNet hyperparameters for the random search.}
\label{tab:grid_search_values}
\centering
\begin{tabularx}{\textwidth}{XX}
\hline
Parameter & Domain\\
\hline
$lr\_features$ & $[1e$-$7, 1e$-$6]$\\
\hline
$lr\_add\_on$ & $[1e$-$7, 1e$-$6]$\\
\hline
$lr\_prot\_vector$ & $[1e$-$7, 1e$-$6]$\\
\hline
$WD$ & $[1e$-$3, 1e$-$2]$\\
\hline
$train\_batch\_size$ & $[20, 40]$\\
\hline
$clst$ & $[0.6, 0.8, 0.9]$\\
\hline
$sep$ & $[-0.1, -0.08, -0.05]$\\
\hline
$l1$ & $[1e$-$5, 1e$-$4, 1e$-$3]$\\
\hline
$num\_prots\_per\_class$ & $[5, 20, 40]$\\
\hline
\end{tabularx}
\end{table}

Given the large number of hyperparameters in ProtoPNet that can be optimized, we investigated only a fraction of them in this work. In particular, we examined the \acrshort{label_LR} at joint epochs, the \acrshort{label_WD}, the batch size of the internal-training subset, the coefficients of the \acrshort{label_ProtoPNet} loss function terms, and the number of prototypes per class. Their possible values are reported in Table~\ref{tab:grid_search_values}. Among the resulting $2592$ configurations, $30$ were randomly selected and used for training.
The remaining hyperparameters were chosen with fixed values instead. The ones different from the original ProtoPNet paper follow: $dropout\_proportion = 0.4$; $add\_on\_layers\_type = bottleneck$; $num\_filters = 512$;
$validation\_batch\_size = 2$;
$push\_batch\_size = 40$;
$warm\_optimizer\_lrs = \{add\_on\_layers: 1e$-$6$, $prototype\_vectors: 1e$-$6\}$; and $last\_layer\_optimizer\_lr = 1e$-$6$.

At training time, we performed data augmentation on the internal-training subset by adding slightly modified copies of already existing data. Typically, this procedure reduces overfitting when training a machine learning model and acts as regularization. 
We adopted the following transformations: (i) images underwent rotation around their center by an angle randomly picked in the range $[-10^\circ,+10^\circ]$; (ii) images were perspective skewed, that is, transforming the image so that it appears as if it was viewed from a different angle; the magnitude was randomly drawn from a value up to $0.2$; (iii) images were stretched by shear along one of their sides, with a random angle within the range $[-10^\circ,+10^\circ]$; images were mirrored (iv) from left to right along y-axis and (v) from top to bottom along x-axis.
Among the presented transformations, those based on a random initialization of certain parameters were repeated ten times each to further augment the number of instances. As a result, considering also the original ones, the number of internal-training images was totally increased by a factor of $33$. 
For such augmentation we exploited the Python Augmentor Library~\cite{bloice2017augmentor}, which has been designed to permit rotations of the images limiting the degree of distortion. 

Differently from the original study, we used fixed LR values instead of an LR scheduler, and we framed the training process within an early stopping (\acrshort{label_ES}) setting rather than a $1000$-epochs one. In particular, we checked the trend of the loss function for \acrshort{label_ES}. We exploited a moving average with $window = 5$ and $stride = 5$ to reduce the influence of noise in contiguous loss values at joint epochs. At every push epoch, a discrete derivative was computed on the two averaged values resulting from the ten joint epochs preceding that push epoch. A non-negative derivative was the condition to be checked. If the condition persisted for the following $30$ joint epochs (patience), ES occurred, and the training process stopped. The considered model was the one saved before the $30$ patience epochs.

Following the random search, we chose the best-performing configuration based on the metrics reported in section~\ref{sec:eval_metrics}. Hence, we re-trained the model on the whole training set with the selected configuration for as many epochs as the average maximum epoch in the \acrshort{label_CV} folds. We then performed a prototype pruning process, as suggested in the workflow of the original paper~\cite{chen2019looks}. We did that to exclude, from the set of learned prototypes, those that potentially regard background and generic regions in favor of more class-specific ones. Finally, we evaluated the final model on test set images.

In the end, we compared \acrshort{label_ProtoPNet} with a simpler, conventional black-box model. Since our ProtoPNet uses ResNet18 as the \acrshort{label_CNN} block, we repeated the classification task with the same pre-processed dataset using a ResNet18 with weights pre-trained on ImageNet.

We conceived the training framework as a fine-tuning of the last convolutional layers. The fine-tuning was performed under the same five-fold CV settings and with the same data augmentation operations. To reduce the overfitting during training, we also inserted a dropout layer before the final \acrshort{label_FC} layer.

Provided that \acrshort{label_ProtoPNet} and ResNet18 have globally different hyperparameters, an independent random search was performed. The subset of investigated hyperparameters follows:
number of re-trained last convolutional layers $=[1, 2, 3, 4, 5, 10, 20]$;
LR $=[1e$-$7,1e$-$6]$;
WD $=[1e$-$3,1e$-$2,1e$-$1]$; and dropout proportion = $[0, 0.2, 0.4]$.
Among the $126$ possible configurations, $50$ were randomly selected for training. 

Following the random search, we selected the top-performing configuration according to the metrics outlined in Section~\ref{sec:eval_metrics}. Accordingly, we re-trained the model on the entire training set with the chosen configuration for a number of epochs equal to the average maximum epoch in the \acrshort{label_CV} folds. 
Lastly, we evaluated the final model on the test set images.

\subsection{Evaluation Metrics}
\label{sec:eval_metrics}
We used both quantitative metrics and a qualitative assessment to evaluate the performance of the models at training time. As for quantitative metrics, we computed the accuracy value and stored it for both the internal training and the internal-validation subsets at each epoch for each \acrshort{label_CV} fold of a given configuration. We then obtained the configuration accuracy with its standard deviation by averaging the best validation accuracy values across the \acrshort{label_CV} folds.

Even though some \acrshort{label_CV} folds might reach high validation accuracy values at some epochs, the overall trend of the validation learning curves could be erratic and noisy over epochs. Hence, we computed the learning curves of accuracy and loss for each configuration and collected them for both internal-training and internal-validation subsets at each \acrshort{label_CV} fold. 
Then, these curves were averaged epoch-wise to obtain an average learning curve and standard deviation values for each epoch. 

We used a qualitative assessment of the average learning curves in combination with quantitative metrics to verify the correctness of the training phase. In this regard, we considered a globally non-increasing or with a high standard deviation trend as unjustifiable.
We then selected the best-performing configuration of hyperparameters based on both the configuration accuracy and the quality assessment. When evaluating the model on the test set, although the classes were balanced and thus the Accuracy metric might have sufficed, we wanted to provide a comprehensive overview of other metrics. Thus, we also assessed its performance through Precision, Recall, F1 score, F2 score, and \acrshort{label_AUROC}.

\subsection{Implementation Environment}
All the experiments in this study ran on the AI@Edge cluster of our Institute, composed by four nodes, each with the following specifications: $1 \times$ NVIDIA\textsuperscript{\tiny\textregistered} A$100$ $40$~GB Tensor Core, $2 \times$ AMD - Epyc $24$-Core $7352$ $2.30$~Ghz $128$~MB, $16$ x DDR$4$-$3200$ Reg. ECC $32$~GB module $= 512$~GB.

We implemented the presented work using Python $3.9.7$ on the CentOS $8$ operating system and back-end libraries of PyTorch (version $1.9.1$, build py$3.9$-cuda$11.1$-cudnn$8005$). 
In addition, to ensure reproducibility, we set a common seed for the random sequence generator of all the random processes and PyTorch functions.

\section{Results}
\label{sec:results_protopnet}
\subsection{Dataset Balancing}
The original dataset consisted of $577$ benign and $637$ malignant masses in the training set and $194$ benign and $147$ malignant masses in the test set. As a result of the cleaning process, we removed $49$ benign and $60$ malignant masses from the training set and $48$ benign and $16$ malignant masses from the test set. Next, based on the more prevalent class in each set, we removed $49$ malignant masses from the training set and $15$ benign masses from the test set to balance the resulting dataset. Therefore, the final number of utilized masses was $528$ for each label in the training set and $131$ for each label in the test set. 

\subsection{Experiment with ProtoPNet}
\begin{figure}[t]
\centering
\includegraphics[width=.75\columnwidth]{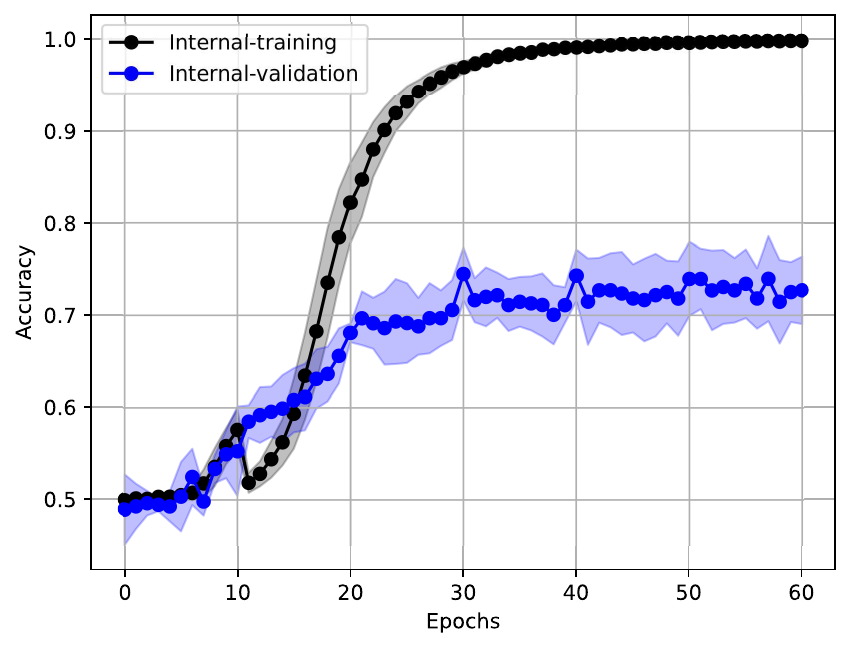}
\caption{Average accuracy curves across the five CV folds for the selected ProtoPNet model's configuration. Shaded regions represent $\pm 1\cdot std$ interval for each epoch.}
\label{fig:average_accuracies}
\end{figure}

As a result of the internal-training and internal-validation split, each \acrshort{label_CV} fold consisted of $844$ and $210$ original images, respectively. Then, as a result of the data augmentation, the internal-training subset consisted of $27852$ images.
The random search with five-fold \acrshort{label_CV} on the specified hyperparameters yielded the results reported in Table~\ref{tab:grid_search_results}. There, values in each configuration belong to the hyperparameter domain of Table~\ref{tab:grid_search_values}, and are listed in the same order. For each configuration, we reported the values of mean and standard deviation accuracy across the \acrshort{label_CV} folds.
Based on those values, the best-performing model was obtained in configuration $28$, which has the following hyperparameter values:
$lr\_features = 1e$-$6$;
$lr\_add\_on = 1e$-$6$;
$lr\_prot\_vector = 1e$-$6$;
$WD = 1e$-$3$;
$train\_batch\_size = 20$;
$clst = 0.8$;
$sep = -0.05$;
$l1 = 1e$-$4$; and
$num\_prots\_per\_class = 20$.
With this model, the validation accuracy was ${0.763\pm0.034}$.
The selected model also satisfied goodness of the learning curves, according to the quality assessment (Fig.~\ref{fig:average_accuracies}).
During the training phase, the \acrshort{label_ES} condition was triggered at epoch $30$. Nevertheless, $60$ epochs are reported in the plot because of the $30$ patience interval epochs.

\begin{table}[t]
\caption{Accuracy results for the random search on ProtoPNet's configurations.}
\label{tab:grid_search_results}
\centering
\begin{tabular}{ll}
\hline
Configuration & $mean\pm std$\\
\hline
$0:[1e$-$6,1e$-$7,1e$-$7,1e$-$3,40,0.6,-0.1,1e$-$4,5]$ & $0.718\pm0.069$\\
\hline
$1:[1e$-$6,1e$-$6,1e$-$6,1e$-$3,20,0.8,-0.1,1e$-$3,40]$ & $0.753\pm0.038$\\
\hline
$2:[1e$-$6,1e$-$7,1e$-$7,1e$-$3,20,0.9,-0.05,1e$-$4,20]$ & $0.746\pm0.043$\\
\hline
$3:[1e$-$6,1e$-$6,1e$-$6,1e$-$3,20,0.9,-0.08,1e$-$5,40]$ & $0.743\pm0.042$\\
\hline
$4:[1e$-$6,1e$-$6,1e$-$6,1e$-$3,20,0.9,-0.05,1e$-$5,40]$ & $0.759\pm0.035$\\
\hline
$5:[1e$-$7,1e$-$6,1e$-$6,1e$-$3,40,0.8,-0.08,1e$-$3,20]$ & $0.706\pm0.056$\\
\hline
$6:[1e$-$7,1e$-$6,1e$-$6,1e$-$2,20,0.8,-0.05,1e$-$5,5]$ & $0.624\pm0.045$\\
\hline
$7:[1e$-$7,1e$-$6,1e$-$6,1e$-$2,20,0.8,-0.1,1e$-$3,20]$ & $0.698\pm0.082$\\
\hline
$8:[1e$-$7,1e$-$6,1e$-$6,1e$-$2,20,0.6,-0.08,1e$-$3,5]$ & $0.700\pm0.037$\\
\hline
$9:[1e$-$7,1e$-$6,1e$-$7,1e$-$3,20,0.6,-0.05,1e$-$5,40]$ & $0.713\pm0.058$\\
\hline
$10:[1e$-$7,1e$-$6,1e$-$7,1e$-$2,40,0.9,-0.05,1e$-$5,5]$ & $0.683\pm0.042$\\
\hline
$11:[1e$-$7,1e$-$6,1e$-$7,1e$-$2,40,0.6,-0.08,1e$-$3,40]$ & $0.697\pm0.057$\\
\hline
$12:[1e$-$7,1e$-$6,1e$-$7,1e$-$2,20,0.6,-0.05,1e$-$5,40]$ & $0.697\pm0.066$\\
\hline
$13:[1e$-$7,1e$-$7,1e$-$6,1e$-$3,40,0.6,-0.08,1e$-$4,5]$ & $0.591\pm0.055$\\
\hline
$14:[1e$-$7,1e$-$7,1e$-$6,1e$-$3,20,0.8,-0.08,1e$-$4,20]$ & $0.683\pm0.067$\\
\hline
$15:[1e$-$7,1e$-$7,1e$-$6,1e$-$2,20,0.9,-0.08,1e$-$3,5]$ & $0.668\pm0.032$\\
\hline
$16:[1e$-$7,1e$-$7,1e$-$7,1e$-$3,40,0.6,-0.05,1e$-$4,5]$ & $0.574\pm0.030$\\
\hline
$17:[1e$-$7,1e$-$7,1e$-$7,1e$-$3,20,0.6,-0.1,1e$-$4,5]$ & $0.679\pm0.045$\\
\hline
$18:[1e$-$7,1e$-$7,1e$-$7,1e$-$2,40,0.6,-0.08,1e$-$3,20]$ & $0.668\pm0.041$\\
\hline
$19:[1e$-$6,1e$-$6,1e$-$6,1e$-$2,20,0.8,-0.05,1e$-$5,5]$ & $0.748\pm0.019$\\
\hline
$20:[1e$-$6,1e$-$6,1e$-$6,1e$-$3,40,0.9,-0.05,1e$-$4,20]$ & $0.736\pm0.039$\\
\hline
$21:[1e$-$6,1e$-$6,1e$-$7,1e$-$3,40,0.6,-0.08,1e$-$5,5]$ & $0.757\pm0.023$\\
\hline
$22:[1e$-$6,1e$-$6,1e$-$7,1e$-$3,20,0.8,-0.05,1e$-$3,20]$ & $0.722\pm0.018$\\
\hline
$23:[1e$-$6,1e$-$6,1e$-$7,1e$-$3,20,0.6,-0.1,1e$-$3,40]$ & $0.762\pm0.036$\\
\hline
$24:[1e$-$6,1e$-$6,1e$-$7,1e$-$2,40,0.6,-0.05,1e$-$4,20]$ & $0.757\pm0.038$\\
\hline
$25:[1e$-$6,1e$-$7,1e$-$6,1e$-$2,40,0.9,-0.1,1e$-$3,20]$ & $0.732\pm0.055$\\
\hline
$26:[1e$-$6,1e$-$7,1e$-$6,1e$-$2,40,0.6,-0.1,1e$-$3,40]$ & $0.745\pm0.028$\\
\hline
$27:[1e$-$6,1e$-$6,1e$-$6,1e$-$3,40,0.8,-0.08,1e$-$5,40]$ & $0.743\pm0.042$\\
\hline
\boldmath{$28:[1e$-$6,1e$-$6,1e$-$6,1e$-$3,20,0.8,-0.05,1e$-$4,20]$} & \boldmath{${0.763\pm0.034}$}\\
\hline
$29:[1e$-$6,1e$-$7,1e$-$6,1e$-$2,20,0.9,-0.1,1e$-$4,20]$ & $0.741\pm0.040$\\
\hline
\end{tabular}
\end{table}

According to the training curves in Fig.~\ref{fig:average_accuracies}, we re-trained the selected model on the training set for $30$ epochs. After pruning, $9$ and $2$ prototypes were removed from the benign and the malignant classes, respectively. As a result, $29$ final prototypes were retained. Then, we assessed this model on the test set.

Finally, regarding the comparison with ResNet18, we obtained the following results. Among the $50$ explored configurations, the best performing model was found with the following hyperparameters: number of re-trained last convolutional layers $=3$, \acrshort{label_LR} $=1e$-$6$, \acrshort{label_WD} $=1e$-$3$, dropout rate $=0.4$. This model reached an average validation accuracy across the five CV folds of $0.776\pm 0.026$.
After re-training the model on the whole training set for $20$ epochs, we evaluated it on the test set images.

\begin{figure}
\centering
\includegraphics[width=.99\textwidth]{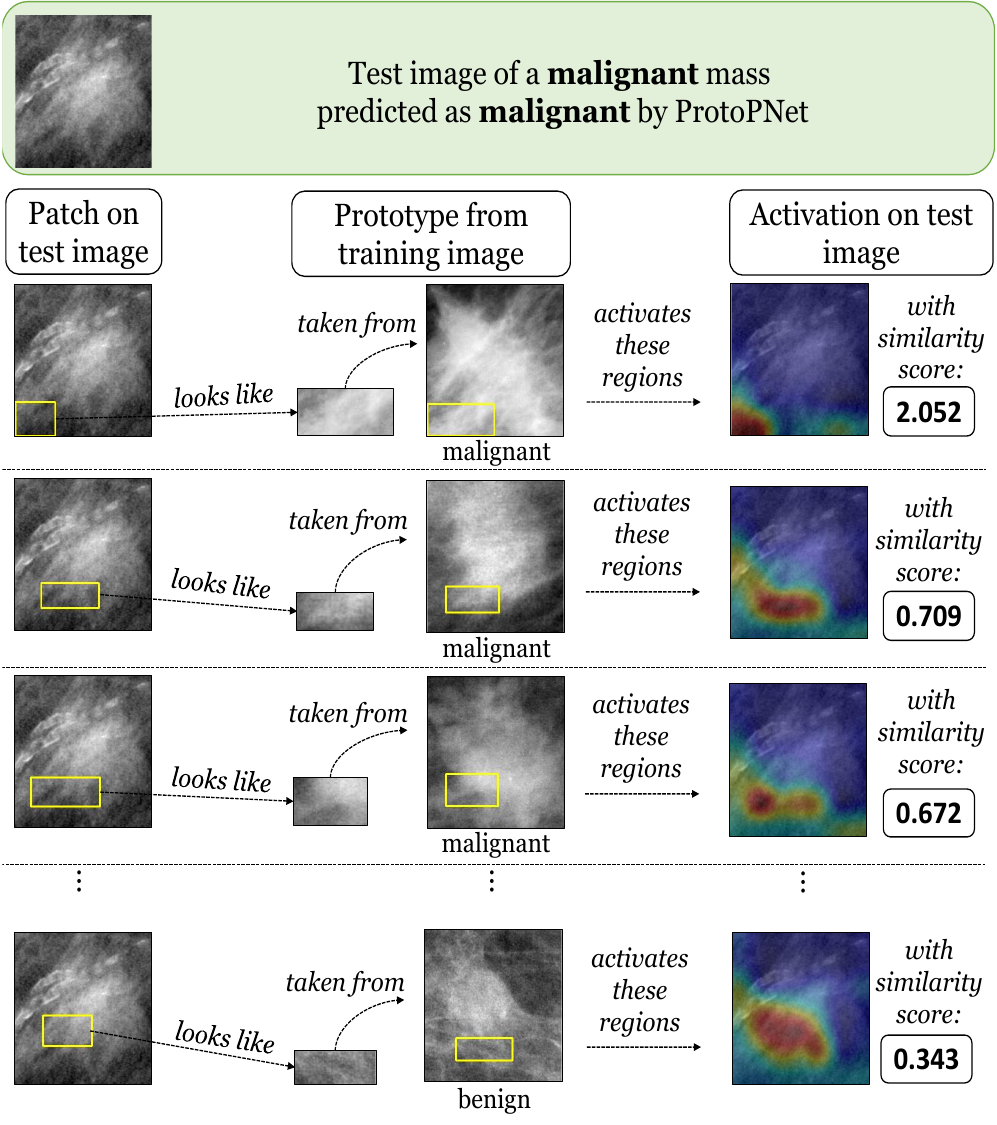}
\caption{The test image of a malignant mass is correctly classified as malignant by the model. Each row of this image represents the activation process of a certain prototype. In the first column there is the patch found on the test image, in the second column the activated prototype is shown together with the training image from which it originated, in the third column is shown the activation map with the corresponding similarity score.}
\label{fig:thislooklikethat_correct}
\end{figure}

The test-set metrics yielded by \acrshort{label_ProtoPNet} and ResNet18 in their independent experiments are reported in Table~\ref{tab:test_results}.
In Fig.~\ref{fig:thislooklikethat_correct}, we report an example of an explanation provided by ProtoPNet for a test image of a correctly classified malignant mass. Similarities with prototypes recognized by the model are listed from top to bottom according to the decreasing similarity score of the activation. Note that the top activated prototypes correctly derive from training images of malignant masses. Instead, towards the lower scores, prototypes originating from other classes might be activated, in this case of benign masses. 

\begin{table}
\caption{Test-set metrics obtained with the validation best-performing models.}
\label{tab:test_results}
\centering
\begin{tabularx}{\textwidth}{XXXXXXX}
\hline
Model & Accuracy & Precision & Recall & F1 & F2 & AUROC\\
\hline
ProtoPNet & $\mathbf{0.685}$ & $0.658$ & $\mathbf{0.769}$ & $\mathbf{0.709}$ & $\mathbf{0.744}$ & $\mathbf{0.719}$\\
ResNet18 & $0.654$ & $\mathbf{0.667}$ & $0.615$ & $0.640$ & $0.625$ & $0.671$\\
\hline
\end{tabularx}
\end{table}

\section{A Clinical Feedback on the Output Prototypes: Expert Radiologist Viewpoint}
\label{sec:clinical_viewpoint}
Specific domain knowledge is necessary to understand and interpret the explanations provided by models such as ProtoPNet when applied to medical images. The validity of provided visual explanations is hardly evaluable by someone without a background in the specific task. Furthermore, explanations can be misleading or confusing when analyzed by non-experts.

\begin{figure}
\centering
\includegraphics[width=.9\textwidth]{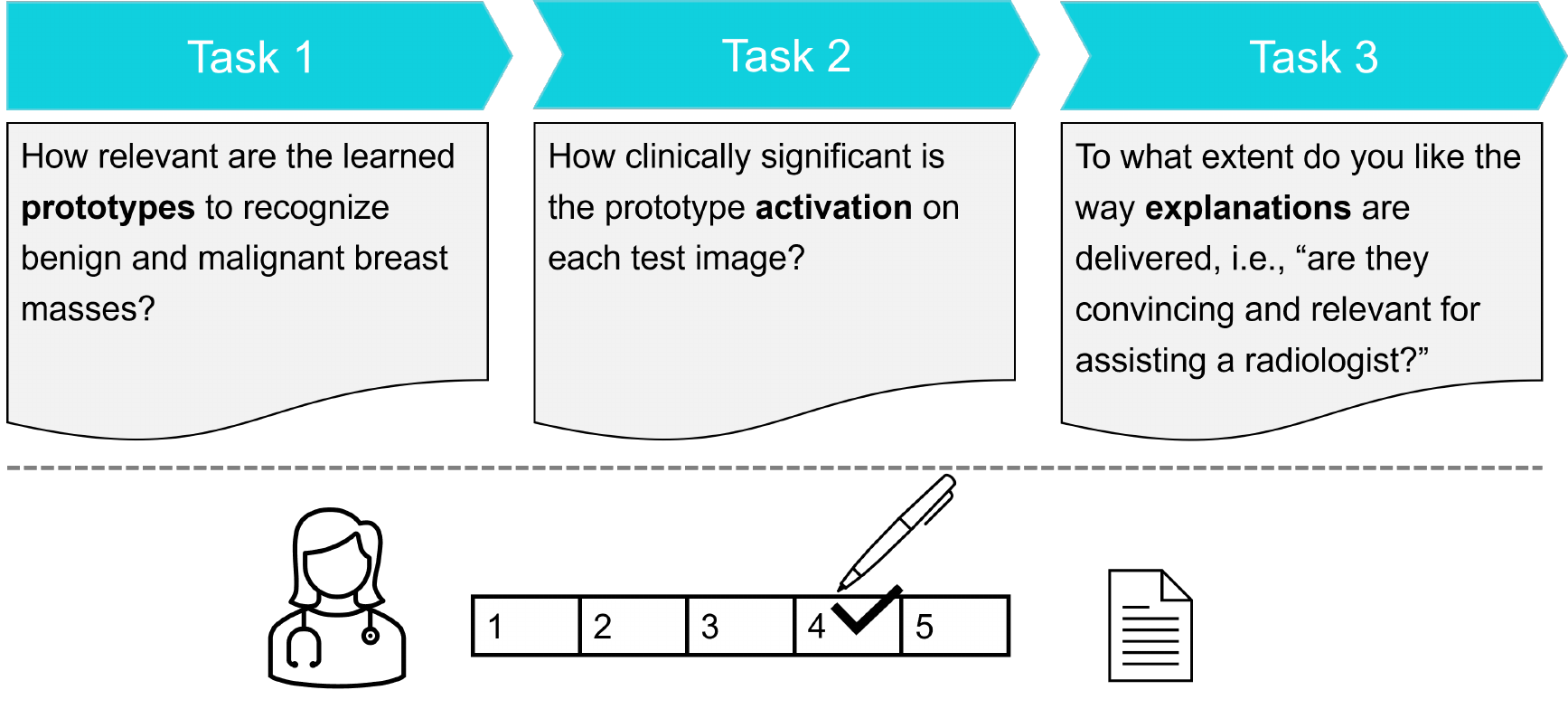}
\caption{An experienced radiologist provided clinical feedback on the goodness of the produced explanations.}
\label{fig:expertFeedback_protopnet}
\end{figure}

When dealing with explainable models, one of the first concerns is to ensure that explanations are based on correct information. Also, for such models to be interpretable and hence helpful in medical practice, their explanations should use intuitions that somewhat resemble the reasoning process of a physician. In this regard, we asked a radiologist with $16$ years of experience for a clinical viewpoint on the outputs of the selected model on a random subset of test images ($15$ benign, $15$ malignant) to prevent selection bias. In particular, we conceived three tasks (i.e., Task 1, 2, and 3 in Figure \ref{fig:expertFeedback_protopnet}).

\subsection{Task 1: How Good Are Prototypes?}
As before stated, \acrshort{label_ProtoPNet} bases the classification outcome and the explanation on patch similarities with a set of learned prototypes. Therefore, we first wish to understand whether good-quality prototypes were learned and used to characterize each class.
This was done in \textbf{Task $1$}. We presented the radiologist with a series of images representing the learned prototypes from both classes and the images from which they were extracted. We asked her to rate how much each prototype was specific for its corresponding class on a scale from one to five. Lower scores would be assigned to generic, not clinically significant prototypes, while class-specific, meaningful prototypes would receive higher scores. As a result of Task $1$, only $50\%$ of the benign prototypes were considered to be of acceptable quality, while about $88\%$ of malignant prototypes were deemed good by the radiologist.
Based on those results, we also ran a sentiment analysis\footnote{
A technique used to determine the emotional tone behind a series of outputs, often used to gain insights into public opinion or customer feedback. In our case, it involves analyzing the ratings given by the expert radiologist to gauge her overall sentiment towards the outputs of our model. Since our scale is from $1$ to $5$, we can directly use these numbers as sentiment scores; $1$-$2$ Negative sentiment (disappointment), 3: Neutral sentiment, $4$-$5$: Positive sentiment. We multiply each rating by its weight (same as rating), sum these values, and then divide by the total weight. The resulting interpretation of scores is as follows; $1.0$ - $2.0$: Indicates a generally negative sentiment (the output did not meet expectations),  $2.1$ - $3.0$: Indicates a mixed or neutral sentiment (outputs were somewhat satisfactory but had room for improvement), and $3.1$ - $5.0$: Indicates a positive sentiment, where the output met or exceeded expectations.
}.

The average sentiment score for benign prototypes was $2.8$, indicating an almost neutral or slightly negative sentiment, whereas the sentiment score for malignant prototypes was $4.4$, indicating a strongly positive sentiment.

\subsection{Task 2: Is Prototype Activation Clinically Valid?}
Next, we would like to check that ProtoPNet was capable of learning a meaningful concept of similarity. In this sense, image regions that the model recognized as similar should contain comparable clinical information.
Therefore, in  \textbf{Task $2$}, we asked the radiologist to rate the activation of the most activated prototype w.r.t each image in the selected subset. For each case, the activated patch on the test image and the corresponding activating prototype were given. The rating was expressed on a scale from one to five. Activations that shared mutual clinical information would receive higher scores.

Regarding Task $2$, among the $30$ activated patches of the test images, $20$ resulted as clinically similar to the activating prototypes according to radiologist's feedback. The average sentiment score for that task was $3.2$, indicating a slightly positive reaction. 

\subsection{Task 3: Are the Overall Explanations Relevant?}
Finally, we wished to figure out the degree of satisfaction in medical end-users for the explanations provided.
This was carried out in  \textbf{Task $3$}. We presented the radiologist with test images, each labeled with the classification yielded by ProtoPNet, along with the explanation based on the two most activated prototypes.
In fact, sorted by decreasing activation, the first two prototypes were those with higher activation scores, but then from the third onward, there tended to be a visible decrease. For this reason and to make it feasible for the radiologist to evaluate in a reasonable time, we limited ourselves to the two most similar cases.
She provided scores on a scale from one to five for the overall satisfaction of such explanations. A lower score would be assigned to explanations that highlighted non-relevant regions or did not highlight regions on which the radiologist would focus. Instead, if the radiologist believed the explanation to be convincing and complete (i.e., all the relevant regions are identified), she would have returned a higher score.

The analysis on Task $3$ showed that the radiologist recognized explanations for benign-predicted masses as sufficiently satisfying only in $50\%$ of the cases. This resulted in a sentiment score of $2.8$, indicating a slightly negative sentiment. On the other hand, explanations for malignant-predicted masses were convincing $89\%$ of the times, resulting in a sentiment score of $4.1$, indicating a positive reception. This also reflects the results of Task $1$ regarding the quality of the learned prototypes.

This investigation of the explanation quality of the proposed method, both on the detection of prototypes and the activations correctness, is preliminary. As a by-product, the expert radiologist's feedback is a precious contribution for the design, in the near future, of other tests to assess both the explanation's correctness and of explanation's acceptance by end-users. 

\section{Discussion and Future Work} 
\label{sec:discussion_protopnet}
Historically, not knowing precisely why \acrshort{label_DL} models provide their predictions has been one of the biggest concerns raised by the scientific community.
Healthcare, in particular, is one of the areas massively impacted by the lack of transparency of such black-box models. That is especially relevant for automatic medical image classification, which medical practice still strives to adopt.
Explainable and interpretable \acrshort{label_AI} might overcome this issue by getting insights into models' reasoning. In this regard, a promising approach is that of ProtoPNet~\cite{chen2019looks}, an explainable-by-design model first introduced in the natural images domain.

Our work aimed at exploring the applicability of prototypical part learning in medical images and, in particular, in the classification of benign/malignant breast masses from mammogram images. We assessed the applicability based on two aspects: the ability of the model to face the task (i.e., classification metrics) and the ability of the model to provide end-users with plausible explanations.
We trained a ProtoPNet model and optimized its hyperparameters in a random search with five-fold \acrshort{label_CV}. Then, we compared its performance to that obtained with an independently optimized ResNet18 model. We selected images from CBIS-DDSM~\cite{lee2017curated}, a publicly available dataset of scanned mammogram images. After, came a cleaning and balancing process to obtain the final study cohort. 
As opposed to the original paper, we utilized a hold-out test independent from the internal-validation subset used at training time to assess the final performance. In addition, we introduced two-dimensional dropout and batch-normalization after each add-on convolutional layer in the ProtoPNet architecture.

Evaluation metrics resulting from the best-performing \acrshort{label_ProtoPNet} model seem mostly higher than with the ResNet18 architecture. In particular, we observed the most substantial improvement in the Recall, which is of considerable interest for this specific task. Indeed, it represents the capacity of the model to detect positive cases: a high Recall means that the model correctly identifies the majority of malignant masses. 
In addition, ProtoPNet provides a level of transparency that is completely missing from ResNet18. That said, it is well known that neural networks often use context or confounding information instead of the information that a human would use to solve the same problem in both medical~\cite{wang2018removing} and non-medical applications~\cite{hu2020stratified}.

We believe a large amount of prior domain knowledge is necessary to evaluate ProtoPNet's explanations. Without domain knowledge, its results are likely to be misinterpreted. Moreover, such knowledge would be necessary to properly select the number of prototypes for each class, instead of empirically derive it from a hyperparameter optimization.
To prevent explanations to be based on irrelevant regions of the images, we asked for the radiologist's viewpoint.
In this regard, she provided some helpful insights into the models' outputs.
From Task $1$, it seems reasonable to assume that \acrshort{label_ProtoPNet} manages to learn more relevant prototypes for malignant masses similar to radiologists. As in actual practice, a suspicious finding (a non-circumscribed contour, whether microlobulated, masked, indistinct, or spiculated), even only in one projection, results easy to detect and justifies a recall for further assessment. On the other hand, a benign judgment requires an accurate bi-dimensional analysis of typical benign findings in both projections and differential diagnoses with overlapping tissue.
From Task $2$, it appears that the model's mathematical concept of similarity differs from how a radiologist would deem two regions clinically similar. The reason behind this may be that the radiologist recalls specific features from past experience, possibly consisting of other exams aside from mammography and biopsy results alone. This is way broader than the dataset the network uses for training, which strictly consists of image-biopsy label pairs.
Finally, from Task $3$, results that explanations for images classified as malignant are, in general, more likely to be more convincing to the radiologist. Notably, this behavior goes in the same direction as the low clinical relevance of benign prototypes from Task $1$.
Overall, the radiologist found ProtoPNet's explanations very intuitive and hence reported a high level of satisfaction. This is remarkably important because we were interested in the right level of abstraction for explanations to foster human interpretability.

Comparing our work with previous studies is not straightforward: no other work with prototypical part learning has been done on the CBIS-DDSM dataset and benign/malignant mass classification task. Nevertheless, we hereafter compare our results with previous works utilizing ResNets on the same dataset and task, albeit some of them in slightly different ways. In the comparison, we report accuracy as the common performance metric across these studies. In our experiments we achieved an accuracy of $0.654$ with ResNet18 and of $0.685$ with ProtoPNet. Among the ones using ResNet18, Arora et al.~\cite{arora2020deep} and Ragab et al.~\cite{ragab2021framework} achieved an accuracy of $0.780$ ad $0.722$, respectively. Instead, among the works using different ResNet architectures, Ragab et al.~\cite{ragab2021framework} achieved an accuracy of $0.711$ and $0.715$ when using ResNet50 and ResNet101, respectively. Tsochatzidis et al.~\cite{tsochatzidis2019deep} deployed ResNet50 obtaining an accuracy of $0.749$. Also Alkhaleefah et al.~\cite{alkhaleefah2020influence} experimented with ResNet50 in different scenarios and achieved accuracy values between $0.676$ and $0.802$. Finally, Ansar et al.~\cite{ansar2020breast} reported an accuracy of $0.637$ by using ResNet50. Although performance metrics reported in the previous works are in line with ours, they are, in general, higher. 

Regarding previous studies adopting a prototypical part learning scheme to the mass classification task, not much work has been done. To the best of our knowledge, the work by Barnett et al.~\cite{barnett2021case} is the only one, even though the authors utilized a different (and private) dataset and a different novel architecture, derived from ProtoPNet. 
For these reasons, a fair comparison may not be feasible. Besides, we achieved an \acrshort{label_AUROC} of $0.719$ with ProtoPNet, which is lower than theirs ($0.840$). 
The authors used images in combination with a dedicated fine-annotation of relevant regions and mass margins, and their model heavily exploits that information for its conclusions. We point out that this is different from our work, where ProtoPNet uses only image-level labels without annotated images to resemble the experimental setup of the original work on bird classification~\cite{chen2019looks}.
This is probably one of the reasons for the performance discrepancies between the two studies. However, fine-annotated images needed in their methodology require a massive intervention by clinical experts.
Also, intending to deploy such models to fast assist radiologists in the classification of a new image, we believe their approach to be too dependent on annotations, therefore, our approach may be preferable.
We likely obtained acceptable results without the complexity of the model and of the dataset of \cite{barnett2021case}.

Interestingly, the performance in \cite{barnett2021case} is somewhat similar to that obtained on the bird classification task of the original work introducing ProtoPNet~\cite{chen2019looks}. The inclusion of information regarding relevant regions and mass margins annotations might have been the key to achieve such high results on the mass classification task. However, our work shows that, by taking the same annotation-free approach of \cite{chen2019looks}, lower results might be obtained for this task. According to our results, without additional information to complement images, the task to be solved is more challenging, and the problem covers a higher level of complexity. Specifically, in images acquired by projection, planes at different depths are fused in a single bi-dimensional representation. That makes object separation especially hard for these images. 
This implies that answering our research question may not be as straightforward as for the ornithology task.

Our work comes with limitations. Firstly, given the large number of hyperparameters in \acrshort{label_ProtoPNet}, we selected a subset of them for the optimization process. Moreover, of all the possible configurations obtainable with the chosen subset of hyper-parameters, we evaluated the model only on a random selection of them. That likely had an impact on the discovery of the optimal configuration.
Secondly, due to the limited size of the utilized dataset, our models were prone to overfitting, which affects the generalization capabilities on new images. That is particularly true for ProtoPNet, where the entire architecture has to be re-trained. That happened even though we took several actions to counteract the issue. Specifically, we selected a shallower ResNet architecture, deployed WD, and introduced dropout and batch-normalization layers. 
In addition, we provided the clinical viewpoint of a single radiologist. We are aware that this clashes somewhat with the subjective nature of such views due to inter-rater variability\cite{kang2022variability}. A group of differently experienced radiologists should have been included to reach more robust conclusions.

Future work would include adopting a different initialization for the filter values, for example, with values learned on the same dataset using the corresponding base architecture instead of those pre-trained on ImageNet. Moreover, in addition to geometrical transformations, one could also exploit intensity-based transformations to try to improve the networks' generalization capabilities on images possibly obtained with different acquisition settings. These may include histogram equalization and random brightness modification. Also, one could utilize a combination of different mammogram image datasets to augment diversity in the data cohort. On top of that, a dataset comprising digital breast tomosynthesis images instead of conventional digital mammogram images could be used. That is a pseudo-3D imaging technique based on a series of low-dose breast acquisitions from different angles, which has the potential to overcome the tissue superposition issue and thus improve the detection of breast lesions.

\section{Summary}
\label{sec:conclusion}
Our research question was to investigate the applicability of \acrshort{label_ProtoPNet} to the automatic classification of breast masses from mammogram images. Although a clear-cut answer might not have been provided, this exploratory work allowed us to assess the advantages and the weak points of this kind of approach. The two aspects we considered to evaluate the applicability of this approach were the classification capabilities and the validity of explanations.
Classification results were acceptable but insufficient for this method to enter the clinical practice. 
Based on the clinical assessment, we may say that explanations provided for malignant masses were highly plausible, valuable, and intuitive to a radiologist. However, this is not true for benign masses yet, and this currently invalidates the applicability of ProtoPNet in real clinical contexts. On the other hand, this behavior is comparable to that of a radiologist, who, typically, finds it easier to recognize malignant masses' characteristics. 
Nevertheless, our findings are promising and suggest that ProtoPNet may represent a compelling approach that still requires further investigation. We believe that training this model on more images or performing a more extensive optimization of the model's architecture may bring improved classification performance. That might also increase the ability of the model to deliver plausible explanations for benign cases.
\chapter{The Role of Causality in Explainable Artificial Intelligence}
\label{chap:review_XAI_causality}
\begin{figure}[h!]
\includegraphics[width=\textwidth]{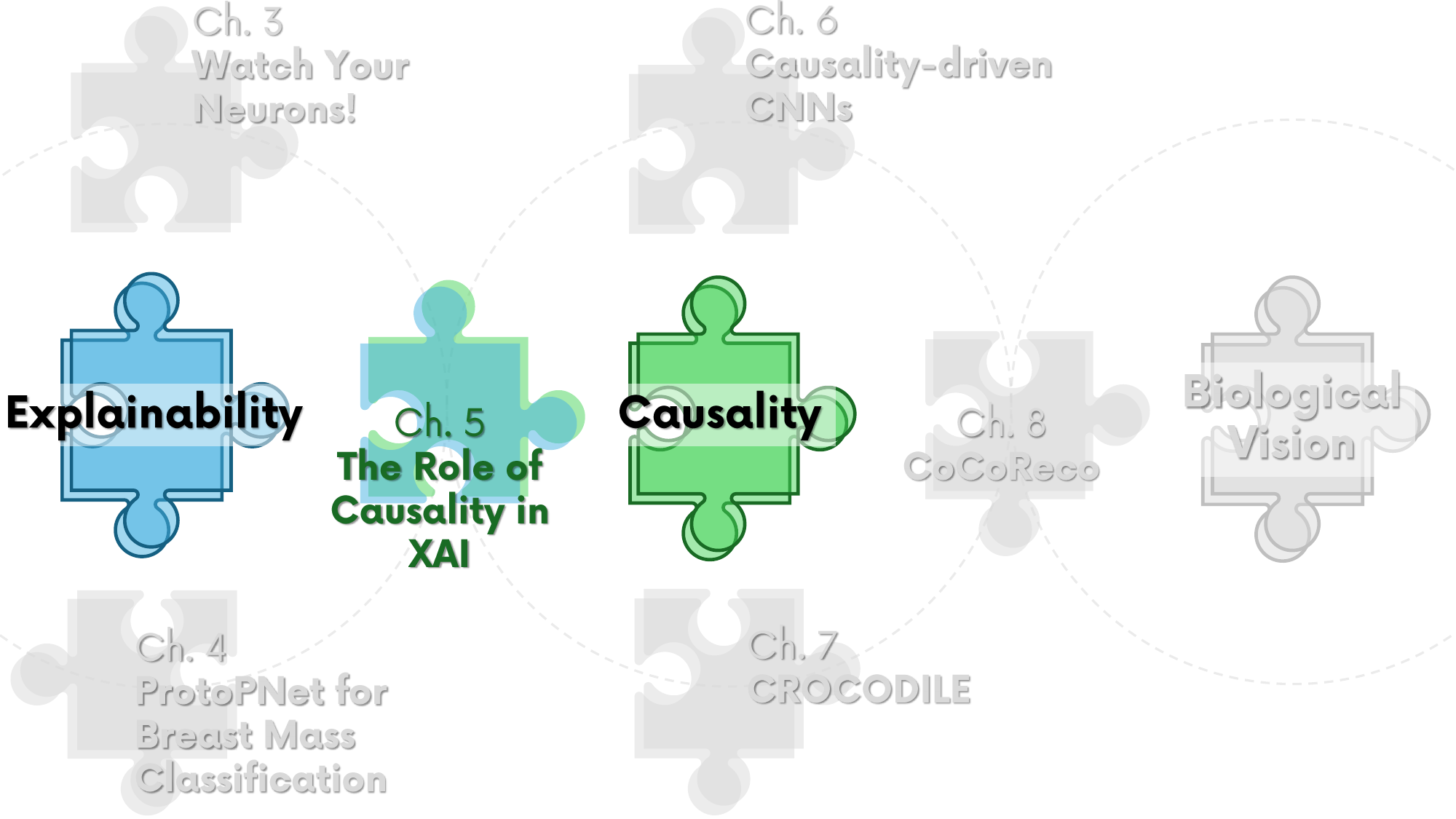}
\end{figure}

After investigating \acrshort{label_XAI} methods in the previous two chapters, we approached other aspects necessary for a more human-aligned \acrshort{label_AI}. Among those, we realized that causality (i.e., the study of cause-effect relationships) could be very closely related to the sphere of \acrshort{label_XAI}. In this chapter, we present a systematic review of the roles of causality in \acrshort{label_XAI}.

Section ~\ref{sec:review_ancientRoots} narrates how causation and explanation are not new concepts since they have always drawn humans' attention. They have been highly intertwined since the ancient Greeks and throughout the philosophy of science. On the other hand, these concepts have evolved in diverse ways in \acrshort{label_AI} and computer science. 

Section \ref{sec:review_rationaleObjective} presents the rationale and objective of this study, describing three pieces of information that led us to start the investigation. We investigated the interdisciplinary literature from both theoretical and methodological viewpoints to understand how and to what extent causality and \acrshort{label_XAI} are intertwined. More precisely, we sought to uncover what kinds of relationships exist between the two concepts and how one can benefit from them, for instance, in building trust in AI systems.

Accordingly, in Section \ref{sec:review_methods}, we describe the structured process we followed to specify the inclusion criteria, detail the information sources, illustrate the search strategy on specified databases, and describe the selection process. Also, this section describes how we extracted information from the publications to answer our research question and, finally, how we collected software solutions in a structured way.

After Section \ref{sec:review_results_coOccurrence}, which presents a high-level analysis of keywords' co-occurrence and provides insights into how the literature relates different research concepts, we present the results to the research question in Section \ref{sec:review_results_researcQuestion}.
Indeed, three main perspectives are identified.
The first way to relate the two fields is to move some critics to \acrshort{label_XAI} under a causal lens, to serve as a watch out. In this regard, a non-negligible subset of publications recognizes causality as a missing component of current XAI research to achieve robust and explainable systems. Other works highlight how the field of XAI (and \acrshort{label_AI} by extension) suffers from certain innate issues, making the problem itself ill-posed. In a similar light, a further branch of works investigates different forms and desiderata of the XAI-produced explanations and their link with the causal theory.
%
The second perspective tries to relate \acrshort{label_XAI} and causality in a pragmatic way and sees the former as a means to get to the latter. Such works believe XAI has the potential to foster scientific exploration for causal inquiry. Indeed, by means of approaches able to identify pursue-worthy experimental manipulations, XAI may help scientists generate hypotheses about possible causal relationships to be tested.
%
The third perspective turns the previous one around, claiming that causality is propaedeutic to \acrshort{label_XAI}. Causal tools and metrics are exploited to implement XAI, and specific XAI approaches are brought back to their formal causal definition to improve generalization capabilities. Among the distinctive ideas of this perspective, getting access to the causal model of a system is a way to intrinsically explain the system itself.

After listing and discussing relevant software solutions used to automate causal tasks in Section \ref{sec:review_software}, we dive into an overall discussion of the conceptualized perspectives and analyze their validity according to new, updated literature in Section \ref{sec:review_discussion}.


The content of this chapter is based on our paper: Carloni Gianluca, Andrea Berti, and Sara Colantonio. "The role of causality in explainable artificial intelligence.", \textit{WIREs Data Mining and Knowledge Discovery}. August 2024 (DMKD-00660.R1, minor revisions). An \textit{ArXiv} preprint before revisions is available at arXiv:2309.09901 (2023)\cite{carloni2023role}.
\begin{figure}
    \centering
    \includegraphics[width=0.99\textwidth]{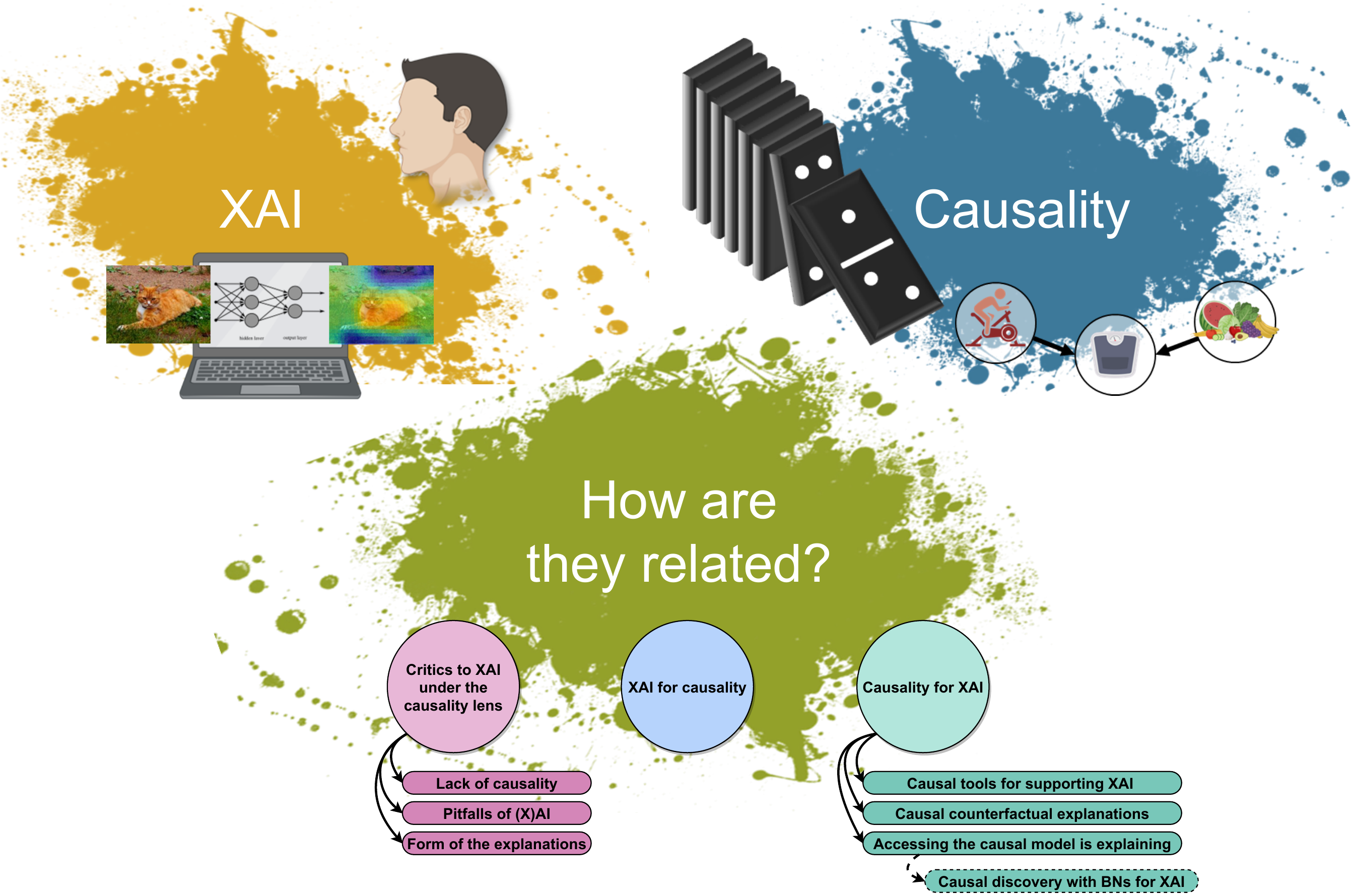}
    \caption{\textbf{Graphical Abstract}: Reviewing the literature to uncover how eXplainable Artificial Intelligence (XAI) and causality are related - the three main perspectives.}
    \label{fig:graphical_abstract}
\end{figure}

\section{Ancient roots}
\label{sec:review_ancientRoots}
The study of causation and explanation can be traced back to the ancient Greek philosophers. Aristotle, for instance, introduced causality as the foundation of explanation and argued that there must be a necessary and sufficient reason for every event \cite{hankinson1998cause}.

As early as the 18th century, the empiricist David Hume formalized causation in terms of sufficient and necessary conditions: an event \textit{c} causes an event \textit{e} if and only if there are event-types \textit{C} and \textit{E} such that \textit{C} is necessary and sufficient for \textit{E}. He, however, remained skeptical about humans' ability to explain and truly know any event. Indeed, he argued that we cannot perceive any necessary connection between cause and effect, but only events occurring in regular succession based on habit \cite{hume2003treatise}.

From the 1950s onward, some others also investigated scientific explanations.
Initially, the “standard model” of explanation was deductive, following the Deductive-Nomological (DN) model by Hempel and Oppenheim \cite{hempel1948studies}. An outcome was implied logically from universal laws plus initial conditions via deductive inference (e.g., explaining the volume of gas via the ideal gas law and some observations such as pressure).
Regarding Hempel's viewpoint on causality, causal explanations are special cases of DN explanations, but not all laws and explanations are causal.

Later, Salmon \cite{salmon1984scientific} developed a model in which good scientific explanations must be statistically relevant to the outcome to be explained. He argued that, in attempting to explain probabilistic phenomena, we seek not merely a high probability but screen for causal influence by removing system components to find ones that alter the probability. Salmon found causality ubiquitous in scientific explanation and was convinced that the time had come to put the “cause” back into “because”. Although remaining vague as to how to attain it, he invited scientists to reconsider the role of causal relations as potentially fundamental constituents of adequate explanations.

\section{Rationale and Objective}
\label{sec:review_rationaleObjective}
What seemed to emerge from the literature at the time of this study is that there was no clear vision of whether there was a dependent relationship between the fields of causality and \acrshort{label_XAI} in the realm of \acrshort{label_AI}. But we had a clue.

Three main pieces of information led us to believe that those could be complementary fields, and, thus, motivated us to start our investigation:
\begin{itemize}
    \item First, the concepts of causality and explanation have been jointly investigated since ancient times (Sec.~\ref{sec:review_ancientRoots}).
    
    \item Second, even though they were born separately in the field of \acrshort{label_AI} (\textit{explanation} as XAI and \textit{causation} as Pearl's causality theory), they share a common goal. Indeed, both fields feature human-centricity in AI systems and aim to ensure true usefulness to humans, be it by explaining in a human-comprehensible way what an AI system did (ref. Section \ref{sec:back_XAI}), or by designing the system in such a way that it reasons like humans (ref. Section \ref{sec:back_causality}).
    
    \item Third, another "canary in the coal mine" for us was the presence of the same "counterfactual" term in both fields\footnote{In a somewhat confusing way, the term "counterfactual" may be encountered not only at the third rung of Pearl's causal hierarchy (Section \ref{sec:back_causality_what}), but also in the \acrshort{label_XAI} literature, where it applies to any instance with an alternative outcome (see Section \ref{sec:back_XAI}). There, a \textit{counterfactual explanation} (\acrshort{label_CFE}) refers to the smallest change in an input that changes the prediction of an ML classifier \cite{wachter2017counterfactual,mothilal2020explaining,moreira2024benchmarking}. This concept is quite distinct from the causal meaning of the term (Section \ref{sec:back_causality_what}). In this regard, as a piece of clarification, we utilize \textit{\acrshort{label_CFE}} and \textit{\acrshort{label_CF}} to address, respectively, the XAI method and the causality concept.}.
\end{itemize}
\noindent Therefore, this review jointly covers these two fields and investigates the role(s) of causality in the world of \acrshort{label_XAI} or, broadly, the relationship between causality and XAI. 

To the best of our knowledge, Chou et al. (2022) \cite{chou2022counterfactuals} were the only ones investigating a somewhat similar question at the time of this study, albeit with a narrower scope. They systematically review current counterfactual model-agnostic approaches (i.e., \acrshort{label_CFE}s) studying how they could promote \textit{causability}. Causability is a relatively new term representing "the extent to which an explanation of a statement to a human expert achieves a specified level of causal understanding with effectiveness, efficiency, and satisfaction in a specified context of use." \cite{holzinger2019causability}. Since causability differs from causality, this is the first (and major) difference with our study, which covers the wide notion of causality itself. Our work also departs from \cite{chou2022counterfactuals} in that they solely investigate CFE methods, while, in our analysis, we consider the whole corpus of \acrshort{label_XAI} literature, which also includes (but is not limited to) \acrshort{label_CFE}s.

\section{Methods}
\label{sec:review_methods}
This review aims at exploring the literature surrounding the relationship between causality and \acrshort{label_XAI}, from both theoretical and methodological viewpoints.
We conducted our work by adopting a structured process that involved the following: (i) specifying the eligibility criteria; (ii) detailing the information sources; (iii) illustrating the search strategy on specified databases; (iv) describing the selection process; (v) conducting a high-level analysis on the cohort of selected studies; (vi) extracting relevant data and information from studies; and (vii) synthesizing results.

We carried out our search on four popular bibliographic databases,
\textit{Scopus}\footnote{\url{https://www.scopus.com/}}, \textit{IEEE Xplore digital library} (s. IEEE)\footnote{\url{https://ieeexplore.ieee.org/}}, \textit{Web of Science} (s. WoS)\footnote{\url{https://clarivate.com/webofsciencegroup/solutions/web-of-science/}}, and \textit{ACM Guide to Computing Literature} (s. ACM)\footnote{\url{https://dl.acm.org/browse}}, utilizing the following query:\\

\noindent\textsc{(causal*) \textbf{AND} (expla*) \textbf{AND} ("xai" OR "explainable artificial intelligence" OR "explainable ai") \textbf{AND} ("machine learning" OR "ai" OR "artificial intelligence" OR "deep learning")}\\

\noindent Elements within brackets had to be present within at least one of the title, abstract, or keywords of the manuscript. Terms ending with the wildcard “$*$” matched all the terms with the specified common prefix.
Note that the third and fourth conjuncts in the above query overlap semantically and syntactically, as most papers containing the former, would also contain the latter in the same fields. Indeed, we verified that relaxing the fourth conjunct had little impact on the number of retrieved papers, resulting in less than $2\%$ additional papers.
Among the obtained publications, we removed duplicates and ensured that only peer-reviewed papers from conference proceedings and journals were included. Then, upon completing the process of screening, eligibility, and inclusion of articles, $51$ publications formed the basis of our review. The obtained final cohort of papers were published between December 2017 and August 2022.

In our study, we first performed a high-level analysis of the final cohort of records regarding keywords co-occurrence, then, we extracted information from the publications to answer our research question, and, finally, we collected any cited software solutions in a structured way. 
The following section begins by describing the technical details of the whole study collection process.

\subsection{Study Selection Process}
\label{sec:appendix_studyselectionprocess}
Although we did not apply any temporal constraint to the search, we adopted some exclusion criteria in the process. We excluded works that were not written in English, articles from electronic preprint archives (e.g., ArXiv\footnote{\url{https://arxiv.org}}), book chapters, and theses.
In addition, we excluded too-short papers and/or papers of poor quality that hindered our ability to extract data meaningfully. We also deemed off-topic those papers that considered causality in the common and everyday sense of the term, not based on theoretical definitions. Indeed, they frequently present few occurrences of the causal domain terms, which were often either poorly contextualized or only present in the abstract/keywords of the article.

Regarding information sources, we selected Scopus, IEEE, Wos, and ACM because they cover a comprehensive range of AI works and provide powerful interfaces for retrieving the required data with limited restrictions.
Conversely, we excluded Google Scholar\footnote{\url{https://scholar.google.com/}}, SpringerLink\footnote{\url{https://link.springer.com/}}, and Nature\footnote{\url{https://www.nature.com/siteindex}} since they do not allow to formulate the query string with the same level of detail as the selected databases do, and, on the other hand, we excluded PubMed\footnote{\url{https://pubmed.ncbi.nlm.nih.gov/}}, since it provides this capability, but its coverage is restricted solely to the medical field.

\subsection{Query Strings}
As for the search strategy on the specified databases, the use of the wildcard made word-matching easier. For instance, \textbf{causal$*$} matched terms like \textit{causal} and \textit{causality}, while \textbf{expla$*$} matched terms such as \textit{explanation(s)}, \textit{explainable}, \textit{explainability}, \textit{explaining}, and \textit{explained}. 

On July 14, 2022, we utilized the research query on the four databases for the first time. We collected the retrieved publications and started analyzing them. Then, on September 5, 2022, we repeated the search in the same settings. This allowed us to refine our cohort of papers with new works that have been published in the meanwhile, therefore enriching our analyses. In general, although we utilized the same research query across the four databases, the actual query string was edited according to the specific syntax of each of them. In this regard, those strings are shown in Table \ref{tab:query_strings}.

\begin{table}
\caption{Query strings used for each database. AB, ABS: abstract; AK, KEY: keywords; TI: title.}
\label{tab:query_strings}
\centering
\resizebox{\textwidth}{!}{%
\begin{tabular}{p{1.1cm}|p{15.75cm}}
\hline
Database & Query string\\
\hline\\
Scopus & \texttt{TITLE-ABS-KEY(causal*) \textbf{AND}\newline
TITLE-ABS-KEY(expla*) \textbf{AND}\newline
TITLE-ABS-KEY(xai OR "explainable artificial intelligence" OR "explainable ai") \textbf{AND}\newline
TITLE-ABS-KEY("machine learning" OR ai OR "artificial intelligence" OR "deep learning")}\\
\hline\\
Web of\newline Science & \texttt{(TI=causal* OR AB=causal* OR AK=causal*)
\textbf{AND}\newline
(TI=expla* OR AB=expla* OR AK=expla*)
\textbf{AND}\newline
(TI=(xai OR "explainable artificial intelligence" OR "explainable ai") OR AB=(xai OR "explainable artificial intelligence" OR "explainable ai") OR AK=(xai OR "explainable artificial intelligence" OR "explainable ai"))
\textbf{AND}\newline
(TI=("machine learning" OR ai OR "artificial intelligence" OR "deep learning") OR AB=("machine learning" OR ai OR "artificial intelligence" OR "deep learning") OR AK=("machine learning" OR ai OR "artificial intelligence" OR "deep learning"))}\\
\hline\\
IEEE Xplore & \texttt{("Document Title":causal* OR "Abstract":causal* OR "Author Keywords":causal*) 
\textbf{AND}\newline
("Document Title":expla* OR "Abstract":expla* OR "Author Keywords":expla*) 
\textbf{AND}\newline
("Document Title":xai OR "Document Title":"explainable artificial intelligence" OR "Document Title":"explainable ai" OR "Abstract":xai OR "Abstract":"explainable artificial intelligence" OR "Abstract":"explainable ai" OR "Author Keywords":xai OR "Author Keywords":"explainable artificial intelligence" OR "Author Keywords":"explainable ai") 
\textbf{AND}\newline
("Document Title":"machine learning" OR "Document Title":ai OR "Document Title":"artificial intelligence" OR "Document Title":"deep learning" OR "Abstract":"machine learning" OR "Abstract":ai OR "Abstract":"artificial intelligence" OR "Abstract":"deep learning" OR "Author Keywords":"machine learning" OR "Author Keywords":ai OR "Author Keywords":"artificial intelligence" OR "Author Keywords":"deep learning")}
\\
\hline\\
ACM & \texttt{(Title:causal* OR Abstract:causal* OR Keyword:causal*)
\textbf{AND}\newline
(Title:expla* OR Abstract:expla* OR Keyword:expla*)
\textbf{AND}\newline
(Title:xai OR Title:"explainable artificial intelligence" OR Title:"explainable ai" OR Abstract:xai OR Abstract:"explainable artificial intelligence" OR Abstract:"explainable ai" OR Keyword:xai OR Keyword:"explainable artificial intelligence" OR Keyword:"explainable ai")
\textbf{AND}\newline
(Title:"machine learning" OR Title:ai OR Title:"artificial intelligence" OR Title:"deep learning" OR Abstract:"machine learning" OR Abstract:ai OR Abstract:"artificial intelligence" OR Abstract:"deep learning" OR Keyword:"machine learning" OR Keyword:ai OR Keyword:"artificial intelligence" OR Keyword: "deep learning")}\\
\hline
\end{tabular}}
\end{table}

\begin{figure}[t]
    \centering
    \includegraphics[width=0.6\textwidth]{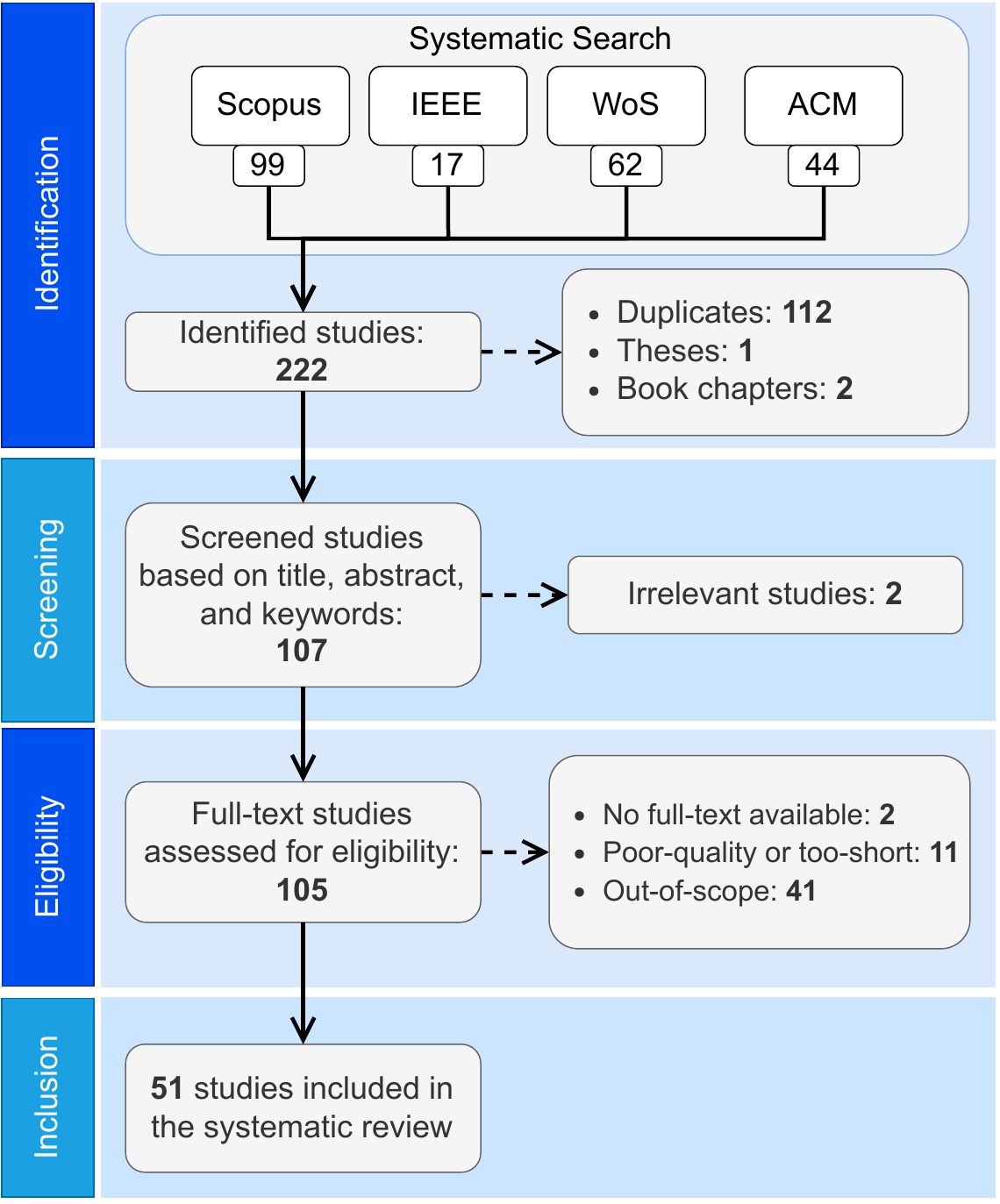}
    \caption{Flowchart of the study collection process, from identification, through screening, to eligibility and inclusion.}
    \label{fig:prisma_flowchart}
\end{figure}

Figure ~\ref{fig:prisma_flowchart} shows the process of identification, screening, eligibility, and inclusion of articles in our work.
From the search, we obtained the following number of records from the four databases: $99$ (Scopus), $17$ (IEEE), $62$ (WoS), and $44$ (ACM). As a result, we collected a total of $222$ publications.
Upon extraction of query results from the databases, we operated the identification phase. For the retrieved records, we extracted the BibTeX files and uploaded them into a popular reference manager application by Elsevier, namely Mendeley\footnote{\url{https://www.mendeley.com/}}, desktop version 1.19.8. We then utilized its \textit{Check for Duplicates} feature to perform duplicate removal. Then, we removed one thesis and two book chapters, according to the defined exclusion criteria. After these steps, the joint output was $107$ publications.

During the screening phase, we examined independently the resulting works by title, abstract, and keywords to verify and ensure that proper results were retrieved by the query.
Whenever both authors deemed a paper irrelevant, it was discarded from the cohort. Specifically, two publications were hereby discarded. Instead, publications for which the authors agreed on the inclusion, together with those on which they disagreed, passed to the next phase.

Next, in the eligibility phase, we first checked for the availability of full-text manuscripts for the records in the cohort. We excluded two studies as we could not access their full text. We then jointly analyzed the available full-text publications to remove papers that were clearly out of scope, together with poor-quality or too-short papers. As a result, we identified $11$ poor-quality or too-short papers and $41$ out-of-scope works. 
Lastly, once we reached a common decision for each of the publications, we collected the final cohort of studies to be included in the review.

\subsection{Keywords' co-occurrence analysis}
Regarding the high-level analysis of the final cohort of records, we constructed a bibliometric network of articles' keywords co-occurrence, by utilizing the Java-based application \textit{VOS Viewer}\footnote{\url{https://www.vosviewer.com/}}.
Bibliometric networks are methods to visualize, in the form of graphs, the collective interconnection of specific terms or authors within a corpus of written text. In our setting, we applied such networks to study the paired presence of articles' keywords within a corpus of scientific manuscripts.

\subsection{Research question analysis}
For each of the papers that were included in the review, we identified the most relevant aspects on a conceptual level. According to the research question, we searched for any theoretical viewpoints and comments on the possible ways in which causality and \acrshort{label_XAI} may relate, including formalization frameworks and insights from AI, cognitive, and philosophical perspectives.

Based on the collected information, we performed a topic clustering procedure to organize the literature in related concepts and gain a global view of the field. Selecting cluster topics for a multidisciplinary field as that of causality in the broad field of \acrshort{label_XAI} proved challenging. Topics that are too general would result in an excessively vague and superficial division of papers and therefore be of little use in answering the research question. On the other hand, topics that are too specific would create many quasi-empty clusters, resulting in an improper division, which lacks abstraction capabilities and prevents an overall view of the field. Therefore, we iteratively refined the clusters during a trial-and-error process.

\subsection{Software tools collection}
During the analysis of the full-text manuscripts, we kept track, in a structured collection, of any cited software solutions (e.g., tools, libraries, packages), whenever they were used to automate causal tasks. Specifically, for each one, we analyzed: (i) the URL of the corresponding web-page; (ii) whether the software was commercial or with an open-source license, according to the Open Source Initiative\footnote{\url{https://opensource.org}}; (iii) the name of the company for cases of commercial software; (iv) the eventual release publication that launched the software; (v) whether the frontend consisted in a command line interface (\acrshort{label_CLI}) or a graphical user interface (\acrshort{label_GUI}); and, finally, (vi) the main field of application and purpose.

\section{Results to the keywords' co-occurrence analysis} 
\label{sec:review_results_coOccurrence}
\begin{figure}
    \centering
    \includegraphics[width=0.6\textwidth]{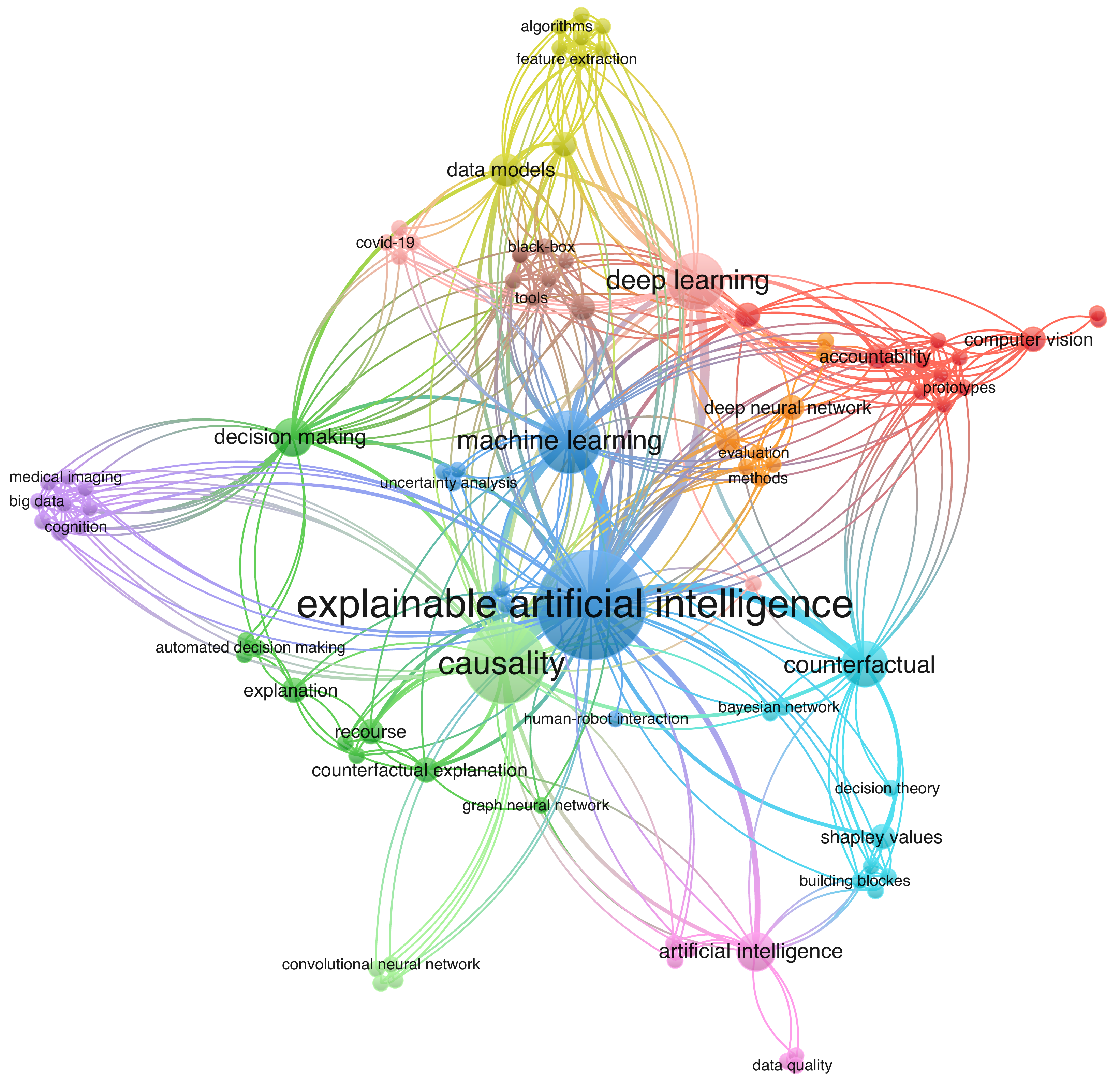}
    \caption{Bibliometric network of papers' keywords for the cohort of publications included in the review.}
    \label{fig:panel_top}
\end{figure}

As a result of the high-level analysis of the final cohort of records, we obtained the bibliometric network shown in Figure \ref{fig:panel_top}.
The items (i.e., nodes) of the network represent terms (specifically, articles' keywords); the link (i.e., edge) between two items represents a co-occurrence relation between two keywords; the strength of a link indicates the number of articles in which two keywords occur together; and, finally, the importance of an item is given by the number of links of that keyword with other keywords and by the total strength of the links of that keyword with other keywords. Accordingly, more important keywords are represented by bigger circles in the network visualization, and more prominent links are represented by larger edges between keywords.

\begin{figure}
    \centering
    \includegraphics[width=\textwidth]{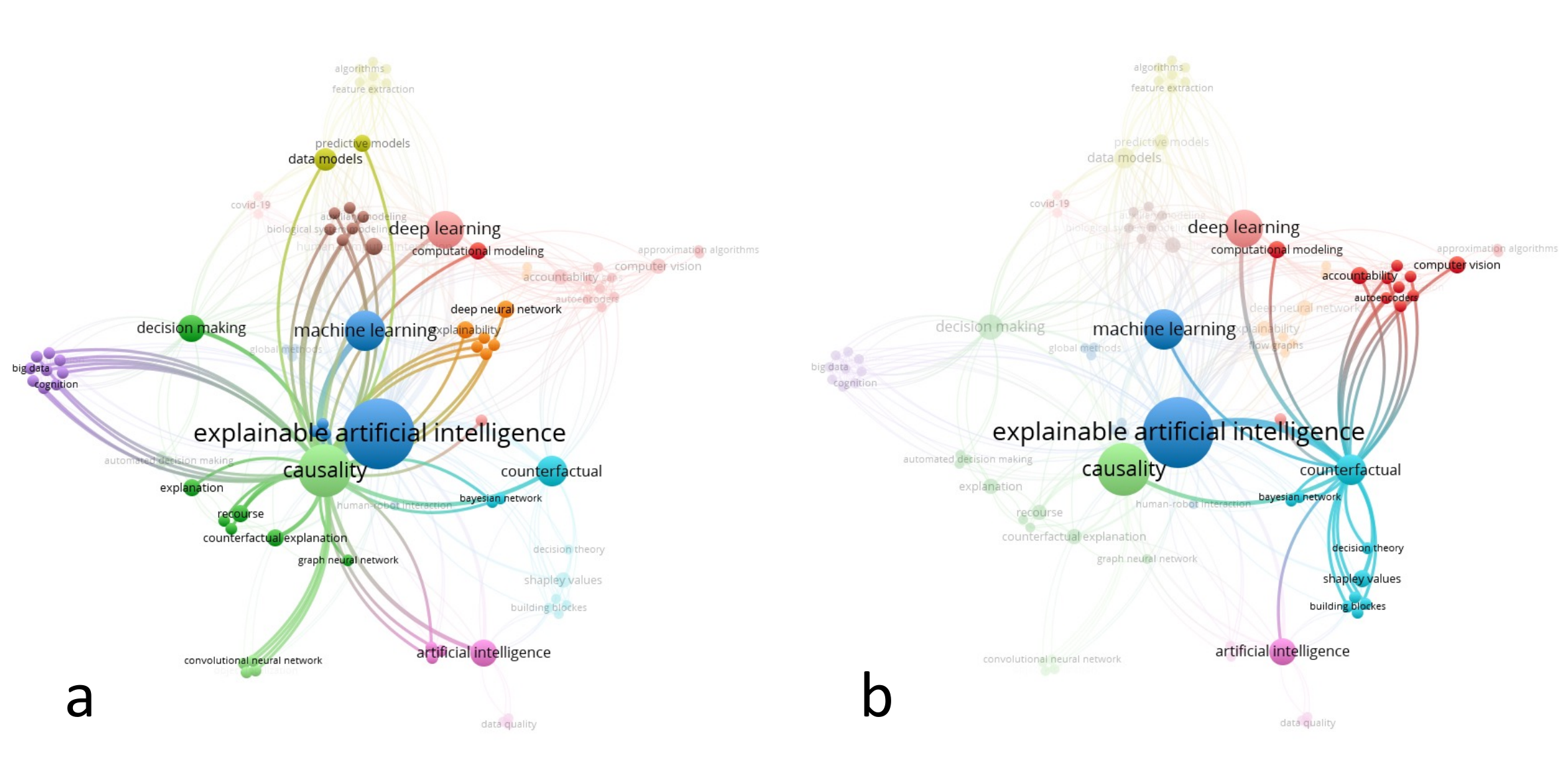}
    \caption{The isolated connections from Figure ~\ref{fig:panel_top} for the terms \textit{causality} (a) and \textit{counterfactual} (b).}
    \label{fig:agglomerato_quadrato}
\end{figure}

This visualization provides insight into how and to what extent the literature relates different research concepts, and it helped us to appreciate the multidisciplinary nature of our research question. Moreover, it is possible to marginalize the scope of specific keywords by identifying the terms to which they relate, as shown in Figs.~\ref{fig:agglomerato_quadrato}a-b for the keywords \textit{causality} and \textit{counterfactual}, respectively.
The relevance and wide scope of the first are justified by the structure of our query, where it was an obligatory search term. Regarding the latter, its scope and relevance represent the central role of the term in both the research fields of causality and \acrshort{label_XAI}. 

\newpage
\section{Results to the research question analysis}
\label{sec:review_results_researcQuestion}
\begin{figure}
    \centering
    \includegraphics[width=0.99\textwidth]{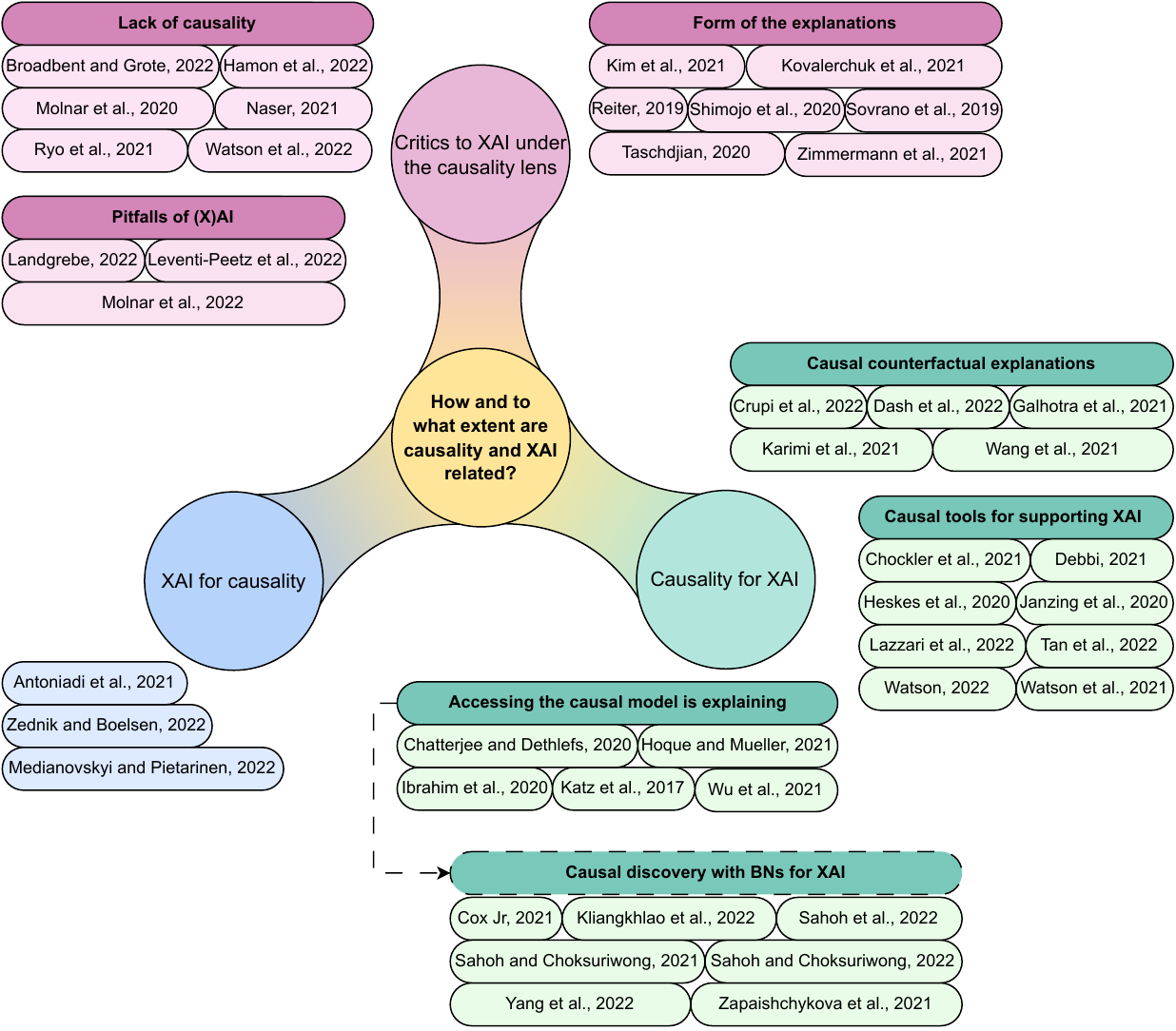}
    \caption{The included studies are classified according to the three main perspectives on how causality and \acrshort{label_XAI} may be related: \textit{Critics to XAI under the causality lens}, \textit{XAI for causality}, and \textit{Causality for XAI}. Next to each of them, are the possible sub-clusters.}
    \label{fig:rq1_clusters}
\end{figure}

This review allowed us to understand how the theory of causality could intertwine with the \acrshort{label_XAI} literature and, specifically, which methodologies and theoretical frameworks could be adopted to approach the bridge between these two fields.
We conceived three main topic clusters of studies, which are presented together with their possible sub-clusters in Figure \ref{fig:rq1_clusters}. Specifically, they embody the following perspectives:

\begin{itemize}
    \item "\textbf{Critics to XAI under the causality lens}"; 
    
    \item "\textbf{XAI for causality}"; 
    
    \item "\textbf{Causality for XAI}". 
\end{itemize}
This procedure led us to identify which of the three possible perspectives is the preferable one in order to correctly combine the two areas of causality and \acrshort{label_XAI}. We discuss them in Sec.~\ref{sec:rq1_ctx} --- \ref{sec:rq1_cfx}.

\subsection{Critics to XAI under the causality lens}
\label{sec:rq1_ctx}
This first perspective utilizes a causal viewpoint to identify some issues in current \acrshort{label_XAI}. The focus of such papers is either: (i) to point out the inability of XAI to consider causality, (ii) to highlight the profound limitations of current (X)AI both on a methodological and a conceptual level, or (iii) to investigate the forms of the produced explanations.

\subsubsection{Lack of causality}
A fundamental aspect that hinders the value of classical \acrshort{label_AI} models' inference and explainability methods is the lack of a foundation in the theory of causality. Indeed, classical \acrshort{label_ML} and \acrshort{label_DL} predictive models are based on the correlation found among training data instead of true causation. This might be of particular concern in specific fields, such as epidemiology, that have always been grounded in the theory of causation\cite{broadbent2022can}. Moreover, this lack of causality makes models more easily affected by adversarial attacks and less valuable for decision-making\cite{molnar2020interpretable}. Since the parameters and predictions of classical data-driven \acrshort{label_AI} models cannot be interpreted causally, they should not be used to draw causal conclusions.

As Naser (2021)\cite{naser2021engineer} points out, meeting specific performance metrics does not necessarily mean that an AI/ML model captures the physics behind a phenomenon. In other words, there is no guarantee that the found correlations map to causal relations between input data and final decisions. For this reason, determining whether such models reflect the true causal structure is crucial\cite{ryo2021explainable}.
This inability of today's ML/DL to grasp causal links reflects also on \acrshort{label_XAI}, constituting a major broad challenge to the ability of AI systems to provide sound explanations.

Hamon et al. (2022)\cite{hamon2022bridging} stress how this poses serious challenges to the possibility of satisfactory, fair, and transparent explanations. 
Regarding the soundness of the generated explanations, Watson et al. (2022)\cite{watson2022agree} demonstrate that they are volatile to changes in model training that are perpendicular to the classification task and model structure. This raises further questions about trust in \acrshort{label_DL} models which just rely on spurious correlations that are made visible via explanation methods.
Since causal explanations cannot be provided for \acrshort{label_AI} yet, explanatory methods are fundamentally limited for the time being. 

\subsubsection{Pitfalls of (X)AI} 
In addition to the weaknesses due to the lack of causality, some works highlight how the fields of \acrshort{label_AI} and \acrshort{label_XAI} may suffer from some innate issues. On a methodological level, Molnar et al. (2022)\cite{molnar2022general} present a number of pitfalls of local and global model-agnostic interpretation techniques, such as in case of poor model generalization, interactions between features, or unjustified causal interpretations.
At a deeper level, some researchers advocate some concerns about XAI based on its very nature.
For instance, Landgrebe (2022)\cite{landgrebe2022certifiable} argues that the human inability to interpret the behavior of deep models in a more objective manner still restricts \acrshort{label_XAI} methods to provide merely a partial, subjective interpretation. Undeniably, deep neural networks solve their classification in a manner that differs completely from the way humans interpret text, language, sounds, and images. For instance, \acrshort{label_CNN}s use features of the input space to perform their classifications, which are different from those humans use. Not only is it true, but what's more, we do not understand how humans themselves classify texts or images or conduct conversations. Indeed, as of now, human or physical behavior can only be emulated by creating approximations, but approximations cannot be understood any more than complex systems can be.

Under similar considerations, Leventi-Peetz et al. (2022)\cite{leventi2022scope} study the scope and sense of explainability in AI systems. In their view, it is impossible or unwise to follow the intention of making every ML system explainable. Indeed, even domain experts cannot always provide explanations for their decisions and, furthermore, on \acrshort{label_AI} systems much higher demands are made than on humans when they have to make decisions.

\subsubsection{Form of the explanations} These works explore different forms, qualities, and desiderata of the explanations produced by \acrshort{label_XAI} methods and their link with causality.
Depending on the application domain, less accurate yet simpler explanations may be preferable to convey a proper understanding of an AI decision. For instance, in Natural Language Generation, a narrative explanation where facts are linked with causal relations is probably a better explanation for narrative-inclined individuals, even though it may not be the most accurate way to describe how the model works \cite{reiter2019natural}. Similarly, in image classification via \acrshort{label_CNN}s, a simpler visualization (e.g., natural dataset examples) may lead to an equal causal understanding of unit activation instead of using complex activation maximization approaches \cite{zimmermann2021well}.

Shimojo et al. (2020)\cite{shimojo2020does} examine what a good explanation is by drawing on psychological evidence regarding two explanatory virtues: (i) the number of causes enforced in an explanation\footnote{This is sometimes referred to as \textit{simplicity} and is conforming with the \textit{Occam's razor} principle, according to which, an event should not be explained by more causes than necessary \cite{jefferys1992ockham}.}, and (ii) the number of effects invoked by cause(s) in an explanation\footnote{This is sometimes referred to as \textit{scope}. Explanations with a broader scope (i.e., correctly predict more events) make humans feel more certain than explanations with a narrower one \cite{johnson2014explanatory}.}. The authors report that, in a user study, the two virtues had independent effects, with a higher impact for the first one.
Similarly, Kim et al. (2021)\cite{kim2021multi} discuss several desiderata of \acrshort{label_XAI} systems, among which, they should adjust explanations based on the knowledge of the explainee, to match their background knowledge and expectations. This is further stated by Kovalerchuk et al. (2021)\cite{kovalerchuk2021survey}, who define as "quasi-explanations" those explanations using terms that are foreign to a certain application domain (e.g., medicine, finance, law), such as distances, weights, and hidden layers, and that consequently do make sense only for the data scientists.
Kim et al. (2021)\cite{kim2021multi} further states that explanations are considered to be \textit{causal} when they arise from the construction of causal models, serving as the basis for recreating a causal inference chain to (i.e., a “recipe” for reconstructing) a prediction. According to the authors, intelligent systems must be able to provide causal explanations for their actions or decisions when they are critical or difficult to understand.  When a causal explanation answers a "why" question, it can be referred to as a \textit{scientific} explanation. In general, answers to questions such as “How does a personal computer work?” are not considered to be scientific explanations. Such answers are still part of a scientific discipline, but they are descriptive rather than explanatory.

Some other works argue that useful explanations are not only causal explanations but many types of non-causal explanations (e.g., semantic, contrastive, justificatory) may help \cite{sovrano2019difference}.
A pilot user study from Taschdjian (2020)\cite{taschdjian2020did} supports this idea revealing that participants preferred causal explanations over the others only when presented in chart form, whilst they resulted as the least favorite choice when in text form.

\subsection{XAI for causality}
\label{sec:rq1_xfc}
Only three papers openly support a pragmatic line of thinking according to which \acrshort{label_XAI} is a basis for causal inquiry. Indeed, such works recognize certain limits of current XAI methods but approach the discussion pragmatically.

Zednik and Boelsen (2022)\cite{zednik2022scientific} discuss the role of post-hoc analytic techniques from XAI in scientific exploration.
The authors show that XAI techniques, such as \acrshort{label_CFE}s, can serve as a tool for identifying potentially pursue-worthy experimental manipulations within a causal framework and, therefore, for recognizing causal relationships to investigate.
In this regard, the authors remark on an asymmetry between the role of CFEs in \textit{industry} and in \textit{science}. The following two hypothetical scenarios clarify this idea:
\begin{itemize}
    \item \textit{industry}: a bank decides whether to accept or reject a loan application based on an \acrshort{label_AI} agent. A CFE for a rejection case has revealed that doubling the client's income would have led to the acceptance of the loan. Here, the AI agent is not trying to model reality, but it is reality itself. Indeed, a change in the client’s income would actually change the application outcome, meaning that \acrshort{label_CFE}s are \textit{perfect} guides to causal inference.
    \item \textit{science}: an AI agent determines the probability of type-2 diabetes based on patients' features. A \acrshort{label_CFE} for a high-probability case has revealed that losing weight would decrease that probability. Here, the AI agent is trying to model the biological reality of the problem, but still, it remains an approximation. Indeed, it is still possible that losing weight does not actually reduce the probability of type-2 diabetes. That is to say that a change in the model’s behavior does not actually change the way the world works, but at best constitutes a changed representation of how the world could possibly work. In this light, CFEs are \textit{imperfect} guides to causal inference.
\end{itemize}
All in all, it is just because the relevant ML models might not perfectly adhere to reality that the generated \acrshort{label_XAI} explanations only foster scientific \textit{exploration} rather than scientific \textit{explanation}. At most, products of XAI may be thought of as starting points to study potentially causal relationships that have yet to be confirmed.

Similarly, Medianovskyi and Pietarinen (2022)\cite{medianovskyi2022explainable} consider the outputs of the current XAI methods, such as \acrshort{label_CFE}s, to be far from conclusive explanations. Rather, they are initial sketches of possible explanations and invitations to explore further. Those sketches must go through validation processes and experimental procedures before satisfactorily answering the "why" questions, long sought after by \acrshort{label_XAI}.

According to the review by Antoniadi (2021)\cite{antoniadi2021current}, XAI can help to shed some light onto causality. Indeed, since causation involves correlation, an explainable ML model could validate the results provided by causality inference techniques. Additionally, \acrshort{label_XAI} can provide a first intuition of (i.e., generate hypotheses about) possible causal relationships that scientists could then test \cite{arrieta2020explainable,lipton2018mythos}.

\subsection{Causality for XAI}
\label{sec:rq1_cfx}
This third perspective is driven by the idea that causality is propaedeutic to XAI. Indeed, these works either: (i) exploit causality-based concepts to support XAI, (ii) restore the causal foundation of \acrshort{label_CFE}s, or (iii) argue that accessing the causal model of a system is intrinsically explaining the system itself.

\subsubsection{Causal tools for supporting XAI}
Such papers interpret the role of causality in \acrshort{label_XAI} in the sense that some causal concepts, such as structural causal model (\acrshort{label_SCM}) and \textit{do}-operator (\ref{sec:back_causality_how_SCM}) and causal metrics, may bring useful tools for explainability and for finding the causes of AI predictions. 
Regarding the use of \textbf{Structural causal models} to foster XAI, Reimers
et al. (2020)\cite{reimers2020determining} reduce DL to a basic level and frame the constitutional structure of a CNN model into an \acrshort{label_SCM}. In this setting, the random variables represent, for instance, the network's weights and the final prediction, while the functions linking the variables are the \textit{training function} (from labeled images to the network's weights), and the \textit{inference function} (from unlabeled images and weights to the prediction).
By doing so, the authors aim to establish whether a feature is relevant to a \acrshort{label_CNN} prediction by leveraging causal inference and Reichenbach's Common Cause Principle\footnote{According to \cite{reichenbach1956direction}, if two variables A and B are dependent, then there exists a variable C that causes A and B. In particular, C can be identical to A or B meaning that A causes B or B causes A.}.

Lazzari et al. (2022)\cite{lazzari2022predicting}, in order to predict employee turnover, utilize the concept of SCM to revisit and equip the Partial Dependence Plot (\acrshort{label_PDP})\footnote{A visual tool introduced by \cite{friedman2001greedy}, commonly used for model-agnostic XAI, that shows the marginal effect of one feature on the predicted outcome of a system.} method with causal inference properties. Their \acrshort{label_SCM}-based \acrshort{label_PDP} can now go beyond correlation-based analyses and reason about causal interventions, allowing one to test causal claims around factors. This, in turn, provides an intuitive visual tool for interpreting the results and achieving the explainability of automatic decisions.

Regarding \textbf{\textit{do}-operator}, some authors employ this concept to bring the theory of Shapley values a step further\footnote{
The Shapley additive explanation (\acrshort{label_SHAP}) \cite{lundberg2017unified} is a post-hoc XAI method that leverages concepts from cooperative game theory to explain individual predictions. It belongs to \textit{local feature attribution} methods, and specifically to \textit{additive} feature attribution methods, which are those whose attributions sum to a specific value, such as the model’s prediction. \acrshort{label_SHAP} values represent the contribution of each feature to the prediction, ensuring a fair and consistent allocation of importance scores. This method is much used in the tabular data domain and provides a unified measure of feature importance, making it easier to compare and interpret the influence of different features across various models.}.
A fundamental component of Shapley values is to evaluate the reference distribution of dropped (i.e., 'out-of-coalition') features, which has implications on how Shapley values are estimated since this helps define the value function. Based on this distribution, the following variants of Shapley values exist \cite{watson2022rational,heskes2020causal}: \textit{marginal} Shapley values (they ignore relations among features and are used to discover the model's decision boundary), \textit{conditional} Shapley values (they consider feature dependencies and condition by observation), and \textit{interventional} Shapley values. The latter was introduced by Janzing et al. (2020) \cite{janzing2020feature} who replaced conventional \textit{conditioning by observation} with \textit{conditioning by intervention} (\textit{do}-operator).

Extending this concept, Heskes et al. (2020)\cite{heskes2020causal} introduce \textit{causal} Shapley values by explicitly considering the causal relationships between the data in the real world to enhance the explanations. Using the interventional distribution is optimal when, with access to the underlying \acrshort{label_SCM}, one seeks explanations for causal data-generating processes. These methods are required when seeking to use \acrshort{label_XAI} for discovery and/or planning, as they seem to provide sensible, human-like explanations that incorporate causal relationships in the real world. 

Finally, some other works borrow \textbf{metrics from the causal theory} to aid XAI. Specifically, they leverage the Probability of Necessity (\acrshort{label_PN}) and Probability of Sufficiency (\acrshort{label_PS}) from Glymour et al. (2016)\cite{glymour2016causal} and the metric of \textit{responsibility} from  Chockler and Halpern (2004)\cite{chockler2004responsibility}.

\begin{tcolorbox}[width=\textwidth,colback=red!5!white,colframe=red!75!black, before upper={\parindent15pt}]    
   PN is the probability that the garden would not have got wet had the sprinkler not been activated ($Y_0=0$), given that, in fact, the garden did get wet ($Y=1$) and the sprinkler was activated ($X=1$). Mathematically, this becomes: $PN = P(Y_0=0|X=1,Y=1)$.
In other words, this probability quantifies to what extent activating the sprinkler is necessary to get the garden wet, and consequently if other factors (e.g., rain) may have caused the wet garden.

PS is the probability that the garden would have got wet had the sprinkler been activated ($Y_1=1$), given that the sprinkler had not in fact been activated ($X=0$), and the garden did not get wet ($Y=0$). Mathematically, this becomes: $PS = P(Y_1=1|X=0,Y=0)$.
In other words, this probability quantifies to what extent activating the sprinkler is sufficient to wet the garden, and consequently, if there may exist scenarios (e.g., hardware malfunctioning) where activating the sprinkler does not wet the garden.

Responsibility is a quantification of causality, attributing to each actual cause its degree of responsibility $\frac{1}{1+k}$, which is based on the size $k$ of the smallest contingency feature set required to obtain a change in the prediction (i.e., creating a counterfactual dependence).
The degree of responsibility is always between $0$, for variables that have no causal influence on the outcome ($k \rightarrow\infty$), and $1$, for counterfactual causes ($k=1$). Responsibility extends the actual causality framework of Halpern and Pearl (2005)\cite{halpern2005causes}.
\end{tcolorbox}

Regarding \acrshort{label_PN} and \acrshort{label_PS}, two works investigate their implications for \acrshort{label_XAI}. Indeed, such probabilities, often addressed as "probabilities of causation", play a major role in all "attribution" questions.
Watson et al. (2021)\cite{watson2021local} formalize the relationship between existing XAI methods and the probabilities of causation. For instance, they highlight the role of PN and PS in feature attribution methods and \acrshort{label_CFE}s. 
Regarding the former, the authors reformulate the theory of Shapley values in their framework and show how the value function (i.e., the payoff associated with a feature subset) precisely corresponds to the PS of a factor.
Regarding the latter, the authors rewrite the \acrshort{label_CFE} optimization problem with an objective based on the PS of the factor with respect to the opposite of the outcome.
Moreover, Tan et al. (2022)\cite{tan2022learning} borrow PN and PS and adapt them to evaluate the necessity and sufficiency of the explanations extracted for a graph neural network (\acrshort{label_GNN}). This makes it possible to conduct a quantitative evaluation of GNN explanations even without ground-truth explanations for real-world graph datasets.

On the other hand, regarding the metric of responsibility, Chockler et al (2021)\cite{chockler2021explanations} propose \textsc{DC-Causal}, a greedy, compositional, perturbation-based approach to computing explanations for image classification. It leverages causal reasoning in its feature masking phase with the goal of finding causes in input images by causally ranking parts of the input image (i.e., superpixels) according to their responsibility for the classification.
In addition to responsibility, Debbi (2021)\cite{debbi2021causal} borrows from \cite{chockler2004responsibility} the concept of blame to compute visual explanations for \acrshort{label_CNN} decisions. The author abstracts the CNN model into a causal model by virtue of similarity in a hierarchical structure, and filters are considered as actual causes for a decision. First, each filter is assigned a degree of responsibility (i.e., weight) as a measure of its importance to the related class. Then, the responsibilities of these filters are projected back to compute the blame for each region in the input image. The regions with highest blame are returned then as the most important explanations.

\subsubsection{Causal counterfactual explanations}
As noted in Sec.~\ref{sec:review_rationaleObjective}, the \textit{counterfactual} concept seems to belong both to the \acrshort{label_XAI} literature and to the causality literature. 
Some authors remark on how \acrshort{label_CFE}s and \acrshort{label_CF} are two separate concepts \cite{crupi2022leveraging} and, strictly speaking, some would not even call the former \textit{counterfactuals}, precisely to contrast the causal perspective \cite{dash2022evaluating}.
Interestingly, however, these two seemingly separate concepts may be bridged in what we could name structural causal explanations. Indeed, the papers in this sub-cluster present methods for generating \acrshort{label_CF} based on their formal causal definition, restoring the causal underpinning to \acrshort{label_CFE}s by using the concept of \acrshort{label_SCM} and Pearl's CF three-step "recipe" (Section \ref{sec:back_causality_how_SCM}).

In their quest to explain an image classifier's output and its fairness using counterfactual reasoning, Dash et al. (2022)\cite{dash2022evaluating} propose \textsc{ImageCFGen}, a system that combines knowledge from an \acrshort{label_SCM} over image attributes and uses an inference mechanism in a generative adversarial network-like framework to generate counterfactual images.
The proposed architecture directly maps to Pearl's three steps: (i) for \textit{abduction}, an encoder infers the latent vector of an input image coupled with its attributes; (ii) for \textit{action}, a subset of desired attributes is changed and, accordingly, the values of their descendants in the SCM are updated; (iii) for \textit{prediction}, a generator takes the latent vector together with the modified set of attributes and produces a counterfactual image.
A subset of work focuses on a specific aim of the \acrshort{label_XAI} research tightly bound with counterfactual reasoning, i.e., \textit{recourse}. Recourse can be seen as the act of recommending a set of feasible actions to assist an individual to achieve a desired outcome. 
Karimi et al. (2021)\cite{karimi2021algorithmic} argue that the conventional, non-causal \acrshort{label_CFE}s are unable to convey a relevant recourse to the end-user of AI algorithms since they help merely understand rather than act (i.e., inform an individual to where they need to get, but not how to get there). Shifting from explanation to \textit{minimal intervention}, the authors leverage causal reasoning (i.e., tools of \acrshort{label_SCM}s and structural interventions) to incorporate knowledge of the causal relationships governing the world in which actions will be performed. This way, the authors are able to compute what they refer to as \textit{structural CF} by performing the \textit{abduction}-\textit{action}-\textit{prediction} steps and provide \textit{algorithmic recourse}.
Galhotra et al. (2021)\cite{galhotra2021explaining} introduce \textsc{Lewis}, a principled causality-based approach for explaining black-box decision-making systems. They propose to achieve \textit{counterfactual recourse} by solving an optimization problem that searches for minimal interventions on a pre-specified set of actionable variables that have a high probability of producing the algorithm's desired future outcome. Notably, the authors propose a GUI that implements \textsc{Lewis}, of which they show a demo in \cite{wang2021demonstration}.
Crupi et al. (2021)\cite{crupi2022leveraging} also contribute to the recourse objective by proposing \textsc{Ceils}, a new post-hoc method to generate causality-grounded \acrshort{label_CFE}s and recommendations. It involves the creation of an \acrshort{label_SCM} in the latent space, the generation of causality-grounded CFEs, and their translation to the original feature space.
  
\subsubsection{Accessing the causal model is explaining}
\label{par:res_rq1_cfx_accessing}
Part of the work relates to the common thought that accessing the causal model of a system intrinsically explains the system itself. Under this view, two fundamental observations are supported:
\begin{itemize}
    \item when a model is built on a causal structure, it is inherently an interpretable model;
    \item making the inner workings of a causal model directly observable, such as through a directed acyclic graph (DAG) (\ref{sec:back_causality_how_DAG}), makes the model inherently interpretable.
\end{itemize}

Much of the causality theory focuses on explaining observed events, that is, inferring causes from effects. According to its retrospective attribution, causality lies at the heart of explanation-based social constructs such as explainability and, therefore, causal reasoning is an important component of \acrshort{label_XAI} \cite{wu2021methods}.

Ibrahim et al. (2020)\cite{ibrahim2020actual} try to fill the lack in the causality literature of automatic and explicit operationalizations to enable explanations. The authors propose an extensible, open-source, interactive tool (Actual Causality Canvas) able to implement three main activities of causality (causal modeling, context setting, and reasoning) in a unifying framework. According to the authors, what Canvas can provide, through answers to causal queries, largely overlaps with the ultimate goal of \acrshort{label_XAI}, which is providing the end-user with explanations of why particular factors occurred.
Hoque and Mueller (2021)\cite{hoque2021outcome} propose Outcome Explorer, an interactive framework guided by causality, that allows expert and non-expert users to select a dataset, choose a \acrshort{label_CD} algorithm for structure discovery (Section \ref{sec:back_causality_how_BN}), generate (and eventually refine) a causal diagram, and interpret it by setting values to the input features to observe the changes in the outcome.
Katz et al. (2017)\cite{katz2017autonomous} propose an \acrshort{label_XAI} system that encodes the causal relationships between actions, intentions, and goals from an autonomous system and explains them to a human end-user with a cause-effect reasoning mechanism (i.e., causal chains).  
Chatterjee and Dethlefs (2020)\cite{chatterjee2020temporal} exploit the representational power of \acrshort{label_CNN}s with attention, to discover causal relationships across multiple features from observed time-series and historical error logs. The authors believe causal reasoning can enhance the reliability of decision support systems making them more transparent and interpretable.

A subset of publications sees CD as the most appropriate way of operationalizing the idea that accessing the causal model of a system intrinsically explains the system itself. In this regard, all of them utilize \textbf{Bayesian networks (\acrshort{label_BN}s)} (Section \ref{sec:back_causality_how_BN}) as the methodological tool.
Since establishing unique directions for edges based on passive evidence alone may be challenging, knowledge-based constraints can help orient arrows to reflect causal interpretations \cite{cox2021information}. In line with this, some works perform \acrshort{label_CD} with \acrshort{label_BN}s in a mixed approach: on the one hand, they leverage knowledge from domain-experts to outline the causal structure of the system (i.e., finding nodes and related edges); on the other hand, they fit the model parameters on observed, real-world data.

Sahoh and Choksuriwong (2022)\cite{sahoh2022proof} propose a new system to support emergency management (e.g., terrorist events) based on the Deep Event Understanding perspective, introduced in an earlier work of theirs \cite{sahoh2021beyond}. Deep Event Understanding aims to model expert knowledge based on the human learning process and offers explanation abilities that mimic human reasoning. Their model utilizes \acrshort{label_BN}s based on social sensors as an observational resource (i.e., text data from Twitter), with prior knowledge from experts to infer and interpret new information. Their approach helps in recognition of an emergency event and in the uncovering of its possible causes, contributing to the explanation of “why” questions for decision-making.

Sahoh et al. (2022)\cite{sahoh2022causal} propose discovering cause-effect \acrshort{label_ML} models for indoor thermal comfort in Internet of Things (IoT) applications. They employ five different CD algorithms and show how these may converge to the ground-truth \acrshort{label_SCM} of the problem variables obtained from domain experts.
Kliangkhlao et al. (2022)\cite{kliangkhlao2022design} introduce a BN model for agricultural supply chain applications, initially constructed from causal assumptions from expert qualitative knowledge, which conventional ML cannot reasonably conceive. Therefore, a data-driven approach using observational evidence is employed to encode these causal assumptions into quantitative knowledge (i.e., parameter fitting). The authors report their system constitutes a framework that is able to provide reasonable explanations of events for decision-makers.

In \cite{zapaishchykova2021interpretable} the authors leverage the respective strengths of \acrshort{label_DL} for feature extraction and \acrshort{label_BN}s for causal inference, achieving an automatic and interpretable system for grading pelvic fractures from \acrshort{label_CT} images. The \acrshort{label_BN} model is constructed upon variables extracted with the neural network, together with a variable from the clinical practice (i.e., patient age). By doing so, the authors believe that the framework provides a transparent inference pipeline supplying fracture location and type, by establishing causal relationships between trauma classification and fracture presence.

Yang et al. (2022)\cite{yang2022interpretable} propose a new process monitoring scheme based on \acrshort{label_BN}s to explain (diagnose) a detected fault and promote decision-making. Their system allows the identification of the root cause (i.e., labeling the abnormal variables)  so that the result of the analysis can be linked to the repairing action, reducing the investigation time. Among one of their use cases, the authors fit a BN model on observed, real-world data for manufacturing fault events. During this CD process, they employ a blacklist obtained from domain experts to exclude causally-unfeasible relationships.

\section{Results of software tools collection}
\label{sec:review_software}
We hereby present a summary of the main data mining software tools collected within the cohort of papers. Table~\ref{tab:software_tools_qualities} comprises tools for
performing \acrshort{label_CD} with \acrshort{label_BN}s (i.e., PySMILE\footnote{\url{https://www.bayesfusion.com/smile/}}, CausalNex\footnote{\url{https://causalnex.readthedocs.io/en/latest}}, bnlearn\footnote{\url{https://www.bnlearn.com}}, CompareCausalNetworks\footnote{\url{https://cran.r-project.org/web/packages/CompareCausalNetworks/}}, CaMML\footnote{\url{https://bayesian-intelligence.com/software/}}, Python Causal Discovery Toolbox\footnote{\url{https://fentechsolutions.github.io/CausalDiscoveryToolbox/html/index.html}}, and Tetrad\footnote{\url{https://htmlpreview.github.io/?https:///github.com/cmu-phil/tetrad/blob/development/docs/manual/index.html}}),
creating and analysing SCMs (i.e., IBM\textsuperscript{\tiny\textregistered} SPSS\textsuperscript{\tiny\textregistered} Amos\footnote{\url{https://www.ibm.com/products/structural-equation-modeling-sem}}, lavaan\footnote{\url{https://cran.r-project.org/web/packages/lavaan/index.html}}, and semopy\footnote{\url{https://semopy.com/}}),
and editing and analyzing DAGs (i.e., DAGitty\footnote{\url{http://www.dagitty.net/}}).
We believe this list of software solutions may be of interest to \acrshort{label_AI} practitioners in helping them save valuable time when choosing the right tool to automate causal tasks.

The most popular choice is an open-source license type, and this reflects the great interest in sharing code and information across the \acrshort{label_AI} research community.
The first benefit of that is flexibility. Researchers often need to access the source code of software implementations to eventually customize its functionalities according to a desired (yet not implemented) purpose. This would be highly unfeasible with closed and commercial software.
Another advantage of having open-source implementations is software security. According to Linus's law, "given enough eyeballs, all bugs are shallow" \cite{raymond1999cathedral}. That is, when all the source code for a project is made open to professionals worldwide, it is more likely that security checks could discover eventual flaws.

Furthermore, Table~\ref{tab:software_tools_qualities} shows that the \acrshort{label_CLI} is the preferred frontend interface across such solutions. This aspect also reflects the AI research community viewpoint. Opting for CLI over the GUI brings some advantages, such as faster and more efficient computing, easier handling of repetitive tasks, lighter memory usage, and availability of the history of commands.
On the other hand, using CLI involves a steeper learning curve associated with memorizing commands and complex arguments, together with the need for correct syntax. This may explain why \acrshort{label_GUI} is preferred in cases where the end-user does not have a programming background. Typical examples of that include physicians in healthcare facilities or product managers in finance companies, who prefer, in general, a more user-friendly product.

\begin{table}
\footnotesize 
\caption{Software tools within the cohort of papers useful to automate causal tasks. BSD: Berkeley Software Distribution, CD: causal discovery, CLI: common line interface, GPL: General Public License, GUI: graphical user interface.}
\label{tab:software_tools_qualities}
\begin{tabularx}{\textwidth}{XXXXX}
\hline
\textbf{Name} & \textbf{License}\newline \textbf{type} & \textbf{Release}\newline \textbf{paper} & \textbf{Frontend}\newline \textbf{interface} & \textbf{Main\newline purpose}\\
\hline
\textit{bnlearn} & Open-source\newline (GPL) & \cite{scutari2010learning} & CLI (R) & BNs for CD\\
\hline
\textit{CaMML} by Bayesian Intelligence Pty Ltd & Open-source\newline (BSD) & n.a. & CLI (Bash)\newline and GUI & BNs for CD\\
\hline
\textit{CausalNex} by QuantumBlack, AI by McKinsey & Open-source\newline (Apache 2.0) & n.a. & CLI (Python) & BNs for CD\\
\hline
\textit{Compare Causal Networks} & Open-source\newline (GPL) & \cite{heinze2018causal} & CLI (R) & BNs for CD\\
\hline
\textit{DAGgity} & Open-source\newline (GPL) & \cite{textor2016robust} & CLI (R) and GUI & Create and analyze causal diagrams\\
\hline
\textit{IBM SPSS Amos} by IBM Corp. & Commercial & n.a. & GUI & Create and analyze SCMs\\
\hline
\textit{lavaan} & Open-source\newline (GPL) & \cite{rosseel2012lavaan} & CLI (R) & Create and analyze SCMs\\
\hline
 \textit{PySMILE} by BayesFusion LLC & Commercial & n.a. & CLI (Python) & BNs for CD\\
\hline
\textit{Python Causal Discovery Toolbox} by Fentech & Open-source\newline (MIT) & \textit{\cite{kalainathan2020causal}} & CLI (Python) & BNs for CD\\
\hline
\textit{semopy} & Open-source\newline (MIT) & \cite{semopy}\newline \cite{meshcheryakov2021semopy} & CLI (Python) & Create and analyze SCMs\\
\hline
\textit{Tetrad} & Open-source\newline (GPL) & \cite{ramsey2018tetrad} & GUI & BNs for CD\\
 \hline
\end{tabularx}
\normalsize
\end{table}

\section{Discussion}
\label{sec:review_discussion}
The concepts of causation and explanation have always been part of human nature, from influencing the philosophy of science to impacting the data mining process for knowledge discovery of today's \acrshort{label_AI}.
To investigate the relationship between causality and \acrshort{label_XAI}, we provided a unified view of the two fields by highlighting which methodologies could be adopted to approach the bridge between them and uncovering possible limitations.
As a result of the analysis, we found and formalized three main perspectives.

The "\textit{Critics to XAI under the causality lens}" perspective analyses how the lack of causality is one of the major limitations of current (X)AI approaches as well as the "optimal" forms to provide explanations.
Regarding the former, traditional AI systems are only able to detect correlation instead of true causation, which affects the robustness of models against adversarial attacks and of the produced explanations. This is of concern since pure associations are not enough to accurately describe causal effects.
Regarding the latter, optimal explanations may be characterized by being expressed according to the explainee's knowledge and domain terminology and being able to explain many effects with few causes. However, it is debated whether causal explanations (i.e., causal inference chains to a prediction) are the only useful ones in the \acrshort{label_XAI} landscape. 
This first perspective states the problem and serves as a watch out.

The "\textit{XAI for causality}" perspective openly claims that \acrshort{label_XAI} may be a basis for further causal inquiry. 
Despite the recognized limits of XAI explanations, they may be pragmatically thought of as starting points to generate hypotheses about possible causal relationships that scientists could then confirm. That is, \acrshort{label_XAI} can only foster scientific exploration, rather than scientific explanation.
Although underrepresented in the final cohort, this perspective suggests a really thoughtful idea in our opinion.

The "\textit{Causality for XAI}" perspective supports the idea that causality is propaedeutic to XAI. 
This is realized in three manners.
First, some causal concepts (i.e., \acrshort{label_SCM} and \textit{do}-operator) are leveraged to revisit existing XAI methods to empower them with causal inference properties. 
Second, the formal causal definition of CF (Sec.~\ref{sec:review_rationaleObjective}) is invoked to generate causal CFEs using the \acrshort{label_SCM} tool, which may also enable recourse. 
Third, and lastly, it is argued that, when a model is built on a causal structure, it is inherently an interpretable model. In a related way, making the inner workings of a causal model directly observable (e.g., through a \acrshort{label_DAG}) makes the model inherently interpretable.

Among the three main perspectives, we believe "\textit{Causality for XAI}" to be the most promising one. Naturally, it comes with limitations. Much work in causal modeling is based on specific and (by far) non-unique causal views of the problems at hand. Interventions and CF make sense as long as the specified causal graph makes sense, which may hinder the generalization of their results. Overall, their causal claims depend on strong and often non-testable assumptions about the underlying data-generating process. 
On the other hand, however, this may be in line with what already happens in our life, and we should not request from AI more than we request from human beings. 
Another weak point is the interpretability of a causal model with hundreds of variables. In this scenario, a DAG would encode too much information and the complexity of the underlying \acrshort{label_SCM} would rise exponentially with the number of modeled variables. This, however, is common to other simpler and more traditional approaches such as Decision Trees with hundreds of nodes. 

We acknowledge three main limitations that may have led us to miss publications that could have potentially been included in the review: (i) the exclusion of non-peer-reviewed e-prints, (ii) the usage of only four databases, and (iii) not having extracted any references from the collected papers to enrich our search. The latter was motivated by the fact that, this being an unexplored field, the papers we collected were sufficient and significant enough to produce a first scenario. Obviously, as with any human-made assignment, the search process for relevant material may have been affected by the cognitive bias of the authors, who have brought their knowledge and assumptions in the study. 

We believe our results could be useful to a wide spectrum of readers, from upper-level undergraduate students to research managers in the industry, and have implications for practice, policy, and future research. Indeed, having a clear view of how the two concepts of causality and \acrshort{label_XAI} are related can benefit both areas individually, as well as the joint research field. Considering our conceptual framework, future publications may be framed in a precise and rigorous way and have the potential to expand (or generate new flavors of) one of the identified perspectives.

\subsection{Is all this still valid in 2024?}
Since our research covers papers up to August 2022, we wanted to assess whether the considerations made were still valid for papers in the latest published literature. Given the exponential interest in researching causality and \acrshort{label_XAI} in the \acrshort{label_AI} domain, systematically covering the new literature that appeared in the last two years would have been not trivial. However, we have been committed to investigating whether our conclusions/suggestions remain the same in 2024 and thus performed a new analysis in August 2024 on selected recent works, specifically those we believe have had a significant impact. In the following paragraphs, we discuss and map them to our findings, highlighting similarities and differences with the developed perspectives.

New evidence enriched our “\textbf{Critics to XAI under the causality lens}” perspective. Specifically, to the “\textit{Lack of Causality}” sub-cluster, Pichler and Hartig (2023)\cite{pichler2023machine} remark how \acrshort{label_XAI} explanations can simply reveal which are the most used features by the model, which may not correspond to true causation, as ML models tend to rely on spurious correlations between outcomes and features. According to the authors, it is often easier for a model to get good predictions for the wrong reasons, as the best predictive model (i.e., smallest prediction error) need not be the true causal model (i.e., correct effect estimates).

In their simulated examples, they dive into discussing causal versus predictive model-building strategies. On the one hand, causal inference aims to establish a correct hypothesis about the causal structure to obtain correct effect estimates, often at the expense of including otherwise uninteresting collinear features to control for confounding. On the other hand, the goal of predictive modeling is to minimize the prediction error, which is achieved by removing collinear features since they increase uncertainties while contributing relatively little to the prediction (i.e., their effects can be “emulated” by other features). Consequently, off-the-shelf post-hoc \acrshort{label_XAI} methods built on purely predictive models fail to capture the causation between observed variables and predictions. This makes it questionable whether received explanations are suitable for guiding people’s actions.
On a similar note, Ghaffarian et al. (2023)\cite{ghaffarian2023explainable} point out how CFEs might miss the intricate causal chains of an alteration by solely looking at direct feature modifications. Indeed, by focusing on the change in one feature at a time, XAI-based counterfactuals might give the impression that factors operate in isolation, whereas, in reality, real-world factors are deeply interconnected. Accordingly, the authors recommend incorporating causal inference methods to identify factors contributing to an event and discuss that in the field of disaster risk management (e.g., response, recovery, and prevention strategies).

Regarding the "\textit{Form of the Explanations}” sub-cluster, Danry et al. (2023)\cite{danry2023don} investigate a novel form of causal explanations via a user study with 204 subjects. Building on the Socratic approach to understanding, they propose AI-Framed Questioning, where an AI system asks participants about the causal link between a reason and the system label. For instance, given the logically invalid statement “Violent videogames cause people to be aggressive in the real world. A gamer stabbed another after being beaten in the online game Counterstrike”, an AI-Framed Questioning would be “if one person played videogames and was aggressive, \textit{does it follow that} everyone that plays videogames will be aggressive?”. Conversely, the basic, affirmative form of explanation that simply tells people what and why, would take the form “if one person played videogames and was aggressive, \textit{it does not follow that} everyone that plays videogames will be aggressive”. The authors found that the latter form may lead to a passive acceptance of the statements. On the other hand, the questioning form of explanation actively elicits users’ thinking, improves human discernment outcomes, and helps them assess whether a statement is logically valid.

As for our “\textbf{XAI for Causality}”, Longo et al. (2024)\cite{longo2024explainable} extensively discuss the open challenges and interdisciplinary research directions of \acrshort{label_XAI}. Notably, they support the idea that uncovering causal connections learned through a model via explanations is a fundamental hope associated with XAI. Indeed, it would be desirable to reach that goal with simpler XAI approaches. However, as already highlighted by other papers in this perspective, attaining a conclusive explanation perfectly adhering to the causal world is challenging. Rather, XAI can be a pragmatic tool to foster causal inquiry. Pichler and Hartig (2023)\cite{pichler2023machine} corroborate that view and suggest using ML and XAI to identify interesting feature patterns, especially if we have a high dimensional dataset, and test them later in a confirmatory analysis (i.e., scientific \textit{exploration} over scientific \textit{explanation}).

Lastly, we augmented our “\textbf{Causality for XAI}” perspective with new works. Within the “\textit{Causal tools for supporting XAI}” sub-cluster, Wang et al. (2023)\cite{wang2023time} propose to leverage the causal discovery tool for better interpretability via XAI methods. Specifically, they adopt Direct-LiNGAM \cite{shimizu2011directlingam} not only for the selection (screening) of input features, but also within model construction to mine causal links between those input features (relationship mining). Indeed, they append the adjacency matrix containing the causal strengths between the selected input features to the input matrix to make the prediction more robust and accurate. This proved to be beneficial to the subsequent \acrshort{label_SHAP} interpretation step, where the authors explored the impact of the selected features on the predictions and quantified the degree of that influence more robustly. Markou et al. (2024)\cite{markou2024framework} also exploit causal reasoning for improved XAI, by constraining the learning phase of a variational autoencoder to generate feasible counterfactual examples. Indeed, in addition to the usual sparsity, validity, and distance loss terms for \acrshort{label_CFE}s, they design proper terms to guide the model in generating examples that satisfy real-world causal constraints. Examples of that include the attribute “age” in a dataset of subjects' demographics ----- a CFE suggesting decreasing “age” will be considered as infeasible, since it violates the logical causal constraint that age can only increase over time. All in all, the authors improved the feasibility of the produced \acrshort{label_CFE}s, which were able to preserve the true logical causal relationships.
As we shall see in Chapter \ref{chap:mulcat_ICCV_ESWA}, two of our recent works (\cite{carloni2023causality} and \cite{carloni2024exploiting}) are mapped to this cluster as we propose causality-driven methods to learn better CNN models and produce more robust \acrshort{label_XAI} explanations. This is achieved by modeling feature co-occurrence and designing attention-inspired modules to weigh feature maps according to their causal influence in the scene. This way, the models learn what to enhance or suppress. As a result, the classification performance increased and the CAM explanations became more reliable and robust, focusing on relevant parts of the image.
Baron (2023)\cite{baron2023explainable} is the next new paper that entered this sub-cluster. The author builds on formalizations of the causal framework by Pearl (2009)\cite{pearl2009causality} and Woodward (2005)\cite{woodward2005making} and introduces “causal certification”, a novel method for determining whether an existing approach to \acrshort{label_XAI} always delivers genuine causal information (i.e., truly knowing the real causes of something). If it does, the method under investigation is said to pass the basic test. Moreover, if the XAI method always provides a complete account of the causal factors that led a model to deliver a particular outcome, it is said to pass also the \textit{complete} test. Notably, the author finds that traditional \acrshort{label_CFE} methods (Sections \ref{sec:back_XAI_methods_CFE} and \ref{sec:review_rationaleObjective}) pass the test for \textit{basic} causal certification, and this guarantees that at least one of the system’s variables constitutes a genuine cause, while they do not pass the \textit{complete} test, and this prevents the user in knowing whether any causal factors have been left out.

In the same work, Baron (2023)\cite{baron2023explainable} goes further and proposes a hybrid two-step approach to overcome the above-mentioned limitation of traditional CFEs. We deem that discussion relevant to enrich our “\textit{Causal counterfactual explanations}” sub-cluster. In the first step, the interventional account of the Pearl-Woodward framework is employed to every combination of the model’s features to find the complete list of causal information. This is achieved by changing each variable while holding the others fixed and seeing which ones change the output, thus revealing \textit{individual} causes (which can flip the outcome by their own) and \textit{parts} of causes (which are variables that need to be varied in concert with others in the set). In the second step, knowledge about the interventions is used to identify the variables that are plausible and actionable among those selected in the previous step. As a result, the user receives the list of factors first outlining the individual causes, then the parts of causes, and then the statement guaranteeing that no causes have been missed. The two steps help the user in causal understanding, contesting decisions, and getting practical advice.

A new line of research emerged in recent works by Von Kügelgen et al. (2023)\cite{von2023backtracking}, Kladny et al. (2024)\cite{kladny2024deep}, and Strobl (2024)\cite{strobl2024counterfactual}, which entered the "\textit{Causal counterfactual explanations}" sub-cluster as an \textit{avant-garde}. Indeed, Von Kügelgen et al. (2023)\cite{von2023backtracking} explore and formalize, for the first time, within the \acrshort{label_SCM} framework, the “backtracking counterfactuals”, an alternative mode of counterfactual reasoning. Unlike traditional \acrshort{label_CF}s, which involve intervening on the causal laws while fixing the values of exogenous variables, the “backtracking
counterfactuals” fix the causal laws while differences to the factual world are "backtracked" to altered initial conditions (i.e., updating upstream exogenous variables). As a result, \acrshort{label_CFE}s grounded on them
tend to be more realistic by staying closer to the observational distribution of the data manifold. However, the authors point out that this new kind of counterfactual formalization is not a replacement of Pearl’s interventional one, but rather their semantics is fit for different reasoning scenarios. Interventional CFs are suitable when the structural information of the system is already known, and one wishes to determine the effects of interventions.
On the other hand, backtracking CFs can be used to determine, diagnose, or even challenge the causes of a given event by tracing back the effects to the causes. Kladny et al. (2024)\cite{kladny2024deep} derive formulations compute those backtracking counterfactuals for deep \acrshort{label_SCM}s and propose “DeepBC”, a family of practical algorithms to efficiently attain them. Lastly, Strobl (2024)\cite{strobl2024counterfactual} proposes a way to operate the novel backtracking formulation in the biomedical domain, specifically in identifying patient-specific root causes of disease. As reported, the backtracking interpretation reveals to be powerful and matches the way clinicians identify causes of diseases by backtracking on factual data given a medically established causal model.

In the end, this new analysis of recent literature revealed our proposed perspectives and sub-clusters were well-designed. We could effortlessly map novel evidence, consolidating our findings while confirming the relative prominence of the perspective. As could be expected, the “XAI for causality” perspective was still under-represented, being enriched by only few papers. On the opposite side, the “Causality for XAI” perspective remained the most attractive account, even accommodating cutting-edge research lines.

\section{Conclusion and Summary}
The concepts of causation and explanation have always been part of human nature, from influencing the philosophy of science to impacting the data mining process for knowledge discovery of today’s AI. In this study, we investigated the relationship between causality and \acrshort{label_XAI} by exploring the literature from both theoretical and methodological viewpoints to reveal whether a dependent relationship between the two research fields exists. 

As a result of the analysis, we found and formalized three main perspectives. The first one considers the lack of causality as a crucial limitation of current (X)AI approaches and explores the "optimal" form of explanations. In the second perspective, a pragmatic view is proposed that considers XAI as a tool to promote scientific exploration for causal inquiry via the identification of pursue-worthy experimental manipulations. Lastly, from the third perspective, the idea that causality is propaedeutic to XAI is supported. That is achieved in three ways: exploiting concepts borrowed from causality to support or improve XAI, utilizing counterfactuals for explainability, and considering accessing a causal model as explaining itself. To complement our analysis, we also provide relevant software solutions to automate causal tasks.

Our work provides a unified view of the two fields of causality and XAI by highlighting potential domain bridges and uncovering possible limitations. The perspectives we developed are a well-designed and versatile scaffold, and this is further supported by the high consistency found when performing the confirmatory analysis on the new literature, which appeared much later than our initial query.
All in all, this work disclosed how causality and XAI may be related in a profound way. The “Causality for XAI” perspective has excellent potential to produce significant scientific results. We suggest readers monitor this trend, as we expect it to flourish the most soon.

As we shall see soon, the perspectives we synthesised in this Chapter influenced our further DL network proposals in Chapters \ref{chap:mulcat_ICCV_ESWA}, \ref{chap:crocodile}, and \ref{chap:cocoreco_ECCV}. Indeed, we build on the lack of causality of current systems (first perspective) and then operationalize the idea that causality is propaedeutic to XAI (third perspective) by exploiting causal inference and feature co-occurrence, drawing attribute causal graphs, injecting prior knowledge, and implementing do-calculus and backdoor adjustment. Overall, we show how such modifications supported or improved models' explainability and robustness.
\chapter{Causality-driven CNNs: Exploiting Feature Co-occurrence in Medical Images}
\label{chap:mulcat_ICCV_ESWA}
\begin{figure}[h!]
\includegraphics[width=\textwidth]{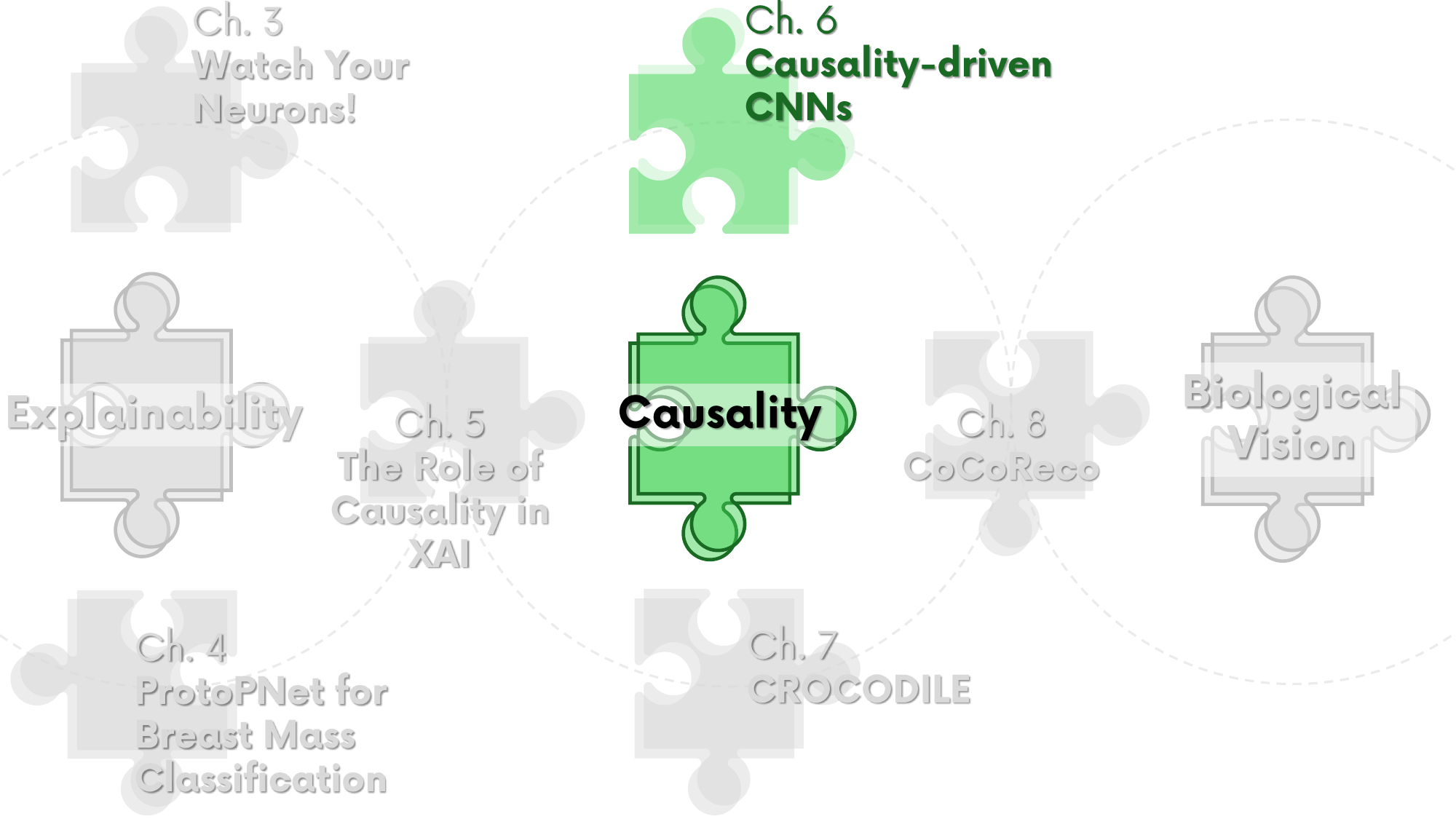}
\end{figure}

Having explored the literature at the intersection of \acrshort{label_XAI} and causality (Chapter \ref{chap:review_XAI_causality}), we wanted to dive experimentally into the latter field. In this chapter, we present our study and development of a novel technique to discover and exploit weak causal signals directly from images via neural networks for classification purposes.
%
%

As we have seen also in Chapter \ref{chap:background}, the concepts of causal inference and causal reasoning have received increasing attention across the \acrshort{label_AI} community in recent years. This trend began with the very first work of the computer scientist Judea Pearl on Bayesian networks and the mathematical formalization of causality, which enabled the creation of computational systems that can automatically model causality  \cite{pearl1985bayesian, pearl2009causality, pearl2018book}. Today, we have encouraging examples of the integration of causality into the \acrshort{label_ML} community \cite{luo2020causal, scholkopf2022causality} and the \acrshort{label_DL} research \cite{berrevoets2024causal}, with extensions to causal representation learning \cite{scholkopf2021toward}, causal discovery under distribution shifts \cite{perry2022causal} and with incomplete data \cite{wang2020causal}. Unfortunately, along this line of research, the processed data are often tabular and structured, simulated instead of real, and accompanied by \textit{a priori} information about the process that generated them.

Unlike tabular data, when it comes to images, their representation does not include any explicit indications regarding objects or patterns. Instead, individual pixels are used to convey a particular scene visually, and image datasets do not usually provide labels describing the objects’ dispositions. Additionally, unlike video frames, a single image cannot reveal the dynamics of the appearance and change of objects in a scene. These critical issues could explain why images have been neglected by research on the tabular causal discovery, where instead, there are established algorithms \cite{spirtes1991algorithm,spirtes2000causation,chickering2002optimal}. 
A particular case would be \textbf{discovering hidden causalities among objects} in an image dataset.

In this work, we propose a way to automatically discover and exploit weak causal signals within images without requiring prior knowledge and use them to enhance CNN classifiers.
By combining a regular \acrshort{label_CNN} with the proposed causality-factors extraction module, we present a new scheme based on feature map enhancement to enable “causality-driven” \acrshort{label_CNN}s. This way, we weigh each feature map according to its causal influence in the scene in an attention-inspired fashion, modeling how the presence of a feature in one part of the image affects the appearance of another feature in a different part of the image.
We frame our system as an automatic cancer diagnosis model from medical images since we study the efficacy of the proposed methods on publicly available datasets of prostate \acrshort{label_MRI} images and breast histopathology slides. 
Our extensive empirical evaluations included also developing different architecture variants and empirically evaluate all the models on two learning regimes: fully-supervised and few-shot learning.
Moreover, we studied the effectiveness of our module when integrated with existing attention-based solutions, and conduct ablation studies.
Besides investigating the quantitative aspect, we also explored the concept of explainability of \acrshort{label_AI} in our evaluation. The results on \acrshort{label_CAM}s demonstrate that our method improves classification and produces more robust predictions by focusing on the relevant parts of the image, thus enhancing reliability, trustworthiness, and user confidence.

This Chapter is structured as follows. First, in Section \ref{sec:preliminaries}, we provide the concepts behind the causal signals' interpretation in images. Then, Section \ref{sec:methods} opens by describing the novelty of our work, namely the methodological framework and the causality-factors extractor module we introduced. We also illustrate the datasets, the training scheme, and the evaluation details. Later, we present our main results in Section \ref{sec:results}, explore the significance of our findings in the general discussion in Section \ref{sec:discussion}, and pull the threads in Section \ref{sec:conclusions}.

The content of this Chapter is based on the following publications:
\begin{itemize}
    \item Carloni, G., Pachetti, E., \& Colantonio, S. (2023). "Causality-driven one-shot learning for prostate cancer grading from MRI". In \textit{Proceedings of the IEEE/CVF international conference on computer vision} (pp. 2616-2624). \cite{carloni2023causality}
    \item Carloni, G., \& Colantonio, S. (2024). "Exploiting causality signals in medical images: A pilot study with empirical results". In \textit{Expert Systems with Applications}, 249, 123433. \cite{carloni2024exploiting}
    
    \item and the corresponding Python/Pytorch code can be found on my GitHub page at:
\url{https://github.com/gianlucarloni/causality_conv_nets}.
\end{itemize} 

\section{Causality Signals in Images}\label{sec:preliminaries}
Lopez-Paz et al. (2017) \cite{lopez2017discovering} first proposed the idea of “causal disposition” as a simple way to understand the hidden causes in images instead of using the methods of do-calculus and causal graphs from Pearl’s framework \cite{pearl2009causality,pearl2018book}. 
In their view, by counting the number $C(A,B)$ of images in which the causal dispositions of artifacts $A$ and $B$ are such that $B$ disappears if one removes $A$, one can assume that artifact $A$ causes the presence of artifact $B$ when $C(A,B)$ is greater than the converse $C(B,A)$. For instance, they argue that the presence of a car causes the presence of a wheel, but not the other way around, because removing the car would make the wheel disappear, but removing the wheel would not make the car disappear. By studying such asymmetries, the authors find the causal direction between pairs of random variables representing features of objects and their contexts in images. Although the causal disposition concept is more primitive than the interventional approach, it could be the only way to proceed with limited a priori information.
This concept leads to the intuition that any causal disposition induces a set of asymmetric causal relationships between the artifacts from an image (features, object categories, etc.) that represent (weak) causality signals regarding the real-world scene. A point of contact with machine vision systems would be to automatically infer such asymmetries from an observed image dataset.

Convolutional neural networks obtain the essential features required for classification not directly from the pixel representation of the input image but through a series of convolution and pooling operations designed to capture meaningful features from the image (ref. Sec. \ref{sec:back_DL_MLP_CNN_cnn}). Convolution layers are responsible for summarizing the presence of specific features in the image and generating a set of feature maps accordingly. Pooling consolidates the presence of particular features within groups of neighboring pixels in square-shaped sub-regions of the feature map.
When a feature map $F^i$ contains only non-negative numbers (e.g., thanks to \acrshort{label_ReLU} functions) and is normalized in the interval $[0,1]$, we can interpret its values as probabilities of that feature to be present in a specific location. For instance, $F^i_{r,c}$ is the probability that the feature $i$ is recognized at coordinates ${r,c}$.

By assuming that the last convolutional layer outputs and localizes to some extent the object-like features, we may modify the architecture of a \acrshort{label_CNN} such that the $n \times n$ feature maps ($F^1,F^2,\dots F^k$) obtained from that layer got fed into a new module that computes pairwise conditional probabilities of the feature maps. The resulting $k \times k$ map would represent the causality estimates for the features and be called \textit{causality map}. 
Given a pair of feature maps $F^i$ and $F^j$ and the formulation that connects conditional probability with joint probability, $P(F^i|F^j) = \frac{P(F^i,F^j)}{P(F^j)}$, \cite{terziyan2023causality} suggest to heuristically estimate this quantity by adopting two possible methods, namely \textit{Max} and \textit{Lehmer}.
The \textit{Max} method considers the joint probability to be the maximal presence of both features in the image (each one in its location):
\begin{equation}
    P(F^i|F^j) = \frac{(\max_{r,c} F^i_{r,c})\cdot (\max_{r,c} F^j_{r,c})}{\sum_{r,c} F^j_{r,c}}
    \label{eq:causality_method_max}
\end{equation}
On the other hand, the \textit{Lehmer} method entails computing 
\begin{equation}
    P(F^i|F^j)_p = \frac{LM_p(F^i \times F^j)}{LM_p(F^j)}
    \label{eq:causality_method_lehmer}
\end{equation}
where $F^i \times F^j$ is a vector of $n^4$ pairwise multiplications between each element of the two $n \times n$ feature maps, while $LM_p$ is the generalized Lehmer mean function \cite{bullen2003handbook} with parameter $p$, which is an alternative to power means for interpolating between minimum and maximum of a vector $x$ via harmonic mean ($p=-1$), arithmetic mean ($p=0$), and contraharmonic mean ($p=1$):
$LM_p(x) = \frac{\sum_{k=1}^n x_k^{p+1}}{\sum_{k=1}^n x_k^p}$.

\begin{figure}
	\centering
		\includegraphics[width=0.99\textwidth]{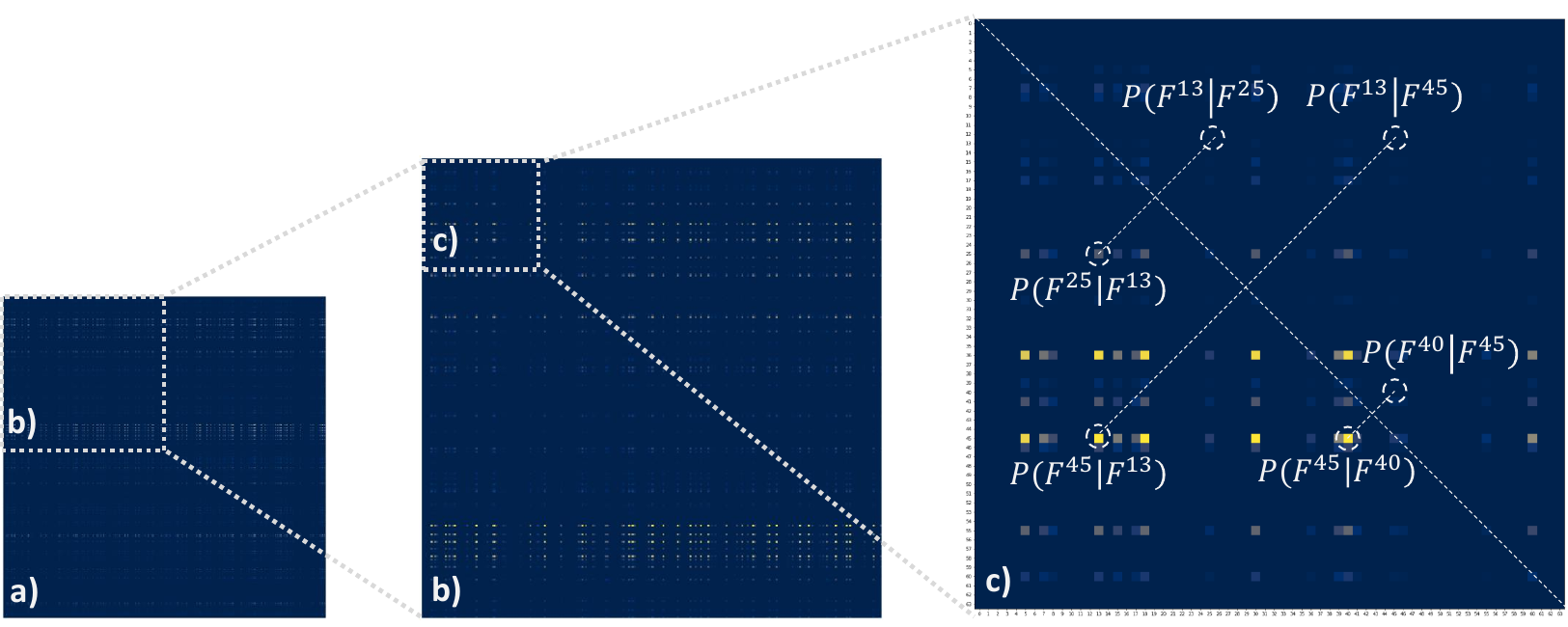}
	  \caption{Zoomed-in visualizations of a sample causality map computed with Eq. \ref{eq:causality_method_max} on $512$ feature maps extracted from an input image. (a) $512 \times 512$ original causality map; (b) $256 \times 256$ zoom-in of (a); (c) $64 \times 64$ zoom-in of (b), where dashed circles indicate exemplar elements and their corresponding elements opposite the main diagonal, representing conditional asymmetries of the type $P(F^i|F^j) \neq P(F^j|F^i)$. We can see that, for instance, $P(F^{25}|F^{13}) > P(F^{13}|F^{25})$, that is $F^{25} \rightarrow F^{13}$, and $P(F^{45}|F^{40}) > P(F^{40}|F^{45})$, that is $F^{45} \rightarrow F^{40}$.}
   \label{fig_cmapvisual}
\end{figure}

Equations \ref{eq:causality_method_max} and \ref{eq:causality_method_lehmer} could be used to estimate asymmetric causal relationships between features $F^i$ and $F^j$, since, in general, $P(F^i|F^j) \neq P(F^j|F^i)$. By computing these quantities for every pair $i$ and $j$ of the $k$ feature maps, the $k \times k$ causality map is obtained. We interpret asymmetries in such probability estimates as weak causality signals between features, as they provide some information on the cause-effect of the appearance of a feature in one place of the image, given the presence of another feature within some other places of the image. Accordingly, a feature may be deemed to be the reason for another feature when $P(F^i|F^j) > P(F^j|F^i)$, that is ($F^i \rightarrow F^j$), and vice versa.  
As an example, Figure \ref{fig_cmapvisual} depicts a causality map to give a visual interpretation of this concept.  

In this work, we integrate a regular CNN with a new causality-extraction module to explore the features and causal relationships between them extracted during training. The previous work that inspired us \cite{terziyan2023causality} is preliminary, and we introduce a novel attention-like scheme based on feature maps enhancement to enable “causality-driven” \acrshort{label_CNN}s, providing an extensive empirical evaluation of the impact of this new introduction on real data.
We hypothesize that it would be possible and reasonable to get some weak causality signals from the individual images of some medical datasets without adding primary expert knowledge and leverage them to better guide the learning phase. 
Ultimately, a model trained in such a manner would exploit weak causal dispositions of objects in the image scene to distinguish the tumor status of a medical image.

\section{Material and Methods}\label{sec:methods}
\subsection{Embedding Causality into \acrshort{label_CNN}s}
\label{sec:embedding_causality}
Usually, a CNN performs image classification based on the final set of (flattened) $n\times n \times k$ feature maps obtained just before the dense layers that constitute the classifier. In the following, we describe how the architecture of such a regular \acrshort{label_CNN} (baseline) might be modified to make the classifier consider the information entailed in the estimated causality map.

\subsubsection{Feature Concatenation}
\begin{figure}
	\centering
		\includegraphics[width=0.99\textwidth]{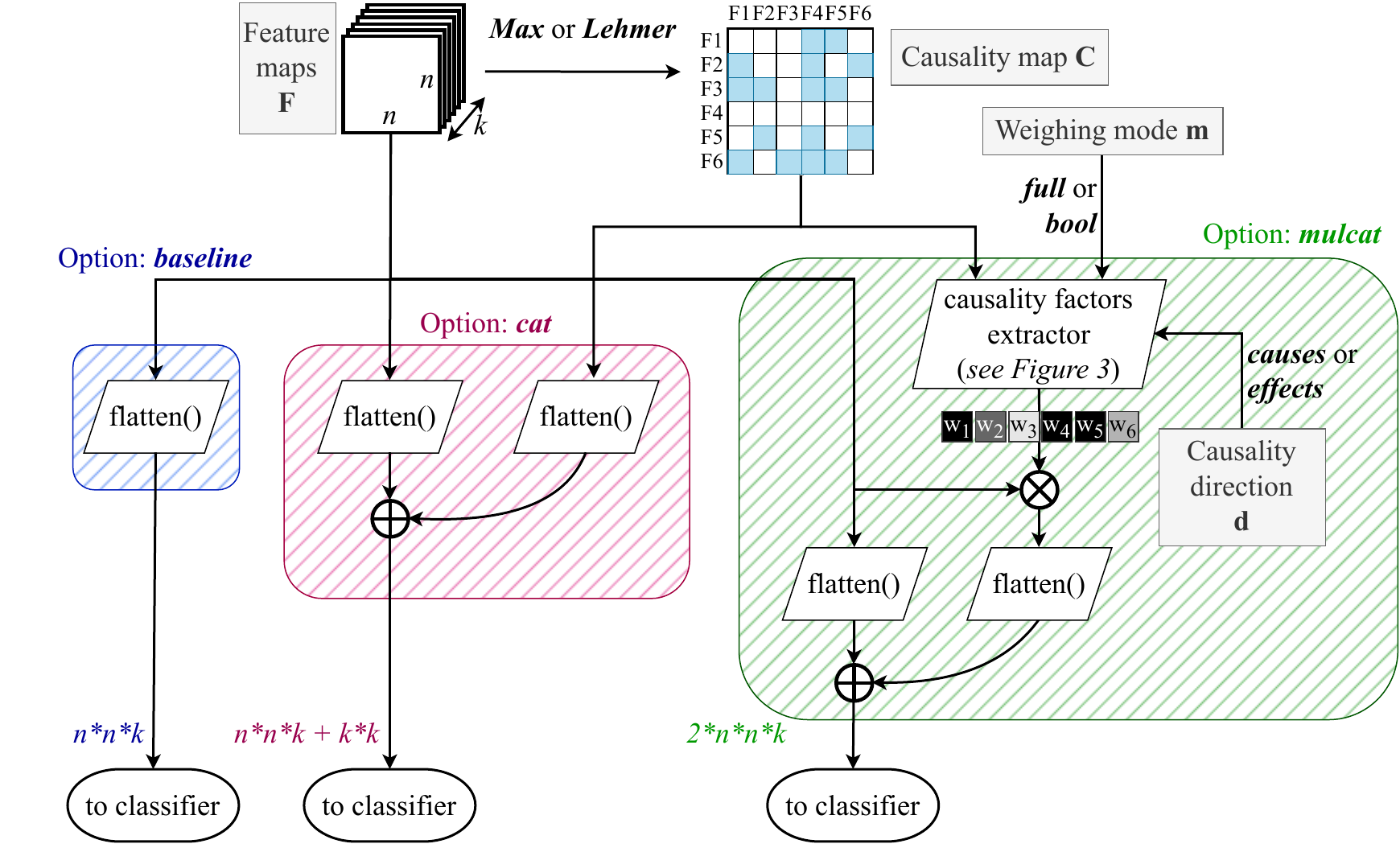}
	  \caption{Overview of the different settings investigated in this work: \textit{Baseline}, \textit{Cat}, and the proposed \textit{Mulcat}. Assuming $k=6$ feature maps as an example, the tensor \textbf{F} of feature maps that are obtained from a \acrshort{label_CNN} just before the classifier can be either flattened and used as they are (Option \textit{baseline}) or can be leveraged to compute the causality map \textbf{C} via the \textit{Max} or \textit{Lehmer} method. Once obtained, \textbf{C} can be flattened as well and concatenated to the feature maps (Option \textit{Cat}) or fed to our proposed causality factors extractor (see Figure \ref{fig:caufacextractor}) to implement the Option \textit{Mulcat}. The latter produces a vector of causality factors that weighs the feature maps obtaining a causality-driven version of them, which is then concatenated to the original ones and fed to the classifier. Weighing mode \textbf{m} and causality direction \textbf{d} are two external signals used to tune the functioning of the system. This image is best seen in color.}
   \label{fig:overview}
\end{figure}

Feature concatenation is a basic (yet popular) way to embed additional information in \acrshort{label_CNN}s. Indeed, by concatenating the flattened causality map to the flattened set of feature maps just before the classifier, \cite{terziyan2023causality} let the CNN learn how these causality estimates influence image classification. That means that in addition to the $n\times n \times k$ features, the \acrshort{label_FC} layers of the classifier will now have a $k \times k$ input, and the weights for the corresponding connections (i.e., actual causality influences) will be learned by back-propagation the same way as other neural network parameters. We will call this method the \textbf{Cat} (\textit{con\textbf{cat}enate}) option (see the magenta box in Figure \ref{fig:overview}). 
\subsubsection{Mulcat: Multiply and Concatenate}
Alternatively, one could enhance or penalize parts of the existing information according to the newly gained one. Our proposition here is a new way to exploit the causality map: this time, it is used to compute a vector of causality factors that multiply (i.e., weighs) the feature maps so that each feature map is strengthened according to its causal influence within the image's scene. 
After multiplication, the obtained causality-driven version of the feature maps is flattened and concatenated to the flattened original ones, producing a $2 \times n\times n \times k$ input to the classifier. We will call this method the \textbf{Mulcat} (\textit{\textbf{mul}tiply and con\textbf{cat}enate}) option (see the green box in Figure \ref{fig:overview}).

\subsubsection{Causality-factors Extractor}
\begin{figure}
	\centering
		\includegraphics[width=0.99\textwidth]{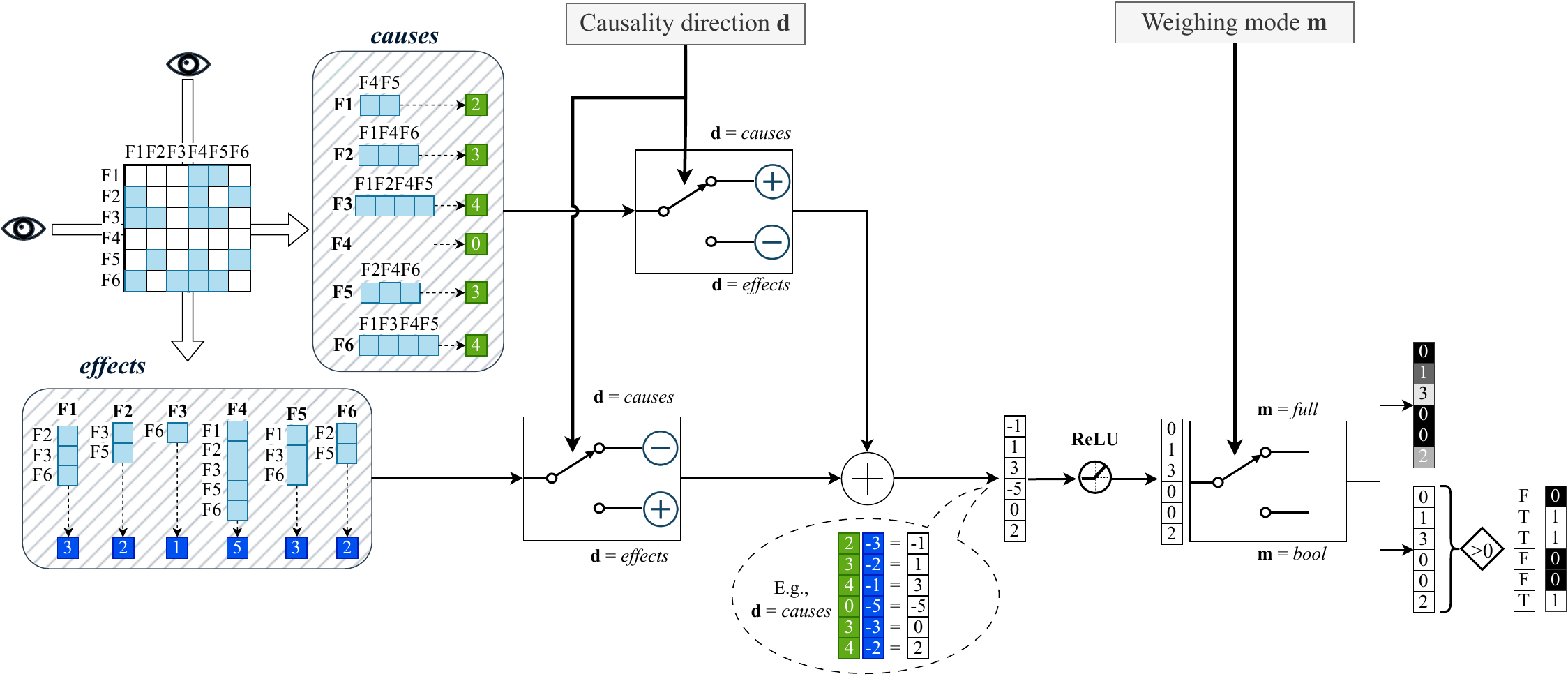}
	  \caption{The internals of the proposed \textit{causality factors extractor} block of Figure \ref{fig:overview} given an example causality map. Cyan squares in the causality map indicate whether the probability value of one element is greater than its element opposite the main diagonal. The \textit{causes} box shows how the causality map is processed row-wise for each feature map: the number of times that feature is a cause of another feature is registered. Similarly, the \textit{effects} box shows how the causality map is processed column-wise for each feature map. Before being summed element-wise, those two vectors are either passed as they are or the sign of their elements is reversed according to the causality direction \textbf{d}. The obtained vector is rectified and then returned as it is or passed through boolean filtering depending on the weighing mode \textbf{m}. This image is best seen in color.}
   \label{fig:caufacextractor}
\end{figure}

At the core of the \textbf{Mulcat} option stands our \textit{causality factors extractor} module, which yields the vector of weights needed to multiply the feature maps (see Figure \ref{fig:caufacextractor}).
The main idea here is to look for asymmetries between elements opposite the main diagonal of the causality map, as they represent conditional asymmetries entailing possible cause-effect relationships (e.g., Figure \ref{fig_cmapvisual}). Indeed, some features may be more often found on the left side of the arrow (i.e., $F\rightarrow$) than on the right side (i.e., $\rightarrow F$). 
Accordingly, the $2$D causality map is processed row-wise and column-wise. In the former case, we register the number of times each feature map $F^i$ was found to cause another feature map $F^j$, that is, $P(F^i|F^j) > P(F^j|F^i)$. This way, we obtain a vector of values that quantify how much those feature maps can be called "\textit{causes}." Conversely, in the column-wise processing, we register the number of times each feature map $F^j$ was found to be caused by another feature map $F^i$, obtaining a vector of values that quantify how much the feature maps can be deemed "\textit{effects}."

\subsubsection{Variants}
At this point, we propose two variants to the model's functioning. We allow an external signal $\textbf{d}$ to represent the causality direction of analysis, which can be either \textit{causes} or \textit{effects}. When \textbf{d} $=$ \textit{causes}, the vector of \textit{causes} (obtained row-wise) is not altered, while the sign is changed to the elements of the \textit{effects} vector (obtained column-wise). Hence, as those two vectors enter a summation point, the difference between \textit{causes} and \textit{effects} is obtained as the weight vector. On the other hand, when \textbf{d} $=$ \textit{effects}, the vector of \textit{effects} is not altered, while it is to the vector of \textit{causes} that the sign is changed. Therefore, the difference between \textit{effects} and \textit{causes} is obtained at the summation point. As a result, the obtained weight vector is rectified to set any negative elements to zero.

In addition, we conceive two variants of the model controlled by another external signal \textbf{m}, that represents the weighing mode and can be one of: 

\begin{itemize}
    \item \textbf{full}. The vector of non-negative causality factors is left at its full count, being returned as it is. As a result of this choice, the model weighs features more according to their causal importance (a feature that is \textit{cause} $10$ times more than another receives $10$ times more weight).

    \item \textbf{bool}. The factors undergo boolean thresholding where all the non-zero factors are assigned a new weight of $1$ and $0$ otherwise. As a result, this choice is more conservative and assigns all features that are most often \textit{causes} the same weight.
\end{itemize}
In the following sections, we describe the data used for our empirical evaluations, the different types of model architectures we utilized, and the implementation details of the training process.

\subsection{Datasets}
To validate our proposed methods, we utilized multiple publicly available medical imaging datasets. To begin with, we exploited the Breast cancer Histopathological Image (BreakHis) dataset \cite{Spanhol15}. On the other hand, we used the dataset from the PI-CAI challenge \cite{Saha23}, comprising multi-parametric MRI (mpMRI) acquisitions of the prostate. 

\subsubsection{BreakHis Dataset}
The dataset has 7909 microscopic images of breast tumor tissues aggregated from 82 subjects at magnification levels of 40, 100, 200, and 400. There are eight classes in this dataset, namely adenosis, tubular adenoma, fibroadenoma, phyllodes tumor, papillary carcinoma, lobular carcinoma, mucinous carcinoma, and ductal carcinoma. In addition, a binary classification was provided, namely, \textit{benign} and \textit{malignant} lesions. In particular, the first four classes represent benign lesions, while the last four represent malignant lesions. We considered the images with a magnification level of 400 for a total of 1819 images. We split this dataset into training (1235 images), validation (218 images), and test (366 images) sets, ensuring class balancing according to the binary classification, i.e., benign and malignant. 

\begin{figure}
	\centering
		\includegraphics[width=0.99\textwidth]{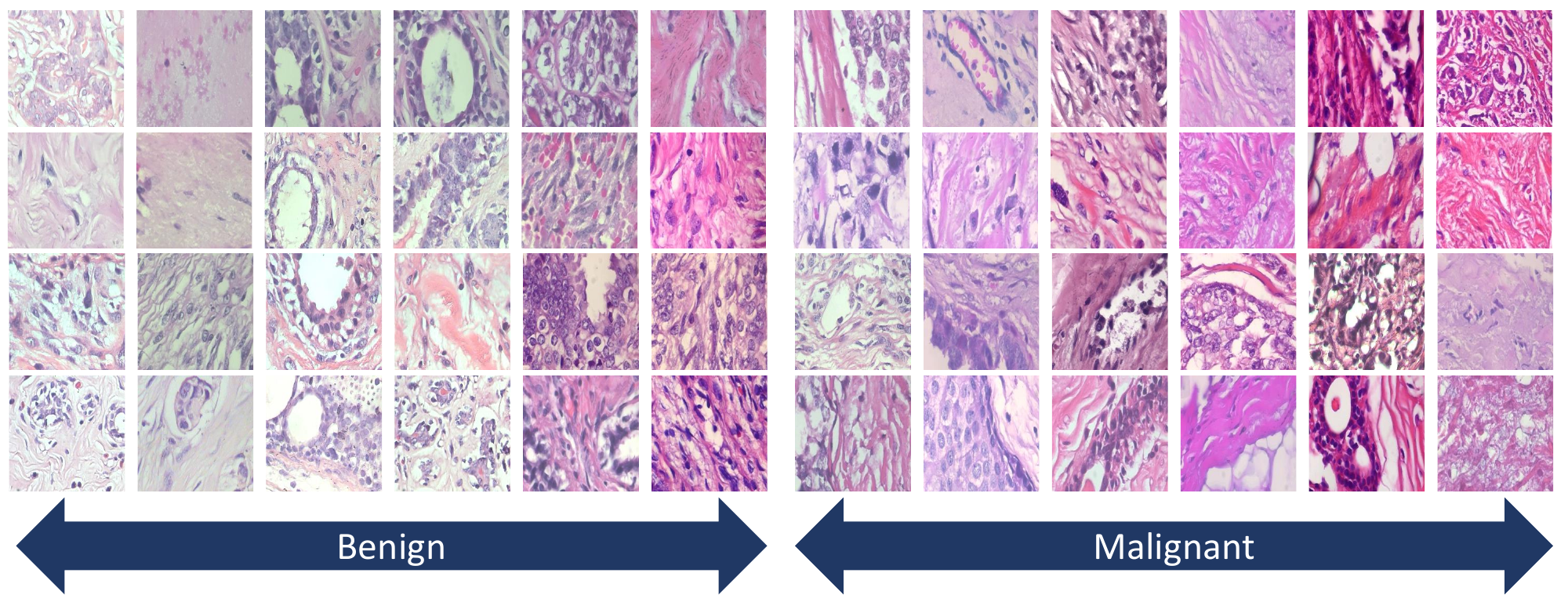}
	  \caption{Some benign and malignant samples from the utilized BreakHis dataset. This image is best seen in color.}
   \label{fig:dataset_breakhis}
\end{figure}

In this study, we utilize the processed version of the BreakHis dataset, curated by \cite{Pereira23} to be used in \acrshort{label_ML} tasks. Indeed, the original images were resized to 224x224 pixels and organized according to binary and multiclass classification tasks. We further resize the images to a consistent 128x128 pixel matrix. Some samples from the utilized dataset are presented in Figure \ref{fig:dataset_breakhis}.

\subsubsection{PI-CAI Dataset}
From the available $1500$ acquisitions, we only selected  T2-weighted (\acrshort{label_T2w}) images. Within this cohort of patients and respective scans, some cases didn't have any tumors (i.e., they had no biopsy examination), while others had cancer lesions. For each of the latter, the dataset contained biopsy reports expressing the severity as Gleason Score (\acrshort{label_GS}). In anatomopathology, a \acrshort{label_GS} of $1$ to $5$ is assigned to the two most common patterns in the biopsy specimen based on the cancer severity. The two grades are then added together to determine the \acrshort{label_GS}, which can assume all the combinations of scores from "$1$+$1$" to "$5$+$5$". Additionally, the dataset included the assigned GS's group affiliation, defined by the International Society of Urological Pathology (\acrshort{label_ISUP}) \cite{Egevad16}, ranging from $1$ to $5$, which provides the tumor severity information at a higher granularity level.
In this study, we included both cancerous and no-tumor patients. From the former case, we only considered lesions with GS $\geq 3+4$ (ISUP $\geq 2$) and selected only the slices containing lesions by exploiting the expert annotations of the disease provided in the dataset. For the latter case, we considered all the available slices. In the end, we obtained a total number of $4159$ images (from $545$ patients), with a balanced distribution over the two classes: $2079$ tumor images vs. $2080$ no-tumor images.
To constitute our subsets, we divided the available images into training ($2830$), validation ($515$), and testing ($814$) subsets. During the splitting process, we ensured patient stratification (i.e., images of the same patient were grouped to prevent data leakage) and class balancing.

\begin{figure}
	\centering
		\includegraphics[width=0.99\textwidth]{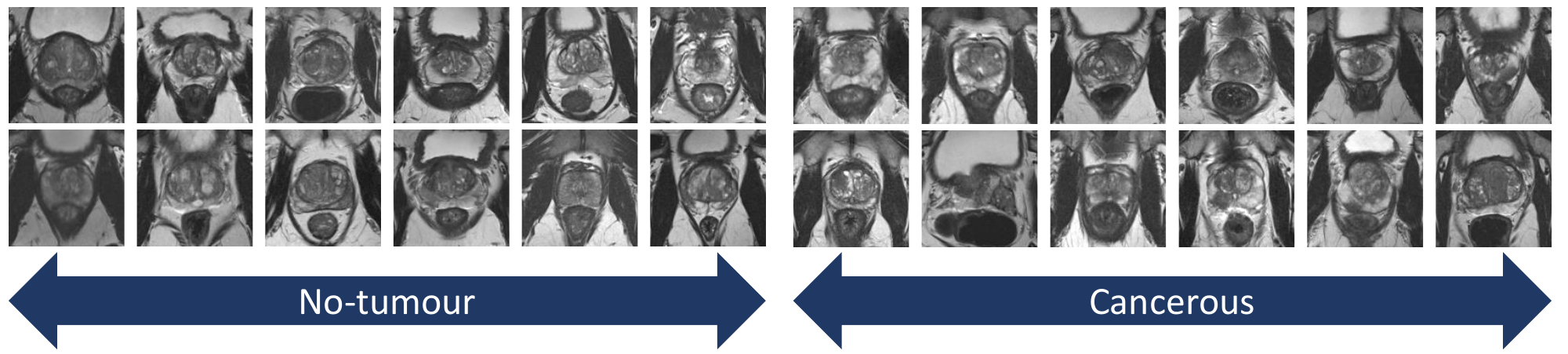}
	  \caption{Some no-tumour and cancerous samples from the utilized PI-CAI dataset.}
   \label{fig:dataset_picai}
\end{figure}

We utilized the provided whole prostate segmentation to extract the mask centroid for each slice. We then standardized the field of view (FOV) at $100$ mm in both $x$ ($FOV_x$) and $y$ ($FOV_y$) directions to ensure consistency across all acquisitions and subsequently cropped each image based on this value around its centroid. To determine the number of rows ($N_{rows}$) and columns ($N_{cols}$) corresponding to the fixed FOV, we utilized the pixel spacing in millimeters along the $x$-axis ($px$) and the $y$-axis ($py$). The relationships used to derive the number of columns and rows are $N_{cols} = \frac{FOV_x}{px}$ and $N_{rows} = \frac{FOV_y}{py}$, respectively.
Furthermore, we resized all the images to a uniform matrix size of $96 \times 96$ pixels to maintain consistent pixel counts. Finally, we performed image normalization using an in-volume method. That involved calculating the mean and standard deviation of all pixels within the volume acquisition and normalizing each image based on these values using the z-score technique. Some samples from the utilized dataset are presented in Figure \ref{fig:dataset_picai}.

\subsection{Architecture and Training}
For each dataset, we built different \acrshort{label_CNN} models to automatically classify input images in the two classes according to their diagnosis labels under full supervision. As for the architectures, we used the popular ResNet18 as the backbone for all the causality-driven models.
To handle images of different sizes in image recognition, many common architectures use an adaptive average pooling layer that outputs a $1\times1$ shape before the classifier. It does this by adjusting its parameters (such as kernel size, stride, and padding) based on the input size. However, this reduces the dimensionality of the feature maps and ignores their 2D structure, which is needed for finding causalities. Therefore, we replaced the \textit{AdaptiveAvgPool2D} layer of the ResNet18 with an identity layer in our experiments.

As described in Section \ref{sec:embedding_causality}, we could integrate the information of the causality map into the CNN classification in different manners. In this work, we developed six types of models for each  dataset and trained them to test the efficacy of the newly proposed \textit{Mulcat} architectures on medical image classification, namely:
\begin{itemize}
    \item \textbf{ResNet18}. This model is a regular ResNet18 architecture to serve as a baseline, where we replaced its \textit{AdaptiveAvgPool2D} layer with an identity layer. See Figure \ref{fig:overview} (blue box) for a visual representation.
    \item \textbf{ResNet18 $+$ Cat}. This is a ResNet18 model we modified to embed the causal information via concatenation as in \cite{terziyan2023causality}. See Figure \ref{fig:overview} (magenta box) for a visual representation. 
    \item \textbf{ResNet18 $+$ Mulcat (full, causes)}. This variant exploits our \textit{causality factors extractor} to obtain weights for the feature maps. In this model, we set the causality direction \textbf{d} = \textit{causes} and the weighing mode \textbf{m} = \textit{full}.
    \item \textbf{ResNet18 $+$ Mulcat (bool, causes)}. It is similar to the previous, but we set the weighing mode to \textbf{m} = \textit{bool}.
    \item \textbf{ResNet18 $+$ Mulcat (full, effects)}. This variant turns the way the set of causality factors is obtained within our \textit{causality factors extractor} by setting the causality direction \textbf{d} = \textit{causes}. We use \textbf{m} = \textit{full} in this model.
    \item \textbf{ResNet18 $+$ Mulcat (bool, effects)}. It is analogous to the previous, but setting the weighing mode to \textbf{m} = \textit{bool}.
\end{itemize}

As shown in Figure \ref{fig:overview}, the different types of models we investigated expose the classifier to a different number of input features. Therefore, the classifier is modified for each type according to the number of new neurons entering the fully-connected layer.

We optimized the way we computed the causality maps (using either the \textit{Max} option (Eq. \ref{eq:causality_method_max}) or the \textit{Lehmer} (Eq. \ref{eq:causality_method_lehmer})) and, for the \textit{Lehmer} option, we tried six different values of its parameter \textit{p}: [$-100, -2, -1, 0, 1, 100$]. Consequently, for each dataset, we trained seven models for each of the five types of causality-driven models, resulting in $35$ causality-driven models plus one baseline model. 
%
We provide the pseudo-code for the algorithms utilized to compute the causality maps and the proposed causality factors in Algorithm \ref{alg:cmap} and Algorithm \ref{alg:cfactors}, respectively. 

\begin{algorithm}
\caption{Causality map computation}\label{alg:cmap}
\begin{algorithmic}[1]
\Require $\textbf{x}\gets$ feature maps [$k$,$n$,$n$], $CM\gets$ computation method, $lehmer\_p \gets$ Lehmer mean power.
\State $\textbf{x} \gets \textbf{x}/max(\textbf{x})$ \Comment{Normalize in range [$0-1$] by dividing for maximum activation across all maps}
\State $\textbf{cmap} \gets \textbf{0}$ \Comment{Size of cmap: [$k$,$k$]}
\If {$CM == $ \textit{Max}} \Comment{Eq. \ref{eq:causality_method_max}}
    \State $\textbf{sumValues} \gets sum(flatten(\textbf{x}),dim=1)$ \Comment{Compute sum of values for each feature}
    \State $\textbf{maxValues} \gets max(flatten(\textbf{x}),dim=1)$ \Comment{Get maximum values of each feature}
    \State $\textbf{prod} \gets outerProduct(\textbf{maxValues},\textbf{maxValues})$ \Comment{Numerator of Eq. \ref{eq:causality_method_max} for each $i$,$j$}
    \State $\textbf{cmap} \gets \textbf{prod}/\textbf{sumValues}$
\Else \If {$CM ==$ \textit{Lehmer}} \Comment{Eq. \ref{eq:causality_method_lehmer}}
    \State $\textbf{x}\gets flatten(\textbf{x})$
    \State $\textbf{crossMat} \gets outerProduct(\textbf{x},\textbf{x})$ \Comment{Pairwise multiplications between each element of the features}
    \State $\textbf{num\_a} \gets sum(exp(\textbf{crossMat},lehmer\_p+1))$
    \State $\textbf{num\_b} \gets sum(exp(\textbf{crossMat},lehmer\_p))$
    \State $\textbf{num} \gets \textbf{num\_a}/\textbf{num\_b}$ \Comment{Numerator of Eq. \ref{eq:causality_method_lehmer}}
    \State $\textbf{den\_a} \gets sum(exp(\textbf{x},lehmer\_p+1))$
    \State $\textbf{den\_b} \gets sum(exp(\textbf{x},lehmer\_p))$
    \State $\textbf{den} \gets \textbf{den\_a}/\textbf{den\_b}$ \Comment{Denominator of Eq. \ref{eq:causality_method_lehmer}}
    \State $\textbf{cmap} \gets \textbf{num}/\textbf{den}$
    \EndIf
\EndIf
\State \textbf{return cmap}
\end{algorithmic}
\end{algorithm}

\begin{algorithm}[t]
\caption{Causality factors extraction}\label{alg:cfactors}
\begin{algorithmic}[1]
\Require $\textbf{x}\gets$ feature maps [$k$,$n$,$n$], $\textbf{cmap}\gets$ causality map [$k$,$k$], $d\gets$ causality direction, $m\gets$ weighing mode.
\State $\textbf{triu} \gets triu(\textbf{cmap},diag=1)$ \Comment{Upper triangular matrix}
\State $\textbf{tril} \gets tril(\textbf{cmap},diag=-1).transpose()$ \Comment{Lower triangular matrix}
\State $\textbf{bool\_ij} = (tril>triu).transpose()$
\State $\textbf{bool\_ji} = (triu>tril)$
\State $\textbf{bool\_matrix} = \textbf{bool\_ij} + \textbf{bool\_ji}$ \Comment{Sum of booleans is the OR logic}
\State $\textbf{by\_col} = sum(\textbf{bool\_matrix}, dim=1)$ \Comment{Obtain the \textit{causes} view}
\State $\textbf{by\_row} = sum(\textbf{bool\_matrix}, dim=0)$ \Comment{Obtain the \textit{effects} view}
\If {$d ==$ \textit{causes}}
    \State $\textbf{mul\_factors} = ReLU(\textbf{by\_col} - \textbf{by\_row})$ \Comment{Difference between \textit{causes} and \textit{effects}}
    \If {$m ==$ \textit{full}}
        \State \textbf{return mul\_factors}
    \Else \If {$m ==$ \textit{bool}}
        \State $\textbf{mul\_factors} = 1*(\textbf{mul\_factors}>0)$ \Comment{Boolean thresholding}
        \State \textbf{return mul\_factors}
    \EndIf
    \EndIf
\Else \If {$d ==$ \textit{effects}}
    \State $\textbf{mul\_factors} = ReLU(\textbf{by\_row} - \textbf{by\_col})$ \Comment{Difference between \textit{effects} and \textit{causes}}
    \If {$m ==$ \textit{full}}
        \State \textbf{return mul\_factors}
    \Else \If {$m ==$ \textit{bool}}
        \State $\textbf{mul\_factors} = 1*(\textbf{mul\_factors}>0)$ \Comment{Boolean thresholding}
        \State \textbf{return mul\_factors}
\EndIf
\EndIf
\EndIf
\EndIf
\end{algorithmic}
\end{algorithm}

Regarding the training phase, we utilized the \acrshort{label_CE} loss as the criterion and \acrshort{label_Adam} as the optimizer, as well as performed data augmentation (random horizontal flip) at training time. We trained the models for $200$ epochs and set up a \acrshort{label_LR} scheduler to decrease the \acrshort{label_LR} during training. Specifically, the scheduler starts by multiplying the LR by $1.0$ after the first epoch, and then this factor linearly decreases to $0.1$ at epoch $200$.
As for models' hyperparameters, we investigated different values of initial LR ($0.01$ and $0.001$) and of weight decay ($0.01$, $0.001$, and $0.0001$). Accordingly, for each dataset, we trained the $36$ models for each of the six combinations of hyperparameters and chose the best-performing model on the validation set.
To prevent our results from being biased due to the random processes of the algorithms, we repeated the entire analysis ($216$ experiments) four times with different starting seeds that govern the random processes of the scripts.

\subsection{Quantitative Evaluation}
During training, we utilized the loss and accuracy obtained by the models on the validation set to track their evolution during epochs, selecting the best-performing one once the training phase ended. Then, we evaluated such selected models on the external never-before-seen test set and reported their accuracy since classes are balanced and we care about the overall performance. In this way, we obtain a quantitative metric to compare the baseline architecture, the \textit{Cat} model, and our proposed \textit{Mulcat} architectures.

Ablation studies remove or damage specific components in a controlled setting to investigate all possible outcomes of system failure, thus understanding the contribution of a component to the overall system. The \textbf{ResNet18} (baseline) models already act as the ablation models for the remaining five types of models. Nevertheless, we wanted to do more than solely remove components. To gauge the significance of the values contained in causality maps and causality factors, we performed an additional test where we distort (i.e., damage) their information. We call these partially ablated versions of the networks \textbf{damaged}.
Concerning the \textbf{ResNet18 $+$ Cat} option, the only contribution of the causality map to the classification resides in the flattened elements that are concatenated to the actual (flattened) feature maps. Therefore, a natural \textit{damaged} network for such a setting would be to create a fictitious causality map filled with random probability values. We called this model the \textbf{ResNet18 $+$ Damaged-Cat}.
On the other hand, when it comes to the \textbf{ResNet18 $+$ Mulcat} option, the key functionality is to extract a vector of meaningful causality factors that serve as weights to the feature maps. Hence, we created the \textbf{ResNet18 $+$ Damaged-Mulcat} model, where we modify that vector to weigh features randomly rather than based on a principled way. This model comes in two variants according to the possible values of the causality factors mode, \textbf{m}. 
Indeed, when \textbf{m} $=$ \textit{full}, the $1 \times k $ vector of causality factors (i.e., weights) is replaced with a random vector of the same size with integer values ranging from $0$ (a feature map is never \textit{cause} of another feature) to $k-1$ (it is \textit{cause} of every other feature). Whereas, when \textbf{m} $=$ \textit{bool}, the values of the weights are randomly assigned to either $0$ or $1$. Since, in this setting, weights are hand-crafted, there is no need to consider the causality direction used; therefore, the \textit{damaged} study we performed is valid for both \textbf{d} $=$ \textit{causes} and \textbf{d} $=$ \textit{effects}.

To observe how the different architectures differ in terms of memory requirements, we track the size of the trained models (in megabytes) and the number of corresponding parameters (in millions). To compute the former, we don't want to rely on the file size of the saved models (e.g., \textit{.pth} files from PyTorch), as the file might be compressed. In fact, we calculate the number of parameters and buffers, multiply them by the element size, and accumulate these numbers.

\subsection{Qualitative Evaluation}
To further investigate the possible benefits of integrating causality into CNNs for medical image classification, we performed \acrshort{label_XAI} experiments on the best-performing model for each type. Specifically, we aimed to obtain \acrshort{label_CAM} for the networks' decisions in all six types of models in our investigation: \textbf{Baseline} model, \textbf{Cat} model, \textbf{Mulcat-full-causes} model, \textbf{Mulcat-bool-causes} model, \textbf{Mulcat-full-effects} model, and \textbf{Mulcat-bool-effects} model. In all these models we assume ResNet18 as the backbone. Since investigating the variability of the visual output when changing the \acrshort{label_XAI} method used is outside the scope of our work, we chose the popular Grad-CAM method \cite{selvaraju2017grad}, implemented in the \textit{pytorch-grad-cam} library \cite{jacobgilpytorchcam}. For the same reason, we selected the last convolutional layer of our architectures as the target layer for which we computed the \acrshort{label_CAM} and performed the analysis with standard parameters. A more systematic analysis would require investigating the CAM output on all layers of the \acrshort{label_CNN} and optimizing the smoothing parameters. 

To evaluate the quality and robustness of the produced \acrshort{label_CAM}s, we followed the following criteria for the two datasets:
\begin{itemize}
    \item \textbf{BreakHis dataset}. To differentiate benign from malignant tumors, pathologists examine breast tissues at different magnification levels. Specifically, at $400\times$ magnification level, as the one used for our experiments, they analyze cytological features, such as shape and size of the nuclei, hyperchromatic nuclei, mitotic cells, and prominent nuclei \cite{young2013wheater}. To highlight cell nuclei, they employ Hematoxylin and Eosin stains, which make the nuclei appear dark purple or blue, while the other structures appear in shades of pink, red, and orange \cite{he2012histology}. For these reasons, we considered good explanations, the ones that focus on regions containing the dark purple/blue structures assumed as the nuclei of the cells.
     
    \item \textbf{PI-CAI dataset}. Based on the classification task, we considered explanations focusing on discriminative regions of the MRI (e.g., prostate gland area) to be better. In contrast, we considered explanations focusing on other structures, such as the rectum, bladder, or lateral muscle bundles, to be of lower quality and robustness.
\end{itemize}

\subsection{Additional Experiments}
We conducted additional experiments on two very common application fronts to further test the effectiveness of our method. On the one hand, we proved that our module is easy to fit into existing convolutional models using other forms of visual attention, thus creating synergy. On the other, we verified its functioning in low-data scenarios, extending its applicability to Few-Shot Learning (\acrshort{label_FSL}) \cite{fink2004object,fei2006one}.

\subsubsection{Integrating Bottleneck Attention Modules}
The bottleneck attention module (\acrshort{label_BAM}) \cite{park2020simple,woo2018cbam} is a popular attention-based mechanism that, given a feature map, learns the attention map along two factorized axes, \textit{channel} and \textit{spatial}, to strengthen the representational power of \acrshort{label_CNN}s. We thus investigated the addition of our module to models that already leveraged \acrshort{label_BAM}. We used the same backbone as above (ResNet18) and placed multiple BAMs located after its layers $1$, $2$, and $3$, to build hierarchical attention. After training with the same strategy as the main study, we compared the performance of BAM-based regular models (\textbf{ResNet18 $+$ BAM}), BAM-based models integrating the \textit{Cat} method (\textbf{ResNet18 $+$ BAM $+$ Cat}), and BAM-based models that integrate our \textit{Mulcat} module (\textbf{ResNet18 $+$ BAM $+$ Mulcat}).

\subsubsection{Few-Shot Learning Experiments}
In addition to fully-supervised studies, in this work, we extended our recent investigation into causality-driven one-shot learning (\acrshort{label_OSL}) \cite{carloni2023causality} to the new BreakHis dataset, to understand how our \textit{Mulcat} methods worked under the shortage of annotated data in the medical imaging domain. To make the analyses consistent, we only considered the \textit{causes} direction in our \textit{Mulcat} models. We adopted the meta-learning strategy and formulated each task (i.e., episode) of the training process as an \textit{N-way 1-shot} classification problem, that is, to classify \textit{N} classes using only \textit{1} support image per class.

Regarding the PI-CAI dataset, we utilized a subset of the data containing lesions and the clinical question was tumor grading (i.e., predict aggressiveness). From a higher-level perspective than that of \acrshort{label_GS} scores and \acrshort{label_ISUP} groups, prostate lesions with GS $\leq3+3$ (ISUP $=1$) and with GS $ =3+4$ (ISUP $=2$) are considered low-grade (LG) tumors, while those with GS $>3+4$ (ISUP $>2$) are high-grade (HG) tumors. We considered only lesions whose GS was $\geq3+4$ (ISUP $\geq2$). As a result, we had eight classes of GS and four classes of ISUP in our dataset. The total number of images was $2049$ (from $382$ patients), which we divided into training (1611), validation ($200$), and testing ($238$) subsets, and resized to $128\times128$. We experimented with two classification scenarios on this dataset.
In the first scenario (\textbf{2-way}), the meta-training data are labeled to the four \acrshort{label_ISUP} classes, and the model is meta-trained by randomly picking \textit{two} of the four classes in each task while distinguishing between LG and HG lesions during meta-testing. 
In the second scenario (\textbf{4-way}), we label meta-training data on the GS, and the model randomly picks \textit{four} of the eight \acrshort{label_GS} classes in each task while distinguishing between four ISUP classes in meta-testing.

Regarding the BreakHis dataset, we used the same subsets as for the main study and considered two scenarios: in the \textbf{2-way} scenario, the meta-training is performed by randomly picking two of the eight classes of aggressiveness in each task, and meta-testing is done on the two high-level classes \textit{benign-vs-malignant}; in the \textbf{4-way} scenario, the meta-training is done on four out of eight random classes and the meta-testing is performed on four most prevalent classes (i.e., ductal carcinoma, fibroadenoma, phyllodes tumor, and tubular adenoma).

To increase the models' robustness to different data selections, we performed 600 meta-training tasks, 600 meta-validation tasks, and 600 meta-testing tasks for each experiment. To cope with the dataset unbalancing, we employed the AUC margin loss (AUCM) \cite{yang2022algorithmic} and the proximal epoch stochastic method (PESG) \cite{guo2020fast}, maximizing the \acrshort{label_AUROC}, which we used as our training and evaluation metric. Specifically, in \textit{2-way} experiments, we computed the binary AUROC, while we calculated the AUROC using the \textit{One-vs-rest} setting in \textit{4-way} experiments. Moreover, we evaluated the binary classification performance of the \textit{4-way} models by computing the AUROC of one significant class versus all the rest (i.e., $1$-vs-$3$): \textit{malignant} (ductal carcinoma) versus \textit{benign} (fibroadenoma, phyllodes tumor, tubular adenoma) for the BreakHis dataset, and \textit{LG} (\acrshort{label_ISUP}=$2$) versus \textit{HG} (ISUP=$3$, ISUP=$4$, ISUP=$5$).
As with the main study, we performed ablation studies by repeating the OSL experiments with the \textit{damaged} version of our \textit{Mulcat} method.

\subsection{Implementation Details}
All the experiments in this study ran on an NVIDIA A$100$ $40$ GB Tensor Core of the AI@Edge cluster of our Institute. We used Python $3.8.15$ and back-end libraries of PyTorch (version $1.13.0$, cuda $11.1$), together with other libraries such as scikit-learn $1.2.0$, grad-cam $1.4.8$, pydicom $2.3.1$, and pillow $9.4.0$. Docker version $20.10.11$ (build dea9396) was installed in the machine. To make results reproducible for each battery of experiments, we set a common seed for the random sequence generator of all the random processes and PyTorch functions. We released the code base for our framework at \url{https://github.com/gianlucarloni/causality_conv_nets}.

\section{Results}
\label{sec:results}
\begin{table}[t]
    \centering
    \caption{Results of the main study and the ablation study for the best-performing models w.r.t the causality setting, the mode of computing the causality factors, and the direction used to encode the causality factors. We report accuracy results on the test set as the mean and standard deviation (in lower script) over four repetitions of the experiments with different seeds. The top half of the table refers to the BreakHis dataset, while the bottom half of it refers to the PI-CAI dataset.}
    \resizebox{\textwidth}{!}{
    \begin{tabular}{lllc}
    \hline
    \textbf{Architecture} & \textbf{Causality factors mode} & \textbf{Causality direction} & \textbf{Test set accuracy} [$\uparrow$] \\
    \hline
    \toprule
    \multicolumn{4}{c}{\textbf{\textit{BreakHis dataset (main study)}}}\\
    \midrule
    ResNet18 & - & - & $88.48_{0.77}$\\
    \hline
    ResNet18 $+$ Cat \cite{terziyan2023causality} & - & - & $85.77_{3.16}$\\
    \hline
    \multirow{4}{*}{ResNet18 $+$ \textbf{Mulcat (ours)}} 
     & Full & Causes & $91.32_{1.63}$ \\
    \cline{2-4}
     & Bool & Causes & $91.06_{1.32}$ \\
    \cline{2-4}
     & Full & Effects & $90.65_{1.14}$ \\
    \cline{2-4}
     & Bool & Effects & $91.59_{0.83}$ \\    
    \hline\\ 
    \multicolumn{4}{c}{\textbf{\textit{BreakHis dataset (ablation study)}}}\\
    \hline
    ResNet18 $+$ Damaged-Cat & - & - & $50.66_{1.93}$ \\
    \hline
    \multirow{2}{*}{ResNet18 $+$ Damaged-Mulcat} 
     & Full & Causes/Effects & $81.57_{0.50}$ \\
    \cline{2-4}
     & Bool & Causes/Effects &  $82.24_{2.15}$ \\    
    \bottomrule 

    \toprule
    \multicolumn{4}{c}{\textbf{\textit{PI-CAI dataset (main study)}}}\\
    \midrule
    ResNet18 & - & - & $68.38_{1.50}$ \\
    \hline
    ResNet18 $+$ Cat \cite{terziyan2023causality} & - & - & $70.07_{1.86}$ \\
    \hline
    \multirow{4}{*}{ResNet18 $+$ \textbf{Mulcat (ours)}} 
     & Full & Causes & $71.82_{1.00}$ \\
    \cline{2-4}
     & Bool & Causes & $70.31_{1.63}$ \\
    \cline{2-4}
     & Full & Effects & $69.96_{0.55}$ \\
    \cline{2-4}
     & Bool & Effects & $71.13_{0.36}$ \\
    \hline\\     
    \multicolumn{4}{c}{\textbf{\textit{PI-CAI dataset (ablation study)}}}\\
    \hline
    ResNet18 $+$ Damaged-Cat & - & - & $52.08_{0.93}$ \\
    \hline
    \multirow{2}{*}{ResNet18 $+$ Damaged-Mulcat} 
     & Full & Causes/Effects & $49.63_{0.88}$ \\
    \cline{2-4}
     & Bool & Causes/Effects &  $67.18_{0.99}$ \\    
    \bottomrule    
    \end{tabular}
    }
    \label{tab:best_results}
\end{table}

\subsection{Main Study}
Table \ref{tab:best_results} shows the results of our main study for both datasets. We report the accuracy metric of the best-performing models on the external test set as the mean and standard deviation over four repetitions of the experiments with different seeds.
Regarding the \textbf{BreakHis dataset}, the baseline models (ResNet18) achieved an accuracy of $88.48$, while the competing method \textit{Cat} performed worse than the baseline, with an accuracy of $85.77$. On the other hand, our proposed \textbf{Mulcat} models, where the causality factors are ultimately computed in different ways depending on the mode \textit{m} and the direction \textit{d}, demonstrate higher performance than both previous choices. For instance, the \textit{full}-\textit{causes} models achieved $91.32$ accuracy, while their \textit{bool} version achieved an accuracy of $91.06$. On the other hand, the \textit{full}-\textit{effects} and \textit{bool}-\textit{effects} models reached accuracies of $90.65$ and $91.59$, respectively.
Regarding the \textbf{PI-CAI dataset}, while the baseline models achieved an accuracy of $68.38$, embedding causality in different forms improved performance. For instance, when the causality map was used with the \textit{Cat} version, the models achieved an accuracy of $70.07$. As for the \textbf{Mulcat} models, they all ranked above baseline, with the \textit{full}-\textit{causes} models that achieved $71.82$ accuracy and the \textit{bool} version achieving an accuracy of $70.31$. On the other hand, the \textit{full}-\textit{effects} models achieved $69.96$ accuracy, while using the \textit{bool} version led the models to reach an accuracy of $71.13$.

\subsection{Ablation Study}
Table \ref{tab:best_results} also shows the results of purposely damaging the information contained in the causality maps and causality factors. These partial ablation studies for the two datasets reveal that, on BreakHis data, the \textbf{Damaged-Cat} models obtained an accuracy of $50.66$, and the \textbf{Damaged-Mulcat} models obtained accuracy values of $81.57$ and $82.24$ when using \textit{full} and \textit{bool} mode, respectively. As for the PI-CAI dataset, the \textbf{Damaged-Cat} models achieved an accuracy of $52.08$, and the \textbf{Damaged-Mulcat} models obtained accuracy values of $49.63$ and $67.18$ when using \textit{full} and \textit{bool} mode, respectively.     

\subsection{Memory Requirements}
The size of the trained models and the number of corresponding parameters for both experiments are given in Table \ref{tab:memory_requirements}. While our \textbf{Mulcat} models increase memory demand by a negligible amount compared to their baseline counterparts ($+0.8\%$), using \textbf{Cat} models results in an overhead of up to approximately $+4.7\%$. 

\begin{table}
    \centering
    \caption{Memory requirements in terms of model size (in megabytes) and number of parameters (in millions) of the ResNet18 baseline models, the ResNet18 backbone with the addition of the \textit{Cat} option, and the ResNet18 backbone with our \textit{Mulcat} module. Lower is better.}
    \resizebox{\textwidth}{!}{
    \begin{tabular}{lcc}
    \hline
    \textbf{Architecture} & \textbf{Model size (MB)} [$\downarrow$] & \textbf{Number of parameters ($\times 10^6$)} [$\downarrow$] \\
    \hline
    \toprule
    \multicolumn{3}{c}{\textit{\textbf{BreakHis dataset} (image size: $3\times128\times128$)}}\\
    \midrule
    ResNet18 & $42.73$ & $11.19$ \\
    \hline
    ResNet18 $+$ Cat \cite{terziyan2023causality} & $44.73$ & $11.72$ \\
    \hline
    ResNet18 $+$ \textbf{Mulcat (ours)} & $42.80$ & $11.21$ \\
    \hline
    
    \multicolumn{3}{c}{\textit{\textbf{PI-CAI dataset} (image size: $1\times96\times96$)}}\\
    \midrule
    ResNet18 & $42.68$ & $11.18$ \\
    \hline
    ResNet18 $+$ Cat \cite{terziyan2023causality} & $44.68$ & $11.70$ \\
    \hline
    ResNet18 $+$ \textbf{Mulcat (ours)} & $42.72$ & $11.19$ \\
    \hline
    \end{tabular}
    }
    \label{tab:memory_requirements}
\end{table}

\subsection{XAI Evaluations}
\begin{figure}
	\centering
		\includegraphics[width=0.99\textwidth]{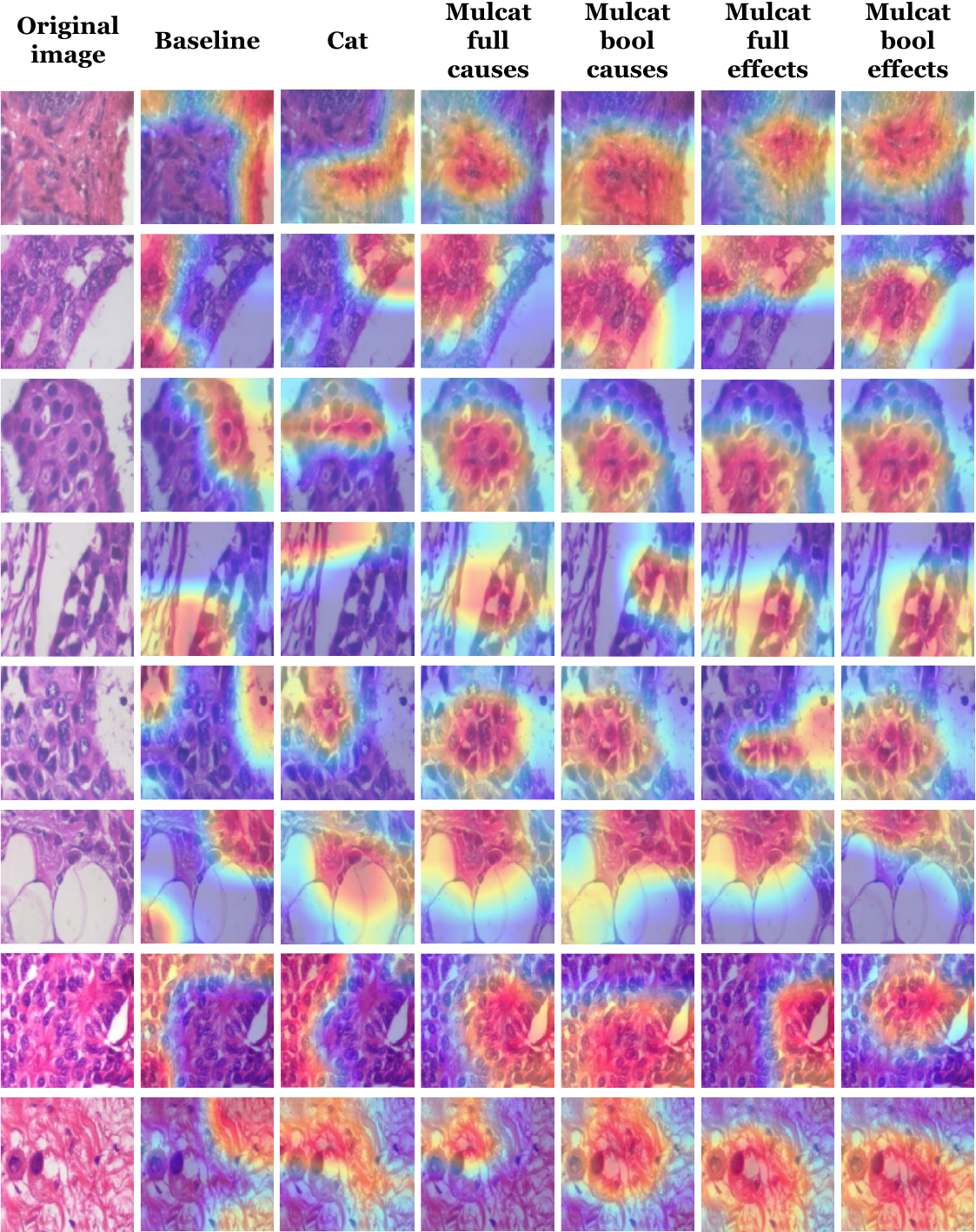}
	  \caption{Visual assessment of class activation maps for different malignant cases of the BreakHis dataset. Each row represents a different biopsy slide, and the columns represent the Grad-CAM outputs for the baseline model, the \textit{Cat} option, and all the proposed \textit{Mulcat} variants. ResNet18 is assumed as the backbone architecture in all models. Best seen in color.}
   \label{fig:CAM_malignant_breakhis}
\end{figure}

\begin{figure}
	\centering
		\includegraphics[width=0.99\textwidth]{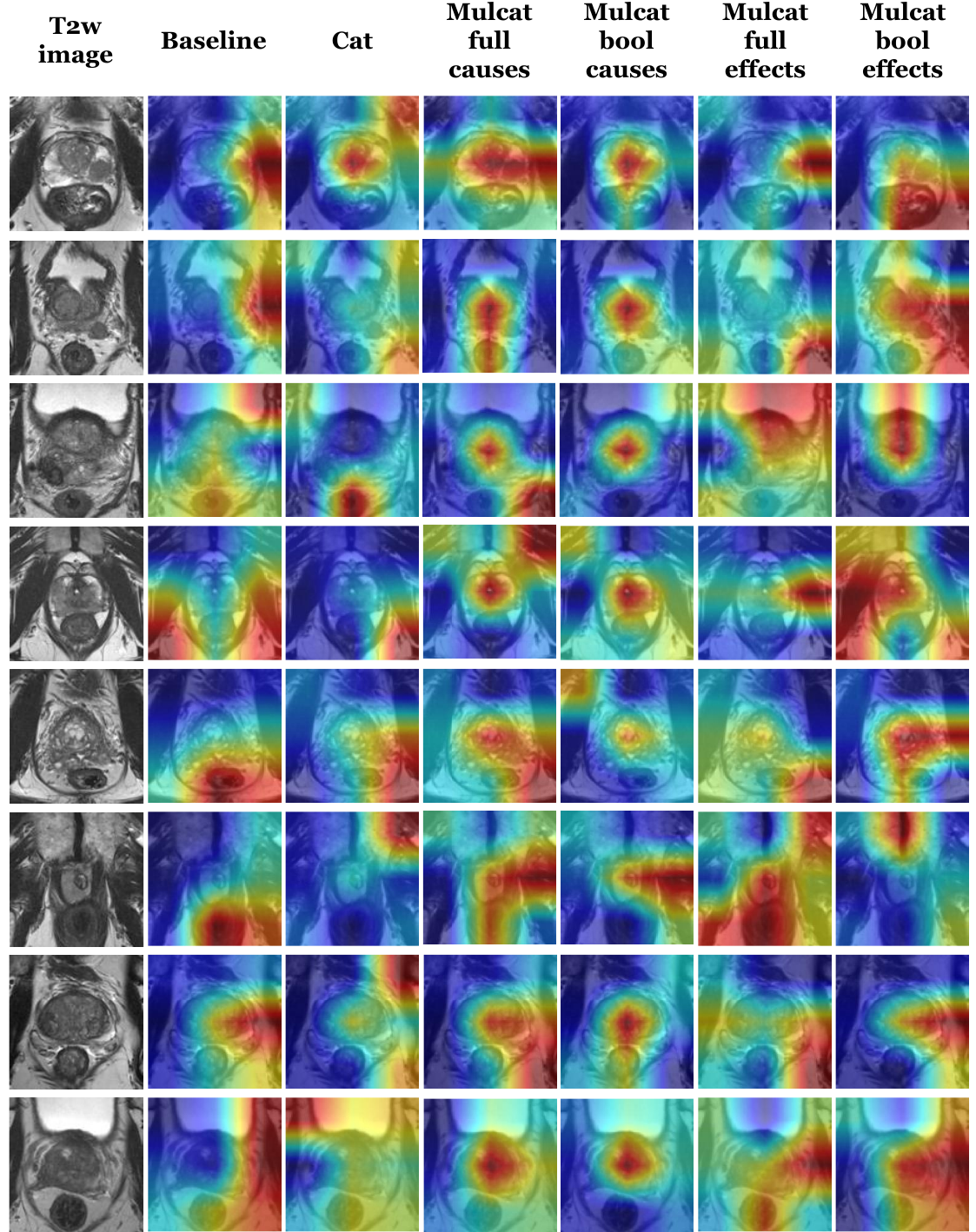}
	  \caption{Visual assessment of class activation maps for \textbf{cancerous} cases of the PI-CAI dataset. Each row represents a different scan, and the columns represent the Grad-CAM outputs for the baseline model, the \textit{Cat} option, and all the proposed \textit{Mulcat} variants. ResNet18 is assumed as the backbone architecture in all models. Best seen in color.}
   \label{fig:CAM_tumor}
\end{figure}

\begin{figure}
	\centering
		\includegraphics[width=0.99\textwidth]{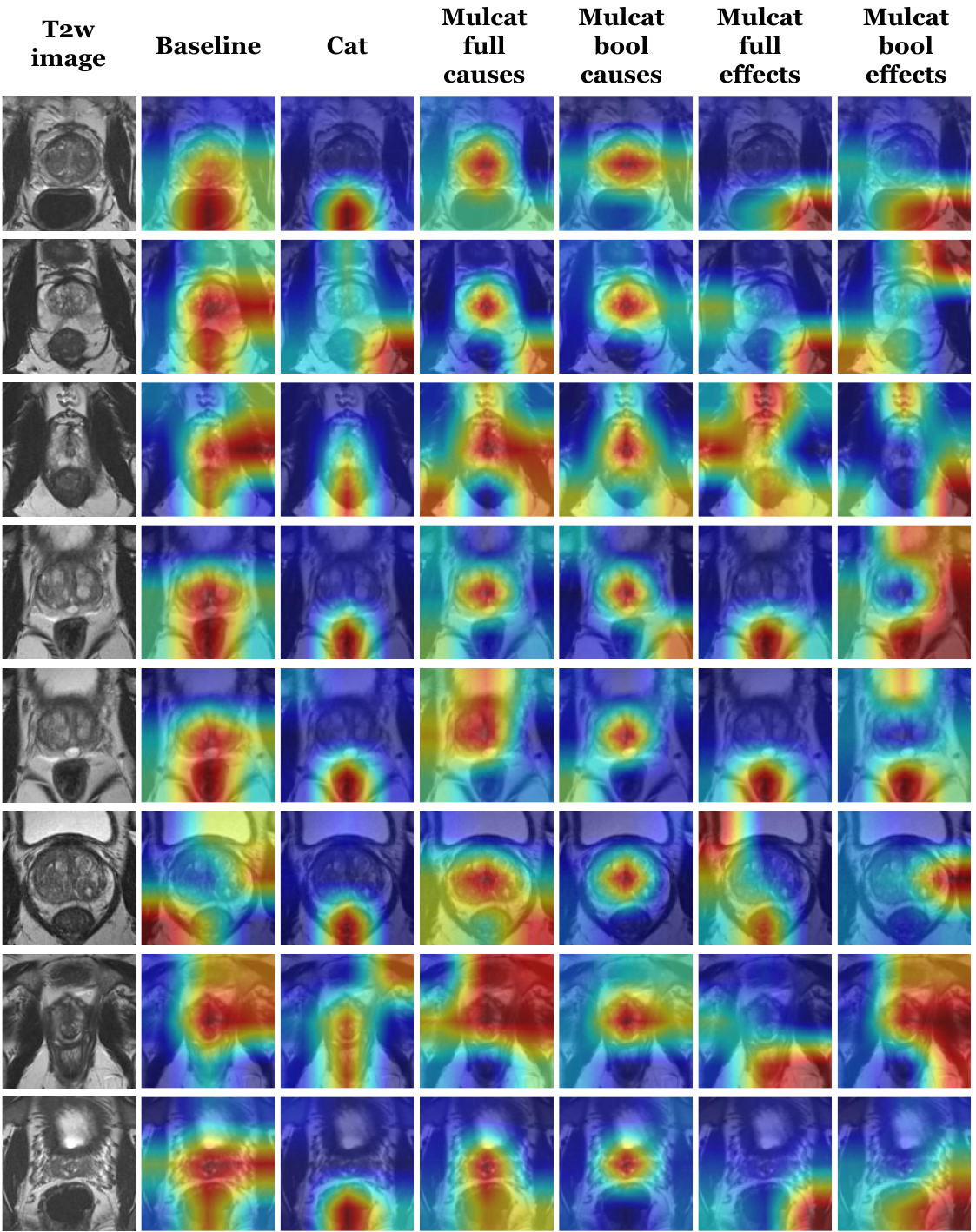}
	  \caption{Visual assessment of class activation maps for \textbf{no-tumor} cases of the PI-CAI dataset. Each row represents a different scan, and the columns represent the Grad-CAM outputs for the baseline model, the \textit{Cat} option, and all the proposed \textit{Mulcat} variants. ResNet18 is assumed as the backbone architecture in all models. Best seen in color.}
   \label{fig:CAM_notumor}
\end{figure}

In addition to the quantitative experiments, we obtained qualitative results for the six models for each dataset by comparison of their \acrshort{label_CAM}s given the same input test images. As an example, Figure \ref{fig:CAM_malignant_breakhis} shows the results for some BreakHis malignant cases for which all the models yielded the same correct prediction. Rows regard different bioptic slides, while columns represent from left to right the original input image, the CAM of the baseline (non-causality-driven) model, the CAM of the \textit{Cat} models, and the \acrshort{label_CAM}s of our proposed \textit{Mulcat} models with their specific settings (i.e., direction \textit{d} and mode \textit{m}).
Similarly, Figures \ref{fig:CAM_tumor} and \ref{fig:CAM_notumor} show results for the PI-CAI dataset on different cancerous and no-tumor cases, respectively. Again, rows represent several scans, while columns represent, from left to right, the original \acrshort{label_T2w} input image and the CAMs of each configuration.

In Section \ref{sec:discussion}, we will give more detailed interpretation of those visual results. Indeed, we discuss when the models focus on regions densely populated by nuclei or lateral zones on BreakHis data from Figure \ref{fig:CAM_malignant_breakhis}, or how the field of view and the anatomic structures influence the models on PI-CAI data for Figures \ref{fig:CAM_tumor} and \ref{fig:CAM_notumor}.

\subsection{Integration with BAMs}
In Table \ref{tab:best_results_BAM}, we show the result of integrating our \textit{Mulcat} module to convolutional networks that utilize BAM attention. Regarding the BreakHis dataset, the regular BAM-based models achieved an accuracy of $89.63$, and utilizing a \textit{Cat} option worsened performance ($85.77$). Conversely, almost all the models that integrated our \textit{Mulcat} modules performed better with a maximum accuracy of $92.07$. We obtained similar results for the PI-CAI dataset (max accuracy: $71.93$).

\begin{table}[t]
    \centering
    \caption{The effect of integrating our \textit{Mulcat} method to models that utilize BAM modules, compared to baseline and competing models. We report accuracy results on the test set as the mean and standard deviation (in lower script) over four repetitions of the experiments with different seeds. The top half of the table refers to the BreakHis dataset, while the bottom half of it refers to the PI-CAI dataset.}
    \resizebox{\textwidth}{!}{
    \begin{tabular}{lllc}
    \hline
    \textbf{Architecture} & \textbf{Causality factors mode} & \textbf{Causality direction} & \textbf{Test set accuracy} [$\uparrow$] \\
    \hline
    \toprule
    \multicolumn{4}{c}{\textbf{\textit{BreakHis dataset}}}\\
    \midrule   
    ResNet18 $+$ BAM \cite{park2020simple} & - & - & $89.63_{1.15}$\\
    \hline
    ResNet18 $+$ BAM $+$ Cat \cite{terziyan2023causality} & - & - & $85.77_{1.72}$\\
    \hline
    \multirow{4}{*}{ResNet18 $+$ BAM $+$ \textbf{Mulcat (ours)}} 
     & Full & Causes & $84.75_{2.01}$ \\
    \cline{2-4}
     & Bool & Causes & $90.65_{2.01}$ \\
    \cline{2-4}
     & Full & Effects & $91.87_{1.15}$ \\
    \cline{2-4}
     & Bool & Effects & $92.07_{0.86}$ \\   
    \bottomrule
    
    \toprule
    \multicolumn{4}{c}{\textbf{\textit{PI-CAI dataset}}}\\
    \midrule
    ResNet18 $+$ BAM \cite{park2020simple} & - & - & $68.00_{0.61}$\\
    \hline
    ResNet18 $+$ BAM $+$ Cat \cite{terziyan2023causality} & - & - & $66.09_{0.41}$\\
    \hline
    \multirow{4}{*}{ResNet18 $+$ BAM $+$ \textbf{Mulcat (ours)}} 
     & Full & Causes & $67.75_{1.99}$ \\
    \cline{2-4}
     & Bool & Causes & $71.25_{0.87}$ \\
    \cline{2-4}
     & Full & Effects & $71.93_{0.37}$ \\
    \cline{2-4}
     & Bool & Effects & $66.52_{1.67}$ \\    
    \bottomrule    
    \end{tabular}
    }
    \label{tab:best_results_BAM}
\end{table}

\subsection{One-Shot Learning Tasks}
The main results of our \acrshort{label_OSL} analysis are reported in Table \ref{tab:best_results_OSL} for both datasets. We report all values as mean and standard deviation \acrshort{label_AUROC} across all the $600$ meta-test tasks.
Regarding the BreakHis dataset, the \textit{2-way} experiment was where our \textit{Mulcat} module improved the models the most. Indeed, while the baseline achieved $0.51$ AUROC, we achieved up to $0.69$ AUROC with the ResNet18$+$\textit{Mulcat-Bool}. In contrast, we found this improvement to be lower in the case of \textit{4-way} experiments. Table \ref{tab:best_results_OSL} also shows the results for \textit{damaged Mulcat} models, which consistently performed worse than their \textit{Mulcat} counterparts.
Concerning the PI-CAI dataset, embedding our \textit{Mulcat} module improved the models in all scenarios, with a more pronounced improvement in the \textit{4-way 1-shot*}, where the models are trained to distinguish four classes (ISUP $2-5$), but the AUROC is computed between \acrshort{label_ISUP} $2$ versus rest.
\begin{table}[t]
    \centering
    \caption{Results of the best-performing models under One-Shot Learning (OSL) settings in terms of AUROC values across all the 600 meta-test tasks as the mean and standard deviation (in lower script). The top half of the table refers to the BreakHis dataset, while the bottom half of it refers to the PI-CAI dataset. *: Trained to distinguish four classes, but the AUROC is computed with \textit{One-vs-rest}.}
    \resizebox{\textwidth}{!}{
    \begin{tabular}{lcccc}
    \hline
    \textbf{Architecture } & \textbf{Causality factors mode} & \textbf{2-way 1-shot} [$\uparrow$] & \textbf{4-way 1-shot} [$\uparrow$] & \textbf{4-way 1-shot*} [$\uparrow$]\\
    \hline
    \toprule
    \multicolumn{5}{c}{\textbf{\textit{BreakHis dataset (OSL study)}}}\\
    \midrule
    ResNet18 & - & $0.51_{0.15}$ & $0.59_{0.06}$ & $0.58_{0.20}$\\
    \hline
    \multirow{2}{*}{ResNet18 $+$ \textbf{Mulcat}} 
     & Full & $0.66_{0.24}$ & $0.59_{0.08}$ & $0.59_{0.15}$ \\
    \cline{2-5}
    & Bool & $0.69_{0.25}$ & $0.57_{0.09}$ & $0.57_{0.15}$ \\
    \hline     
    \multicolumn{5}{c}{\textbf{\textit{BreakHis dataset (OSL ablation study)}}}\\
    \hline
    \multirow{2}{*}{ResNet18 $+$ Damaged-Mulcat} 
     & Full & $0.51_{0.12}$ & $0.52_{0.06}$ & $0.52_{0.12}$ \\
    \cline{2-5}
     & Bool & $0.66_{0.25}$ & $0.50_{0.05}$ & $0.56_{0.13}$ \\ 
    \bottomrule
    
    \toprule
    \multicolumn{5}{c}{\textbf{\textit{PI-CAI dataset (OSL study)}}}\\
    \midrule
    ResNet18 & - & $0.54_{0.14}$ & $0.58_{0.07}$ & $0.59_{0.12}$\\
    \hline
    \multirow{2}{*}{ResNet18 $+$ \textbf{Mulcat}} 
     & Full & $0.55_{0.14}$ & $0.61_{0.07}$ & $0.71_{0.12}$ \\
    \cline{2-5}
    & Bool & $0.56_{0.14}$ & $0.61_{0.07}$ & $0.71_{0.12}$ \\
    \hline     
    \multicolumn{5}{c}{\textbf{\textit{PI-CAI dataset (OSL ablation study)}}}\\
    \hline
    \multirow{2}{*}{ResNet18 $+$ Damaged-Mulcat} 
     & Full & $0.53_{0.14}$ & $0.55_{0.06}$ & $0.55_{0.11}$ \\
    \cline{2-5}
     & Bool & $0.54_{0.14}$ & $0.57_{0.07}$ & $0.61_{0.12}$ \\    
    \bottomrule
    
    \end{tabular}
    }
    \label{tab:best_results_OSL}
\end{table}

\section{Discussion}\label{sec:discussion}
In this work, we presented a new method for automatically classifying medical images that use weak causal signals in the image to model how the presence of a feature in one part of the image affects the appearance of another feature in a different part of the image. Our plug-and-play \textit{Mulcat} module leverages causality maps in a new way and extracts multiplicative factors that eventually weight feature maps according to their causal influence in the scene. Our results seem to indicate that this lightweight, attention-inspired mechanism makes it possible to exploit weak causality signals in medical images to improve neural classifiers without any additional supervision signal.

In our main study, we assessed the effectiveness of our method under a fully-supervised learning scheme. In general, all the models obtained with our \textbf{Mulcat} implementation achieved higher performance than the \textit{baseline} (ResNet18) on the test set with both datasets (see Table \ref{tab:best_results}). This superiority ranged from a minimum of $+2.31$\% to a maximum of $+5.03$\%. On the other hand, utilizing the \textbf{Cat} option from \cite{terziyan2023causality} resulted in worse performance than most of our \textit{Mulcat} model and, with BreakHis data, even of the \textit{baseline}.
We found that most best-performing models used the \textit{Lehmer} method to get the causality map. Nevertheless, this choice comes with the drawback of necessitating more memory than the \textit{Max} method. We experimented with six different integer values for the parameter \textit{p} to sample the range of possible values. A possible improvement would be to let the network itself learn the parameter \textit{p} by back-propagation instead of giving it a fixed value beforehand.

To further confirm the numerical results of our studies, we conducted partially ablating studies on the actual influence of the causality factors on generating useful causality-driven feature maps. As anticipated, when we damage the causal weights by replacing them with random vectors, the accuracy of the final model is lower than its \textit{main study} counterpart (see Table \ref{tab:best_results}).
The \textit{Damaged Cat} performed worse than the \textit{None (baseline)} because the network was likely to be confused by the large number of random values that were concatenated to the actual extracted features. We expected that concatenating a random vector, not trained in back-propagation, would be worse than concatenating nothing at all.
In the \textit{Damaged Mulcat}, the weights multiplying the features maps are random, and, being untrained, they are re-computed at each iteration without any optimization from previous iterations. This results in scenarios where depending on the multiplication factors, irrelevant features are amplified while important ones are suppressed. When \textit{Damaged Mulcat} is used with the \textit{full} option, this behavior is more pronounced (random weights can have very high values, up to $k-1$), and performance is low (even lower than baseline for PI-CAI data) because the network assigns a lot of importance to these potentially incorrect features. In contrast, when the \textit{bool} option is used, this behavior is mitigated (random weights have a maximum value of $1$), so the degree of confusion of the network is reduced, and performance is higher.
Experiencing reduced performance when the causality maps and factors are completely ablated or partially corrupted suggests that our module is computing something significant. This observation indicates that, even if weak, the causality signals learned during training assist the network to perform better.

Although we notice the improvement of our \textbf{Mulcat} models over the baseline from a quantitative point of view, it seems that the different combinations of mode \textbf{m} and direction \textbf{d} behaved roughly the same way.
Thus, we wanted to investigate the potential benefits of our proposition on a different level. We deepened the analysis and found that significant differences can emerge on the \acrshort{label_XAI} side, supporting the role of causality in explainability (Chapter \ref{chap:review_XAI_causality}). 
On BreakHis data, the \textbf{Mulcat-full} and \textbf{Mulcat-bool} models manage to focus on regions densely populated by nuclei, which is what pathologists do at this magnification level. Instead, \textit{Baseline} models often pay attention to lateral zones or regions less critical for the malignancy classification. The latter behavior is also observed in the \textit{Cat} models, which frequently focus on lateral, small, and/or irrelevant portions for classification purposes (see Figure \ref{fig:CAM_malignant_breakhis}). 
As for the PI-CAI dataset, where the field of view is larger and comprises many different anatomic structures other than the prostate, we noticed a trend that \textbf{Mulcat-full-causes} and \textbf{Mulcat-bool-causes} are consistently more focused on the discriminative parts of the image (e.g., prostate gland area). Conversely, the other options led to models that often looked at the rectum, bladder, or lateral muscle bundles (see Figures \ref{fig:CAM_tumor} and \ref{fig:CAM_notumor}). This fact confirms our hypothesis that using causes and not effects allows the network to obtain more faithful results.

Among the methods that exploit the information of causality maps, \textbf{Cat} proves to be one of the worst. That is evident both quantitatively and qualitatively. The reason for this behavior could be the considerable complexity added to the model to account for all the combinations of feature maps. In fact, on the classifier, the number of input neurons goes from $n\times n\times k$ to $n\times n\times k + k\times k$, which with high $k$ results in thousands of additional connections (e.g., $512\times 512=262144$ new neurons for a ResNet18).
The overhead induced by the \textit{Cat} method is quantitatively confirmed by the memory requirements summarized in Table \ref{tab:memory_requirements}. Instead, our \textit{Mulcat} method increases the memory demand by a negligible amount compared to the baseline, promoting it as a low-cost improvement of regular architectures.

We conducted additional experiments to further add evidence of improved performance through our method. First, we showed that our \textit{Mulcat} module can be easily integrated into existing architectures, such as attention-based \acrshort{label_BAM} networks, and can create synergy in improving performance (see Table \ref{tab:best_results_BAM}). Indeed, utilizing \textit{Mulcat} led to an increase of up to $+2.72\%$ over the regular BAM-ResNet18 and of $+7.34\%$ over the BAM-ResNet18 that used \textit{Cat}, for the BreakHis dataset. Similarly, BAM-ResNet18 which utilized our \textit{Mulcat} option on the PI-CAI dataset achieved up to $+5.78\%$ and $+8.83\%$ w.r.t regular BAM-ResNet18 and \textit{Cat} BAM-ResNet18, respectively.

Second, to tend towards a more generalized demonstration of classification problems, it was interesting to understand how our \textit{Mulcat} method worked in practical application situations, such as the shortage of annotated data in the medical imaging domain. Thus, we performed \acrshort{label_OSL} experiments, both in \textit{2-way} and \textit{4-way} settings (see Table \ref{tab:best_results_OSL}). Our findings suggest that using \textit{Mulcat} can be an effective choice even in low-data scenarios. Indeed, when performing binary classification over the BreakHis data (i.e., \textit{benign-vs-malignant}), ResNet18$+$Mulcat achieved an accuracy up to $+35.2\%$ compared to the baseline, while this improvement was broadly reduced when performing \textit{4-way} experiments. On the other hand, it is on the \textit{4-way} scenarios with PI-CAI data that ResNet18$+$Mulcat outperformed the baseline the most, with an increase of $20.3\%$ accuracy for the \textit{4-way 1-shot*} setting. 

One of the limitations of our work is that we used only ResNet18 as the backbone architecture to extract latent representations for the different implementations, although this is consistent with the pilot nature of our study. 
Moreover, we acknowledge that our methods consider potential causal relationships in pairs rather than among more than two features. That, of course, can lead to suboptimal results, given the impossibility of excluding confounders. In future experiments, we would be interested in extending the operation to more variables and devising variations inspired by the classic PC algorithms of the literature on causal discovery in tabular data \cite{spirtes1991algorithm}.

There could be other directions worth exploring starting from our work, both on the application and architectural level.
It would be interesting to draw inspiration from multi-depth visual attention and extract the causality map from the inner layers on the backbone in addition to the last. Moreover, new methods to combine causal information besides concatenation and weighting of feature maps could be proposed, and experimenting with ensemble methods would be another direction. For more insights and discussion, see Section \ref{sec:conclusions_futureDirections} of the Conclusions chapter.

\section{Summary}
\label{sec:conclusions}
In this work, we introduced a novel technique to discover and exploit weak causal signals directly from medical images via neural networks for classification purposes. Our method consists of a \acrshort{label_CNN} backbone and a causality-factors extractor module, which computes weights for the feature maps to enhance each feature map according to its causal influence in the image's scene in an attention-inspired fashion. We developed different architecture variants and empirically evaluated all of our models on two public datasets of medical images for cancer diagnosis. Moreover, we verified that our module can create synergies when introduced in existing attention-based architectures, and we verified its applicability to few-shot learning settings. 

Our findings demonstrate how minor modifications to traditional models can enhance them. Indeed, our lightweight module can be easily integrated into regular CNN classification systems and produce better models without requiring additional trainable parameters. It enhances the overall classification results and makes the model focus more precisely on the critical regions of the image, leading to more accurate and robust predictions.
This aspect is crucial in medical imaging, where accurate and reliable classification is essential for effective diagnosis and treatment planning. Nevertheless, what we propose in this work may have a broader significance, such as non-medical tasks or application to other data types, such as videos.
We believe that the new elements we introduce with our work are a way to connect machine vision and causal reasoning in a novel way, especially when no prior knowledge of the data is available, adding a unique dimension to the framework. 
\chapter{CROCODILE: Causality aids RObustness via COntrastive DIsentangled LEarning}
\label{chap:crocodile}
\begin{figure}[h!]
\includegraphics[width=\textwidth]{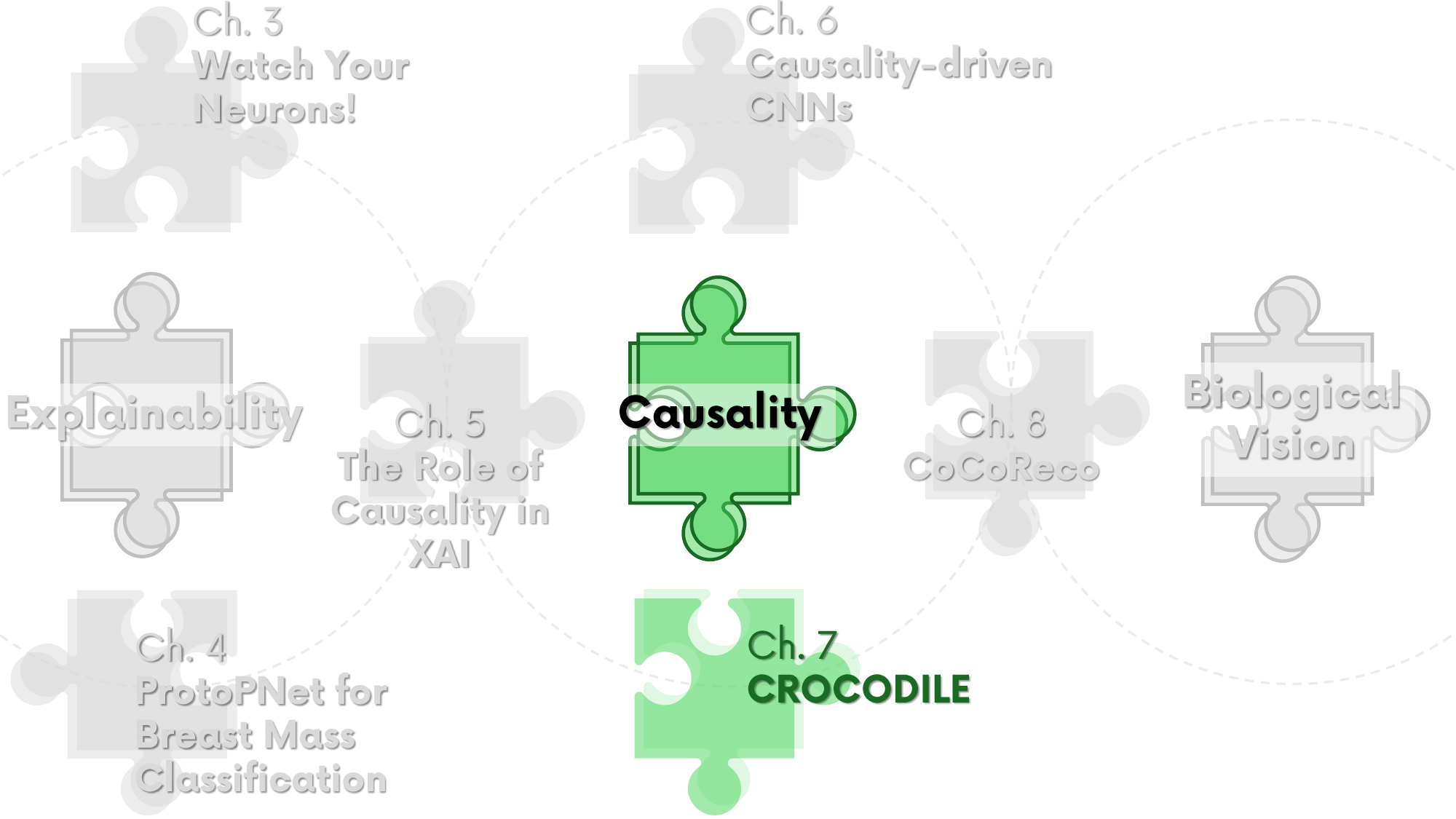}
\end{figure}

As we have also seen in Section \ref{sec:back_causality}, \acrshort{label_ML}/\acrshort{label_DL} image classifiers often struggle with domain shift, leading to significant performance degradation in real-world applications.
Domain shift bias is the problem of models performing not consistently across \textit{in-distribution} (\acrshort{label_ID}) and \textit{out-of-distribution} (\acrshort{label_OOD}) data. The former are independent and identically distributed (\acrshort{label_iid}) to the data on which the model was trained. Conversely, data are OOD when their distribution essentially differs from the source one, such as chest X-rays (\acrshort{label_CXR}) coming from a different hospital than the training one \cite{pooch2020can,cohen2020limits,zhang2022learning}.
Conversely, humans learn robust, causal, and transportable mechanisms and rapidly adapt to evolving scenarios. When you learned about the existence of gravity as a child while playing, you realized that it was also important on other objects - which were not perfectly identical to the first toy you dropped. When you later grew up and learned to drive a car, you learnt the general mechanism, not the specific skill of knowing how to drive that individual car model.
However, traditional \acrshort{label_ML} models still tend to rely on spurious correlations seen during training for predicting the outcome and spectacularly fail when those shortcut associations are not present in OOD data, for instance, due to variations in scanner settings, image artifacts, or patient demographics \cite{castro2020causality,zhang2020coping,degrave2021ai,robinson2021can,sanchez2022causal,bercean2023breaking,hartley2023neural,vilouras2023group}. For this reason, the field of domain generalization (\acrshort{label_DG}) has searched for ways to make \acrshort{label_DL} models learn robust features that could generalize better to unseen domains \cite{li2021domain,kim2022broad,ouyang2022causality,wang2022generalizing,zunaed2024learning}.

Conceptually, we could think of a set of features that causally determine the outcome and are invariant to shifts in irrelevant/spurious attributes, as well as a separate set of features that are spuriously correlated with the outcome but do not have a causal effect. Some works have proposed using tools from causal inference to achieve this disentanglement \cite{wang2021causal,sui2022causal,nie2023chest,qu2024causality}. The common idea is that using the causal instead of the spurious features would allow a model to learn the underlying mechanism and be more robust on new data.
However, these efforts try to model domain shifts implicitly, with a scope limited to the disease prediction task, disregarding the wealth of information on possible domain shifts from different source data sets.

In this work, we introduce the CROCODILE framework and advance this causal/spurious feature disentanglement on a cross-domain level by leveraging information from different datasets in a contrastive learning setting. We conceive a domain-prediction branch along the disease-prediction branch to instill domain awareness into the model's representations. Moreover, we propose a new way to inject background medical knowledge, effectively designing a task-prior to guiding learning and fostering DG.
As our findings reveal, the model relies less on spurious correlations, learns the mechanism bringing from images to prediction better, and outperforms baselines on \acrshort{label_OOD} data. We apply our method to multi-label lung disease classification from \acrshort{label_CXR}), utilizing over 750000 images from four datasets. Our bias-mitigation method improves domain generalization, broadening the applicability and reliability of deep learning models for a safer medical image analysis.

The content of this Chapter is based on the following publication
\begin{itemize}
    \item Carloni, G., Tsaftaris, S. A., \& Colantonio, S. (2024, October). "CROCODILE: Causality aids RObustness via COntrastive DIsentangled LEarning". In \textit{International Workshop on Uncertainty for Safe Utilization of Machine Learning in Medical Imaging}. MICCAI 2024. Lecture Notes in Computer Science, vol 15167. Springer, Cham. \url{https://doi.org/10.1007/978-3-031-73158-7_10} (\cite{carloni2024crocodile}).
    
    \item and the corresponding Python/Pytorch code can be found on my GitHub page at: \url{https://github.com/gianlucarloni/crocodile}.
\end{itemize} 

\section{Methodology}
\subsection{A Causal Viewpoint on Medical Image Classification}
\begin{figure}[t]
\includegraphics[width=\textwidth]{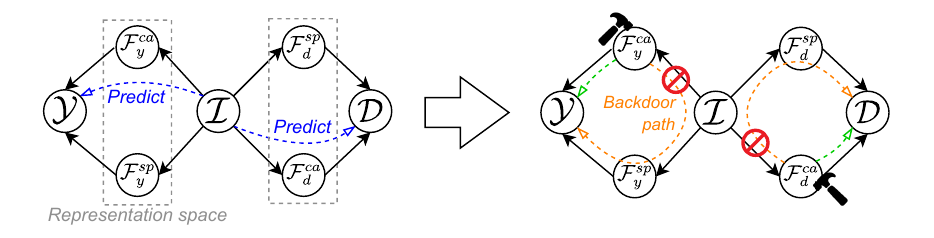}
\caption{A causal view on classifying medical images $\mathcal{I}$ coming from different domains $\mathcal{D}$ for the presence of diseases $\mathcal{Y}$. By applying the latent causal intervention (hammer), the backdoor path through the spurious features is cut off.}
\label{fig:introduction}
\end{figure}
We define a structural causal model (SCM) \cite{pearl2010causal} for medical image classification in Fig \ref{fig:introduction}.
Given the input images $\mathcal{I}$, such as \acrshort{label_CXR}s, and the disease classification $\mathcal Y$, we obtain two sets of features via feature extraction.
We denote $\mathcal F^{ca}_y$ the causal features that truly determine the outcome (e.g., the patchy airspace opacification typical in pneumonia).
Similarly, we denote $\mathcal F^{sp}_y$ the spurious features, determined by data bias's confounding effect, which are unrelated to a disease (e.g., metal tokens on the image corners).
Ideally, $\mathcal Y$ should be caused only by $\mathcal F^{ca}_y$, but is naturally confounded by $\mathcal F^{sp}_y$, as both types of features 
usually coexist in medical data. Unfortunately, conventional models tend to learn the correlation $P(\mathcal{Y}|\mathcal{F}^{ca}_y)$ via the shortcut (backdoor) path $\mathcal{F}^{ca}_y \leftarrow \mathcal{I} \rightarrow \mathcal F^{sp}_y \rightarrow \mathcal Y$ instead of the desired $\mathcal F^{ca}_y \rightarrow \mathcal Y$. As we detail next, we exploit the \textit{do-calculus} from causal theory \cite{pearl2014interpretation} on the causal features to block the backdoor path, estimating $P(\mathcal{Y}|do(\mathcal{F}^{ca}_y))$.
Following the same idea, we conceive two other sets of features extracted from $\mathcal{I}$, this time concerning the trivial task of predicting from which source domain come the data $\mathcal{D}$: $\mathcal F^{ca}_d$ would be the causal features that are relevant to distinguish different domains, and $\mathcal F^{sp}_d$ the spurious (confounding, irrelevant) features. 

\subsection{Disease-branch and Domain-branch}
\begin{figure}
\includegraphics[width=\textwidth]{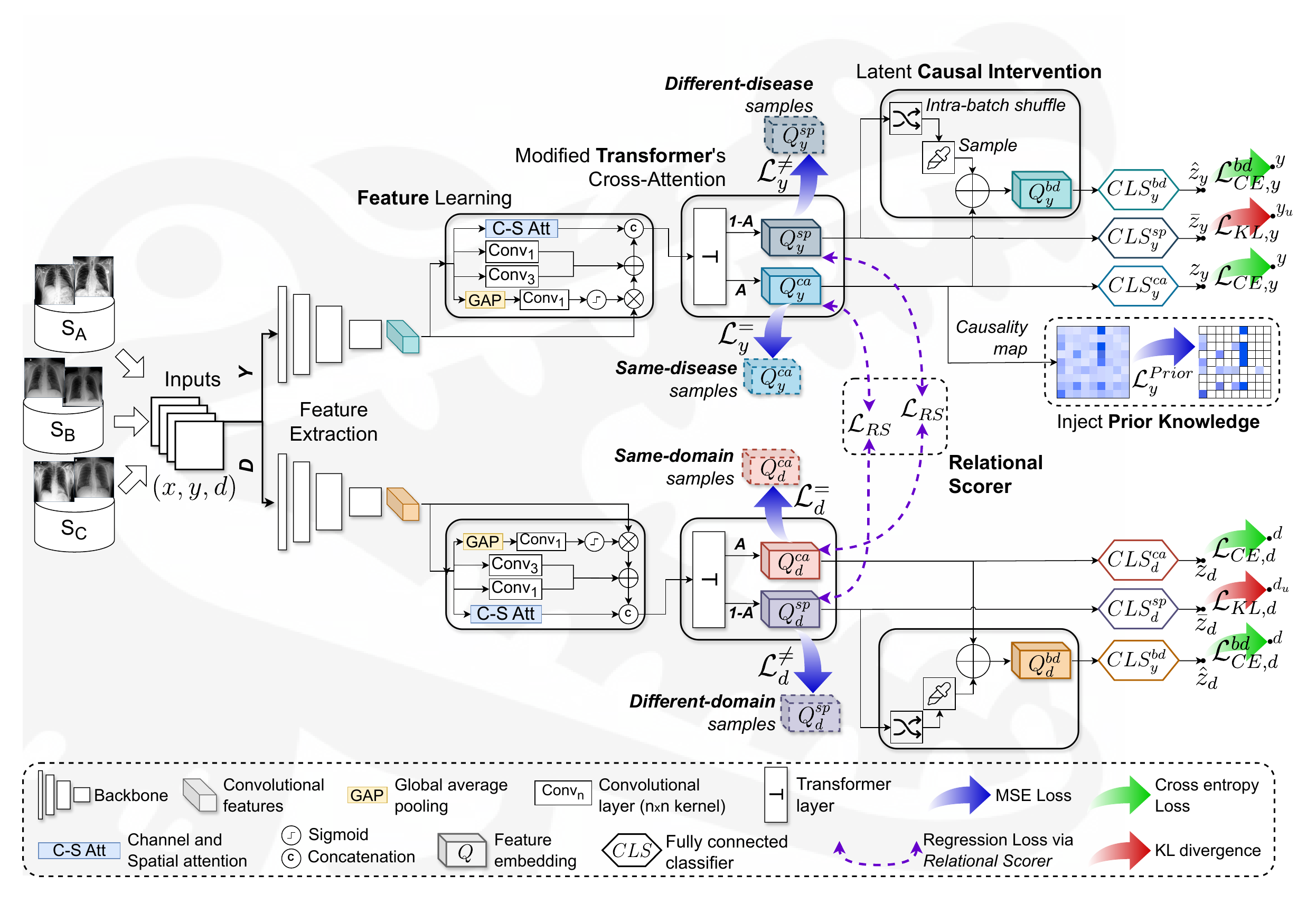}
\caption{CROCODILE involves two branches to learn robust, invariant features for predicting the labels from medical images (e.g., multi-label findings from \acrshort{label_CXR}s) while disregarding confounding features.
We disentangle \textit{causal} features determining the label from \textit{spurious} features associated with the label due to domain shift.
We exploit images from multiple domains in a contrastive learning scheme and propose a new way to inject prior knowledge. Best seen in color.}
\label{fig:crocodile_framework}
\end{figure}

We present our overall framework in Fig \ref{fig:crocodile_framework}. 
A \textit{disease prediction} branch learns to extract useful image features to predict the medical finding (e.g., pneumothorax or atelectasis in a \acrshort{label_CXR}), regardless of the different domains. On another parallel branch for \textit{domain prediction}, the image features that are useful for the trivial task of predicting the domain the images come from are learned (regardless of the different diseases). The architecture is trained end-to-end. Each branch involves a feature extraction backbone followed by a block to enhance features
via channel- and spatial- attention \cite{pan2022integration}.
Then, a Transformer network \cite{vaswani2017attention} yields the set $\mathbf{A}$ of attention scores, typically in the range 0-1, that identifies the portion of the input that is causally relevant to the task of interest (i.e., \textit{what knowledge does the network use to make predictions}). Given an arbitrary set $\mathbf{A}$, we modify the Transformer's cross-attention mechanism to yield also the complementary set $\mathbf{1}-\mathbf{A}$ ($\mathbf{1}$ is the all-one matrix), representing the trivial/spurious aspects of the input. This way, we encode disentangled causal and spurious feature embeddings, $Q^{ca}$ and $Q^{sp}$, by modulating the features by $\mathbf{A}$ and $\mathbf{1}-\mathbf{A}$, respectively.
Finally, three classifiers connect the features $Q$ to the classification logits $z$.
In the following sections, we design specific contrastive learning losses and introduce a novel way to inject prior knowledge about the medical task.

\subsection{Feature Disentanglement and Causal Intervention}
For each branch, we need to make $Q^{ca}$ and $Q^{sp}$ capture the authentic and trivial aspects from the input samples. To achieve the correctness of the predictions, we impose two \acrshort{label_CE} loss terms, $\mathcal L_{CE,y}$ and $\mathcal L_{CE,d}$, over the classification logits $z_y$ and $z_d$ from the causal features $Q^{ca}_y$ and $Q^{ca}_d$, supervised by the disease labels $y$ and domain labels $d$, respectively.

To make $Q^{sp}$ features encode the trivial patterns that are unnecessary for classification, we push its predictions $\bar{z}_y$ and $\bar{z}_d$ evenly to all respective categories. We define the uniform classification losses $\mathcal L_{KL,y}$ and $\mathcal L_{KL,d}$ as the \acrshort{label_KL}-divergence between the spurious features and the respective uniform distribution ($y_u$ or $d_u$).

To alleviate the confounding effect, we implement the backdoor adjustment by performing a latent causal intervention \cite{sui2022causal,nie2023chest}: we stratify the spurious features appearing from training data and pair the causal set of features with those stratified spurious features to compose the \textit{intervened} graph. This way, we fit the concept of \textit{borrowing from others} (i.e., "if everyone has it, it is as if no one has it").
We impose \acrshort{label_CE} losses $\mathcal L_{CE,y}^{bd}$ and $\mathcal L_{CE,d}^{bd}$ between the logits $\hat{z}_y$ and $\hat{z}_d$ obtained from the corresponding intervened features $Q^{bd}$ and the same ground-truth label for the causal features. This way, we push the predictions of such intervened images to be invariant and stable across different stratifications due to shared causal features. Practically, we approximate this operation with an intra-batch shuffling of $Q^{sp}$ followed by random sampling (with $0.3$ drop probability) and addition to $Q^{ca}$.
By combining the supervised \acrshort{label_CE} loss, the \acrshort{label_KL} loss, and the backdoor CE loss for each branch, we obtain the two following equations:

\begin{equation}
\label{eq:L_y}
\mathcal{L}_y = - (
    \lambda_1
    \underbrace{y^\top\!\!\log (z_y)}_{\mathcal{L}_{CE,y}}
    +
    \lambda_2
    \underbrace{KL(y_{u}, \bar{z}_y)}_{\mathcal{L}_{KL,y}}
    +
    \lambda_3
    \underbrace{y^\top\!\!\log (\hat{z}_y)}_{\mathcal{L}_{CE,y}^{bd}}
    )
\end{equation}
\begin{equation}
\label{eq:L_d}
\mathcal{L}_d = - (
    \lambda_4
    \underbrace{d^\top \log (z_d)}_{\mathcal{L}_{CE,d}}
    +
    \lambda_5
    \underbrace{KL(d_{u}, \bar{z}_d)}_{\mathcal{L}_{KL,d}}
    +
    \lambda_6
    \underbrace{d^\top \log (\hat{z}_d)}_{\mathcal{L}_{CE,d}^{bd}}
    )
\end{equation}

\subsection{Contrastive Learning}
To attain cross-domain robustness, we posit there should also exist an alignment between the \textit{causal} features that determine the \textit{disease} and the \textit{spurious} features for the \textit{domain} prediction task. And the converse should also be true.
For instance, we want the regions of the image that determine the presence of pneumonia to be unrelated to what contributes to discerning different domains (e.g., spurious metal tokens). Conversely, the image aspects determining which domain the image comes from should be unrelated to what determines disease prediction.

\begin{figure}[t]
\includegraphics[width=\textwidth]{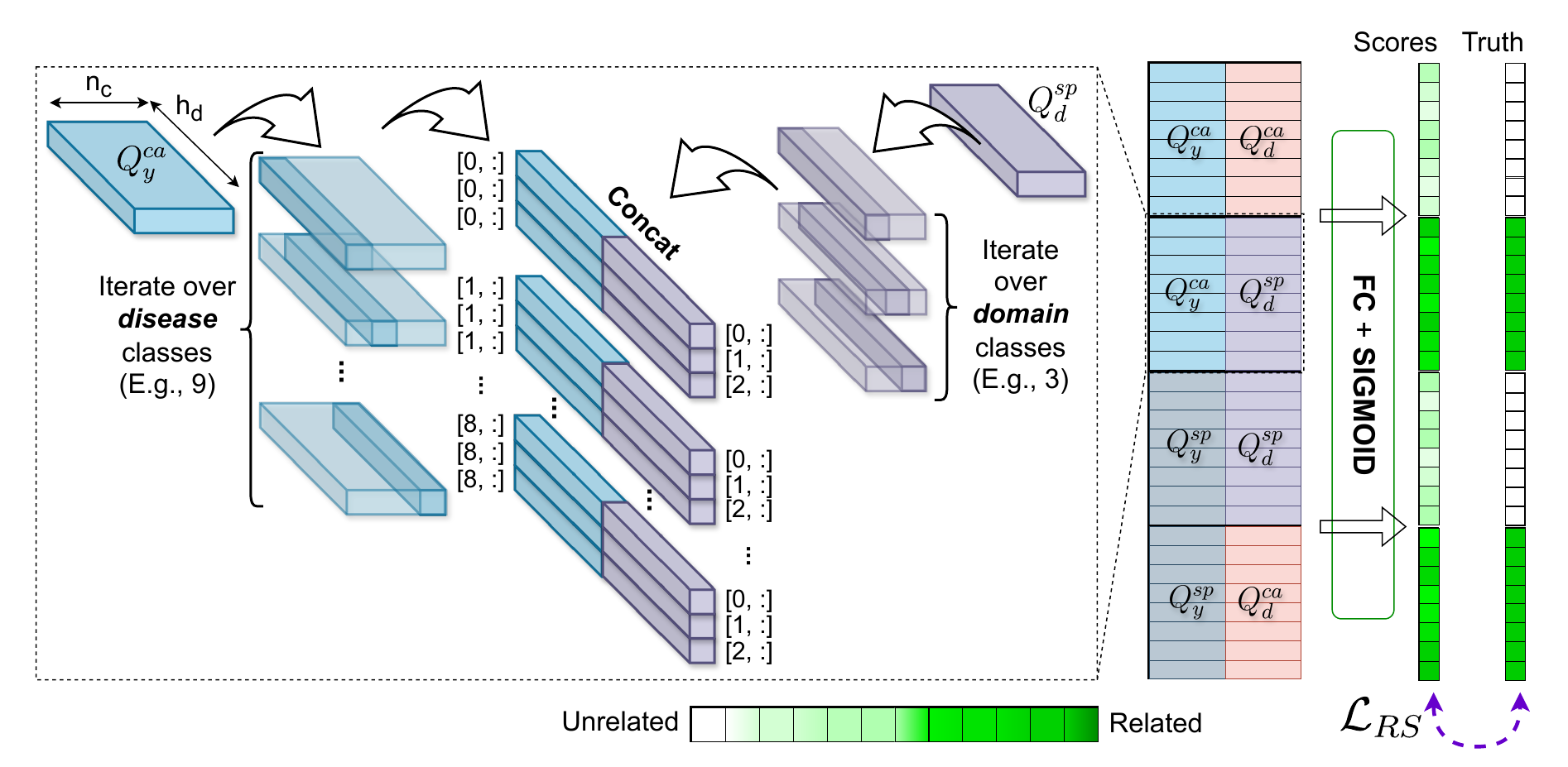}
\caption{Our \textit{Relational Scorer} stratifies and concatenates every combination of \textit{causal} and \textit{spurious} features across both tasks. With a fully connected layer and a consecutive sigmoid($\cdot$), it maps each pair to a \textit{relational score} between 0 and 1. We use an \acrshort{label_MSE} loss regressing the relational scores to the ground truth. The model \textit{learns to compare} the four sets of disentangled features. Best in color.}
\label{fig:relational_scorer}
\end{figure}

However, we are interested in measuring the \textit{relational} alignment rather than the structural similarity of the representations. Matched (mismatched) pairs should "inform" ("repel") each other. Therefore, inspired by the concept of \textit{learning to compare} \cite{sung2018learning,cao2022learning}, we design a new module named \textbf{Relational Scorer} (\acrshort{label_RS}) to learn which image representations' pairings are semantically related and which are not (Fig \ref{fig:relational_scorer}).
Our RS stratifies and combines each possible cross-branch pairing $p\in P=\{Q_y^{ca} \times Q_d^{ca} \cup Q_y^{ca} \times Q_d^{sp}\cup Q_y^{sp} \times Q_d^{sp} \cup Q_y^{sp} \times Q_d^{ca} \}$ and then maps them to a \textit{relational score} between $0$ and $1$.
We use an \acrshort{label_MSE} loss regressing the relational scores $r$ to the ground truths $r^{GT}$: matched pairs have a similarity of $1$, and the mismatched pair have a similarity of $0$.

Although this problem may seem to be a \textit{classification} problem with label space \{$0$, $1$\}, we are predicting relation scores, which can be considered a \textit{regression} problem (with $r^{GT} \in \{0, 1\}$ generated by construction).
We set the ground truth to $1$ for the $Q_y^{ca}$-$Q_d^{sp}$ and $Q_y^{sp}$-$Q_d^{ca}$ pairings, and $0$ otherwise.
The resulting regression loss term is:
\begin{equation}
    \mathcal{L}_{RS}= - \lambda_7 \sum_{i=1}^{|P|}(r_i-r_i^{GT})^2
    \label{eq:L_RS}
\end{equation}

Moreover, we conceive other loss terms to enforce consistency/separation of medical image representations in a contrastive setting at a \textit{batch} level:
\begin{itemize}
    \item $\mathbf{\mathcal L_y^=}$: samples exhibiting a \textbf{common} radiological \textbf{finding} should lie close in \textit{disease-causal} feature space $Q_y^{ca}$, regardless of the source domain.
    \item $\mathbf{\mathcal L_y^{\neq}}$: samples exhibiting \textbf{different} radiological \textbf{findings} should lie close in \textit{disease-spurious} feature space $Q_y^{sp}$, regardless of the source domain.
    \item $\mathbf{\mathcal L_d^=}$: samples from the \textbf{same dataset} should lie close in \textit{domain-causal} feature space $Q_d^{ca}$, regardless of the diseases.
    \item $\mathbf{\mathcal L_d^{\neq}}$: samples from \textbf{different datasets} should lie close in \textit{domain-spurious} feature space $Q_d^{sp}$, regardless of the diseases.
\end{itemize}

We implement each of such terms via an \acrshort{label_MSE} loss between the representation $Q$ of each sample in the batch and the corresponding average representation $\Tilde{Q}$ of samples with the same/different label:
\begin{equation}
\label{eq:L_y_batch}
\mathcal{L}_y^{batch} = - (
    \lambda_8
    \underbrace{\sum_{y \in \mathcal{Y}}(Q_y^{ca}-\Tilde{Q}_y^{ca})^2
    }_{\mathcal{L}_y^=}
    +
    \lambda_9
    \underbrace{\sum_{y \in \mathcal{Y}}(Q_y^{sp}-\Tilde{Q}_{not(y)}^{sp})^2
    }_{\mathcal{L}_y^{\neq}}
    )
\end{equation}
\begin{equation}
\label{eq:L_d_batch}
\mathcal{L}_d^{batch} = - (
    \lambda_{10}
    \underbrace{\sum_{d \in \mathcal{D}}(Q_d^{ca}-\Tilde{Q}_d^{ca})^2
    }_{\mathcal{L}_d^=}
    +
    \lambda_{11}
    \underbrace{\sum_{d \in \mathcal{D}}(Q_d^{sp}-\Tilde{Q}_{not(d)}^{sp})^2
    }_{\mathcal{L}_d^{\neq}}
    )
\end{equation}
\noindent where $\mathcal{Y}$ and $\mathcal{D}$ are the possible disease and domain labels seen in the batch.
To compute those losses correctly, we design a custom sampler favoring consistent batches where the class prevalence is respected. 

\subsection{Injecting Prior Knowledge}
\begin{figure}[t]
\includegraphics[width=\textwidth]{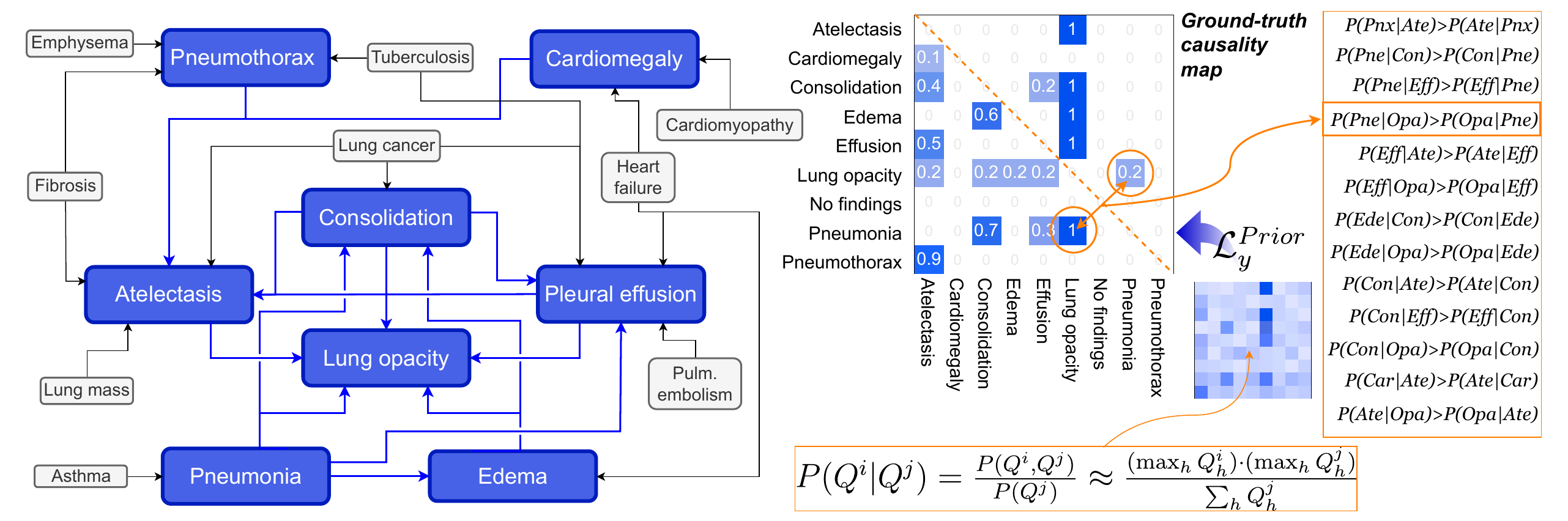}
\caption{Causal graphical model among the CXR findings of interest (blue) and the ground-truth \textit{causality map} defined from that graph. Gray boxes represent additional findings or risk factors (not investigated in this study) that might be associated with the desired ones.}
\label{fig:prior_knowledge}
\end{figure}

Motivated by the high interclass similarity and hierarchical structure of \acrshort{label_CXR} findings \cite{rajaraman2020training,wang2017chestx}, we propose a new method to inject prior (medical) knowledge into the model to guide its learning (Fig.~\ref{fig:prior_knowledge}). Differently from solutions as \textit{conditional training} \cite{pham2021interpreting}, which rely on data, our proposal is desirable to capture semantic priors without relying on data.
We define a causal graph representing the relationship between the \acrshort{label_CXR} findings and propose a novel formulation of the \textit{causality map} concept \cite{carloni2023causality,carloni2024exploiting} to model the co-occurrence of CXR findings in the images.
As we have seen, each $Q_y^{ca}$ representation has shape $n_c \times h$, where $n_c$ is the number of classes (e.g., nine CXR findings) and $h$ is the hidden dimension of the embeddings.
After normalizing $Q_y^{ca}$ by their global maximum batch-wise, they lie in the range $0$-$1$, and we interpret their values as probabilities of the \acrshort{label_CXR} findings to be present in the image. Indeed, given two embeddings $Q^i$ and $Q^j$, to compute the effect of the former on the presence of the latter, we estimate the ratio between their joint and marginal probabilities as:
\begin{equation}
    P(Q^i|Q^j) = \frac{P(Q^i,Q^j)}{P(Q^j)} \approx \frac{(\max_{h} Q^i_{h})\cdot (\max_{h} Q^j_{h})}{\sum_{h} Q^j_{h}}, \forall i, j \in  1 \leq i,j \leq n_c   
\label{eq:cmap}
\end{equation}
\noindent thus obtaining the relationships between embeddings $Q^i$ and $Q^j$, since, in general, $P(Q^i|Q^j) \neq P(Q^j|Q^i)$. By computing these quantities for every pair $i,j$, we obtain the $n_c \times n_c$ map $C_y$. We interpret asymmetries across estimates opposite the main diagonal in $C_y$ as causality signals between their activation. Accordingly, the representation of a CXR finding causes the activation of another when $P(Q^i|Q^j) > P(Q^j|Q^i)$, that is $Q^i \rightarrow Q^j$.  
We design our \textbf{Task-Prior loss} as an MSE loss to push the causality map $C_y$ obtained from the learned representations to the ground-truth causality map $C_y^{GT}$, which we defined by estimating frequencies based on medical knowledge about the possible co-occurrence of CXR findings:
\begin{equation}
    \mathcal{L}_y^{Prior}= - \lambda_{12} (C_y-C_y^{GT})^2
    \label{eq:L_prior}
\end{equation}

Overall, the training objective of our CROCODILE framework is defined as the sum of the losses defined in Equations \ref{eq:L_y}, \ref{eq:L_d}, \ref{eq:L_RS}, \ref{eq:L_y_batch}, \ref{eq:L_d_batch} and \ref{eq:L_prior}:
\begin{equation}
    \mathcal L_{TOT} = \mathcal L_y + \mathcal L_d + \mathcal L_{RS} + \mathcal L_y^{batch} + \mathcal L_d^{batch} + \mathcal L_y^{prior}.
\end{equation}

\section{Experimental Setup}
\subsection{Datasets and Pre-Processing}
We classify eight radiological findings (plus the \textit{No finding} class) from frontal CXR images of four popular data sets in both \acrshort{label_ID} and \acrshort{label_OOD} settings. After cleaning, the number of images for each set is: $112110$ for ChestX-ray14 \cite{wang2017chestx}, $183453$ for CheXpert \cite{irvin2019chexpert}, $95452$ for PadChest \cite{bustos2020padchest}, and $365737$ for MIMIC-CXR \cite{johnson2019mimic}.
For the first dataset, we create the \textit{Lung opacity} class as OR logic across the \textit{consolidation}, \textit{effusion}, \textit{edema}, \textit{pneumonia}, and \textit{atelectasis} classes.
We resize the images to $320 \times 320$ and adjust their contrast in the range $0$-$255$.
\subsection{Experiments}
For ID experiments, we combine images of ChestX-ray14, CheXpert, and PadChest, split them into 80-20\% train and validation sets, and assess the multi-label classification performance via the area under the \acrshort{label_ROC} curve (\acrshort{label_AUC}) and the average precision (\acrshort{label_AP}) score for each category, in addition to their mean values (averaged across categories).
Indeed, as we cared about ranking the labels, we chose AP because it measures precision at varying recall levels, and it is typically used in tasks when the labels have different degrees of importance, such as real-world CXR findings. In addition, we employed ROC-AUC to summarize the classifier's performance across all possible classification thresholds.
We test the best-performing \acrshort{label_ID} model on the external, never-before-seen MIMIC-CXR dataset to evaluate \acrshort{label_OOD} generalization abilities.

\subsection{Training Procedure and Parameters}
In all the experiments, we adopted ResNet50 backbones, \acrshort{label_Adam} optimizer, a \acrshort{label_LR} of 1e-6, a batch size of $12$, and trained the model in early-stopping on a multi-node multi-\acrshort{label_GPU} cluster with NVIDIA 64 GB cards\footnote{In general, ResNet18 was our go-to solution in most of our experiments (e.g., Chapters 4, 6) to achieve a reasonable tradeoff between performance, input/batch size, and GPU memory consumption when working on the “AI@Edge” cluster (equipped with 40GB VRAM GPU cards). Here, CROCODILE represents the only exception since we could work on the “Leonardo” supercomputer, equipped with 64GB cards, which allowed us to switch to ResNet50, a more powerful (yet heavier) backbone.}.
For the lambda hyperparameters, we tried out different values that would counterbalance the average values of the losses with unit weights. We thus conducted a random grid search and chose the following: set $\lambda_{1},\lambda_{3},\lambda_{4},\lambda_{6}$ to $1$; $\lambda_{2}$ to $10$; $\lambda_{9},\lambda_{11}$ to $15$; $\lambda_{8},\lambda_{10}$ to $25$; $\lambda_{5}$ to $80$; and
$\lambda_{7},\lambda_{12}$ to $100$.
We compare to a regular ResNet50 architecture, a ResNet50 version of Nie et al. \cite{nie2023chest} corresponding to discarding domain-branch and task-prior information from our method, our method without contrastive learning (\acrshort{label_CL}) ($\mathcal L_{RS}$, $\mathcal L_y^{batch}$, $\mathcal L_d^{batch}$), and our method without the task prior ($\mathcal L_y^{prior}$).

\section{Results and Discussion}
\subsection{Performance on ID and OOD data}
The results of our \acrshort{label_ID} and \acrshort{label_OOD} investigations (Table \ref{tab:results_crocodile}) reveal our method is behind its ablated versions and Nie et al.\cite{nie2023chest} on \acrshort{label_iid} data (\acrshort{label_ID}) while is the best-performing model on the external never-before-seen data (\acrshort{label_OOD}). Notably, our method is the most effective in reducing the ID-to-OOD drop in performance.
\begin{table}[h]
    \centering
    \caption{The \acrshort{label_AUC}/\acrshort{label_AP} scores obtained on each CXR finding on ID and OOD data. \acrshort{label_CL}: contrastive learning, TP: task prior. ID-OOD drop is the average percent drop in scores from ID to OOD settings.}
    \begin{tabular}{lccccc}
        \hline
        Finding & ResNet50 \cite{he2016deep} & Nie et al. \cite{nie2023chest} & \textbf{Ours} \textit{w/o} CL & \textbf{Ours} \textit{w/o} TP & \textbf{Ours}\\
        \hline
        \multicolumn{6}{c}{\textbf{In-distribution} (ID) data}\\
        \hline
        Atelectasis & 65.74/24.98 & 76.81/30.04 & \textbf{77.13}/30.26 & 77.07/30.37 & 77.04/\textbf{30.37}\\
        Cardiomegaly & 81.53/51.21 & 92.43/56.56 & \textbf{92.92}/\textbf{56.60} & 92.29/56.20 & 92.27/56.17\\
        Consolidation & 69.74/8.71 & 80.89/13.85 & 80.62/\textbf{14.10} & \textbf{81.13}/13.82 & 81.10/13.86\\
        Edema & 77.34/17.62 & 88.49/\textbf{23.01} & 88.21/22.53 & \textbf{88.73}/22.02 & 88.72/22.05\\
        Effusion & 77.69/51.26 & 88.68/56.31 & \textbf{89.08}/56.46 & 88.92/\textbf{56.65} & 88.93/56.65\\
        Lung opacity & 69.81/39.27 & 81.20/44.62 & \textbf{81.20}/\textbf{44.66} & 80.60/44.10 & 80.55/44.08\\
        No finding & 68.75/68.08 & \textbf{80.14}/73.46 & 79.68/\textbf{73.47} & 79.38/73.22 & 79.35/73.22\\
        Pneumonia & 67.76/20.74 & 78.05/\textbf{26.13} & \textbf{79.15}/25.73 & 77.65/24.86 & 77.63/24.85\\
        Pneumothorax & 78.86/32.78 & 89.87/\textbf{38.17} & \textbf{90.25}/37.69 & 88.79/37.02 & 89.86/37.03\\
        \textit{Mean} [$\uparrow$] & 73.02/34.96 & 84.06/\textbf{40.24} & \textbf{84.25}/40.17 & 83.95/39.81 & 83.94/39.81\\  
        \multicolumn{6}{c}{\textbf{Out-of-distribution} (OOD) data}\\
        \hline
        Atelectasis & 62.79/31.56 & 74.02/36.69 & 74.11/36.63 & 74.15/\textbf{36.89} & \textbf{74.18}/36.83\\
        Cardiomegaly & 61.43/31.84 & 71.44/36.22 & 71.86/36.42 & \textbf{72.82}/37.16 & 72.80/\textbf{37.17}\\
        Consolidation & 66.41/7.20 & 77.01/11.97 & 77.38/\textbf{12.53} & 77.46/12.13 & \textbf{77.80}/12.07\\
        Edema & 74.04/36.12 & 84.52/40.48 & 83.95/40.46 & \textbf{85.43}/41.43 & 85.39/41.45\\
        Effusion & 75.10/59.66 & 86.16/64.60 & 86.04/64.87 & 86.01/64.85 & \textbf{86.49}/\textbf{64.99}\\
        Lung opacity & 56.92/28.52 & 67.86/33.49 & 67.43/33.10 & 68.30/33.83 & \textbf{68.31}/\textbf{33.85}\\
        No finding & 67.39/63.72 & 78.53/68.66 & 78.72/68.99 & \textbf{78.78}/69.02 & 78.74/\textbf{69.05}\\
        Pneumonia & 53.64/7.47 & 63.96/12.29 & 64.62/12.52 & 65.01/12.76 & \textbf{65.03}/\textbf{12.80}\\
        Pneumothorax & 64.72/12.39 & 74.89/16.76 & 75.41/17.65 & 75.48/17.70 & \textbf{76.11}/\textbf{17.72}\\
        \textit{Mean} [$\uparrow$] & 64.71/30.94 & 75.38/35.68 & 75.50/35.91 & 75.94/36.20 & \textbf{76.09}/\textbf{36.21}\\
        \hline
        ID-OOD drop& 11.38/11.50 & 10.33/11.33 & 10.38/10.60 & 9.54/9.07 & \textbf{9.35}/\textbf{9.04}\\
        \hline
    \end{tabular}
    \label{tab:results_crocodile}
\end{table}

This significant result points to a necessary trade-off between in-domain accuracy and out-of-domain robustness on real-world data, supporting recent work \cite{teney2024id}.
As expected, models not contrasting the information from the two branches (Nie et al. (2023)\cite{nie2023chest} and ours without \acrshort{label_CL}) find associations that make them perform better on the ID data, where they remain faithful. Then, however, they fail to perform as well on \acrshort{label_OOD} data, where many spurious correlations due to the domain no longer exist, suggesting those associations are still based mainly on shortcut features. 
On the contrary, adopting our contrastive learning scheme first leads to lower performance on \acrshort{label_ID} data (as if the representation power on such data were ‘spoiled’ compared to the above). Still, it leads to better results on \acrshort{label_OOD} data. This suggests that our method learns image-to-prediction mechanics that are more transportable and generalizable, relying less on confounding factors and breaking down barriers between domains.
\subsection{The Importance of Task Prior}
Moreover, injecting prior task knowledge helped the model with specific findings. For instance, we know \textit{effusion} is likely an effect of \textit{pneumonia} or \textit{consolidation} and one of five aspects defining \textit{lung opacity}. We also know that patients with heart failure typically feature both \textit{cardiomegaly} and \textit{effusion}, but there is no causal effect of one aspect onto the other (Fig \ref{fig:prior_knowledge}). Thus, when the model was equipped with this knowledge during training, it learned to pay attention to such co-occurrences more and ultimately could detect more \textit{effusion} cases in \acrshort{label_OOD} data, possibly disregarding the confounding effect of heart failure.

\subsection{A Psychological Perspective}
In a recent psychology review paper, Goddu and Gopnik (2024) \cite{goddu2024development} take an interventionist perspective on causality, emphasizing that understanding the notion of intervention and its consequences is crucial to developing a fully depersonalized and decontextualized causal reasoning ability. Accordingly, "[...] human causal understanding is distinguished by its \textbf{depersonalized (objective)} and \textbf{decontextualized (general)} representations.".
We argue that this perspective is psycho-evolutionary support for our CROCODILE investigation: \textit{objective} representations correspond to separating spurious and genuine causal features, and \textit{general} representations means they are not domain/dataset dependant but are \acrshort{label_OOD}-generalizable.

\subsection{Limitations and Future Work}
Among the limitations of this work, we have utilized the same architecture type for feature extraction on the two branches, and we implicitly optimized the network on ID validation data. On the one hand, there could be better feature extraction backbones that could be used instead of the ResNet50 we used in this study. On the other hand, it could be the case that the desired domain-branch information and disease-branch information have different degrees of abstraction and, therefore, require different extraction models in order to avoid using overly simple or complex ones. Although we were satisfied with the results obtained in the current settings, future work could try different backbones and allow an ID test set.

\section{Summary}
This Chapter has presented the CROCODILE framework, a new approach to enhance a medical image classifier's generalization and \acrshort{label_OOD} robustness, addressing the problem of removing confounders.
Our solution learns what to focus on/suppress by borrowing from multiple sub-disciplines: latent causal intervention, graphical models, causality maps, feature disentanglement, the \textit{learning to compare} idea, and enforcing representation consistency.
Our bias-mitigation proposal is general and can be applied to tackle domain shift bias in other computer-aided diagnosis applications, fostering a safer and more generalizable medical \acrshort{label_AI}.
\chapter{CoCoReco: Connectivity-Inspired Network for Context-Aware Recognition}
\label{chap:cocoreco_ECCV}
\begin{figure}[h!]
\includegraphics[width=\textwidth]{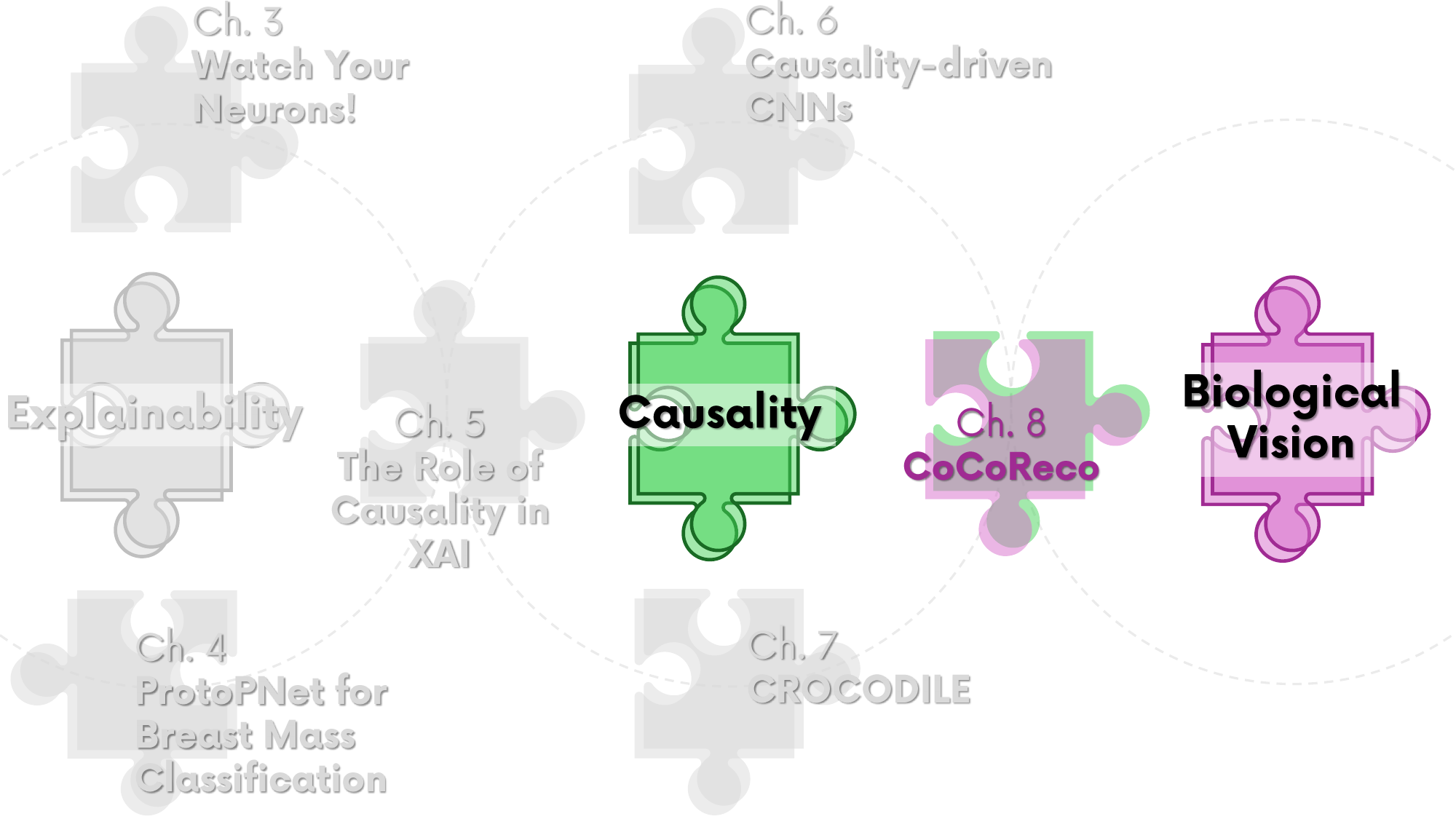}
\end{figure}

In previous chapters, we experimented with how causality can help \acrshort{label_DL} models to be better performing and explainable (Ch. \ref{chap:mulcat_ICCV_ESWA}), as well as more robust to domain variations (Ch. \ref{chap:crocodile}). These investigations allowed me to develop new ideas to bring artificial reasoning closer to human reasoning.
Still, they somehow took computer vision for granted, and we started asking how humans achieve visual recognition in the first place.
In this regard, we became interested in another fundamental aspect of achieving \acrshort{label_AI}-Humans alignment: the study of biological vision.

Accordingly, in Section \ref{sec:back_biological}, we reviewed fundamental notions and recent trends in the study of human vision - what the ventral and dorsal streams are and how they communicate, how top-down modulation occurs, the existence of subcortical pathways, and the importance of context in vision. This can foster Human-inspired Computer Vision (\acrshort{label_HCV}).

In this Chapter, we focus on the effect of incorporating circuit motifs found in biological brains to address visual recognition. Accordingly, the aim of this pilot and exploratory work is twofold, and we make the following key contributions:
\begin{itemize}
    \item \textbf{1)} Proposing a novel biologically motivated neural network for image classification. We design \textbf{CoCoReco}, a \textbf{co}nnectivity-inspired and \textbf{co}ntext-aware \textbf{reco}gnition network. \acrshort{label_CoCoReco} is a convolutional model conceptually inspired by the above-mentioned mechanism of human vision and numerically based on recent connectomic studies. Motivated by the connectivity of human (sub)cortical streams, we implement bottom-up and top-down modulations that mimic the extensive connections between visual and cognitive areas.

    \item \textbf{2)} Presenting a new plug-and-play module to model context awareness. Our Contextual Attention Block (\acrshort{label_CAB}) can be added to any traditional feed-forward architecture to improve recognition by modeling feature co-occurrence (i.e., context) in the real world. \acrshort{label_CAB} infers weights that multiply the feature maps according to their causal influence on the scene, modeling the co-occurrence of objects. We place our module at different bottlenecks to infuse a hierarchical context awareness. We validate \acrshort{label_CoCoReco} on image classification experiments on benchmark data and find consistent improvements in model performance and robustness of the produced explanations. 
\end{itemize}
  
The content of this Chapter is based on the following
\begin{itemize}
    \item publication: 
    Carloni, G., Colantonio, S. (2024, September). "Connectivity-Inspired Network for Context-Aware Recognition". In \textit{International Workshop on Human-inspired Computer Vision} (HCV), ECCV 2024. Cham: Springer Nature Switzerland AG. (Pre-print at \cite{carloni2024connectivity}),
    
    \item and the corresponding Python/Pytorch code can be found on my GitHub page at: \url{https://github.com/gianlucarloni/CoCoReco}.
\end{itemize} 

\section{Methods}
\label{sec:methods_cocoreco}
\subsection{Architecture design}
After reviewing the relevant literature, our next main contribution in this study is the design of the \textbf{Co}nnectivity-inspired \textbf{Co}ntext-aware \textbf{Reco}gnition network (\textbf{\acrshort{label_CoCoReco}}), depicted in Fig. \ref{fig:cocoreco}. It is a dual-branched architecture for image classification inspired by the human ventral and dorsal streams and the tectopulvinar pathway. Moreover, we conceive a top-down modulation of the bottom-up representations from the pre-frontal cortex (\acrshort{label_PFC}) and extensive afferent and efferent projections based on connectome studies.
As we shall see later, we placed our \acrshort{label_CAB} at different bottlenecks to infuse hierarchical context awareness into the model.
We design \acrshort{label_CoCoReco} as a from-scratch \acrshort{label_DNN} because evidence points to the opportunity to simplify DNNs to align with visual streams better, with smaller and less complex DNNs being more brain-like than many of the best-performing ImageNet models \cite{schrimpf2018brain}.

\subsubsection{Multi-branched Network}
\begin{figure}[tb]
  \centering
  \includegraphics[width=\textwidth]{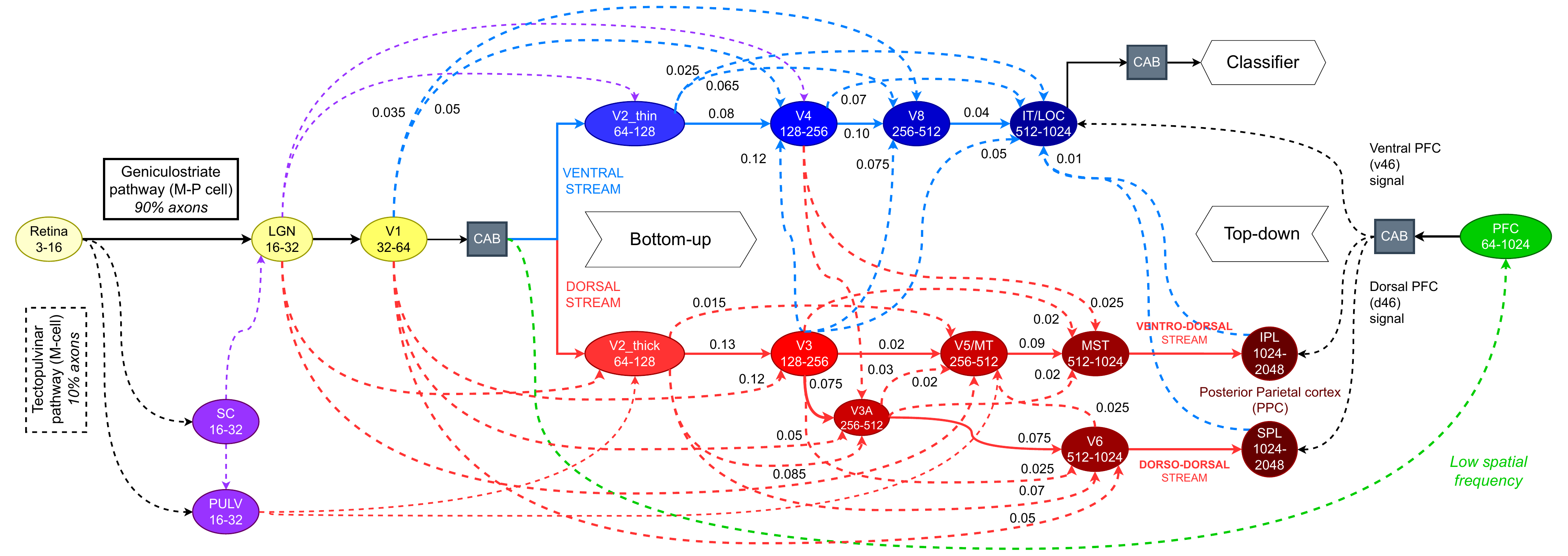}
  \caption{Overview of our \textbf{Co}nnectivity-inspired \textbf{Co}ntext aware \textbf{Reco}gnition network. The internals and rationale of the \acrshort{label_CAB} module are presented in Sec. \ref{sec:cocoreco_cab} and Fig. \ref{fig:cab_module}.
  Other abbreviations: lateral geniculostriate nucleus (LGN), superior colliculus (SC), pulvinar (PULV), classifier (CLS).
  Best seen in color.}
  \label{fig:cocoreco}
\end{figure}

Instead of modeling a single hierarchy of concentrical representations, we implemented a multi-branched convolutional architecture, which considers that shape information is processed ubiquitously in different human brain regions.
In this way, we have operationalized a proper \textit{vision-for-perception} schema based on both the ventral and dorsal streams.
We designed each brain area as a feature layer made of (i) a 2D convolutional layer with a kernel size of 3, stride of 1, and padding of 1; (ii) a \textit{ReLU} activation function; and (iii) a \textit{MaxPool2d} layer of kernel 2 and stride 2. Each convolutional layer's input and output channels are depicted at the corresponding feature node in Fig. \ref{fig:cocoreco} (e.g., the \textit{V1} layer has 32 input and 64 output channels).
We route 90 percent of the retinal signal to the \acrshort{label_LGN} layer and 10 percent to the \acrshort{label_SC} and pulvinar layers, emulating the division of axons. Moreover, those layers are dominated by \acrshort{label_M_cell}s. Thus, we model this faster and coarser information by increasing the convolutions' kernel size to 5 and padding to 2.

\subsubsection{Biologically Plausible Skip Connections}
To model the passing of information from one visual area to other areas, we design skip connections with a proper projection layer. 
Although skip connections are a common technique in computer vision (e.g., concatenation in DenseNets, summation in ResNets), our design offers technical novelties for their biological plausibility in deciding where to place them and how much to weigh their contribution.  
The number and size of feature maps can be higher or lower depending on whether the information is conveyed in a forward (bottom-up) or backward (top-down) pass. For instance, evidence suggests that the V1 signal is transmitted not only to the visual area V2 (which directly follows V1 layer in the model), but also to later areas like V4 or V8, which have more but smaller feature maps compared to V1, according to the hierarchy of constructed representations. To achieve information pass, we thus need to adjust the number and size of earlier representations. To this end, we design proper \textit{projection layers} made of a trainable convolutional layer adjusting the number of feature maps, followed by (i) a 2D average pooling, if it is a \textit{forward} pass, or (ii) a bilinear upsampling if it is a \textit{backward} pass. 
Recent studies on effective connectivity (EC) between brain regions reveal not only if two brain areas are anatomically/functionally connected but also expose the (relative) strength of such connections \cite{rolls2023multiple}. Therefore, we employ the EC measure found in that work as numerical estimates for the strength of the forward/feedback connections described above. As a result, we achieve a weighted projection by multiplying the output of the projection layers by the estimated weight.

\subsubsection{Information Flow From Input to Prediction}
The visual information flows within the network by following the connections in Fig. \ref{fig:cocoreco}. Soon after the activation of the \textit{retina}, the first firing is sent to the \textit{superior colliculus} and \textit{pulvinar} (M-cell pathway taking 10\% of retinal axons) and to the \textit{LGN} (P-cell pathway taking 90\% of retinal axons). The \textit{pulvinar} receives from the retina and the superior colliculus itself.

At this point, the signal is sent to the primary visual cortex \textit{V1} via the joint M-and-P cell pathway. Then, a fast M pathway sends low spatial frequency information directly to the high-level areas in the \textit{PFC}. Contextually, the \textit{V1} information is sent to the ventral and dorsal streams. Regarding the former, a combination of \textit{V1} signal and \textit{LGN} afferent signal activates the thin-striped \textit{V2}; regarding the dorsal counterpart, a combination of \textit{V1}, \textit{LGN}, and \textit{pulvinar} signals activates the thick-striped \textit{V2}.

On the dorsal stream, the \textit{V3} is activated by a combination of the incoming \textit{thick V2} and afferent \textit{V1}. This \textit{V3} signal participates in the activation of the \textit{V4} unit on the ventral stream together with the \textit{thin V2}, the \textit{LGN}, and the \textit{V1} afferent signals.

Now, the \textit{V8} units are activated by a combination of incoming \textit{V4} features and afferent signals from \textit{V1}, \textit{thin V2}, and \textit{V3}. Also, the \textit{V3a} layer receives a combination of \textit{V3}, \textit{V1}, \textit{thick V2}, and \textit{V4}, and projects to the \textit{V6} units. Thanks to similar primary and secondary connections, also the \textit{V5mt} and \textit{mst} layers are activated.
On the parietal cortex, the inferior and superior parietal lobes receive not only from the bottom-up dorsal stream but also from a top-down modulation by the prefrontal cortex.

Finally, all the signals converge to the inferior temporal cortex (\textit{IT}). Here, we find the bottom-up processes from the ventral and dorsal streams and the top-down modulations from the prefrontal and parietal cortices.
The actual decision is made at the network's fully connected layer, which takes the flattened \textit{IT} features as input and returns the class logits.

What we have described above is the general information flow considering primary/secondary streams and feed-forward/feedback connections. However, a final component emerges from Fig. \ref{fig:cocoreco} and must be considered. Indeed, we present our last major contribution in the next subsection.

\subsection{Contextual Attention Blocks}
\label{sec:cocoreco_cab}
We present a new plug-and-play module to inject context awareness into the model. In fact, our \acrshort{label_CoCoReco} solution also models another fundamental aspect of human vision: context.
\subsubsection{Causality Maps: Modeling Feature Co-occurrence}
\begin{figure}[tb]
  \centering
  \includegraphics[width=\textwidth]{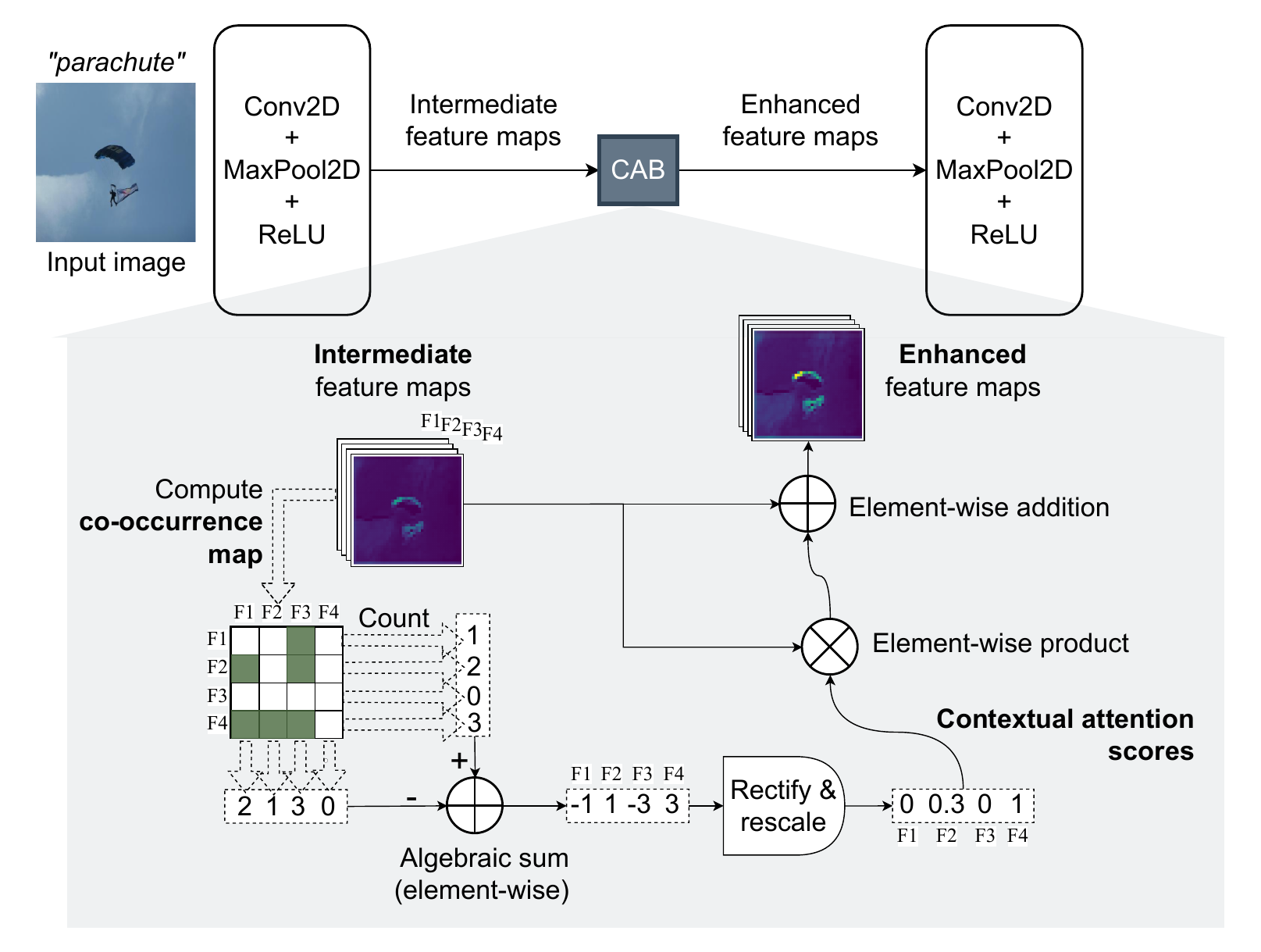}
  \caption{Our Contextual Attention Block (\acrshort{label_CAB}) integrated into a general feed-forward network. As shown, CAB is placed at the convolutional bottleneck of the model. Given intermediate feature maps, the module computes corresponding contextual attention scores through a rectified and rescaled version of the weights obtained from the co-occurrence map.}
  \label{fig:cab_module}
\end{figure}

To conceive our \acrshort{label_CAB}, depicted in Fig. \ref{fig:cab_module}, we get inspiration from our prior works \cite{carloni2023causality,carloni2024exploiting} as a way to model the co-occurrence of different objects in the image scene.
Under that setting, the $F1$, $F2$,...$Fk$ feature maps obtained from the last convolutional layer are used to compute pairwise conditional probabilities, resulting in a $k \times k$ co-occurrence map, called \textit{causality map}. Asymmetries in such probability estimates provide some information on the cause-effect of the appearance of a feature in one place of the image, given the presence of another feature within some other places of the image.

However, our implementation differs from those in the computation of attention scores (we rescale their value to avoid value instability) and in how the enhanced feature maps are embedded to the original ones (we add them element-wise instead of concatenating them to avoid increasing parameter overhead).

\subsubsection{Multiple CABs for Hierarchical Contextual Attention}
To construct a hierarchical attention mechanism, we propose \textbf{placing multiple CAB modules} at different network bottlenecks. Indeed, we use CAB on the output of: (i) the V1 layer, representing a coarse and global context, of (ii) the \acrshort{label_PFC}, representing a semantically rich, goal-oriented, context for top-down information flow, and finally of (iii) the \acrshort{label_ITC}/LOC layer, which is where the final representation for the object is constructed for recognition.

As a result, intermediate feature maps get enhanced by our CAB modules, and we obtain three new feature sets, as detailed in the following pseudocodes:
\begin{itemize}
\item $\mathbf{v1} \leftarrow \mathbf{v1} + extractCausalityFactors(computeCausalityMap(\mathbf{v1}))$   
\item $\mathbf{pfc} \leftarrow \mathbf{pfc} + extractCausalityFactors(computeCausalityMap(\mathbf{pfc}))$
\item $\mathbf{it} \leftarrow \mathbf{it} + extractCausalityFactors(computeCausalityMap(\mathbf{it}))$
\end{itemize}

\subsection{Dataset}
We conceive image classification experiments on \textbf{ImagenetteV2} data, a popular and freely available\footnote{https://github.com/fastai/imagenette} dataset composed of a subset of 10 easily classified classes from ImageNet (tench fish, English springer dog, cassette player, chain saw, church, French horn, garbage truck, gas pump, golf ball, and parachute).
The dataset comes with three different versions thereof: original size, '320 px', and '160 px'; the latter two have their shortest side resized to that size, with their aspect ratio maintained. We utilize the version at 320x320 image resolution and normalize the input channels to standard ImageNet practice (mean=[0.485, 0.456, 0.406], std=[0.229, 0.224, 0.225]). We use the official train-val splits and further split the validation set into actual validation and external test sets in proportions 60-40. As a result, we obtain 9469 training examples, 2355 validation examples, and 1570 test examples. Each category has a different training image count, with a minimum of 858, an average of 947, and a maximum of 993. We use the training set to learn the models, the validation set to conduct internal assessment and hyperparameter tuning, and finally, we test the trained model on the external test data. 

\subsection{Total Training Objective}
As for the total training objective of our model, it is a composition of a cross-entropy loss $\mathcal{L}_{\text{CE}}$ for classification correctness and a series of \acrshort{label_MSE} loss terms for class-based causality map alignment.
Indeed, we propose a novel loss, the \textbf{mini-batch loss}, to push the causality map of samples belonging to the same category closer so that we foster class-based map alignment. We implement the mini-batch loss $\mathcal{L}_{\text{mini-batch}}$ as an \acrshort{label_MSE} loss between the causality map of each sample and the average causality map of samples of the same class found in the minibatch during training.
On top of that, we compute the mini-batch loss at each of the three CAB levels described in Sec. \ref{sec:cocoreco_cab}.

Let $\mathcal{C}$ be the set of classes, and let $x_i$ represent the $i$-th sample in the mini-batch. The causality map corresponding to sample $x_i$ for a given feature set (e.g., "V1", "PFC", or "IT") is denoted by $\mathcal{M}^{\text{feature}}(x_i)$. We define the total loss as a weighted sum of the cross-entropy loss and three mini-batch loss terms:

\begin{equation}
\label{eq:cocoreco_loss}
  \mathcal{L}_{\text{total}} = \mathcal{L}_{\text{CE}} + w_{V1} \mathcal{L}_{\text{mini-batch}}^{V1} + w_{PFC} \mathcal{L}_{\text{mini-batch}}^{PFC} + w_{IT} \mathcal{L}_{\text{mini-batch}}^{IT}  
\end{equation}

where the mini-batch loss for each feature set is defined as:

\begin{equation}
\mathcal{L}_{\text{mini-batch}}^{\text{feature}} = \frac{1}{B} \sum_{i=1}^{B} \left\| \mathcal{M}^{\text{feature}}(x_i) - \frac{1}{|C_i|} \sum_{j \in C_i} \mathcal{M}^{\text{feature}}(x_j) \right\|_2^2
\end{equation}

In this equation:
\begin{itemize}
    \item $C_i \subseteq \{1, 2, \dots, B\}$ is the set of indices corresponding to the samples in the mini-batch that belong to the same class as $x_i$,
    \item $\mathcal{M}^{\text{feature}}(x_j)$ is the causality map of the $j$-th sample, computed using a specific feature set (e.g., "V1", "PFC", or "IT"),
    \item $\|\cdot\|_2$ denotes the $L_2$ norm (Euclidean distance),
    \item $|C_i|$ is the cardinality of the set $C_i$,
    \item $w_{V1}, w_{PFC}, w_{IT}$ are the weights associated with the mini-batch losses computed using the "V1", "PFC", and "IT" features, respectively.
\end{itemize}

We learn the model parameters by optimizing the criterion in Equation \ref{eq:cocoreco_loss} via the Adam optimizer and use a learning rate of 0.0006, a weight decay of 0.0001, and a batch size of 64. Upon exploration of possible values, we found $w_{V1}=0.7$, $w_{PFC}=0.5$, and $w_{IT}=0.1$ to work well.

\subsection{Experiments and Evaluation}
As in most of the previous Chapters, we opted for total accuracy as the go-to evaluation metric to assess the overall classifier goodness. However, we wanted to study its reliability in recognizing positive cases while minimizing false positives and false negatives when considering a slight class imbalance. Thus, we also used the macro-averaged F1-score\footnote{Averaging the unweighted mean per label; in other words, computing the F1 score for each class individually and then averaging them; see \textit{sklearn.metrics} documentation here: \url{https://scikit-learn.org/stable/modules/generated/sklearn.metrics.classification_report.html})}, also in anticipation of when CoCoReco would be trained on medical images. 

We compare our \acrshort{label_CoCoReco} model with two ablation versions and one baseline model. Indeed, to assess the importance of contextual awareness, we remove the \acrshort{label_CAB} modules from the network. Moreover, to assess the importance of bottom-up and top-down modulations (projections), we remove all skip connections between the different visual areas. Finally, we study the effect of removing the two-branched design, and we train a separate from-scratch \acrshort{label_CNN} architecture of the same depth as our CoCoReco (seven convolutional layers, from retina to IT) but with only one branch, representing the traditional bottom-up hierarchy of representations along the ventral stream.
To prevent bias and gain statistics, we repeated all experiments ten times with different random seeds and reported the mean and standard deviation of the computed metrics.

\section{Results and Discussion}
\subsection{Numerical Performance}
Table \ref{tab:results} summarizes our numerical findings. The former shows how our \acrshort{label_CoCoReco} architecture consistently achieves the highest accuracy and F1-score among the investigated models for ten different random seeds. Indeed, we took the mean and standard deviation values across multiple runs and compared CoCoReco to its ablated versions (i.e., without CAB modules and without projections), as well as to the baseline single-branched network. 
\begin{table}[tb]
    \centering
    \caption{Accuracy and F1-scores obtained by our CoCoReco and ablated/baseline models on the external, never-before-seen test set of ImagenetteV2 data at resolution 320x320. From left to right, the columns represent our CoCoReco architecture, its ablated version with the \acrshort{label_CAB} removed, its ablated version with no bottom-up nor top-down projections between different areas, and the baseline from-scratch single-branched \acrshort{label_CNN}. Values are given by mean (standard deviation) from models obtained by training with ten different random seeds. The higher (lower) the mean (standard deviation), the better.}
    \begin{tabular}{ccccc}
       \hline
       & \textbf{CoCoReco (ours)} & Without CAB & Without projections & Baseline CNN \\ 
       \hline
       Accuracy & \textbf{74.6 (0.63)} & 73.8 (0.89) & 73.2 (0.81) & 71.1 (1.0)\\
       F1-score & \textbf{74.4 (0.71)} & 73.3 (0.91) & 73.1 (0.87) & 71.2 (1.1)\\
       \hline
    \end{tabular}
    \label{tab:results}
\end{table}

\subsection{Qualitative Performance}
To further assess the benefits of using our proposal, we conducted post-hoc class activation mapping (CAM). We found the explanations produced by our \acrshort{label_CoCoReco} to be better than those of the competing methods. Generally, they are more robust and focused on the salient object of classification without being affected by confounder aspects of the image that are spuriously associated with the outcome.
\begin{figure}[h]
  \centering
  \includegraphics[width=0.99\textwidth]{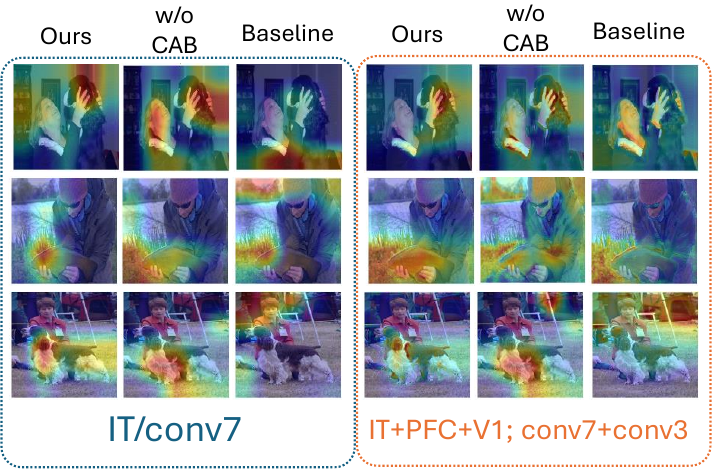}
  \caption{GradCAM activations for three test images for CoCoReco, ablation, and baseline models. The left panel (blue) shows the outputs when the last convolutional layer before the classifier is chosen as the target layer for the GradCAM computation. The panel on the right (orange) shows the outputs for the same images when a combination of target layers is chosen.}
  \label{fig:gradCAM}
\end{figure}

\subsubsection{GradCAM on Last Convolutional Layer}
Figure \ref{fig:gradCAM} (left panel) shows the models' outputs on three example test images when the last convolutional layer before the classifier is chosen as the target layer for the GradCAM computation\footnote{
We need to specify the target layer to compute the CAM for. Some common choices in popular architectures are: model.backbone for FasterRCNN, model.layer4[-1] for ResNet18 and 50, model.features[-1] for VGG and DenseNet.
}. 
In the first row, the test image for class "English Springer Dog" depicts an indoor scene where a blonde woman holds the dog. The baseline CNN (third column) seems to arrive at the correct prediction by looking at seemingly unimportant features.
The ablated CoCoReco (without the CAB modules) in second column correctly identifies some dog features but also includes the woman in its reasoning, that is, it is confounded by attributes of people that frequently coexist in photos of domestic animals.
Finally, our \acrshort{label_CoCoReco} (first column) pays less attention to people attribute and mainly focuses on dog features.

On the same line, when presented with photographs of the fishing domain for the "Tench Fish" class (second row), the baseline and ablated model are still confounded, for instance, by the commonly present tree and grass features. Conversely, our model tends to focus on the texture and shape features of the tench fish while paying little attention to other confounding features to arrive at its conclusion.

\subsubsection{GradCAM on Combinations of Convolutional Layers}
The GradCAM activation in the above analysis is a great tool, but it is limited to a single, final target layer before the classifier. On the other hand, we know humans can merge information from representations at different semantic levels before attaining a classification. In that regard, and in line with the defined CoCoReco architecture, we could utilize not a single target layer but a list with several layers, and then average the produced CAM, effectively obtaining a global-local modulation of the activations. Indeed, we run the pipeline with a combination of inner layers:
\begin{itemize}
    \item For CoCoReco and its ablation without CAB, we selected the combination of \textit{V1}, \textit{PFC}, and \textit{IT} layers as our target.
    \item For the baseline model, we selected the third and seventh layers which corresponded to the same backbone depth of V1 and IT.
\end{itemize}

The result of such combined-layers CAM is shown in the right panel of Fig. \ref{fig:gradCAM}. In these new visualizations, the silhouette of important regions is generally more defined and has a much finer pixel-wise granularity. Returning to the example of the woman holding up the dog, the highlighted features can now represent specific fingers, wrist, neck, ear features, or fur and paws one.

Overall, the improved focus abilities registered by CoCoReco with only the last layer CAM holds true even with the present combined-layers CAM. That suggests \acrshort{label_CoCoReco} has learned important representations in the earlier layers corresponding to V1 areas and semantic-rich \acrshort{label_PFC}. Conversely, the poor quality of explanations produced from \textit{conv7} of the baseline model is confirmed when they are produced jointly from \textit{conv7} and \textit{conv3}.

\subsection{Limitations}
Although our method has already shown performance and interpretability gains, we could explore further in future works. We envisage performing a more extensive optimization of both architecture design (e.g., number and position of \acrshort{label_CAB} modules) and model learning (e.g., hyperparameter grid search or Bayesian optimization\cite{lee2023bayesian}) and validating it on larger datasets.

\section{Summary}
This research bridges neuroscience and computer vision with novel elements, advancing human-inspired computer vision and enriching the \acrshort{label_AI}-neuroscience dialogue.
First, we presented key concepts of the human visual system by performing a detailed neuroscientific literature review, which is essential to inform the AI practitioner. 
Then, we introduced \acrshort{label_CoCoReco}, a novel biologically motivated neural network architecture based on human visual mechanisms for image classification. It incorporates multiple visual pathways and top-down modulation, and recent connectivity studies inspired it.
Lastly, we presented the \acrshort{label_CAB}, a plug-and-play module to improve context awareness. CAB is versatile and can be placed at desired bottlenecks in any traditional feed-forward architecture.
Incorporating circuit motifs found in biological brains to address visual recognition proved promising, and we contributed to enriching \acrshort{label_HCV}.
\chapter{Conclusions}
\label{chap:conclusions}
Giving machines the ability to see and understand visual information can help us manage and explore the rapidly increasing availability of digital images.
That is paramount in medical imaging applications, where a huge corpus of radiological, dermoscopic, or histological images is generated daily to aid patient care.
In this context, new \acrshort{label_AI} techniques based on \acrshort{label_DL} are making high-quality computer vision possible because they can learn effective, high-level representations of visual data from large datasets. Specifically, \acrshort{label_CNN}s have shown remarkable results in recent visual perception and representation research, where they advanced the state of the art in many medical image analysis tasks. Nevertheless, \acrshort{label_DL}-based solutions pose non-trivial engineering challenges in their adoption, such as their lack of interpretability and inability to grasp cause-effect relationships, ultimately leading to reduced trust and usability from human users. 

\section{Closing Remarks}
In this thesis, we investigated ways of rendering such solutions more aligned with human reasoning, capabilities, and demands and proposed novel approaches for efficient, explainable, and robust medical image classification. We approached the investigation from three essential perspectives: explainability, causality, and biological inspiration. In particular:
\begin{itemize}
    \item we studied and proposed solutions to the need for interpretations when humans are presented with opaque \acrshort{label_DL} decision models, highlighting failure cases and expectation alignments in the medical imaging domain;
    \item we offered engineered solutions integrating causal reasoning into \acrshort{label_DL} to boost models' performance and explainability, as well as generalization in out-of-distribution scenarios; and
    \item we developed innovative approaches for computational models inspired by the biological vision to attain human-aligned object recognition in visual data.
\end{itemize}

Chapters \ref{chap:introduction} and \ref{chap:background} provided an introduction and background knowledge about DL, medical image analysis \& classification, \acrshort{label_XAI}, causal \acrshort{label_ML}/\acrshort{label_DL}, and visual information processing in humans.

In Chapter \ref{chap:XAI_wyn}, we tackled the problem of visualizing the representations learned by pre-trained \acrshort{label_DNN}s for different categories. We investigated the feature-visualization approach named Activation Maximization and stress-tested it first on natural images. We extensively studied the effect of varying optimization settings (e.g., input size, iterations' number, seed initialization) and explored using \acrshort{label_AM} to discover dataset biases. Our findings revealed a new, insightful scenario where colorful and detail-rich objects are represented \textit{compositionally} on a gray, average background when the input size is exaggerated over the training one. That generates an image where salient patterns from different angles and scales co-exist, depicting the panel of possible features that activate the said neuron. Also, we gave evidence of the entanglement of different visual concepts in the learned neuron representations and of potential biases and shortcuts at the training dataset level (e.g., patterns of human hands or fingers appear in the neuron activation of human-handled or grasped objects, like nails or hammers). We then empirically showed how difficult it is to employ such methods effectively in medical imaging scenarios and proposed novel mitigation strategies pivoting on medical image regularities, such as symmetry, pixel intensity, noise characteristics, and frequency spectrum. Our transfer-learning and fine-tuning experiments on \acrshort{label_CXR} images highlighted the urgent need for more sophisticated and effective explanation methods to meet human expectations, laying the groundwork for the subsequent chapter.

Chapter \ref{chap:XAI_protopnet} addressed the issue of \textit{post}-hoc \acrshort{label_XAI} method not reproducing the exact calculations of the original model when approximating the outcome. Indeed, we investigated \textit{explaining-by-design}, that is, leveraging powerful \textit{ante}-hoc, inherently interpretable models where training, inference, and explanation of the outcome are intrinsically linked. The goal was to study the applicability of prototypical part learning for clinical diagnosis vision models. We proposed using ProtoPNet as an automatic breast mass classification system from CBIS-DDSM mammogram images to resemble the radiologist's behavior in recognizing patches and comparing them with experience when looking at a new case. Some critical issues can arise when adopting technologies developed for computer vision in natural images onto medical X-ray scans, including the reduced amount of information (single-channel grayscale vs RGB), flat and depth-less projection, anatomical/acquisition variations, and data scarcity. We performed architectural modifications and rigorously optimized our model's hyperparameters via 5-fold \acrshort{label_CV}; then, we evaluated it on benign-vs-malignant classification on a hold-out test set. Experiments showed that the best-performing ProtoPNet model achieved $0.72$ \acrshort{label_AUROC} and performed better than its ResNet18-backbone counterpart over most of the metrics, with the most substantial gains in Recall ($+25\%$), which is of considerable interest for this task (i.e., correctly identifying as many malignant masses as possible). Notably, we asked an experienced radiologist for clinical feedback on (i) the quality and relevance of learned prototypes to characterize classes, (ii) the clinical significance of prototype activations, and (iii) the degree of satisfaction in the way the model combines (i) and (ii) to deliver specific explanations. That revealed never-before-stated observations: first, ProtoPNet managed to learn more relevant prototypes for malignant masses similar to radiologists; second, the model’s mathematical concept of similarity may differ from how a radiologist would deem two regions clinically similar; and third, explanations for images classified as malignant are more likely to be more convincing to the radiologist.

In Chapter \ref{chap:review_XAI_causality}, we address the previously unclear relationship between explanation and causation in \acrshort{label_AI}, specifically, between \acrshort{label_XAI} and causality, by providing a comprehensive review that bridges this critical gap in the literature. Our investigation uncovered how the two concepts share ancient roots but have evolved separately in computer science and how they, on the other hand, are both human-centric and aim for actual usefulness to users.
We explored the multi-disciplinary literature of the field extensively (collected 222 papers) and systematically (performed a PRAMA-structured review). We performed a high-level analysis of the keywords’ co-occurrence via bibliometric networks that revealed how multidisciplinary the literature is, and we summarized popular software tools used to automate causal tasks (e.g., performing \acrshort{label_CD} with \acrshort{label_BN}s, creating \acrshort{label_SCM}s, analyzing \acrshort{label_DAG}s). More importantly, we proposed three main perspectives as a scaffolding organization for past, current, and future research.
The \textit{"Critics to \acrshort{label_XAI} under the causality lens"} perspective serves as a "Watch out!". Causality is recognized as a missing component of current \acrshort{label_XAI} research to achieve robust and explainable systems. Innate issues of (X)AI, such as the inability to distinguish correlation from causation, are also highlighted. Moreover, such papers discuss different forms (e.g., visual, textual, contrastive) and desiderata (e.g., alignment with the explainee's knowledge, ability to explain many effects with few causes) of the XAI-produced explanations and their link with the causal theory. 
The \textit{"XAI for causality"} perspective relates XAI and causality pragmatically by seeing XAI as a basis for further causal inquiry. In this light, despite their limitations, XAI explanations have the potential to foster scientific exploration, thus being starting points to generate hypotheses about possible causal relationships to be then confirmed in pursue-worthy experiments.
The \textit{"Causality for XAI"} perspective, instead, reverses the previous one, claiming that causality is propaedeutic to XAI (in three ways). First, causal metrics or concepts, such as the \acrshort{label_SCM} and the do-operator, are employed to boost existing XAI solutions with causal inference abilities. Second, causal \acrshort{label_CFE}s are generated by invoking the formal causal definition of CF from Pearl to improve generalization capabilities. Lastly, models are built on causal structures to be inherently interpretable, and exposing the inner workings of the causal model of a system through \acrshort{label_DAG}s makes it explainable.

Chapter \ref{chap:mulcat_ICCV_ESWA} explores the challenge of applying causal reasoning to \acrshort{label_DL}-based classification on real, structured image data with no \textit{a priori} information on their generation process. Building on the causal disposition concept, we investigated the feasibility of discovering and exploiting hidden causal signals among objects represented in visual scenes to enhance regular \acrshort{label_CNN} classifiers. We proposed \textit{Mulcat}, a novel attention-inspired feature enhancement scheme that weights each feature map according to its causal contribution to the scene. This is done by computing the complete feature co-occurrence set and then the \textit{causality factors}, i.e., scores, for each feature map, effectively quantifying how the presence of a feature in one part of the image affects the appearance of other features in different parts. We evaluated the efficacy of the proposed methods on publicly available datasets of prostate cancer \acrshort{label_MRI} images and breast cancer histology slides. Extensive experiments on multiple architecture variants, attention schemes (e.g., plain, \acrshort{label_BAM}-based), and learning schemes (e.g., fully supervised, few-shot) showed that our lightweight module helped the models perform better than the baselines. That included not only increased accuracy values but also explainability gains. The models could focus more precisely on the critical regions of the image (e.g., regions densely populated by nuclei in histology slides and prostate gland area in \acrshort{label_MRI}s), leading to more accurate and robust predictions, which is essential in medical imaging applications for effective diagnosis and treatment planning.

In Chapter \ref{chap:crocodile}, we examined a significant limitation of \acrshort{label_CNN}s and \acrshort{label_ML}-based models in general, precisely their vulnerability to distribution shift or, broadly, to \acrshort{label_OOD} data.
We investigated robust \acrshort{label_DL}-based medical image classification under distribution shift, brought up by images from different sources.
Our proposed CROCODILE is a new training framework for bias-mitigated prediction with improved generalization on \acrshort{label_OOD} scenarios. Our approach learns what to focus on or suppress by building on causal concepts and several other \acrshort{label_DL} disciplines. We design a dual-branched architecture to leverage information from multiple domains (e.g., datasets), learn how to separate causal and spurious features of the input, perform a causal intervention in the latent embedding space, and thus remove the effect of confounders in medical images. Also, we propose the \textit{Relational Scorer} module to learn structural relationships among the different representations obtained from input instances. Causal features that determine the disease would thus align with spurious features unrelated to domain prediction, while features important to discriminate domains would be discouraged from discriminating diseases. On top of that, we introduced \textit{contrastive learning} terms that enforce separation or consistency among representations from the same/different disease/domain. Lastly, we proposed a novel approach to \textit{inject background knowledge} about the task into model training, capturing semantic priors without relying on data. Experiments on \acrshort{label_ID} and \acrshort{label_OOD} showed CROCODILE first performs lower than baselines on \acrshort{label_iid} data (\acrshort{label_ID}), but then is the best-performing solution on external data (\acrshort{label_OOD}) and maximally reduced the percentage performance drop when passing from \acrshort{label_ID} to \acrshort{label_OOD}.

Lastly, Chapter \ref{chap:cocoreco_ECCV} presented our final exploration of ways to align \acrshort{label_DL} to human demands, capabilities, and reasoning, that is, biological inspiration. We bridged neuroscience and computer vision in this preliminary study and made two main contributions. 
We proposed \acrshort{label_CoCoReco}, a new biologically motivated neural network architecture based on human visual mechanisms for image classification. It includes multiple visual pathways that separate feature representations and encode ventral/dorsal information flows, and top-down modulations that we numerically based on recent connectivity studies.
Moreover, we proposed and evaluated the \acrshort{label_CAB} block, a plug-and-play module to automatically model attention scores based on causal dispositions and contextual information inspired by human context awareness. CAB is versatile and can be placed at desired bottlenecks in any traditional feed-forward network. We showed through experimental evaluation that our \acrshort{label_CoCoReco} not only achieved the highest metrics (e.g., max $+4.93\%$ accuracy over baseline) but also produced more robust and object-focused visual explanations. We argued that incorporating circuit motifs found in biological brains to address visual recognition can be a promising direction, and we contributed to enriching human-inspired computer vision.

\section{Future Directions}
\label{sec:conclusions_futureDirections}
Our investigation led to practical mitigations of existing problems in adopting deep models for image recognition, mostly for medical image classification.
Due to the thriving research throughput in \acrshort{label_DL} and computer vision, multiple aspects discussed in this dissertation are worth further investigating in light of recent developments in the field. The following paragraphs report on the possible expansions of our work in that regard.

\paragraph{In-the-Loop and Diffusion-based Activation Maximization} In Chapter \ref{chap:XAI_wyn}, we have shown how feature visualization, particularly \acrshort{label_AM}, is a useful diagnostic tool to visualize learned features and discover biases in post-deployment scenarios.
While applying this approach to models trained to diagnose diseases based on medical images has been challenging, we foresee interesting future work.
We envision leveraging AM \textit{during} the training process of \acrshort{label_DNN}s to act as a regularizer against dataset bias, distribution shift, and shortcut learning.
Every so often during training, say every 5 epochs, an AM procedure is triggered - the feedforward information flow in the network being trained is inverted, and new maximally activating images are computed for specific classes of interest.
By designing appropriate task-prior loss terms, that process could represent an effective \textit{in-the-loop} tool to help the model steer towards learning more robust features and fight dataset bias.
Combining developments in diffusion models (\acrshort{label_DM}s)\cite{ho2020denoising,rombach2022high,mamaghan2024diffusion} with \acrshort{label_AM} represents another attractive research direction worth exploring. The key idea of generative \acrshort{label_AI} based on \acrshort{label_DM}s is to create synthetic images by an iterative denoising process starting from pure Gaussian noise.
In a way, we can think of AM as doing the same - starting from a random noise sample, an iterative process builds a synthetic image. We posit that DMs can be utilized in support of \acrshort{label_AM}, for instance, enforcing structural priors on the noise image during the iterations, thus conceiving a specific loss to act as a prior. Also, DMs can be trained to produce AM visualizations for downstream tasks.

\paragraph{Evolving ProtoPNet} In Chapter \ref{chap:XAI_protopnet}, we presented the first investigation of prototypical-part learning on breast mass classification. Although this post-hoc \acrshort{label_XAI} method is general, there are some future research directions that could improve it. We could exploit recent advances in state-of-the-art object detectors (such as YOLOv9) to \textit{identify} lesions before entering ProtoPNet. We could design a teacher-student framework where ProtoPNet is the student model that learns from optimal pseudo labels produced by a global image classification (i.e., teacher) model. 
Other advancements would be combining several ProtoPNet models with different base architectures together in an ensemble fashion or choosing a Vision Transformer architecture~\cite{dosovitskiy2020image} instead of a CNN model at the core of ProtoPNet. In addition, from a broader point of view, we could build a causality-driven ProtoPNet embedding causal representation learning and prior structural knowledge to produce explanations grounded in causality instead of correlation.

\paragraph{Multi-depth Causality Maps and Feature Pruning} In Chapter \ref{chap:mulcat_ICCV_ESWA}, we took a causal viewpoint on visual scene representation and proposed a new method to exploit feature maps' co-occurrence for image recognition.
An extension of our feature enhancement scheme could rely upon multi-depth visual attention \cite{jetley2018learn,yan2019melanoma,schlemper2019attention}. Indeed, that learning method extracts feature maps from the convolutional encoder at different depths and aggregates the obtained local and global information in the classification stage. By taking inspiration from that, one could compute \textit{causality maps} not only from the last convolutional layer of the network, but also at multiple higher resolutions by extracting feature maps from the internal layers. We believe this local-global synergy would increase the capability of our framework to capture long-range causal dependencies from visual inputs.
Additionally, visualizations such as that in Figure \ref{fig_cmapvisual} suggest one could potentially conceive a low-cost self-supervised feature pruning based on similarities of features across rows and columns of causality maps. That could help to disregard redundant features and consequently lower model complexity in a data-driven way.
Ultimately, we foresee a possible integration of our methods within the convolutional block of generative models, such as GANs and DMs, to guide the generation of more realistic images.

\paragraph{Patient-based and Scanner-based Priors for Robust Learning}
In Chapter \ref{chap:crocodile}, we presented CROCODILE, a novel two-branched framework that learns robust invariant features for medical diagnosis under confounding features due to domain-shift bias. The system's first part involves feature extraction via convolutional backbones, for which we utilized ResNet50. A future direction is to investigate the effects of trying different types of backbones and purposely use different choices for the \textit{disease} branch and the \textit{domain} branch to adapt the architecture to the abstraction level of the processed information. 
Another major contribution of our chapter was proposing a new way to inject prior medical knowledge. In this regard, we hypothesized a causal graphical model among the \acrshort{label_CXR} findings and constructed the corresponding ground-truth causality map. Of course, that approximates reality, and the findings we observed are valid as long as the supposed graph is valid. However, as the British statistician George Box said in the late 1970s, \textit{"All models are wrong, but some are useful."}. Thus, we could involve experienced radiologists to design a more comprehensive causal graphical model of CXR findings. To attain the same goal, we could even exploit recent advancements in the reasoning capabilities of \acrshort{label_LLM}s. 
Moreover, similarly to drawing the causal graph of abnormal findings, we could study the data generation process of radiological images and model the graph under other perspectives, such as patient demographics and personal attributes, past \acrshort{label_EHR} records and clinical history, or radiography equipment parameters and acquisition settings. Overall, we expect integrating patient-based and scanner-based priors into the learning process would regularize the model towards more robust and transportable features.

\paragraph{Content-Style Disentanglement and Memory Banking} Chapter \ref{chap:cocoreco_ECCV} presented our novel biologically motivated \acrshort{label_DNN} for image classification and a powerful plug-and-play module based on objects' causal co-occurrence to consider context. We achieved that by building on cognitive and neuroscientific literature regarding the visual information flow in humans, the specific connectivity of brain regions, and bottom-up/top-down modulations. A future research direction may include the definition of separate information paths encoding \textit{invariant content} from \textit{specific style} when looking at a scene. Humans can indeed associate both an aircraft flying in the sky and a wooden toy model on a shelf with the "airplane" class, disregarding differences in shape, texture, and color. For this reason, modeling feature disentanglement, specifically content-style disentanglement, could benefit robustness and generalization.
Moreover, extending inspirations to visual object search and top-down modulation could lead to new bio-inspired computer vision mechanisms. We could conceive a neural model where the early visual representations carrying low spatial frequency information, then sent to later stages, enter a "memory bank" in the latent space where they are compared (e.g., via \acrshort{label_MSE} loss or Cosine distance Loss) to representations of similar objects. This way, the number of object representations required for matching is minimized and object recognition is facilitated.
Finally, an interesting potential way forward is implementing Hebbian Learning training, which leverages the biological learning rule found in the human brain: “Neurons that fire together wire together.” 

\subsection{Exod-IA: the Forbidden One}
We close the dissertation by tracing a fun connection with the \textit{Yu-Gi-Oh!} card "\textit{Exodia the Forbidden One}." Exodia is one of the most iconic cards in the franchise and is notoriously part of a combo that automatically leads the player to win. Its effect reads:
\begin{quote}
If you have \textit{"Right Leg of the Forbidden One"}, \textit{"Left Leg of the Forbidden One"}, \textit{"Right Arm of the Forbidden One"} and \textit{"Left Arm of the Forbidden One"} in addition to this card in your hand, you win the Duel.
\end{quote}

Here, we ironically take inspiration from the five Exodia cards and propose an idea for future developments, which we could call \textbf{Exod-IA}\footnote{the diphthong "IA" represents "Intelligenza Artificiale" (i.e., "Artificial Intelligence" in Italian).}. Much like collecting the five cards in the game, we foresee merging current methods and future advancements of the approaches we proposed in the five operative chapters of this thesis (\ref{chap:XAI_wyn} to \ref{chap:cocoreco_ECCV}, excluding the review in \ref{chap:review_XAI_causality}).
\begin{figure}
\includegraphics[width=\textwidth]{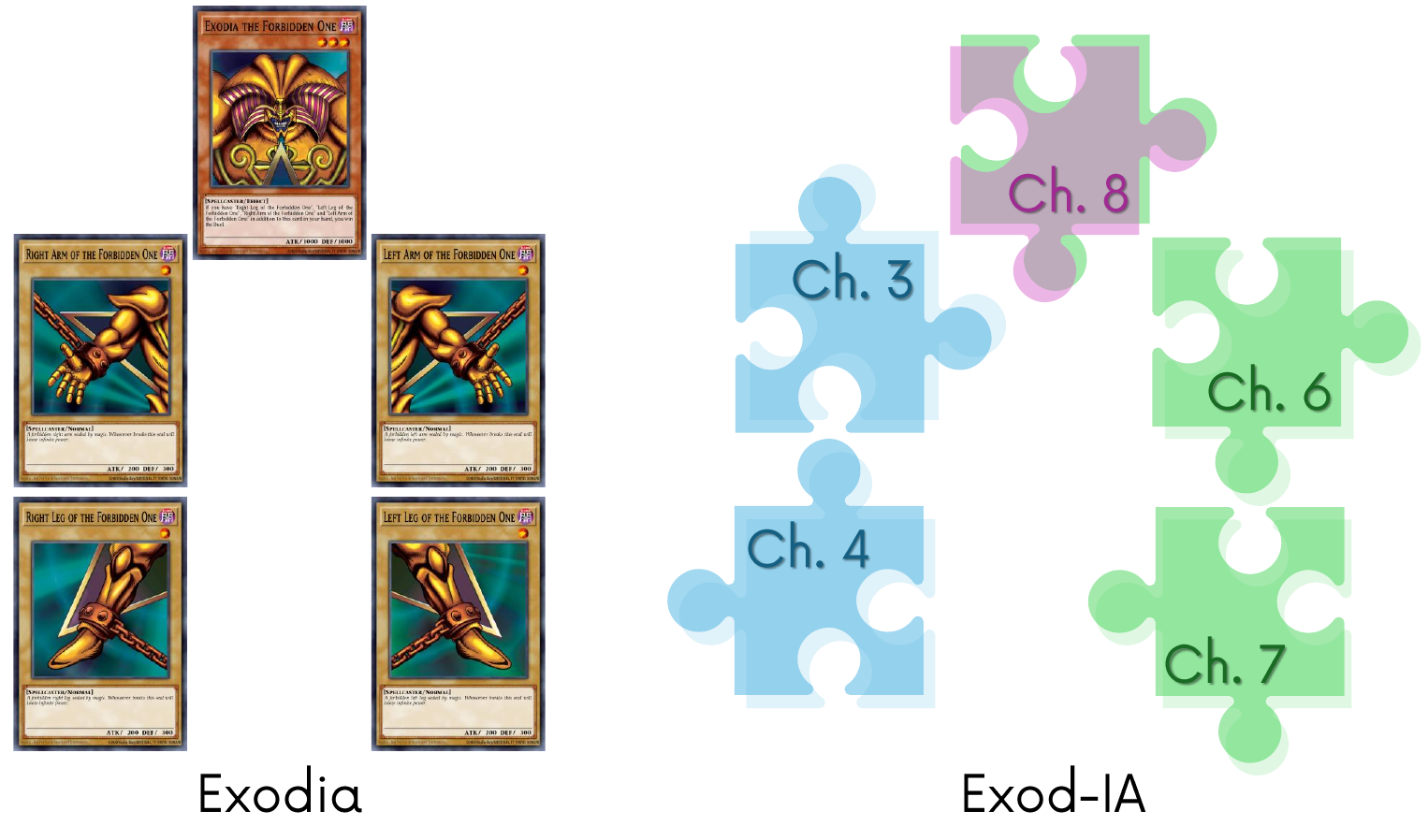}
\caption{Exod-IA: the Forbidden One.}
\label{fig:exodia}
\end{figure}

From Chapter \ref{chap:crocodile}, we take our proposed \textit{CROCODILE} framework as a suitable starting point for the Exod-IA network, given its solid inductive bias, modular nature, and demonstrated efficacy in tackling \acrshort{label_OOD} scenarios. Since its first stage involves extracting features via convolutional backbones, we could plug several of our contributions in there. From Chapter \ref{chap:mulcat_ICCV_ESWA}, we could exploit the \textit{Mulcat} option and utilize the \textit{causality factors extractor} to obtain a causality-driven set of features. Inspired by what we did in Chapter \ref{chap:cocoreco_ECCV}, we could place those feature-map enhancement blocks at multiple bottleneck locations to guide model learning at different resolutions. Additionally, we could conceive top-down modulations of shallower representations by deeper ones in a kind of bio-inspired backward recursiveness.

The next pivotal idea would be leveraging a prototypical part-learning scheme (as in Chapter \ref{chap:XAI_protopnet}) and merging it into our wider architecture. Placing the prototype layer right after CROCODILE's feature extraction or feature learning blocks would be wrong, as at that point in the architecture the feature representations are entangled (i.e., causal and spurious sets are not separated). Instead, we foresee having causally grounded prototypes by inserting a prototype learning layer after causal feature separation. It would map the high-dimensional causal embeddings to a prototype space. Here, the prototypes can be used to regularize the embedding space, encouraging the embeddings to align closely with the learned prototypes, enhancing interpretability and robustness. We could then design a hybrid loss function that combines the standard \acrshort{label_CE} loss used in classification with the prototype similarity loss. This way, the Transformer layers would be trained to not only classify the input based on embeddings but also maximize the similarity to the correct prototypes, ensuring the embeddings are both discriminative and interpretable.

Under this setting, we would then draw on \acrshort{label_AM} (Chapter \ref{chap:XAI_wyn}), specifically its \textit{in-the-loop} advancement we proposed in this section. By incorporating it into the training loop, we could regularize the feature space by generating synthetic samples that force the model to generalize, thus preventing over-reliance on dataset-specific biases. Also, exposing the network to diverse, maximally activating inputs different from training data, discouraging the network from taking shortcuts.
Moreover, we could initiate an \acrshort{label_AM} procedure (e.g., at the end of every 5 epochs) for each class of interest to generate a batch of images that maximize the activation of the causal features learned by the Transformer layer. We then could either incorporate them into the training loop for a few iterations (i.e., exposing the model to its own high-confidence synthetic examples) or use them to refine the prototypes (e.g., updating the prototypes to not only be similar to actual causal features extracted from real images but also to these synthetic examples). 

\begin{figure}[t]
\includegraphics[width=\textwidth]{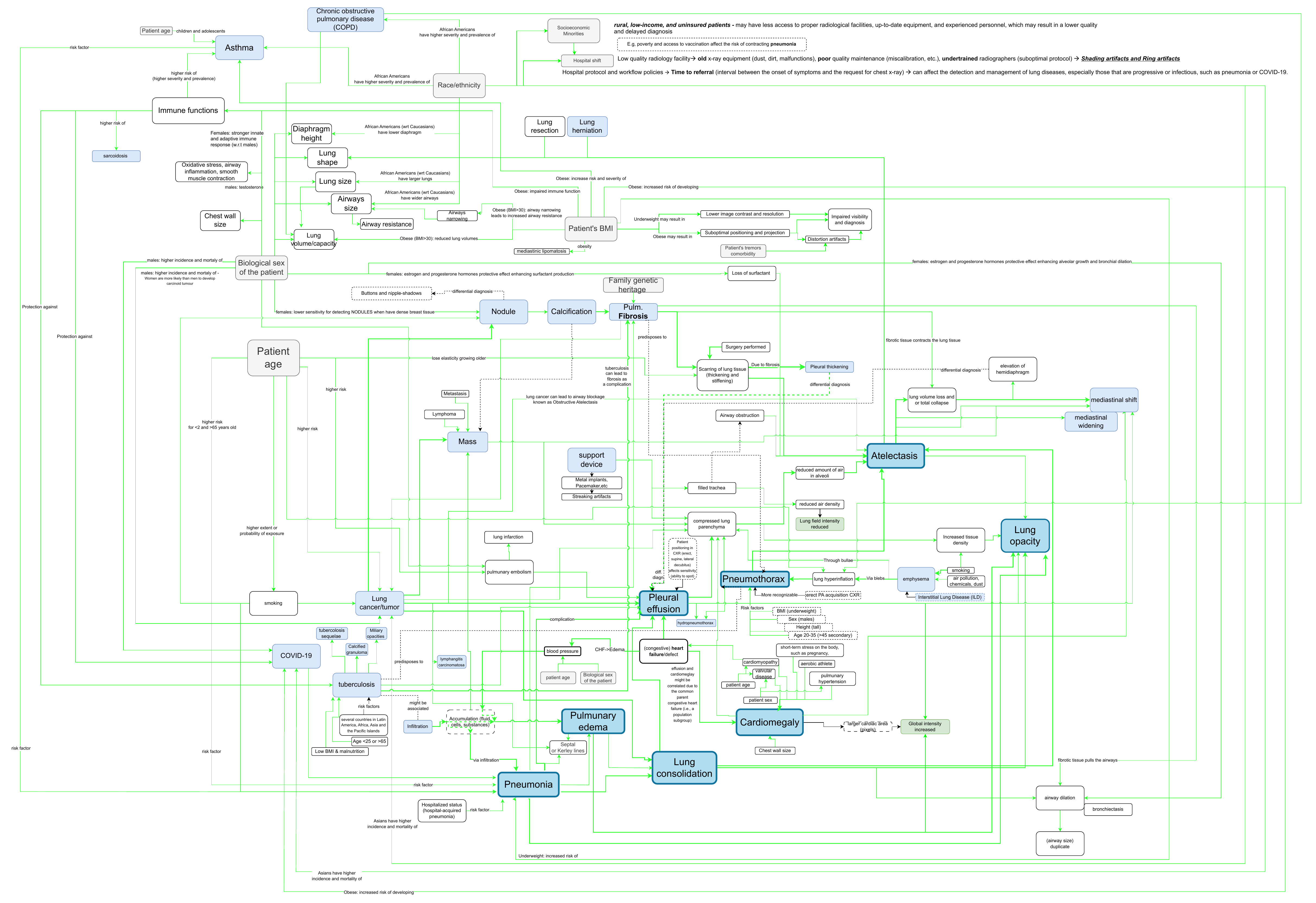}
\caption{Extended causal graphical model of the data-generating process of CXR findings from patient-specific attributes and demographic information.}
\label{fig:prior_radiology_extended}
\end{figure}

\begin{figure}[b]
\includegraphics[width=0.99\textwidth]{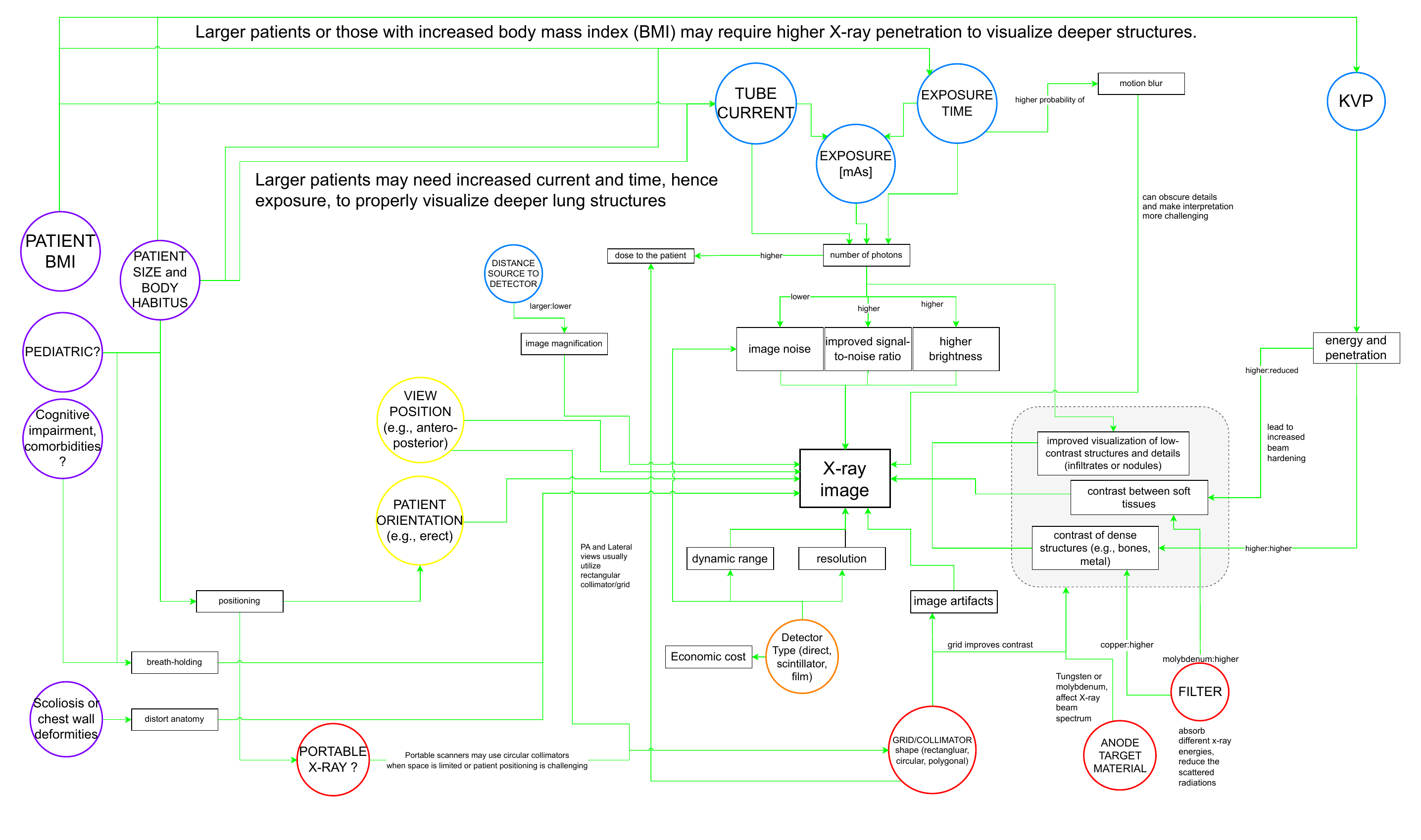}
\caption{Extended causal graphical model of the data-generating process of CXR images from patient-specific attributes, radiography settings, and acquisition parameters.}
\label{fig:prior_radiography_extended}
\end{figure}

On a final note, our proposal of injecting prior (medical) knowledge into the model's training (Chapter \ref{chap:crocodile}) proved effective. We could extend our understanding of the data generation process of radiological images and model the graph from new perspectives. In this regard, Figures \ref{fig:prior_radiology_extended} and \ref{fig:prior_radiography_extended} give a glimpse of our very recent investigations on the possible relationships among patient demographics, personal attributes, radiography equipment parameters, and acquisition settings. Thus, access to well-organized metadata and the complete set of DICOM tags associated with medical imaging datasets will be crucial.

In this section, we gave closing remarks, discussed interesting future developments, and argued a possible way to combine our major contributions into a single framework.
In the author's view, this thesis constitutes a step forward to attain a more human-aligned \acrshort{label_DL}. 
\newpage
\cleardoublepage
\phantomsection
\addcontentsline{toc}{chapter}{\bibname}
\tiny	
\bibliographystyle{plain}
\bibliography{thesis}
\end{document}